%% file: main.tex
\pdfoutput=1
\documentclass[
    acmtog,
    nonacm
]{acmart}

\usepackage{tabularx}
\usepackage{xcolor}
\usepackage{tikz}
\usetikzlibrary{arrows.meta,calc}
\usepackage[noabbrev,capitalize]{cleveref}

\usepackage{algorithm}
\usepackage{algpseudocode}
\usepackage{multirow}
\usepackage{booktabs}
\usepackage{subcaption}

\input{commands.tex}

\title[CiGS: Conduction in Gaussian Splats]{CiGS: Conduction in Gaussian Splats for Interior Temperature Field from Thermal Images}

\author{Liu Zhenyuan}
\affiliation{
  \institution{ETH Z{\"u}rich}
  \city{Z{\"u}rich}
  \country{Switzerland}
}
\email{zhenyliu@ethz.ch}

\author{Bharath Seshadri}
\affiliation{
  \institution{ETH Z{\"u}rich}
  \city{Z{\"u}rich}
  \country{Switzerland}
}
\email{seshadri@arch.ethz.ch}

\author{Bailey Miller}
\affiliation{
  \institution{Carnegie Mellon University}
  \city{Pittsburgh}
  \country{United States}
}
\email{baileymark.miller@gmail.com}

\author{George Kopanas}
\affiliation{
  \institution{Google DeepMind}
  \city{Los Angeles}
  \country{United States}
}
\email{george@runwayml.com}

\author{Ioannis Gkioulekas}
\affiliation{
  \institution{Carnegie Mellon University}
  \city{Pittsburgh}
  \country{United States}
}
\email{igkioule@andrew.cmu.edu}

\author{Bernd Bickel}
\affiliation{
  \institution{ETH Z{\"u}rich}
  \city{Z{\"u}rich}
  \country{Switzerland}
}
\email{bickelb@ethz.ch}

\begin{document}

\input{0-teaser}
\input{0-abstract}

\maketitle
\input{1-intro}
\input{2-related-work}

\input{3-method}

\input{4-results}

\input{5-conclusion}

\bibliographystyle{ACM-Reference-Format}
\bibliography{references}

\appendix
\input{7-appendix}

\end{document}

%% file: commands.tex
\usepackage{cirl}
\usepackage{xspace}
\makeatletter
\DeclareRobustCommand\onedot{\futurelet\@let@token\@onedot}
\def\@onedot{\ifx\@let@token.\else.\null\fi\xspace}

\makeatother

\newcommand{\x}{\mathbf{x}}

\newcommand{\q}{\mathbf{q}}

\newcommand{\bN}{\mathbf{N}}

\newcommand{\bc}{\mathbf{c}}

\newcommand{\bn}{\mathbf{n}}
\newcommand{\bp}{\mathbf{p}}

\newcommand{\bx}{\mathbf{x}}
\newcommand{\by}{\mathbf{y}}

\newcommand{\dd}{\mathrm{d}}

\newcommand{\green}[2]{
    G( {#1} , {#2} )
}
\newcommand{\poisson}[2]{{
    \frac{\partial \green{#1}{#2}}{\partial \bn_{#2}}
}}
\newcommand{\adjpoisson}[2]{{
    \frac{\partial \green{#1}{#2}}{\partial \bn_{#1}}
}}
\newcommand{\normalderiv}[1]{
    \frac{\partial u(#1)}{\partial \bn_{#1}}
}

\usepackage{xspace}
\newcommand{\methodname}{our method\xspace}

\usepackage{gensymb}

%% file: 0-teaser.tex
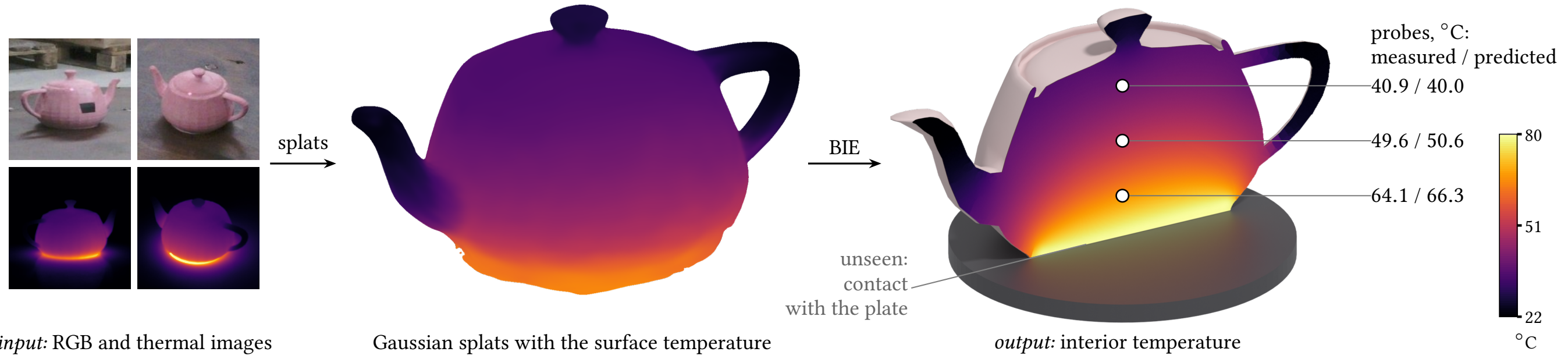
\begin{teaserfigure}
 \centering
 \input{figures/teaser3/layout} \begin{tikzpicture}[every node/.style={inner sep=0, outer sep=0}, lab/.style={font=\footnotesize, anchor=north, align=center}, stp/.style={font=\footnotesize, align=center},
   arr/.style={-{Stealth[length=4pt]}, line width=0.6pt}, prb/.style={circle, draw=black, line width=0.5pt, fill=white, minimum size=4pt, inner sep=0}]
   \pgfmathsetlengthmacro{\tu}{1.133*\textwidth}
   \pgfmathsetlengthmacro{\wp}{0.26*\tu}\pgfmathsetlengthmacro{\hp}{\wp/\teaserAspect}\pgfmathsetlengthmacro{\wt}{0.068*\tu}\pgfmathsetlengthmacro{\gap}{0.004*\tu}
   \pgfmathsetlengthmacro{\yin}{0.5*\hp - \wt - 0.5*\gap}\pgfmathsetlengthmacro{\ymid}{0.5*\hp}
   \pgfmathsetlengthmacro{\xin}{2*\wt + \gap}\pgfmathsetlengthmacro{\xsp}{\xin + 0.045*\tu}\pgfmathsetlengthmacro{\xcu}{\xsp + \wp + 0.045*\tu}
   \node[anchor=south west] (T1) at (0, \yin) {\includegraphics[width=\wt]{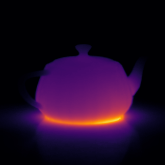}};
   \node[anchor=south west] (R1) at (0, \yin + \wt + \gap) {\includegraphics[width=\wt]{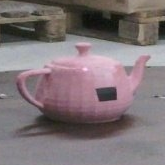}};
   \node[anchor=south west] (T2) at (\wt + \gap, \yin) {\includegraphics[width=\wt]{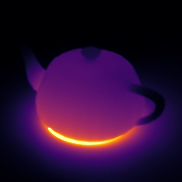}};
   \node[anchor=south west] (R2) at (\wt + \gap, \yin + \wt + \gap) {\includegraphics[width=\wt]{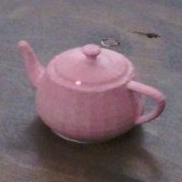}};
   \node[anchor=south west] (SP) at (\xsp, 0) {\includegraphics[width=\wp]{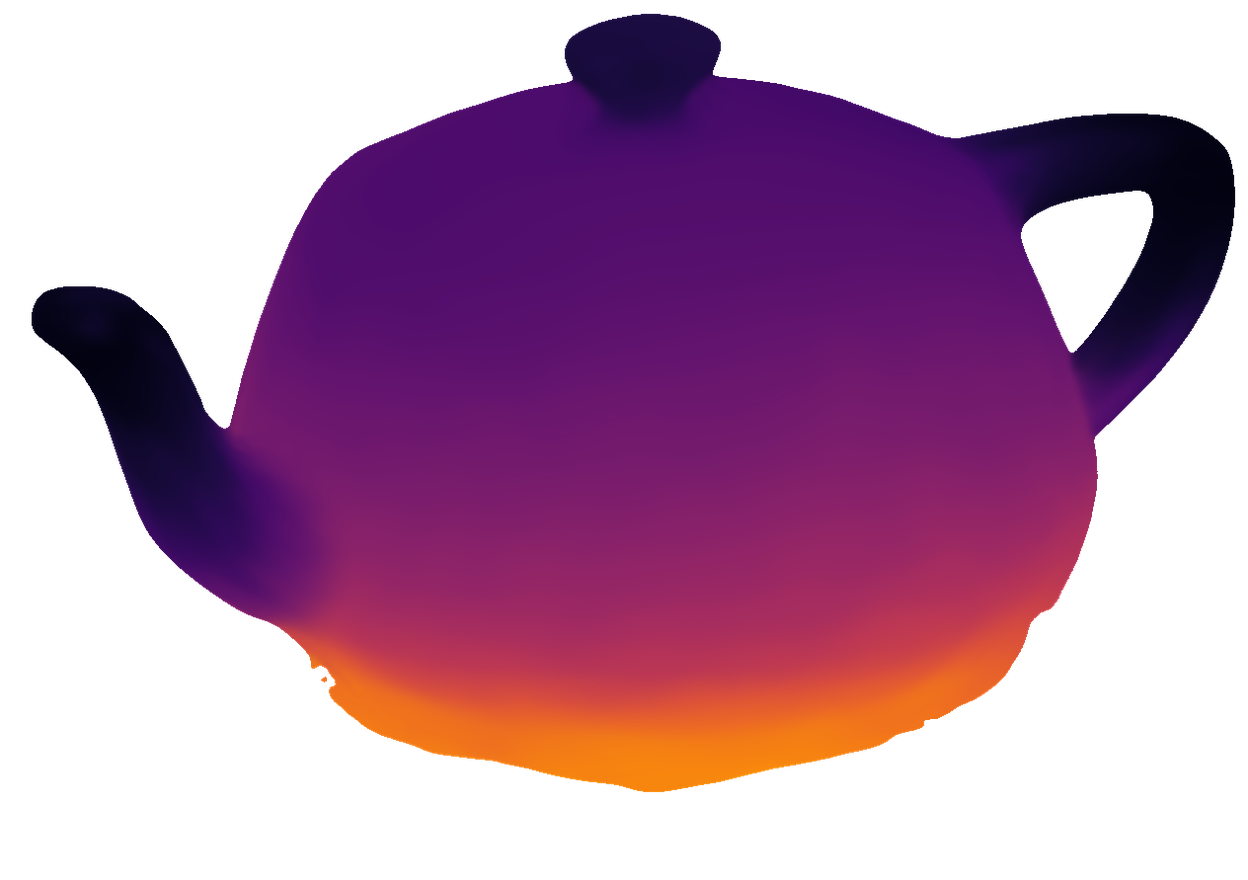}};
   \node[anchor=south west] (CU) at (\xcu, 0) {\includegraphics[width=\wp]{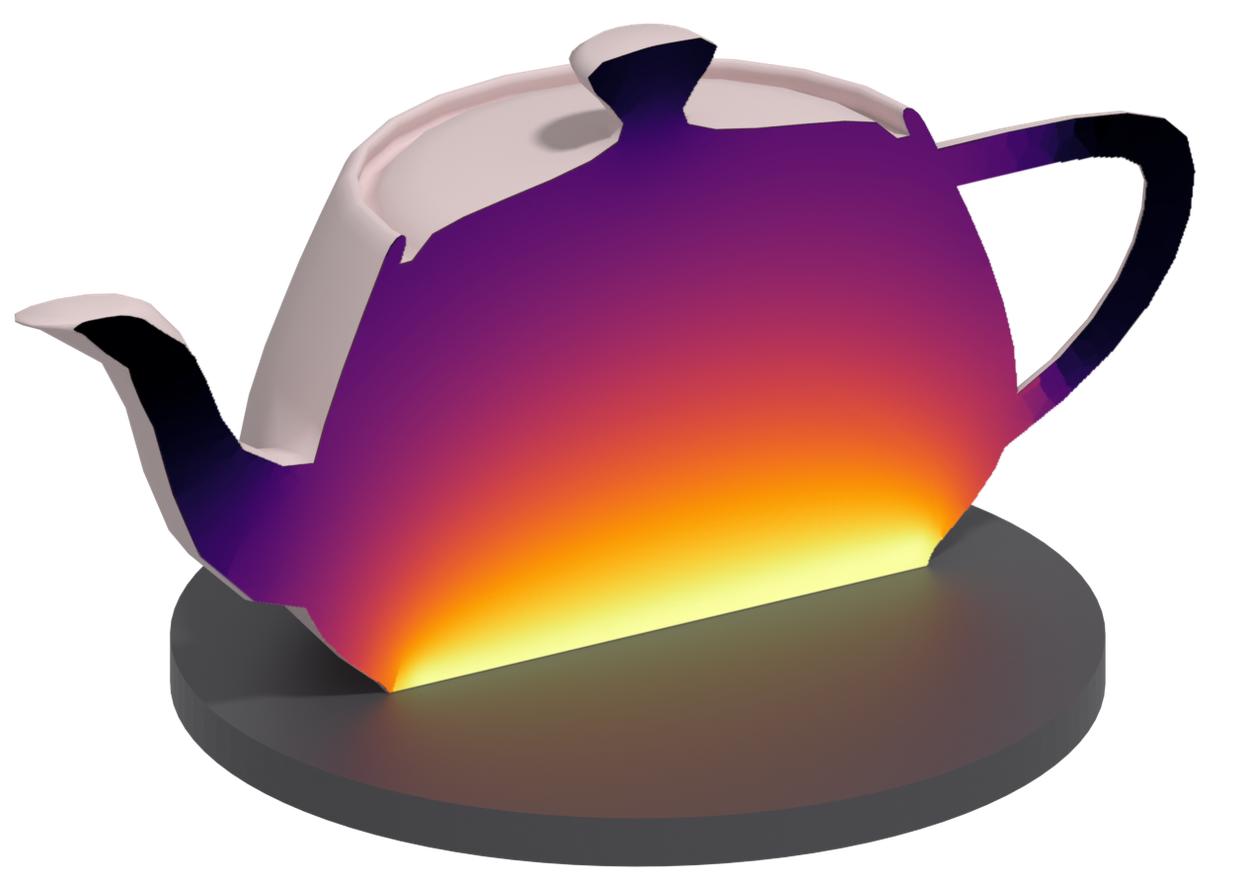}};
   \draw[arr] (\xin + 1.5*\gap, \ymid) -- (\xsp - 0.5*\gap, \ymid) node[midway, stp, anchor=south, yshift=3pt] {splats};
   \draw[arr] (\xsp + \wp + 0.5*\gap, \ymid) -- (\xcu - 0.5*\gap, \ymid) node[midway, stp, anchor=south, yshift=3pt] {BIE};
   \coordinate (UNP) at (\xcu + \teaserCx*\wp, \teaserCy*\hp);
   \node[stp, anchor=east, text=black!60, align=right] (UN) at (\xcu + 0.05*\wp, 0.12*\hp) {unseen:\\contact\\with the plate};
   \draw[black!50, line width=0.4pt] ([xshift=1pt]UN.east) -- (UNP);
   \pgfmathsetlengthmacro{\xl}{\xcu + \wp + 0.012*\tu}
   \coordinate (PA) at (\xcu + \teaserPAx*\wp, \teaserPAy*\hp); \coordinate (PB) at (\xcu + \teaserPBx*\wp, \teaserPBy*\hp); \coordinate (PC) at (\xcu + \teaserPCx*\wp, \teaserPCy*\hp);
   \node[stp, anchor=west] (LA) at (\xl, \teaserPAy*\hp) {$\teaserPAm$ / $\teaserPAp$};
   \node[stp, anchor=west] (LB) at (\xl, \teaserPBy*\hp) {$\teaserPBm$ / $\teaserPBp$};
   \node[stp, anchor=west] (LC) at (\xl, \teaserPCy*\hp) {$\teaserPCm$ / $\teaserPCp$};
   \foreach \q in {A, B, C} {\draw[black!55, line width=0.35pt] (P\q) -- (L\q.west); \node[prb] at (P\q) {};}
   \node[stp, anchor=south west, align=left] at ([yshift=3pt]LC.north west) {probes, $^\circ$C:\\measured / predicted};
   \node[anchor=south west] (CB) at ([xshift=0.016*\tu]LA.south east |- CU.south) {\includegraphics{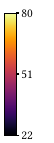}};
   \node[anchor=north, font=\scriptsize] at ([yshift=-2pt, xshift=-0.004*\tu]CB.south) {$^\circ$C};
   \node[lab] at (\wt + 0.5*\gap, -0.004*\tu) {\emph{input:} RGB and thermal images};
   \node[lab] at (\xsp + 0.5*\wp, -0.004*\tu) {Gaussian splats with the surface temperature};
   \node[lab] at (\xcu + 0.5*\wp, -0.004*\tu) {\emph{output:} interior temperature};
 \end{tikzpicture}
  \caption{%
  CiGS infers the interior temperature of an object from registered visible-spectrum and thermal images, without building a mesh.
  From $360$ RGB and thermal images of a cast plaster teapot on a hot plate,
  we reconstruct Gaussian splats that carry the surface temperature.
  Then, by discretizing a boundary integral equation (BIE), we predict the interior temperature,
  and compare with the true readings from
  the three cast-in probes.
  The readings agree with our prediction to $\qty{\teaserProbeMAE}{\degreeCelsius}$ on average.}
  \label{fig:teaser}
\end{teaserfigure}

%% file: figures/teaser3/layout.tex
\def\teaserAspect{1.4186}
\def\teaserPAx{0.5110}\def\teaserPAy{0.4018}\def\teaserPAm{64.1}\def\teaserPAp{66.3}
\def\teaserPBx{0.5110}\def\teaserPBy{0.5696}\def\teaserPBm{49.6}\def\teaserPBp{50.6}
\def\teaserPCx{0.5110}\def\teaserPCy{0.7374}\def\teaserPCm{40.9}\def\teaserPCp{40.0}
\def\teaserCx{0.4334}\def\teaserCy{0.2535}
\def\teaserProbeMAE{1.4}

%% file: 0-abstract.tex
\begin{abstract}
We present a method for inferring the interior temperature distribution of a solid object from registered multi-view visible-spectrum and thermal images, even when the thermal views cover only part of its surface.
We formulate steady-state heat conduction as a Laplace boundary value problem and solve it with a direct boundary integral equation discretized over a point cloud: a Gaussian reconstruction of the visible-spectrum images supplies the boundary geometry, and the thermal images supply the boundary data.
We do not build surface or volume meshes at any stage in our method.
Instead, we sample collocation points from rendered depth without the pixels at depth discontinuities, and use a density subtraction that keeps the interior evaluation accurate up to the surface.
The whole chain is differentiable, and we use it to recover the pose and depth of heating pipes, insulation defects, and the shape of  cracks from thermal images.
We validate our method through experiments on boundary value problems with Dirichlet-only and mixed Dirichlet--Robin boundaries, using synthetic and real captured data on scenes with challenging topology. These experiments show that our method matches analytical solutions and has better accuracy than FEM solvers.
\end{abstract}

%% file: 1-intro.tex
\section{Introduction}

Thermal cameras measure the infrared radiation emitted and reflected by a surface, providing a contactless means of measuring an object's apparent surface temperature.
Thermography is used in applications such as industrial manufacturing inspection, building energy audits, and electronics cooling. However, in these applications, the heat distribution and temperatures \emph{inside} the object reveal more critical information than only surface captures.
For example, a floor-heating pipe embedded within a concrete slab, or insulation that has settled behind a facade, shows only as a faint pattern on the surface the camera sees.
To infer the interior from the surface temperature, we need two ingredients: a heat conduction model and the shape of the object.
For an object without an existing shape representation such as a mesh or CAD model, we need to reconstruct the shape from the images.

For the heat conduction model, we use the following: in a thermal steady state, the interior temperature of a homogeneous isotropic solid object without internal heat sources satisfies the Laplace equation \cite{carslawConductionHeatSolids1959}, subject to appropriate boundary conditions on the object's surface: surface with known temperatures modeled as Dirichlet boundaries, insulated or convecting faces can be modeled as Neumann or Robin boundaries.
The finite element method (FEM) solves this boundary value problem on a volume mesh of the object.
Boundary element methods (BEM) and, more recently, grid-free Monte Carlo solvers \citep{sawhneyMonteCarloGeometry2020, sawhneyWalkStarsGridFree2023, sugimotoPracticalWalkonBoundaryMethod2023} can solve such boundary value problems efficiently and without volumetric meshing.
However, all of these methods require the object's surface geometry to be given in advance, as a mesh or an implicit surface.
Obtaining a simulation-ready mesh from images, though, takes a chain of separate tools:
an image-based mesh reconstruction tool, a repair pass that removes stray pieces and closes holes, and, for a finite-element solver, volumetric meshing where the run time grows steeply as its elements shrink (\cref{ssec:e2e-comparison}).
Radiance field methods, neural or Gaussian, instead take the images as they come and reconstruct thermally consistent scenes directly from the captured views \citep{yeThermalNeRFNeuralRadiance2024, hassanThermoNeRFMultimodalNeural2025, luThermalGaussianThermal3D2024}, but they treat temperature as a learned appearance attribute and do not connect the representation to the heat equation, so they cannot predict the temperature inside the object.

We present a method that bridges this gap: a Gaussian radiance field \cite{zhangRaDeGSRasterizingDepth2024} reconstructs the object from visible-spectrum images, and the surface it renders is sampled into the point cloud on which a boundary integral equation (BIE) is discretized.
Thermal images set the temperature at the surface locations they observe, while a physical prior, such as insulation or convective cooling, applies in the remaining areas. A single solve of the direct BIE then yields the temperature at any interior point.
We do not implement meshing at any stage, surface points require no connectivity, and the interior is never discretized.
For accuracy, we sample points from the rendered surface, with a depth-slope filter that rejects pixels that straddle two surfaces, and a density subtraction keeps the interior evaluation accurate up to the surface.
Our method is also differentiable, so the gradient of a loss on the predicted temperature reaches the boundary conditions and the geometry, and the same solve recovers what the camera cannot see.

We evaluate \methodname on synthetic scenes with known ground truth and on a real capture, across two settings: \emph{temperature interpolation}, where thermal images cover the entire object surface (pure Dirichlet BVP), and \emph{incomplete thermal capture}, where only one thermal view is available and the unseen geometry is assigned an insulating (Neumann) or convective (Robin) condition (mixed BVP), and compare against a mesh-based pipeline built from the same images.
Skipping the meshing saves most of the time after the reconstruction: tetrahedralization alone takes the mesh pipeline $78$ to $518$\,s per scene, while our whole chain takes about half a minute ($31$ against $210$\,s on \textsc{Armadillo}), and its error is on par or lower (\cref{ssec:e2e-comparison}).
In particularly complex meshes, such as those containing thin, strut-like geometry, the mesh pipeline breaks down: it either thins the struts excessively, losing up to half the volume, or closes the openings between them. For the \textsc{V-bunny} shown in \cref{fig:meshers}, FEM yields no solution for 10 to 44\% of the true interior.
\input{figures/meshweak-vbunny/meshers-snippet}

We model the thermal conduction with the Laplace equation,
so the object must be homogeneous, isotropic, and at steady state, with no source generated within the problem domain.
A heat source embedded in the object, such as a heating pipe under a floor, can be modeled as an interior boundary (\cref{ssec:pipe}).
When a clean CAD model of the object is available, FEM is mature and remains the better tool.
Our method targets scenarios that have no CAD model and are known only through their images: one reconstruction provides both the surface the solver discretizes and the boundary data read from the thermal images. Through differentiability, we show that our model recovers an embedded floor-heating pipe, insulation defects behind a facade and, with a Gaussian splat flow, the shape of a crack in a propeller hub (\cref{fig:splat-flow}). Furthermore, we validate the method on a real capture of a heated solid-cast teapot whose embedded temperature probes measured the interior temperature.
\input{figures/splat-flow/splat-flow-snippet}

In summary, our contributions are:
\begin{itemize}
    \item The first boundary integral solver that runs on a Gaussian reconstruction: it infers the 3D interior temperature of an object from RGB and thermal images without building a surface or volume mesh, and is differentiable end to end;
    \item A collocation point cloud suitable for BIEs, extracted from the rendered depth of the reconstruction, with a depth-slope filter that rejects the pixels at depth discontinuities;
    \item A dataset of eighteen scenes with visible-spectrum and thermal views and reference interior fields, and a real capture with embedded temperature probes.
\end{itemize}

%% file: figures/meshweak-vbunny/meshers-snippet.tex
\begin{figure}[t]
\centering
\definecolor{mwpurple}{HTML}{9467BD}%
\begin{tikzpicture}[every node/.style={inner sep=0, outer sep=0}, pn/.style={anchor=north west, draw=black!30, line width=0.4pt}]
  \pgfmathsetlengthmacro{\pw}{(\columnwidth-21pt)/5}\def\gap{2pt}
  \node[anchor=north west] (T0) at (0,0) {\includegraphics[width=\pw]{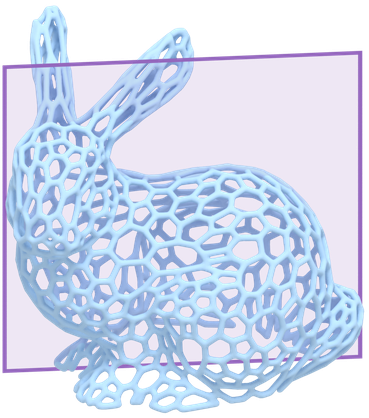}};
  \node[pn] (T1) at ([yshift=-\gap]T0.south west) {\includegraphics[width=\pw]{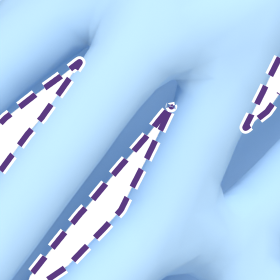}};
  \node[anchor=north west] (T2) at ([yshift=-\gap]T1.south west) {\includegraphics[width=\pw, height=\pw]{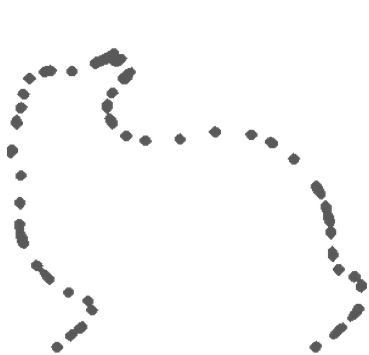}};
    \node[anchor=north west] (R0) at ([xshift=\gap]T0.north east) {\includegraphics[width=\pw]{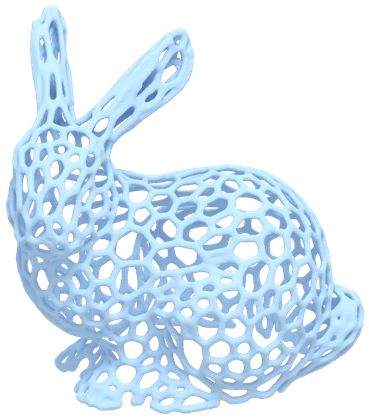}};
    \node[pn] (R1) at ([yshift=-\gap]R0.south west) {\includegraphics[width=\pw]{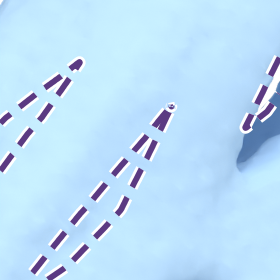}};
    \node[anchor=north west] (R2) at ([yshift=-\gap]R1.south west) {\includegraphics[width=\pw]{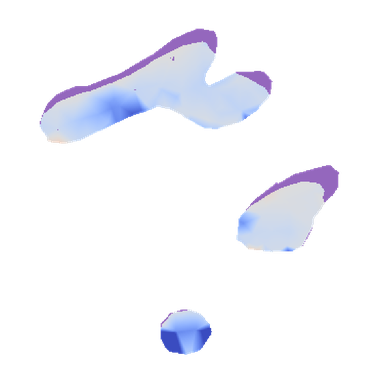}};
    \node[anchor=north west] (M0) at ([xshift=\gap]R0.north east) {\includegraphics[width=\pw]{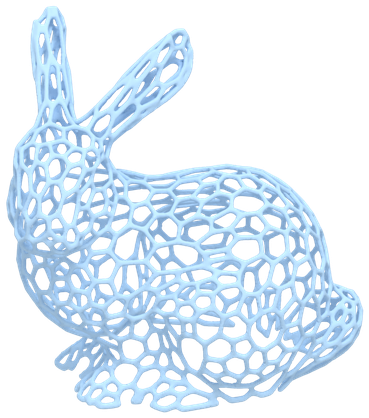}};
    \node[pn] (M1) at ([yshift=-\gap]M0.south west) {\includegraphics[width=\pw]{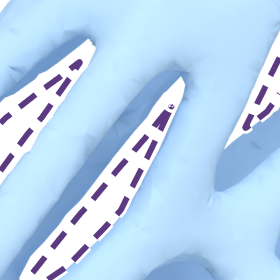}};
    \node[anchor=north west] (M2) at ([yshift=-\gap]M1.south west) {\includegraphics[width=\pw]{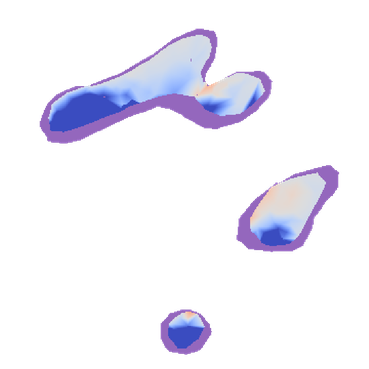}};
    \node[anchor=north west] (P0) at ([xshift=\gap]M0.north east) {\includegraphics[width=\pw]{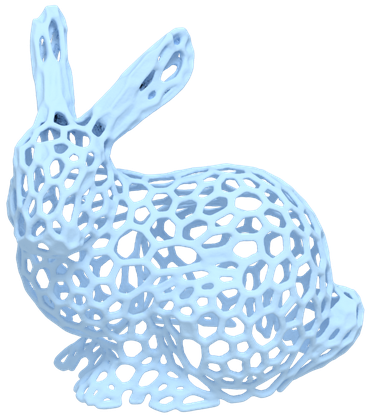}};
    \node[pn] (P1) at ([yshift=-\gap]P0.south west) {\includegraphics[width=\pw]{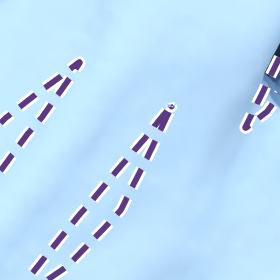}};
    \node[anchor=north west] (P2) at ([yshift=-\gap]P1.south west) {\includegraphics[width=\pw]{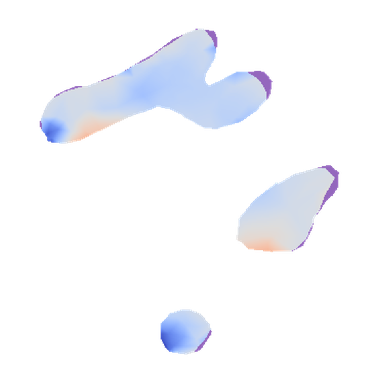}};
    \node[anchor=north west] (O0) at ([xshift=\gap]P0.north east) {\includegraphics[width=\pw]{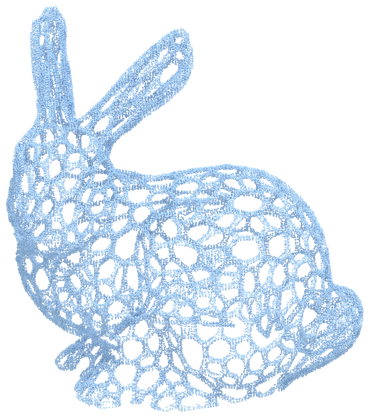}};
    \node[pn] (O1) at ([yshift=-\gap]O0.south west) {\includegraphics[width=\pw]{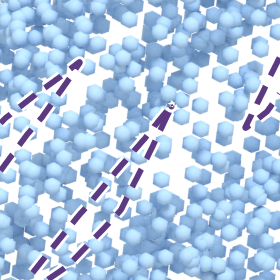}};
    \node[anchor=north west] (O2) at ([yshift=-\gap]O1.south west) {\includegraphics[width=\pw]{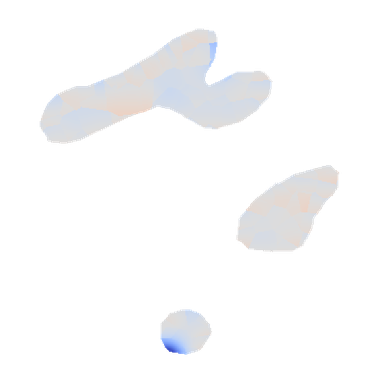}};
  \foreach \n in {T2, R2, M2, P2, O2} {\draw[mwpurple, line width=0.5pt] (\n.south west) rectangle (\n.north east);}
  \foreach \c in {T, R, M, P, O} {
    \begin{scope}[shift={(\c0.south west)}, x={($(\c0.south east)-(\c0.south west)$)}, y={($(\c0.north west)-(\c0.south west)$)}]
      \draw[black!60, line width=0.5pt] (0.3156,0.7465) rectangle ++(0.0656,0.0578);
      \coordinate (wl-\c) at (0.3156,0.7465); \coordinate (wr-\c) at (0.3812,0.7465);
    \end{scope}
  }
  \draw[black!60, line width=0.5pt, densely dotted] (wl-T) -- (T1.north west) (wr-T) -- (T1.north east);
  \begin{scope}[shift={(T2.south west)}, x={($(T2.south east)-(T2.south west)$)}, y={($(T2.north west)-(T2.south west)$)}]
    \draw[mwpurple, line width=0.6pt] (0.2334,0.7236) rectangle ++(0.14,0.1453);
    \coordinate (st) at (0.3734,0.8689); \coordinate (sb) at (0.3734,0.7236);
  \end{scope}
  \draw[mwpurple, line width=0.5pt, densely dotted] (st) -- (R2.north west) (sb) -- (R2.south west);
  \foreach \p/\x/\y in {R2/0.88/0.465, M2/0.90/0.485, P2/0.90/0.485} {
    \begin{scope}[shift={(\p.south west)}, x={($(\p.south east)-(\p.south west)$)}, y={($(\p.north west)-(\p.south west)$)}]
      \node[font=\scriptsize, anchor=south east, align=right, inner sep=0] at (0.97,0.03) {geometry\\[-1.5pt]error};
      \draw[-{Stealth[length=2.2pt, width=2pt]}, line width=0.45pt] (\x,0.285) -- (\x,\y);
    \end{scope}
  }
  \foreach \n/\t in {T0/{ground truth}, R0/{RaDe-GS}, M0/{MILo}, P0/{TopoSurfel}, O0/{ours}} {\node[font=\footnotesize, anchor=base] at ([yshift=3pt]\n.north) {\t};}
  \foreach \n/\t in {T0/{full view}, T1/{close-up}, T2/{interior}} {\node[font=\footnotesize, rotate=90, anchor=base] at ([xshift=-3pt]\n.west) {\t};}
  \foreach \n/\t in {R2/{$0.0077$}, M2/{$0.023$}, P2/{$0.0094$}, O2/{$0.0019$}} {\node[font=\scriptsize, anchor=base] at ([yshift=-7.5pt]\n.south) {\t};}
  \node[font=\scriptsize, anchor=base] at ([yshift=-7.5pt]T2.south) {MAE};
  \node[anchor=north] (CB) at ([yshift=-11pt]$(M2.south)!0.5!(P2.south)$) {\includegraphics{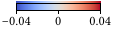}};
  \foreach \p/\t in {R2/FEM, M2/FEM, P2/FEM, O2/BIE} {\node[font=\scriptsize, anchor=north west] at ([xshift=2pt, yshift=-2pt]\p.north west) {\t};}
  \node[font=\scriptsize] at ($(T2.south west)!0.41!(T2.north east)$) {slice};
  \node[font=\scriptsize, anchor=east] at ([xshift=-4pt, yshift=-3pt]CB.north west) {temperature error};
\end{tikzpicture}
\caption{
Existing image-based mesh reconstruction algorithms either thin the struts or fill the openings between them. The finite element solve on these meshes has large errors or gives no valid solutions on part of the geometry (bottom row); our boundary integral solve builds no mesh, and performs robustly.}
\label{fig:meshers}
\end{figure}
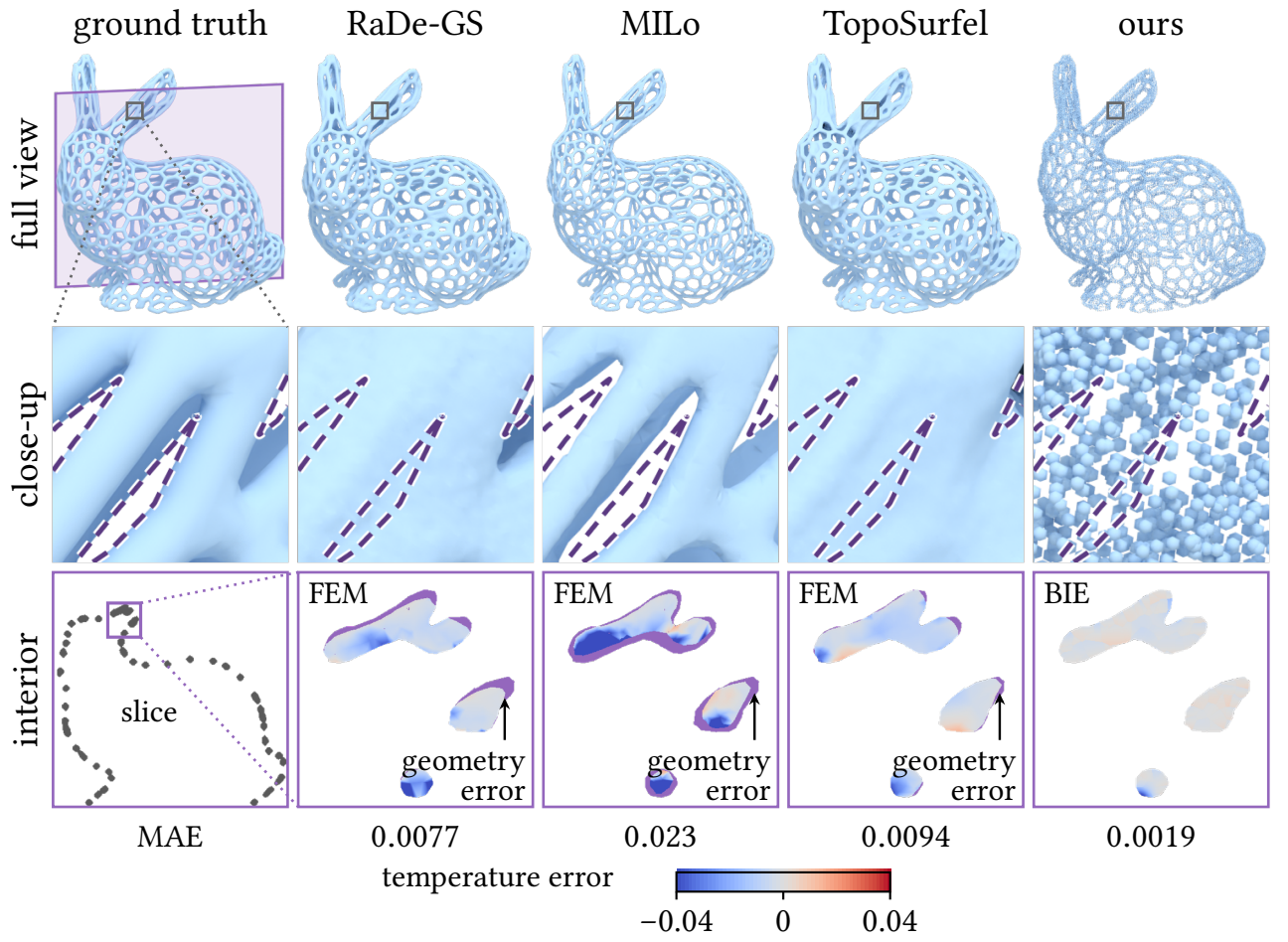

%% file: figures/splat-flow/splat-flow-snippet.tex
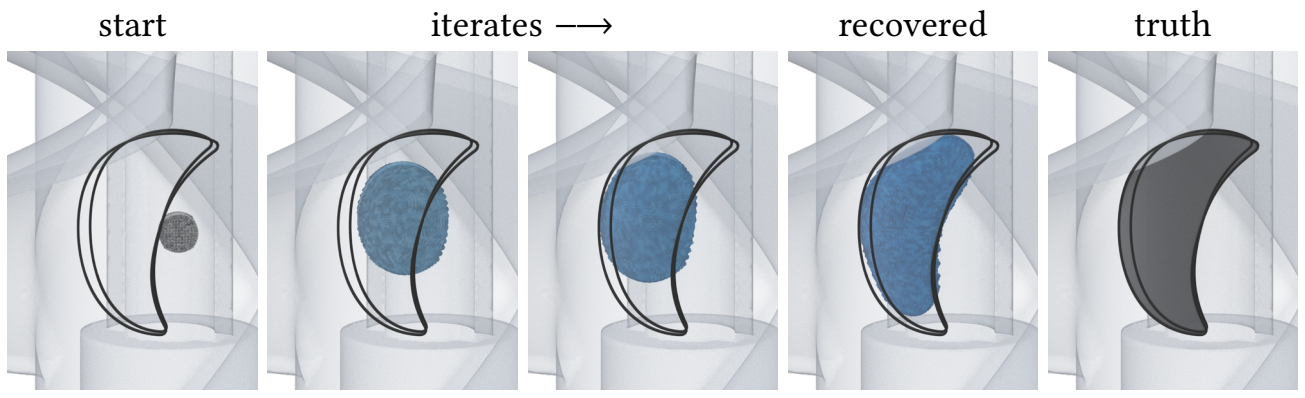
\begin{figure}[t]
\centering
\begin{tikzpicture}[every node/.style={inner sep=0, outer sep=0}, lab/.style={font=\footnotesize, anchor=base}]
  \pgfmathsetlengthmacro{\ph}{(\columnwidth-8pt)/(5*0.74)}\def\gap{2pt}
  \node[anchor=south west] (A) at (0,0) {\includegraphics[height=\ph]{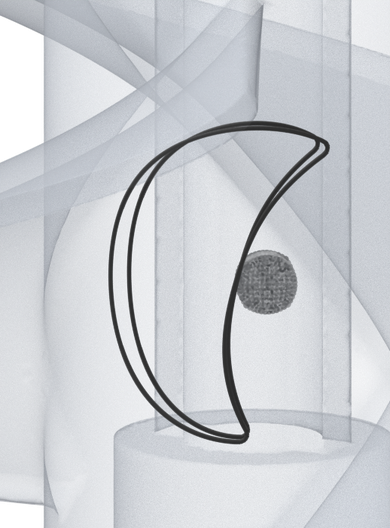}};
  \node[anchor=south west] (B) at ([xshift=\gap]A.south east) {\includegraphics[height=\ph]{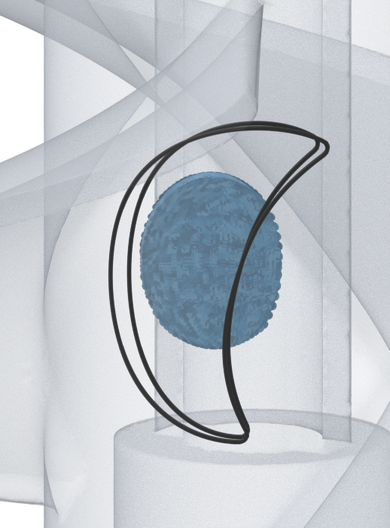}};
  \node[anchor=south west] (C) at ([xshift=\gap]B.south east) {\includegraphics[height=\ph]{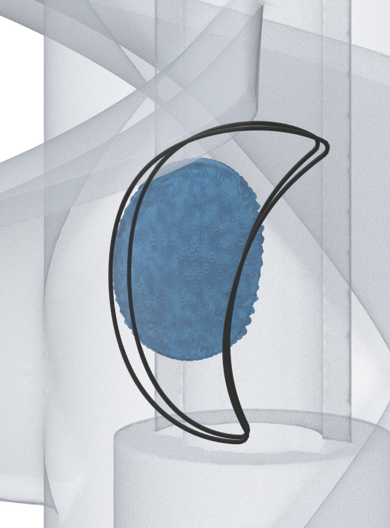}};
  \node[anchor=south west] (D) at ([xshift=\gap]C.south east) {\includegraphics[height=\ph]{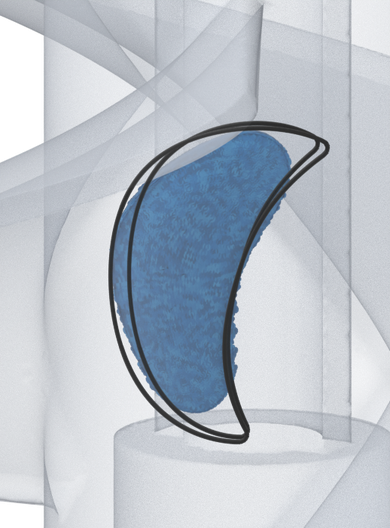}};
  \node[anchor=south west] (E) at ([xshift=\gap]D.south east) {\includegraphics[height=\ph]{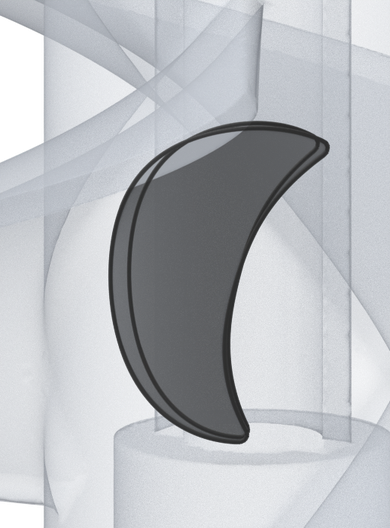}};
  \foreach \n/\t in {A/start, D/recovered, E/truth} {\node[lab] at ([yshift=3pt]\n.north) {\t};}
  \node[lab] at ([yshift=3pt]$(B.north)!0.5!(C.north)$) {iterates $\longrightarrow$};
\end{tikzpicture}
\caption{Our method can recover a crack hidden inside a propeller hub from thermal images (\cref{ssec:crack}), without meshing at any step.}
\label{fig:splat-flow}
\end{figure}

%% file: 2-related-work.tex
\section{Related Work}

\paragraph{Thermal conduction}

Finite element packages, commercial \cite{ansysAnsys2024, comsol} and open \cite{barattaDOLFINxNextGeneration2023, elmer}, are mature modeling tools for heat conduction. But they start with defining geometry and generating a mesh, which the measurements do not directly provide.

Monte Carlo random walks solve the Dirichlet problem at arbitrary query points without a volume mesh \cite{mullerContinuousMonteCarlo1956}; Sawhney and Crane \shortcite{sawhneyMonteCarloGeometry2020} introduced this method to computer graphics research. Subsequent works added Neumann and Robin conditions \cite{sawhneyWalkStarsGridFree2023, millerWalkinRobinWalk2024}, spatially varying coefficients \cite{sawhneyGridfreeMonteCarlo2022}, derivatives for inverse problems \cite{millerDifferentialWalkSpheres2024, yilmazerSolvingInversePDE}, coupling to Monte Carlo rendering for the other heat transfer modes \cite{batiCouplingConductionConvection2023}, variance reduction \cite{qiBidirectionalFormulationWalk2022, millerBoundaryValueCaching2023}, unbounded domains \cite{nabizadehKelvinTransformationsSimulations2021} and a walk on the boundary itself \cite{sugimotoPracticalWalkonBoundaryMethod2023}.
All of them take a representation of the boundary as input.
Inferring the unmeasured side of a conducting body from the temperature and flux on its measured side is the Cauchy problem of steady conduction, which \citeauthor{lesnicIterativeBEMCauchy1997}~\shortcite{lesnicIterativeBEMCauchy1997} solve by BEM on a meshed boundary; \citeauthor{beckInverseHeatConduction1985}~\shortcite{beckInverseHeatConduction1985} treat the inverse heat conduction problem, its ill-posedness and its regularization in general.
Our incomplete thermal capture avoids it by assuming a condition on the unseen side.

\paragraph{Boundary element methods}
Boundary element methods (BEM) discretize only the boundary of a linear elliptic problem, at the price of dense and often ill-conditioned systems \cite[Ch. 3--4]{sauterBoundaryElementMethods2011}; we follow this textbook treatment of the boundary integral operators throughout.
Graphics researchers have used them on authored surfaces, for deformable objects \cite{jamesArtDefoAccurateReal1999} and sound propagation \cite{jamesPrecomputedAcousticTransfer2006, chewAcousticReliefs2025}, and Chen et al. accelerate the solver with a sparse inverse Cholesky preconditioner~\shortcite{chenLightningfastMethodFundamental2024}, which they later extend to nonsymmetric systems as an inverse LU factorization~\shortcite{chenLightningfastBoundaryElement2025}.
There is also work that fits implicit surfaces to captured points with a boundary integral \cite{chen3DReconstructionFast2024} that uses the same kernel mollification as ours; we solve a physical boundary value problem on the surface rather than fitting the surface itself.
Engineering has solved potential and elasticity problems on scattered boundary points for decades, with least-squares interpolants \cite{chatiBoundaryNodeMethod1999, liBoundaryCloudMethod2002} or fundamental solutions collocated on the boundary \cite{chenSingularBoundaryMethod2010}; the regularization error of such kernels can be corrected in closed form \cite{bealeSimpleMethodComputing2016} or by extrapolation \cite{bealeExtrapolatedRegularization2024}, and on reconstructed geometry we found neither to pay (\cref{sec:discretization}).
These solvers assume that the points are given, exact, and weighted; ours come from an image-based reconstruction, so where the points lie, what area each stands for, and which of them can be trusted become part of the method.
In graphics, the Gauss-formula reconstructions of \citeauthor{luSurfaceReconstructionGauss2019}~\shortcite{luSurfaceReconstructionGauss2019} and \citeauthor{linSurfaceReconstructionGauss2023}~\shortcite{linSurfaceReconstructionGauss2023} evaluate the Laplace double layer on a point cloud with area weights, and the fast winding number of \citeauthor{barillFastWindingNumbers2018}~\shortcite{barillFastWindingNumbers2018} supplies the Voronoi areas and the winding number our membership filter reads.

\paragraph{Thermal radiation and imaging}

Thermal radiation propagates at infrared wavelengths, longer than those of visible light on the electromagnetic spectrum. Physically based rendering has been applied to thermal radiation \cite{pedroaguerreThRendRayTracing2021, batiCouplingConductionConvection2023}, including precomputed radiative transfer for interactive simulation \cite{freudePrecomputedRadiativeHeat2023} and its inverse by differentiable photon tracing \cite{freudeInverseSimulationRadiative2025}.
Closer to our setting, NTR-Gaussian \cite{yangNTRGaussianNighttime2025} attaches emissivity, convection, and heat-capacity parameters to 4D Gaussians to predict nighttime surface temperature, but their scope stays on the surface and does not solve boundary integrals.
\citeauthor{narayananShapeHeatConduction2024}~\shortcite{narayananShapeHeatConduction2024} recover an object's shape from thermal videos of heat conducting through it, with a surface approximation of the transport in the volume; we solve the steady conduction in the volume itself.

\paragraph{Gaussian Splatting and Thermal Radiance Fields}
Neural radiance fields \cite{mildenhallNeRFRepresentingScenes2020} and 3D Gaussian splatting \cite{kerbl3DGaussianSplatting2023a} reconstruct scenes from images.
Gaussian surfels do so with a surface definition \cite{huang2DGaussianSplatting2024, daiHighqualitySurfaceReconstruction2024}.
The surfel itself is the surface element of point-based rendering \cite{pfisterSurfelsSurfaceElements2000, zwickerSurfaceSplatting2001}; we keep the primitives for rendering and move the physics to points extracted from their rendered depth (\cref{ssec:area-surfels}).
Surfaces are extracted from splats by sampling the density field \cite{guedonSuGaRSurfaceAligned2024}, by fusing rendered depth across views \cite{zhangRaDeGSRasterizingDepth2024, chenPGSRPlanarBased2024}, which treats the rendered depth rather than the primitive centers as the model's surface estimate, or, for a point cloud, by sampling the primitives directly \cite{stuart3DGSToPC2025}; physics on splats has mostly meant volumetric simulation with the Gaussians as particles \cite{xiePhysGaussianPhysicsIntegrated2024}, not a boundary solve.
Their thermal extensions \cite{hassanThermoNeRFMultimodalNeural2025, luThermalGaussianThermal3D2024, yeThermalNeRFNeuralRadiance2024} handle the low texture and resolution of infrared images but treat temperature as appearance and do not solve for it inside the object.
HiGS \cite{zhenyuanInverseRadiative2025} is the closest work in representation and the farthest in problem: it reconstructs Gaussian primitives from the same kind of capture and recovers how the surface emits, absorbs and reflects, but it does not model conduction.
We assume a purely emissive surface and solve for the interior temperature field.

%% file: 3-method.tex
\section{Steady-state heat conduction as a Laplace problem}
\label{sec:laplace}
\input{figures/overview2/overview-snippet}
\paragraph{Overview and paper organization}
In the remainder of the paper, we first introduce the necessary background on BIEs in \cref{sec:laplace}.
In \cref{sec:discretization}, we extract collocation points from a Gaussian reconstruction and discretize the BIE on them.
In \cref{sec:solver}, we discuss the boundary conditions, solving the system and the differentiability of our system.
In \cref{sec:results}, we evaluate the method on synthetic scenes, a real capture and three inverse problems.
\Cref{fig:overview} illustrates our pipeline.

Given the surface temperature of a solid, its normal derivative, or a combination of the two, we seek the steady-state temperature inside.
On the domain $\Omega \subset \mathbb{R}^3$ with boundary $\Gamma = \partial \Omega$, the temperature $u: \Omega \to \mathbb{R}$ satisfies
\begin{align} \label{eq:laplace-problem}
\Delta u(\bx) & = 0,
\end{align}
with $\Delta$ the Laplace operator, under a boundary condition on $\Gamma$: Dirichlet where the temperature $u$ is known, Neumann where the flux $\partial u / \partial \bn_\bx$ is, and in general the Robin condition
\begin{align} \label{eq:robin-problem}
a(\bx) u(\mathbf{x}) + b(\bx) & \normalderiv{\bx}  = g(\bx) \;\;\; \forall \bx \in \Gamma,
\end{align}
\newsavebox{\notationinsetbox}\sbox{\notationinsetbox}{\input{figures/notation/notation-inset}}%
\noindent\begin{minipage}[c]{\dimexpr\linewidth-\wd\notationinsetbox-6pt\relax}
with known $a, b, g: \Gamma \to \mathbb{R}$, which contains the other two ($a = 1, b = 0$ and $a = 0, b = 1$); we write everything for it.
In our setting a thermal camera fixes $u = g$ where it sees the surface, and a Robin condition holds elsewhere (inset).
\end{minipage}\hspace{6pt}\begin{minipage}[c]{\wd\notationinsetbox}\usebox{\notationinsetbox}\end{minipage}\par

\paragraph{Direct boundary integral formulation}
The boundary value Laplace problem admits several boundary integral equation (BIE) formulations, which differ in the unknown on the boundary \cite[Ch. 3]{sauterBoundaryElementMethods2011}: the two \emph{indirect} ones posit a fictitious single-layer or double-layer density on $\Gamma$ whose potential reproduces $u$; the \emph{direct} one follows from Green's representation theorem and carries the boundary temperature and flux themselves.
We use the direct formulation because it is the most accurate of the three on our clouds (\cref{fig:bie-comp}); \cref{sec:bie-formulations} gives the indirect forms for comparison.
Let $q = \partial u / \partial \bn$ denote the outward normal flux on $\Gamma$.
Green's representation theorem writes the interior temperature from the two boundary quantities,
\begin{align}
\label{eq:evaluate-solution}
u(\bx) = \underbrace{\int_{\Gamma} G(\bx, \by)\, q(\by) \;\dd A_\by}_{\text{single-layer potential}} - \underbrace{\int_{\Gamma} \poisson{\bx}{\by}\, u(\by) \;\dd A_\by}_{\text{double-layer potential}}, \quad \bx \in \Omega,
\end{align}
with $G(\bx, \by) = 1 / (4\pi \lVert \bx - \by \rVert)$ the free-space Green's function of the Laplacian and $\poisson{\bx}{\by} = \partial G / \partial \bn_\by$ its normal derivative at the source point, the double-layer kernel, sometimes also known as Poisson kernel.
Letting $\bx$ approach $\Gamma$ from inside, the double layer jumps by half its density and the representation becomes an equation on the boundary \cite[Ch. 3]{sauterBoundaryElementMethods2011},
for all $\bx \in \Gamma$,
\begin{align}
\label{eq:single-layer-potential}
\tfrac{1}{2}\, u(\bx) + \int_{\Gamma} \poisson{\bx}{\by}\, u(\by) \;\dd A_\by - \int_{\Gamma} G(\bx, \by)\, q(\by) \;\dd A_\by = 0 .
\end{align}
The boundary condition \cref{eq:robin-problem} relates $u$ and $q$ at every point, so \cref{eq:single-layer-potential} has one unknown per point: $q$ where the temperature is prescribed, $u$ where a flux or a Robin exchange is, with $q = (g - a u) / b$.
A constant temperature with zero flux satisfies it exactly, because the double-layer kernel integrates to $-\tfrac{1}{2}$ over a closed surface seen from a point on it and to $-1$ from a point inside; we use both identities below.
Once both boundary quantities are solved, we evaluate the temperature inside with \cref{eq:evaluate-solution}.

\input{figures/subtraction-idea/subtraction-idea-snippet}

\paragraph{Density subtraction}
On the boundary, both integrals of \cref{eq:single-layer-potential} exist although their kernels are singular at $\by = \bx$, since on a smooth surface they grow only as the inverse of the distance.
However, inside as $\bx$ approaches the surface the double-layer kernel of \cref{eq:evaluate-solution} peaks sharply, as the inverse square of the distance, around the boundary point $\by^\ast$ nearest to $\bx$, and any quadrature of that peak is inaccurate (\cref{fig:subtraction}).
Following \citeauthor{carvalhoAsymptoticApproximationsClose2020}~\shortcite[Eqs.~(2.8)--(2.10)]{carvalhoAsymptoticApproximationsClose2020}, we subtract the density at $\by^\ast$ before integrating.
The subtraction costs nothing because the double-layer potential of a constant density is known exactly: $-\poisson{\bx}{\by}\,\dd A_\by$ is the solid angle that the surface element $\dd A_\by$ subtends at $\bx$, divided by $4\pi$, and a closed surface subtends the full sphere from any point inside, so $\int_\Gamma \poisson{\bx}{\by}\,\dd A_\by = -1$.
For any constant $u_0$ and every $\bx \in \Omega$, \cref{eq:evaluate-solution} therefore equals
\begin{align}
\label{eq:density-subtraction}
u(\bx) = \int_{\Gamma} G(\bx, \by)\, q(\by) \;\dd A_\by - \int_{\Gamma} \poisson{\bx}{\by}\, \big(u(\by) - u_0\big) \;\dd A_\by + u_0,
\end{align}
and with $u_0 = u(\by^\ast)$ the subtracted density vanishes where the kernel peaks, so the remaining integrand is only weakly singular, like those on the boundary (\cref{fig:subtraction-idea}).

\section{From Gaussian primitives to collocation points}
\label{sec:discretization}

\paragraph{Background: Gaussian primitives}
\label{ssec:background-surfels}
Gaussian splatting \cite{kerbl3DGaussianSplatting2023a} represents a scene as a set of anisotropic 3D Gaussian primitives, each with a center $\bx_i$, a covariance $\Sigma_i = R_i S_i S_i^{\top} R_i^{\top}$ (rotation $R_i$, scales $S_i = \operatorname{diag}(s^1_i, s^2_i, s^3_i)$), an opacity $o_i$ and a view-dependent color $\bc_i$.
To render the primitives,
we first project the primitives to the image plane,
sort them by depth and
alpha-blend them front to back.
Each contributes to a pixel with its opacity there, attenuated by the opacity of the primitives in front of it.
The primitives are optimized against posed photographs by an image loss; we train on the visible-spectrum images only, so the geometry never depends on the thermal data.
The two need not even be captured together: the reconstruction can come from a separate visible-spectrum capture, as long as the thermal views are registered to it.

The vanilla Gaussian splat renderer does not have the notion of surfaces.
We use RaDe-GS \cite{zhangRaDeGSRasterizingDepth2024} that integrates the notion of surfaces as part of its rendering pipeline through depth maps.
Per primitive and pixel, it computes the depth of maximal density along the camera ray and the corresponding normal, and blends the normals and depths in the same fashion as it blends colors through alpha blending to obtain the normal and depth maps.
During training, the normal and depth maps are supervised with a depth--normal consistency term as in 2DGS \cite{huang2DGaussianSplatting2024} and the multi-view photometric and geometric consistency terms as in PGSR \cite{chenPGSRPlanarBased2024}.
Per training view $k$ we use its depth map $D_k(\bp)$ and normal map $\bN_k(\bp)$ at pixel $\bp$; the depth is the median depth, where the accumulated opacity reaches $\tfrac{1}{2}$.
RaDe-GS is the Gaussian primitive reconstruction algorithm that gives the best depth and normal maps as far as our experiments go.
But any reconstruction that can render consistent depth and normal maps, such as 2DGS \cite{huang2DGaussianSplatting2024} or PGSR \cite{chenPGSRPlanarBased2024}, can take its place.

\paragraph{Point cloud extraction}
\label{par:extraction}
The boundary integral equation \cref{eq:single-layer-potential} holds at every point of the boundary; to solve it, we enforce it at a finite set of collocation points, each standing for a patch of the surface with its position, normal and area.
A point cloud provides exactly that and, unlike a mesh, needs no extra effort for connectivity, such as ensuring watertightness or manifoldness.
For Gaussian primitives, it might seem natural to use the centers of the primitives as the collocation points, but they are not surface samples: a pixel alpha-blends several primitives along its ray, only the blend is constrained, and most centers settle below its median-depth surface, so boundary operators on them see a body systematically smaller than the real one.
We therefore extract the collocation cloud from the rendered depth and normal maps instead.

\input{figures/unproject/unproject-snippet}
In every training view we unproject each pixel $\bp$ of the object interior to the depth $D_k(\bp)$ and carry the rendered normal $\bN_k(\bp)$, oriented toward the camera (\cref{fig:unproject}), skipping a pixel whose normal meets the viewing direction at a grazing angle, with a cosine below $0.35$ (the \emph{frontality test}, \cref{fig:front-sweep}, right).
A pixel belongs to the object interior if the object covers it fully in the image, it lies at least four pixels inside the object mask and its rendered opacity exceeds $0.9$, so that no pixel on the silhouette mixes the object with the background; under one view the margin is one pixel, so that the single thermal image still reads most of the surface it sees.
The samples of all views are then merged on a regular grid of cubic cells one target spacing wide: the samples in a cell average into one point, those with opposite normals separately, so that the two faces of a wall thinner than a cell stay apart, and a cell needs at least two samples to make a point, so that a single stray pixel cannot make one on its own.

\paragraph{Filtering}
At a discontinuity of the depth map, such as the rim of a strut or a self-occluding contour, a pixel straddles two surfaces (\cref{fig:gate}).
Its median depth belongs to one of the surfaces, but its radiance and its rendered normal blend both, so the blended normal looks plausible; only the finite-difference normal of the depth map, which turns away from the camera across the step, gives the pixel away.
We therefore keep a pixel only if this depth-slope normal passes the frontality test as well, a second test we call the depth-slope filter; depth estimation knows these pixels as the source of flying points, which \citeauthor{bianModelingDepthAmbiguity2026}~\shortcite{bianModelingDepthAmbiguity2026} trace to the same straddling.
Pixels rejected by this filter would also yield an inaccurate temperature for any point sampled on them. We therefore apply the same depth-slope filter when selecting the pixels from which a point’s Dirichlet datum is fused, falling back to the unfiltered pixels when none pass.
The depth-slope filter lowers the error on every scene with full coverage and on nearly all under one view, the most on thin structures (\cref{fig:gate} and \cref{tab:ablation}).
The last filter, the membership filter, removes the points far from the boundary.
We do this because
we observe that a few isolated pixels of the depth maps land behind the surface, and unprojecting these pixels leads to points below the surface.
The double-layer sum over every other point is the winding number at a point without its own patch, $\tfrac{1}{2}$ on a clean closed cover, near $0$ for a floater and near $1$ for a buried point, and we keep the points between $0.05$ and $0.95$.
It removes about two percent of the points and lowers the full-coverage error by a sixth on average and up to half on the smooth solids (\cref{tab:ablation}).

\paragraph{Discretization of the BIEs}
The extracted cloud is a quadrature of the boundary: point $j$ stands for a patch of area $A_j$ around $\bx_j$ with outward normal $\bn_j$, on which the temperature $u_j$ and the flux $q_j$ are constant; following \citeauthor{barillFastWindingNumbers2018}~\shortcite{barillFastWindingNumbers2018}, we take $A_j$ as the area of the Voronoi cell of $\bx_j$ among its $20$ nearest neighbors, projected onto its tangent plane.\label{ssec:area-surfels}
Following the Nystr\"om method \cite[Sec.~12.2]{kressLinearIntegralEquations2014}, the points are the collocation points and the integrals of \cref{eq:evaluate-solution,eq:single-layer-potential} become weighted sums, the point-cloud quadrature the fast dipole sums of \citeauthor{chen3DReconstructionFast2024}~\shortcite{chen3DReconstructionFast2024} use for the winding number.
Following \citeauthor{bealeSimpleMethodComputing2016}~\shortcite{bealeSimpleMethodComputing2016}, we regularize the kernels: with $r = \lVert \bx - \by \rVert$ and $z = r/\varepsilon$,
\begin{align}
\label{eq:regularized-green}
G(\bx, \by) = \frac{\operatorname{erf}(z)}{4\pi r}, \qquad
\poisson{\bx}{\by} = -\,\frac{(\by - \bx)\cdot\bn_\by}{4\pi r^3}\,\Big(\operatorname{erf}(z) - \tfrac{2z}{\sqrt{\pi}}\, e^{-z^2}\Big),
\end{align}
\noindent\begin{minipage}[b]{\dimexpr\linewidth-108pt-6pt\relax}
which agree with the free-space kernels for $r \gg \varepsilon$ and stay finite at $r = 0$, where $G(\bx, \bx) = 1/(2\pi^{3/2}\varepsilon)$ and the double-layer kernel vanishes by symmetry (inset, with $\bx$ on the inward normal of $\by$).
\end{minipage}\hspace{6pt}\begin{minipage}[b]{108pt}\includegraphics{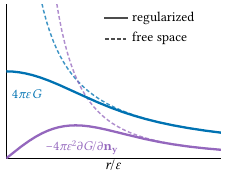}\end{minipage}\par
\input{figures/gate2/gate-snippet}
\noindent That finite value is the self-term of the single-layer sum: with $\varepsilon$ of the order of the spacing, the $j = i$ entry $G(\bx_i, \bx_i)\,q_i A_i$ is the quadrature of the singular kernel over the point's own patch.
The radius does more than make the diagonal finite: a flat-disk self-term with unmollified neighbors is $1.4$ times less accurate (median of eighteen scenes), since the mollification also integrates the nearest neighbors.
The double-layer diagonal follows from the identity above, $K_{ii} = -\tfrac{1}{2} - \sum_{j \neq i} K_{ij}$, so that every row integrates a constant exactly, the rigid-body trick of the boundary element literature \cite[Eq.~(3.11)]{brebbiaBoundaryElementTechniques1984}.
With $V_{ij} = G(\bx_i, \bx_j) A_j$ and $K_{ij} = \poisson{\bx_i}{\bx_j} A_j$ for $j \neq i$, \cref{eq:single-layer-potential} at collocation point $i$ becomes
\begin{align}
\label{eq:gs-single-layer-potenial}
\tfrac{1}{2}\, u_i + \sum_{j} K_{ij}\, u_j - \sum_{j} V_{ij}\, q_j = 0,
\end{align}
one linear equation per point once the boundary condition has eliminated $u_i$ or $q_i$.
The regularization error is $O(\varepsilon)$ near the surface \cite{bealeSimpleMethodComputing2016};
we measure it in \cref{ssec:eps-study}, and set $\varepsilon = h$, the mean spacing of the points.

\paragraph{Evaluating at the interior}
The interior field of \cref{eq:evaluate-solution} becomes a weighted sum over the solved boundary data, but not the naive one: as a point sum, the double-layer kernel fails wherever $\bx$ is closer to the surface than the spacing, since its singularity makes the nearest point's kernel evaluation blow up.
To alleviate the problem, we use the discrete form of density subtraction (\cref{eq:density-subtraction}).
With $u_0$ as the temperature of the collocation point $j^\ast$ nearest to $\bx$:
\begin{align}
\label{eq:gs-eval-solutions}
u(\bx) = \sum_j G(\bx, \bx_j)\, q_j A_j - \sum_j \poisson{\bx}{\bx_j}\,(u_j - u_{j^\ast})\, A_j + u_{j^\ast}.
\end{align}
As can be seen in \cref{fig:subtraction}, without the subtraction the error concentrates in a band one spacing wide along the surface.
However, comparing the density subtraction solution with the smooth reference,
there are also ``faint seam'' artifacts where $j^\ast$ changes.
One possible way to alleviate this artifact is
the higher-order harmonic density interpolation \cite{perezarancibiaHarmonicDensityInterpolation2019}.

\input{figures/subtraction/subtraction-snippet}

\section{Solver pipeline} \label{sec:solver}
The solver pipeline (\cref{fig:overview}) reconstructs the geometry, prescribes the boundary conditions and solves the BIE; the first step trains RaDe-GS and extracts the collocation cloud as in \cref{par:extraction}.

\paragraph{Prescribing the boundary conditions}
\label{par:bcs}%
The boundary conditions are the coefficients $a_i, b_i, g_i$ of \cref{eq:robin-problem} at each collocation point $i$.
A point is \emph{seen} by a thermal view if it projects inside the object mask, away from the silhouette, and the rendered depth agrees with it.
Every seen point carries a measured temperature $T_i$, read from the pixels it was extracted from, if they face the thermal camera: they must pass the frontality test of the extraction (\cref{par:extraction}).
The camera also sees surface it faces more obliquely, $26$ to $49\%$ of the seen points under one view on our scenes.
Such a point is still seen, since its surface is observed, and takes the read of its nearest seen neighbor whose pixels pass the test.
In most experiments the point is Dirichlet ($a_i = 1$, $b_i = 0$, $g_i = T_i$); on the real capture (\cref{ssec:real-capture}) the imaged surface exchanges heat with the room, so its points are Robin and the measurements are the data the unknown parameters are fitted to.
Each pixel value is converted to temperature before any fusion, since \cref{eq:laplace-problem} is linear in $T$ and the radiance is not.
Our renders emit a radiance proportional to $T^4$ \cite[Ch. 1]{modestRadiativeHeatTransfer2021}, so on the synthetic scenes we invert the fourth power over the temperature range of the render.
A single thermal view provides the data unless stated otherwise; with more views the read fuses every view that sees a point.
We take the mean temperature of the point's pixels in all these views, each pixel weighted by the cosine of its frontality test.

The points seen by no thermal image need an assumption of how the boundary exchanges heat with the ambient air, otherwise the problem would be ill-posed.
A Robin condition with user-given $a_i, b_i, g_i$ can model  insulating ($a_i = 0$, $b_i = 1$, $g_i = 0$) or convective surfaces, where the surface exchanges heat with the ambient air in proportion to their temperature difference \cite{narayananShapeHeatConduction2024}.
For convection, $a_i = 1$, $g_i$ is the ambient temperature and $b_i$, the Robin coefficient, is the ratio of the solid's thermal conductivity to the surface's heat transfer coefficient, taken from convection correlations \cite[Ch.~9]{incroperaFundamentalsHeatMass2007}, or engineering standards \cite{iso6946}, or fitted to measurements for real captures (\cref{ssec:real-capture}).

\paragraph{Solving}
\label{par:solving}
\Cref{eq:gs-single-layer-potenial} is one dense linear system, $M\psi = g$, in the unknown half $\psi$ of the boundary data.
Using the technique proposed by Chen et al.~\shortcite{chenLightningfastBoundaryElement2025}, we never need to form the dense system.
Every product with the matrix regenerates the mollified kernels in one fused GPU kernel,
and a preconditioned Krylov method solves the system.
Systems with only temperature data become symmetric positive definite after a diagonal rescaling and are solved by conjugate gradients \cite{hestenesMethodsConjugateGradients1952} with a sparse inverse Cholesky preconditioner \cite{schaferSparseCholeskyFactorization2021, chenLightningfastMethodFundamental2024}, the others by GMRES \cite{saadGMRESGeneralizedMinimal1986} with the sparse inverse LU preconditioner of \citeauthor{chenLightningfastBoundaryElement2025}~\shortcite{chenLightningfastBoundaryElement2025}.
The memory cost is linear in $N$, so the collocation count has almost no cap.
Each point corresponds to one row of the linear system, whose unknown is the flux where the temperature is known and the temperature elsewhere (inset).
Rows with a temperature unknown (blue) also carry it outside the sums, as the $\tfrac{1}{2}u_i$ of \cref{eq:gs-single-layer-potenial}, and keep the unobserved surface well behaved at any $\varepsilon$.
\newsavebox{\rowsinsetbox}\sbox{\rowsinsetbox}{\input{figures/rows/rows-inset}}%
\par\noindent\begin{minipage}[b]{\dimexpr\linewidth-\wd\rowsinsetbox-6pt\relax}
Rows with a flux unknown (purple) carry it only inside the single-layer sum, where a flux that alternates from point to point nearly cancels, so these rows grow ill-conditioned as the spacing shrinks.
They need damping, a constant $\gamma$ added to their diagonal, the identity term they lack.
\end{minipage}\hspace{6pt}\begin{minipage}[b]{\wd\rowsinsetbox}\usebox{\rowsinsetbox}\end{minipage}\par
\noindent As \citet{bealeSimpleMethodComputing2016} derive for this mollifier, at a point on the surface, the mollified single layer falls short of the exact one by $\varepsilon/(2\sqrt{\pi})$ times the local density, and the damping makes up half of that deficit, $\gamma = \varepsilon/(4\sqrt{\pi})$; their correction adds all of it back, which on a flux row is the same diagonal term, twice as large.
Tying $\gamma$ to $\varepsilon$ is important here.
A fixed $\gamma$ grows relative to the self-term, $V_{ii} = A_i/(2\pi^{3/2}\varepsilon)$, as the spacing shrinks. Under one view, the flux is singular along the border between the observed and the unobserved surface, and there a fixed $\gamma$ biases the temperature more at every refinement.
The solve returns the temperature on the unseen part of the surface directly, and it needs no potential evaluation on the boundary.

\paragraph{Differentiability}
\label{ssec:differentiability}
Everything above is differentiable: the matrix $M$ and right-hand side $g$ are smooth in the point positions, normals, areas and coefficients, the solve is linear, and \cref{eq:gs-eval-solutions} is a weighted sum.
Differentiating $\psi = M^{-1} g$ with respect to a parameter $\theta$ by the chain rule would differentiate a factorization.
We avoid it by using the adjoint method.
Considering a loss function $\mathcal{L}(\psi)$, such as the squared error between the solved surface temperature and the measured surface temperature (\cref{ssec:pipe,ssec:facade,ssec:crack}),
we have
\begin{align}
\frac{\dd \mathcal{L}}{\dd \theta} = \lambda^{\top}\Big(\frac{\partial g}{\partial \theta} - \frac{\partial M}{\partial \theta}\,\psi\Big), \qquad M^{\top}\lambda = \frac{\partial \mathcal{L}}{\partial \psi}.
\end{align}
This means, for the gradient, we need
one additional solve with the transposed matrix, regardless of the number of parameters.
We wrap the solve as one differentiable PyTorch \cite{paszkePyTorchImperativeStyle2019} operation whose backward pass is this adjoint solve, by the same Krylov method with the transposed products and the transposed preconditioner; the remaining term $\lambda^{\top}(\partial g/\partial\theta - \partial M/\partial\theta\,\psi)$ is summed over all point pairs by two fused GPU kernels with the closed-form derivatives of the mollified kernels, $\partial G/\partial r = -\big(\operatorname{erf}(z) - \tfrac{2z}{\sqrt{\pi}} e^{-z^2}\big)/(4\pi r^2)$, so neither the matrix nor an autograd graph of it is ever stored.
Discrete decisions (which pixels pass the filters, which points are seen) are frozen within an optimization step.
\Cref{fig:gradcheck} checks it against finite differences on the surface temperature of a convectively cooled body with respect to the translation of a hot sphere hidden inside: the two derivative maps agree to a relative difference of $5.0 \times 10^{-5}$.

\input{figures/gradcheck/gradcheck-snippet}

We use the gradient for two kinds of unknowns: the boundary coefficients of a region (\cref{ssec:real-capture,ssec:facade}) and the shape of a boundary no camera sees (\cref{ssec:pipe,ssec:crack}).
One caution applies to shape: the regularized kernel is resolved only up to the spacing, so per-point moves receive a gradient that is rough on the scale of one cell, while smooth parametrizations (a rigid pose, a height field under a blur, a displacement field smoothed by a Gaussian, a few coefficients per region) receive a clean one; every shape result in \cref{sec:results} uses one of those.

%% file: figures/overview2/overview-snippet.tex
\begin{figure*}
\centering
\begin{tikzpicture}[every node/.style={inner sep=0, outer sep=0}, lab/.style={font=\footnotesize, anchor=north, align=center}, stp/.style={font=\scriptsize, align=center}, arr/.style={-{Stealth[length=4pt]}, line width=0.6pt}, cam/.style={fill=black!65}]
  \def\H{0.165\textwidth}\def\g{0.034\textwidth}\def\gc{0.075\textwidth}
  \def\camera#1{\fill[cam, rounded corners=0.6pt] ([xshift=-0.009\textwidth, yshift=-0.0065\textwidth]#1) rectangle ([xshift=0.011\textwidth, yshift=0.0065\textwidth]#1);
    \fill[cam] ([xshift=-0.009\textwidth, yshift=0.003\textwidth]#1) -- ([xshift=-0.017\textwidth, yshift=0.006\textwidth]#1) -- ([xshift=-0.017\textwidth, yshift=-0.006\textwidth]#1) -- ([xshift=-0.009\textwidth, yshift=-0.003\textwidth]#1) -- cycle;}
  \node[anchor=south west, draw=black!40, line width=0.3pt] (R3) at (0.026\textwidth, 0.090\textwidth) {\includegraphics[width=0.05\textwidth]{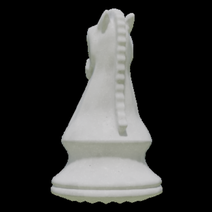}};
  \node[anchor=south west, draw=black!40, line width=0.3pt] (R2) at (0.013\textwidth, 0.077\textwidth) {\includegraphics[width=0.05\textwidth]{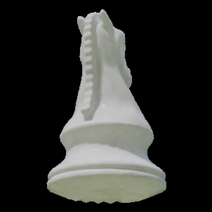}};
  \node[anchor=south west, draw=black!40, line width=0.3pt] (R1) at (0, 0.064\textwidth) {\includegraphics[width=0.05\textwidth]{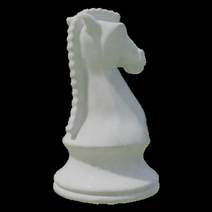}};
  \node[anchor=south west] (Sp) at ([xshift=0.052\textwidth]R3.east |- 0,0) {\includegraphics[height=\H]{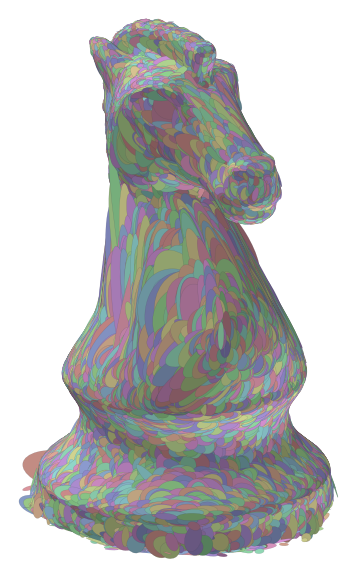}};
  \node[anchor=south west] (C) at ([xshift=\gc]Sp.south east) {\includegraphics[height=\H]{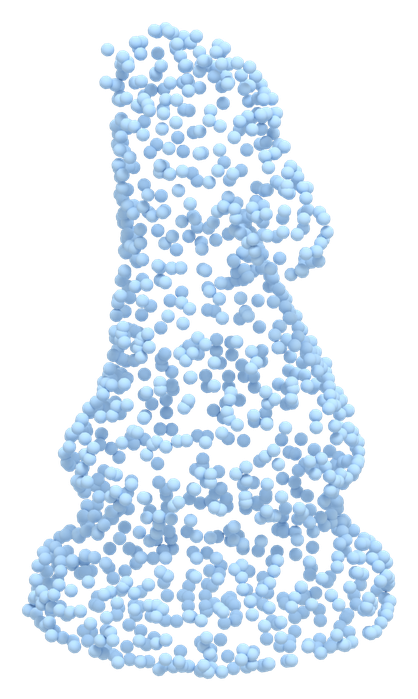}};
  \node[anchor=south west] (B) at ([xshift=\g]C.south east) {\includegraphics[height=\H]{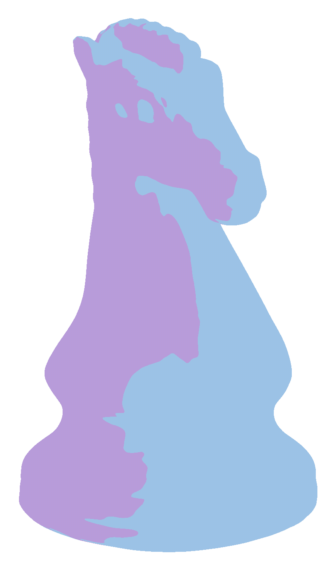}};
  \node[anchor=south west] (F) at ([xshift=\gc]B.south east) {\includegraphics[height=\H]{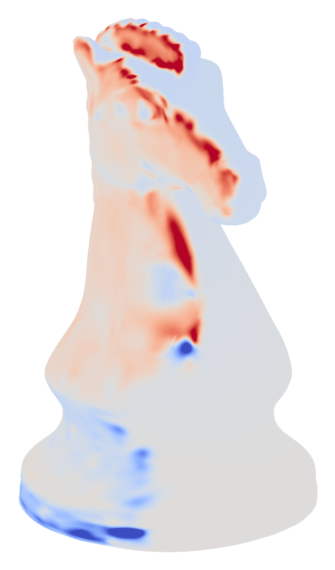}};
  \node[anchor=south west] (S) at ([xshift=\g]F.south east) {\includegraphics[height=\H]{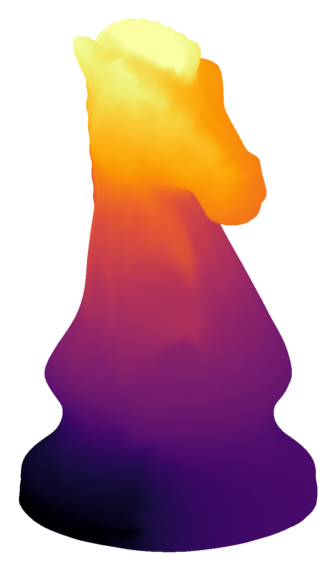}};
  \node[anchor=south west] (I) at ([xshift=0.006\textwidth]S.south east) {\includegraphics[height=\H]{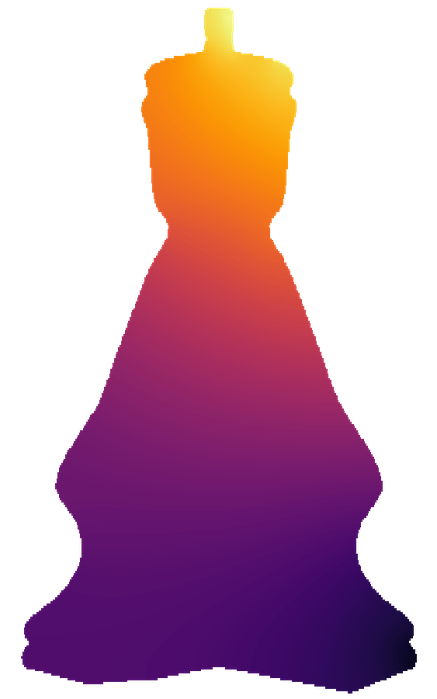}};
  \coordinate (cr) at ([xshift=0.030\textwidth, yshift=0.020\textwidth]Sp.east); \camera{cr}
  \node[anchor=south] (Dm) at ([xshift=-0.009\textwidth, yshift=0.016\textwidth]cr) {\includegraphics[height=0.064\textwidth]{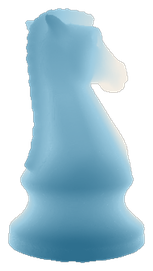}};
  \node[anchor=south west] (Nm) at ([xshift=-0.019\textwidth]Dm.south east) {\includegraphics[height=0.064\textwidth]{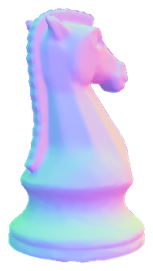}};
  \draw[black!55, line width=0.4pt, densely dashed] ([yshift=0.0065\textwidth]cr) -- ([yshift=0.016\textwidth]cr);
  \node[font=\scriptsize, anchor=south] at ([yshift=1pt]$(Dm.north)!0.5!(Nm.north)$) {depth, normals};
  \coordinate (ct) at ([xshift=0.042\textwidth, yshift=0.020\textwidth]B.east); \camera{ct}
  \node[anchor=south, draw=black!40, line width=0.3pt] (T) at ([xshift=0.003\textwidth, yshift=0.016\textwidth]ct) {\includegraphics[width=0.046\textwidth]{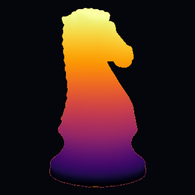}};
  \draw[black!55, line width=0.4pt, densely dashed] ([yshift=0.0065\textwidth]ct) -- (T.south);
  \node[font=\scriptsize, anchor=south] at ([yshift=1pt]T.north) {thermal image};
  \coordinate (tob) at ($(B.south west)!0.41!(B.south east) + (B.south west)!0.56!(B.north west) - (B.south west)$);
  \coordinate (tun) at ($(B.south west)!0.627!(B.south east) + (B.south west)!0.309!(B.north west) - (B.south west)$);
  \draw[arr, black!75, line width=0.5pt] ([yshift=0.006\textwidth]T.west) to[out=200, in=40] (tob);
  \node[stp, anchor=north east] at ([xshift=-0.003\textwidth]T.north west) {Dirichlet};
  \node[stp, anchor=west] (rb) at ([xshift=0.010\textwidth, yshift=0.046\textwidth]B.south east) {Robin};
  \draw[arr, black!75, line width=0.5pt] (rb.west) to[out=180, in=0] (tun);
  \node[font=\scriptsize, anchor=south west, align=left] at ([xshift=0.004\textwidth, yshift=0.004\textwidth]B.south east) {\textcolor[HTML]{B89CDA}{\rule{4pt}{4pt}}\,observed\\[1pt]\textcolor[HTML]{9CC3E6}{\rule{4pt}{4pt}}\,unobserved};
  \draw[arr] ([xshift=0.003\textwidth, yshift=-0.012\textwidth]R3.east) to[out=0, in=160] ([xshift=-0.002\textwidth]Sp.west);
  \draw[arr] ([xshift=0.002\textwidth]Sp.east) -- ([xshift=-0.002\textwidth]C.west);
  \draw[arr] ([xshift=0.002\textwidth]C.east) -- ([xshift=-0.002\textwidth]B.west);
  \draw[arr] ([xshift=0.002\textwidth]B.east) -- ([xshift=-0.002\textwidth]F.west);
  \draw[arr] ([xshift=0.002\textwidth]F.east) -- ([xshift=-0.002\textwidth]S.west);
  \node[lab] at ([yshift=-0.004\textwidth]R1.south -| R2.south) {visible-spectrum\\images};
  \node[lab] at ([yshift=-0.004\textwidth]Sp.south) {Gaussian splats};
  \node[lab] at ([yshift=-0.004\textwidth]C.south) {collocation cloud};
  \node[lab] at ([yshift=-0.004\textwidth]B.south) {boundary conditions};
  \node[lab] at ([yshift=-0.004\textwidth]F.south) {BIE solve};
  \node[lab] at ([yshift=-0.004\textwidth]S.south east) {surface and interior temperature};
\end{tikzpicture}
\caption{Overview. We first reconstruct Gaussian splats from the visible-spectrum images and render a depth and a normal map per view, whose pixels are unprojected and merged into the collocation cloud (\cref{par:extraction}). The thermal image fixes the temperature on the observed part of the surface (Dirichlet), and convection holds on the unobserved part (Robin). The direct boundary integral equation on the cloud returns the flux $q_j$ at every point, from which the temperature follows on the unobserved surface and everywhere inside.}
\label{fig:overview}
\end{figure*}

%% file: figures/notation/notation-inset.tex
\begin{tikzpicture}[x=0.955cm, y=0.955cm, font=\scriptsize, every node/.style={inner sep=1pt}]
  \node[anchor=south west, inner sep=0] at (-1.299,-0.924) {\includegraphics[width=2.6093cm, height=1.8106cm]{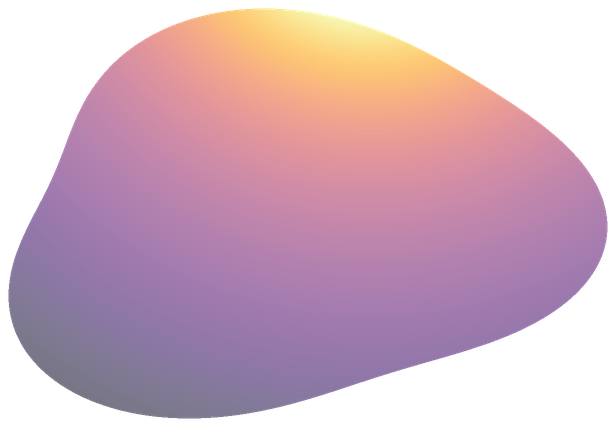}};
  \draw[line width=1.7pt, draw={rgb,255:red,184;green,156;blue,218}] (1.397,0.143) -- (1.387,0.163) -- (1.377,0.182) -- (1.366,0.201) -- (1.354,0.220) -- (1.342,0.239) -- (1.328,0.257) -- (1.315,0.274) -- (1.300,0.292) -- (1.285,0.309) -- (1.270,0.325) -- (1.254,0.342) -- (1.238,0.358) -- (1.221,0.373) -- (1.204,0.389) -- (1.187,0.404) -- (1.170,0.418) -- (1.152,0.433) -- (1.134,0.447) -- (1.116,0.461) -- (1.098,0.474) -- (1.080,0.488) -- (1.062,0.501) -- (1.043,0.514) -- (1.025,0.526) -- (1.006,0.539) -- (0.988,0.551) -- (0.969,0.564) -- (0.950,0.576) -- (0.932,0.588) -- (0.913,0.600) -- (0.894,0.611) -- (0.875,0.623) -- (0.856,0.635) -- (0.837,0.647) -- (0.818,0.658) -- (0.799,0.670) -- (0.780,0.681) -- (0.761,0.693) -- (0.741,0.704) -- (0.721,0.716) -- (0.701,0.727) -- (0.681,0.738) -- (0.660,0.750) -- (0.640,0.761) -- (0.618,0.772) -- (0.597,0.783) -- (0.575,0.795) -- (0.553,0.806) -- (0.531,0.816) -- (0.508,0.827) -- (0.485,0.838) -- (0.461,0.848) -- (0.437,0.858) -- (0.412,0.868) -- (0.387,0.878) -- (0.362,0.887) -- (0.336,0.896) -- (0.310,0.905) -- (0.283,0.913) -- (0.256,0.921) -- (0.228,0.929) -- (0.200,0.936) -- (0.172,0.942) -- (0.143,0.948) -- (0.114,0.954) -- (0.085,0.959) -- (0.055,0.963) -- (0.025,0.967) -- (-0.005,0.970) -- (-0.035,0.972) -- (-0.065,0.974) -- (-0.096,0.975) -- (-0.126,0.975) -- (-0.157,0.975) -- (-0.188,0.974) -- (-0.218,0.972) -- (-0.249,0.969) -- (-0.279,0.966) -- (-0.309,0.961) -- (-0.339,0.957) -- (-0.368,0.951) -- (-0.398,0.944) -- (-0.426,0.937) -- (-0.455,0.929) -- (-0.483,0.921) -- (-0.510,0.911) -- (-0.537,0.901) -- (-0.564,0.891) -- (-0.589,0.879) -- (-0.614,0.867) -- (-0.639,0.855) -- (-0.663,0.842) -- (-0.685,0.828) -- (-0.708,0.814) -- (-0.729,0.799) -- (-0.750,0.784) -- (-0.770,0.769) -- (-0.789,0.753) -- (-0.807,0.736) -- (-0.825,0.720) -- (-0.841,0.703) -- (-0.857,0.686) -- (-0.872,0.669) -- (-0.887,0.651) -- (-0.900,0.634) -- (-0.913,0.616) -- (-0.925,0.598) -- (-0.937,0.581) -- (-0.948,0.563) -- (-0.958,0.545) -- (-0.967,0.528) -- (-0.976,0.510) -- (-0.985,0.493) -- (-0.993,0.476) -- (-1.001,0.459) -- (-1.008,0.442) -- (-1.015,0.425) -- (-1.021,0.409) -- (-1.028,0.393);
  \draw[line width=1.7pt, draw={rgb,255:red,156;green,195;blue,230}] (-1.028,0.393) -- (-1.034,0.376) -- (-1.040,0.360) -- (-1.046,0.345) -- (-1.052,0.329) -- (-1.058,0.313) -- (-1.064,0.298) -- (-1.069,0.283) -- (-1.075,0.268) -- (-1.081,0.253) -- (-1.087,0.239) -- (-1.093,0.224) -- (-1.100,0.210) -- (-1.106,0.195) -- (-1.113,0.181) -- (-1.120,0.166) -- (-1.127,0.152) -- (-1.134,0.137) -- (-1.141,0.122) -- (-1.149,0.107) -- (-1.157,0.092) -- (-1.165,0.076) -- (-1.173,0.061) -- (-1.181,0.045) -- (-1.190,0.029) -- (-1.198,0.012) -- (-1.207,-0.004) -- (-1.215,-0.022) -- (-1.223,-0.039) -- (-1.232,-0.057) -- (-1.240,-0.075) -- (-1.247,-0.094) -- (-1.255,-0.113) -- (-1.262,-0.132) -- (-1.269,-0.151) -- (-1.275,-0.172) -- (-1.281,-0.192) -- (-1.287,-0.213) -- (-1.291,-0.234) -- (-1.295,-0.255) -- (-1.298,-0.277) -- (-1.301,-0.299) -- (-1.302,-0.321) -- (-1.302,-0.344) -- (-1.302,-0.366) -- (-1.300,-0.389) -- (-1.297,-0.412) -- (-1.293,-0.435) -- (-1.288,-0.458) -- (-1.282,-0.480) -- (-1.275,-0.503) -- (-1.266,-0.526) -- (-1.256,-0.548) -- (-1.244,-0.570) -- (-1.232,-0.592) -- (-1.218,-0.613) -- (-1.203,-0.634) -- (-1.186,-0.654) -- (-1.169,-0.674) -- (-1.150,-0.694) -- (-1.130,-0.712) -- (-1.109,-0.730) -- (-1.087,-0.748) -- (-1.064,-0.765) -- (-1.039,-0.781) -- (-1.014,-0.796) -- (-0.988,-0.810) -- (-0.961,-0.824) -- (-0.933,-0.837) -- (-0.905,-0.849) -- (-0.876,-0.860) -- (-0.846,-0.870) -- (-0.815,-0.880) -- (-0.784,-0.888) -- (-0.753,-0.896) -- (-0.721,-0.903) -- (-0.689,-0.909) -- (-0.657,-0.914) -- (-0.624,-0.919) -- (-0.591,-0.922) -- (-0.558,-0.925) -- (-0.526,-0.927) -- (-0.493,-0.928) -- (-0.460,-0.928) -- (-0.427,-0.928) -- (-0.395,-0.927) -- (-0.362,-0.925) -- (-0.330,-0.923) -- (-0.298,-0.920) -- (-0.267,-0.916) -- (-0.236,-0.912) -- (-0.205,-0.907) -- (-0.175,-0.902) -- (-0.145,-0.897) -- (-0.116,-0.891) -- (-0.087,-0.885) -- (-0.059,-0.878) -- (-0.031,-0.872) -- (-0.004,-0.865) -- (0.023,-0.858) -- (0.049,-0.850) -- (0.075,-0.843) -- (0.101,-0.835) -- (0.126,-0.828) -- (0.150,-0.820) -- (0.174,-0.812) -- (0.198,-0.805) -- (0.221,-0.797) -- (0.244,-0.789) -- (0.267,-0.782) -- (0.289,-0.775) -- (0.312,-0.767) -- (0.334,-0.760) -- (0.355,-0.753) -- (0.377,-0.746) -- (0.399,-0.739) -- (0.420,-0.732) -- (0.442,-0.725) -- (0.463,-0.719) -- (0.485,-0.712) -- (0.507,-0.706) -- (0.528,-0.699) -- (0.550,-0.693) -- (0.572,-0.687) -- (0.594,-0.680) -- (0.616,-0.674) -- (0.639,-0.668) -- (0.662,-0.661) -- (0.685,-0.654) -- (0.708,-0.648) -- (0.731,-0.641) -- (0.754,-0.634) -- (0.778,-0.626) -- (0.802,-0.619) -- (0.826,-0.611) -- (0.850,-0.603) -- (0.874,-0.595) -- (0.898,-0.586) -- (0.922,-0.577) -- (0.946,-0.568) -- (0.970,-0.558) -- (0.994,-0.548) -- (1.018,-0.537) -- (1.042,-0.526) -- (1.065,-0.514) -- (1.088,-0.502) -- (1.111,-0.490) -- (1.134,-0.477) -- (1.156,-0.463) -- (1.177,-0.449) -- (1.198,-0.434) -- (1.219,-0.419) -- (1.239,-0.404) -- (1.258,-0.387) -- (1.276,-0.371) -- (1.294,-0.354) -- (1.311,-0.336) -- (1.327,-0.318) -- (1.342,-0.299) -- (1.356,-0.280) -- (1.369,-0.261) -- (1.381,-0.241) -- (1.392,-0.221) -- (1.402,-0.200) -- (1.410,-0.179) -- (1.418,-0.158) -- (1.424,-0.136) -- (1.429,-0.115) -- (1.433,-0.093) -- (1.436,-0.071) -- (1.437,-0.049) -- (1.437,-0.027) -- (1.436,-0.005) -- (1.434,0.016) -- (1.430,0.038) -- (1.426,0.060) -- (1.420,0.081) -- (1.413,0.102) -- (1.406,0.123) -- (1.397,0.143);
  \draw[line width=0.75pt] (1.395,0.000) -- (1.392,0.024) -- (1.387,0.048) -- (1.380,0.072) -- (1.373,0.096) -- (1.363,0.119) -- (1.352,0.142) -- (1.340,0.165) -- (1.327,0.187) -- (1.313,0.209) -- (1.297,0.231) -- (1.281,0.252) -- (1.263,0.273) -- (1.245,0.293) -- (1.226,0.313) -- (1.206,0.332) -- (1.186,0.351) -- (1.165,0.369) -- (1.144,0.387) -- (1.122,0.405) -- (1.100,0.422) -- (1.078,0.439) -- (1.055,0.456) -- (1.032,0.472) -- (1.009,0.488) -- (0.986,0.504) -- (0.963,0.519) -- (0.940,0.534) -- (0.917,0.549) -- (0.893,0.564) -- (0.870,0.579) -- (0.846,0.594) -- (0.822,0.608) -- (0.798,0.623) -- (0.774,0.637) -- (0.750,0.652) -- (0.726,0.666) -- (0.701,0.681) -- (0.676,0.695) -- (0.650,0.709) -- (0.625,0.723) -- (0.598,0.737) -- (0.572,0.751) -- (0.545,0.765) -- (0.517,0.778) -- (0.489,0.791) -- (0.460,0.804) -- (0.430,0.817) -- (0.400,0.829) -- (0.369,0.841) -- (0.338,0.853) -- (0.306,0.864) -- (0.273,0.874) -- (0.240,0.884) -- (0.206,0.893) -- (0.171,0.901) -- (0.136,0.909) -- (0.100,0.915) -- (0.064,0.921) -- (0.028,0.926) -- (-0.009,0.930) -- (-0.046,0.932) -- (-0.084,0.934) -- (-0.121,0.935) -- (-0.158,0.934) -- (-0.196,0.933) -- (-0.233,0.930) -- (-0.270,0.926) -- (-0.307,0.921) -- (-0.343,0.915) -- (-0.379,0.907) -- (-0.414,0.899) -- (-0.448,0.889) -- (-0.482,0.878) -- (-0.515,0.867) -- (-0.547,0.854) -- (-0.578,0.840) -- (-0.608,0.825) -- (-0.637,0.810) -- (-0.665,0.793) -- (-0.691,0.776) -- (-0.717,0.758) -- (-0.741,0.740) -- (-0.764,0.721) -- (-0.785,0.701) -- (-0.806,0.682) -- (-0.825,0.661) -- (-0.843,0.641) -- (-0.860,0.620) -- (-0.876,0.599) -- (-0.891,0.578) -- (-0.904,0.556) -- (-0.917,0.535) -- (-0.929,0.514) -- (-0.940,0.493) -- (-0.950,0.472) -- (-0.960,0.451) -- (-0.969,0.431) -- (-0.977,0.410) -- (-0.985,0.390) -- (-0.993,0.370) -- (-1.001,0.350) -- (-1.008,0.330) -- (-1.015,0.311) -- (-1.023,0.292) -- (-1.030,0.273) -- (-1.037,0.254) -- (-1.045,0.236) -- (-1.052,0.217) -- (-1.060,0.199) -- (-1.068,0.180) -- (-1.077,0.162) -- (-1.086,0.143) -- (-1.095,0.125) -- (-1.104,0.106) -- (-1.114,0.088) -- (-1.123,0.069) -- (-1.133,0.049) -- (-1.144,0.030) -- (-1.154,0.010) -- (-1.164,-0.010) -- (-1.174,-0.031) -- (-1.184,-0.052) -- (-1.194,-0.073) -- (-1.204,-0.095) -- (-1.213,-0.117) -- (-1.222,-0.139) -- (-1.230,-0.163) -- (-1.238,-0.186) -- (-1.244,-0.210) -- (-1.250,-0.234) -- (-1.255,-0.259) -- (-1.258,-0.284) -- (-1.261,-0.309) -- (-1.262,-0.335) -- (-1.262,-0.360) -- (-1.260,-0.386) -- (-1.256,-0.412) -- (-1.252,-0.438) -- (-1.245,-0.464) -- (-1.237,-0.489) -- (-1.227,-0.515) -- (-1.215,-0.540) -- (-1.201,-0.565) -- (-1.185,-0.589) -- (-1.168,-0.613) -- (-1.149,-0.636) -- (-1.128,-0.658) -- (-1.105,-0.680) -- (-1.081,-0.701) -- (-1.055,-0.721) -- (-1.027,-0.740) -- (-0.998,-0.758) -- (-0.968,-0.775) -- (-0.936,-0.791) -- (-0.902,-0.806) -- (-0.868,-0.820) -- (-0.833,-0.832) -- (-0.797,-0.843) -- (-0.760,-0.853) -- (-0.722,-0.862) -- (-0.683,-0.869) -- (-0.645,-0.875) -- (-0.606,-0.880) -- (-0.566,-0.884) -- (-0.527,-0.886) -- (-0.487,-0.887) -- (-0.448,-0.888) -- (-0.408,-0.887) -- (-0.369,-0.885) -- (-0.331,-0.882) -- (-0.293,-0.878) -- (-0.255,-0.874) -- (-0.218,-0.868) -- (-0.181,-0.862) -- (-0.145,-0.856) -- (-0.110,-0.848) -- (-0.075,-0.841) -- (-0.041,-0.833) -- (-0.008,-0.824) -- (0.024,-0.815) -- (0.056,-0.806) -- (0.087,-0.797) -- (0.118,-0.788) -- (0.148,-0.778) -- (0.177,-0.769) -- (0.206,-0.759) -- (0.234,-0.750) -- (0.262,-0.741) -- (0.290,-0.732) -- (0.317,-0.723) -- (0.344,-0.714) -- (0.371,-0.705) -- (0.398,-0.697) -- (0.425,-0.688) -- (0.452,-0.680) -- (0.479,-0.672) -- (0.506,-0.664) -- (0.533,-0.656) -- (0.560,-0.648) -- (0.587,-0.640) -- (0.615,-0.632) -- (0.643,-0.624) -- (0.671,-0.616) -- (0.699,-0.608) -- (0.728,-0.599) -- (0.757,-0.591) -- (0.786,-0.581) -- (0.815,-0.572) -- (0.844,-0.562) -- (0.874,-0.552) -- (0.903,-0.541) -- (0.932,-0.530) -- (0.961,-0.518) -- (0.990,-0.505) -- (1.018,-0.492) -- (1.047,-0.478) -- (1.074,-0.464) -- (1.101,-0.449) -- (1.128,-0.433) -- (1.153,-0.416) -- (1.178,-0.399) -- (1.202,-0.381) -- (1.225,-0.362) -- (1.247,-0.343) -- (1.267,-0.323) -- (1.286,-0.303) -- (1.304,-0.282) -- (1.321,-0.260) -- (1.336,-0.238) -- (1.349,-0.215) -- (1.361,-0.192) -- (1.371,-0.169) -- (1.379,-0.145) -- (1.386,-0.121) -- (1.391,-0.097) -- (1.395,-0.073) -- (1.397,-0.049) -- (1.397,-0.024) -- (1.395,-0.000) -- cycle;
  \draw[densely dashed, line width=0.45pt] (-0.120,-0.220) -- (0.906,-0.540);
  \draw[-{Stealth[length=3pt]}, line width=0.55pt] (0.906,-0.540) -- (1.067,-0.962) node[anchor=east, xshift=-1pt, yshift=1pt] {$\bn_\by$};
  \filldraw[fill=white, line width=0.45pt] (0.906,-0.540) circle (1.3pt); \node[anchor=south east] at (0.862,-0.481) {$\by$};
  \fill (-0.120,-0.220) circle (1.2pt); \node[anchor=north east] at (-0.138,-0.238) {$\bx$};
  \node at (-0.42, 0.30) {$\Omega$};
  \node[anchor=east] (G) at (-1.176,0.385) {$\Gamma$};
  \draw[-{Stealth[length=2.6pt]}, black!50, line width=0.35pt] (G.south) to[bend right=40] (-1.231,0.037);
  \node[anchor=south west] (D) at (-0.075,1.045) {$u = g$};
  \draw[-{Stealth[length=2.6pt]}, black!50, line width=0.35pt] (D.west) to[bend right=40] (-0.602,0.918);
  \node[anchor=north] (R) at (-0.300,-0.998) {$a\,u + b\,q = g$};
  \draw[-{Stealth[length=2.6pt]}, black!50, line width=0.35pt] ([xshift=9pt]R.north) to[bend right=30] (0.128,-0.869);
  \node[anchor=south west, inner sep=0] at (1.599,-0.652) {\includegraphics[width=3pt, height=1.0790cm]{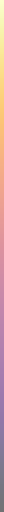}};
  \draw[line width=0.3pt] (1.599,-0.652) rectangle (1.710,0.478);
  \node[anchor=west, inner sep=0.6pt] at (1.739,0.478) {$1$};
  \node[anchor=west, inner sep=0.6pt] at (1.739,-0.652) {$0$};
  \node[anchor=south, inner sep=0.6pt] at (1.654,0.522) {$u$};
\end{tikzpicture}%

%% file: figures/subtraction-idea/subtraction-idea-snippet.tex
\begin{figure}[h]
\centering
\input{figures/subtraction-idea/sp-anchors}%
\definecolor{sppurple}{HTML}{603896}\definecolor{spblue}{HTML}{1E5A96}%
\newcommand{\spK}{\textcolor{sppurple}{\partial G/\partial \bn_\by}}%
\begin{tikzpicture}[every node/.style={inner sep=0, outer sep=0}, eq/.style={font=\scriptsize}, pt/.style={font=\scriptsize}]
  \def\pw{0.3\columnwidth}\def\gap{0.05\columnwidth}
  \node[anchor=south west] (A) at (0, 0) {\includegraphics[width=\pw]{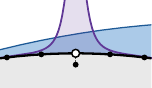}};
  \node[anchor=south west] (B) at ([xshift=\gap]A.south east) {\includegraphics[width=\pw]{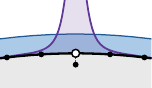}};
  \node[anchor=south west] (C) at ([xshift=\gap]B.south east) {\includegraphics[width=\pw]{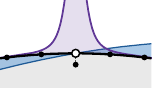}};
  \node at ($(A.east)!0.5!(B.west)$) {$=$};
  \node at ($(B.east)!0.5!(C.west)$) {$+$};
  \coordinate (bl) at ([yshift=-8.5pt]A.south);
  \node[eq, anchor=base, opacity=0] (F1) at (bl) {$-\!\int_\Gamma$$\,\spK$$\,\textcolor{spblue}{u}\,\dd A_\by$};
  \node[eq, anchor=base west] (f1a) at (F1.base west) {$-\!\int_\Gamma$};
  \node[eq, anchor=base west] (f1k) at (f1a.base east) {$\,\spK$};
  \node[eq, anchor=base west] at (f1k.base east) {$\,\textcolor{spblue}{u}\,\dd A_\by$};
  \node[pt, anchor=north, text=sppurple] at ([xshift=1pt, yshift=-1.5pt]f1k.south) {$\to \infty$ as $\bx \to \bx_{j^\ast}$};
  \node[eq, anchor=base] at (B.south |- bl) {$-\!\int_\Gamma \spK\,\textcolor{spblue}{u_0}\,\dd A_\by$};
  \node[eq, anchor=base east, opacity=0] (F3) at (C.south east |- bl) {$-\!\int_\Gamma \spK\,($$\textcolor{spblue}{u - u_0}$$)\,\dd A_\by$};
  \node[eq, anchor=base west] (f3a) at (F3.base west) {$-\!\int_\Gamma \spK\,($};
  \node[eq, anchor=base west] (f3u) at (f3a.base east) {$\textcolor{spblue}{u - u_0}$};
  \node[eq, anchor=base west] at (f3u.base east) {$)\,\dd A_\by$};
  \node[font=\scriptsize, anchor=south west] at ([xshift=2pt, yshift=1.5pt]B.south west) {$\textcolor{spblue}{u_0} = u(\bx_{j^\ast})$};
  \begin{scope}[shift={(A.south west)}, x={($(A.south east)-(A.south west)$)}, y={($(A.north west)-(A.south west)$)}]
    \node[pt, anchor=north] at ([yshift=-2.2pt]\spXx, \spXy) {$\bx$};
    \node[pt, anchor=north west] at ([xshift=2.6pt, yshift=-1.4pt]\spSx, \spSy) {$\bx_{j^\ast}$};
    \node[pt, anchor=west, text=sppurple] at ([xshift=4pt, yshift=-3pt]\spKx, \spKy) {kernel};
  \end{scope}
  \begin{scope}[shift={(C.south west)}, x={($(C.south east)-(C.south west)$)}, y={($(C.north west)-(C.south west)$)}]
    \coordinate (xs) at (\spSx, \spSy);
    \node[pt, anchor=north east] at ([xshift=-2.6pt, yshift=-1.4pt]xs) {$\bx_{j^\ast}$};
  \end{scope}
  \draw[-{Stealth[length=3pt, width=2.4pt]}, spblue, line width=0.5pt] ([yshift=0.5pt]f3u.north) to[out=90, in=-40] ([xshift=1.6pt, yshift=-1.6pt]xs);
\end{tikzpicture}
\caption{Density subtraction.
When evaluating the double layer integral of \cref{eq:evaluate-solution} at a point $\bx$ close to the nearest collocation point $\bx_{j^\ast}$, the singularity of the Poisson kernel might lead to unbounded high values.
The kernel grows without bound as $\bx$ approaches the surface.
By subtracting the constant part of the density $u$, the remainder $u - u_0$ vanishes under the peak, weakening the singularity at $\bx_{j^\ast}$.}
\label{fig:subtraction-idea}
\end{figure}
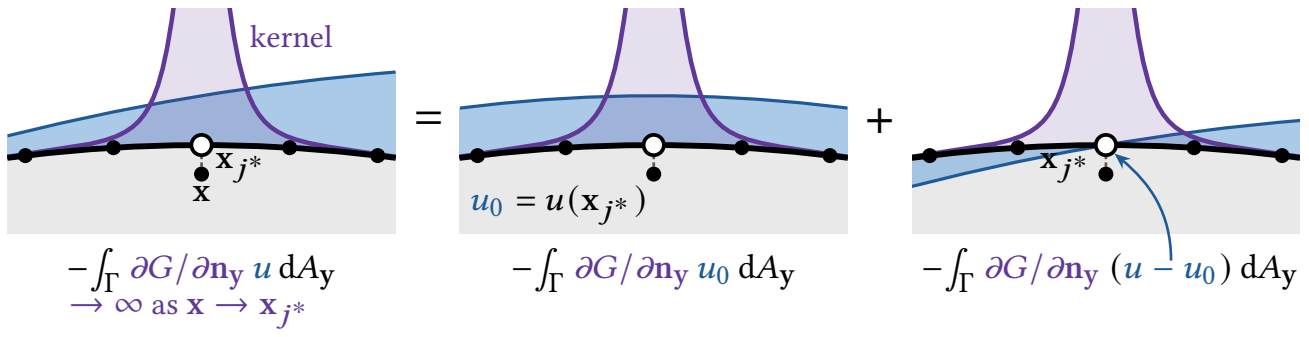

%% file: figures/subtraction-idea/sp-anchors.tex
\def\spXx{0.5000}\def\spXy{0.2627}\def\spSx{0.5000}\def\spSy{0.3922}\def\spKx{0.5708}\def\spKy{0.9490}\def\spUx{0.2727}\def\spUy{0.5380}%

%% file: figures/unproject/unproject-snippet.tex
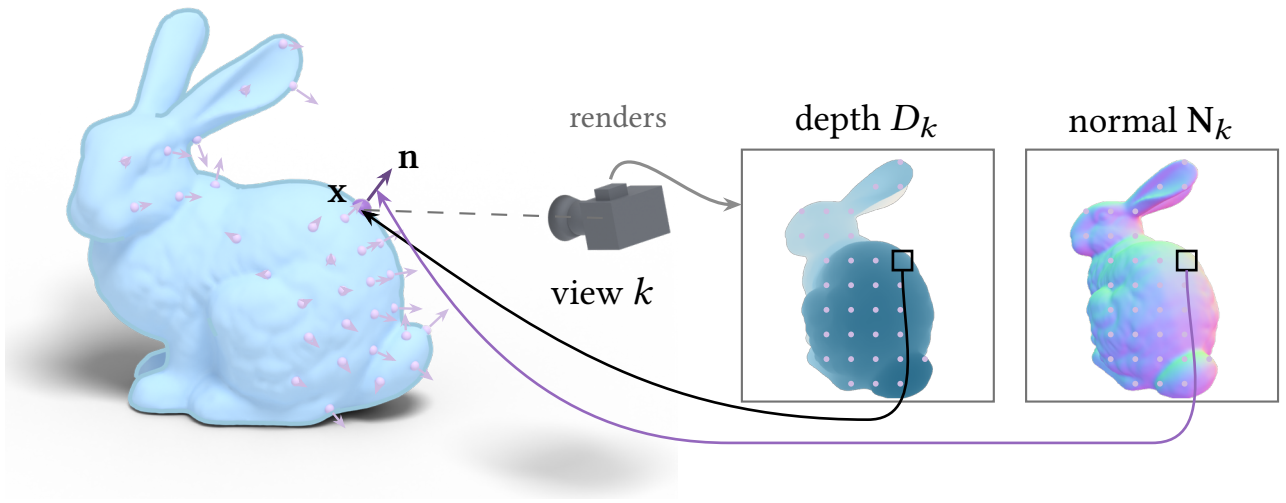
\begin{figure}
\centering
\definecolor{slicepurple}{HTML}{9467BD}%
\begin{tikzpicture}[every node/.style={inner sep=0, outer sep=0}]
  \node[anchor=south west] (O) at (0,0) {\includegraphics[width=0.52\columnwidth]{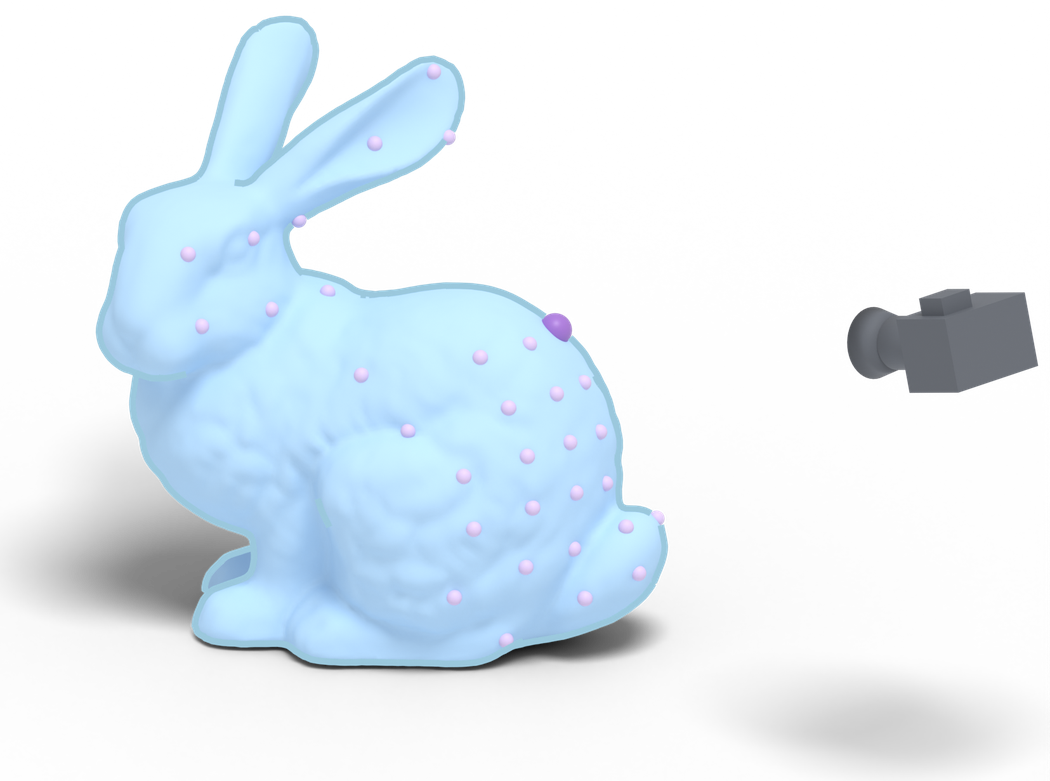}};
  \node[anchor=south west, draw=black!55, line width=0.4pt] (D) at ([xshift=0.05\columnwidth, yshift=0.075\columnwidth]O.south east) {\includegraphics[width=0.195\columnwidth]{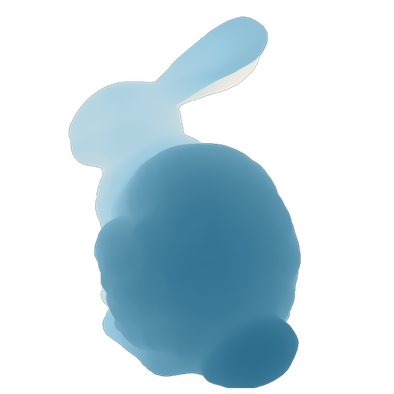}};
  \node[anchor=south west, draw=black!55, line width=0.4pt] (N) at ([xshift=0.025\columnwidth]D.south east) {\includegraphics[width=0.195\columnwidth]{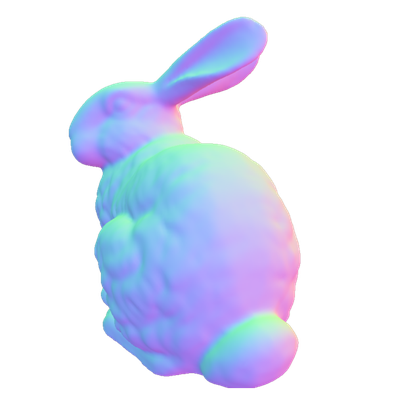}};
  \node[font=\footnotesize, anchor=south] at ([yshift=2pt]D.north) {depth $D_k$};
  \node[font=\footnotesize, anchor=south] at ([yshift=2pt]N.north) {normal $\mathbf{N}_k$};
  \begin{scope}[shift={(O.south west)}, x={($(O.south east)-(O.south west)$)}, y={($(O.north west)-(O.south west)$)}]
    \coordinate (X) at (0.5287,0.5774); \coordinate (Xn0) at (0.6230,0.7449); \coordinate (Xn) at ($(X)!0.5!(Xn0)$); \coordinate (B) at (0.8891,0.5611);
    \draw[black!55, line width=0.4pt, dashed] (B) -- (X);
    \input{figures/unproject/unproject-faint}
    \draw[-{Stealth[length=3.3pt]}, line width=0.6pt, slicepurple!70!black] ($(X)!2.2pt!(Xn)$) -- (Xn);
    \node[font=\footnotesize, anchor=south east] at ([xshift=-2.5pt,yshift=1.5pt]X) {$\mathbf{x}$}; \node[font=\footnotesize, anchor=south west] at ([xshift=1pt]Xn) {$\mathbf{n}$};
    \node[font=\footnotesize, anchor=north] at ([yshift=-0.045\columnwidth]B) {view $k$};
  \end{scope}
  \begin{scope}[shift={(D.south west)}, x={($(D.south east)-(D.south west)$)}, y={($(D.north west)-(D.south west)$)}] \coordinate (pd) at (0.6375,0.5596); \input{figures/unproject/unproject-faint-pix}\end{scope}
  \begin{scope}[shift={(N.south west)}, x={($(N.south east)-(N.south west)$)}, y={($(N.north west)-(N.south west)$)}] \coordinate (pn) at (0.6375,0.5596); \input{figures/unproject/unproject-faint-pix}\end{scope}
  \foreach \p in {pd, pn} { \draw[line width=0.5pt] ([xshift=-1.8pt,yshift=-1.8pt]\p) rectangle ([xshift=1.8pt,yshift=1.8pt]\p); }
  \draw[-{Stealth[length=3.5pt]}, line width=0.6pt, black!45] ([xshift=0.006\columnwidth, yshift=0.03\columnwidth]B) to[out=75, in=180] ([yshift=0.055\columnwidth]D.west);
  \node[font=\scriptsize, text=black!60, anchor=south] at ([xshift=0.012\columnwidth, yshift=0.070\columnwidth]B) {renders};
  \coordinate (Dl) at ($(D.south)+(0,-0.014\columnwidth)$); \coordinate (Nl) at ($(N.south)+(0,-0.032\columnwidth)$); \coordinate (Wl) at ($(D.south west)+(-0.01\columnwidth,-0.032\columnwidth)$);
  \draw[-{Stealth[length=3.5pt]}, line width=0.5pt] ([yshift=-1.8pt]pd) to[out=-90, in=0] (Dl) to[out=180, in=-35] (X);
  \draw[-{Stealth[length=3.5pt]}, line width=0.5pt, slicepurple] ([yshift=-1.8pt]pn) to[out=-90, in=0] (Nl) -- (Wl) to[out=180, in=-60] ($(X)!0.5!(Xn)$);
\end{tikzpicture}
\caption{Unprojection. We extract a collocation point from the rendered maps. For training view $k$ the reconstruction renders a depth map $D_k$ and a normal map $\mathbf{N}_k$ (right). A pixel of the object interior is unprojected along its ray to its depth, which gives the point $\mathbf{x}$ on the rendered surface, and the normal map gives its normal $\mathbf{n}$ at the same pixel.}
\label{fig:unproject}
\end{figure}

%% file: figures/unproject/unproject-faint.tex
\draw[-{Stealth[length=2.6pt]}, line width=0.45pt, slicepurple!45] ($(0.1918,0.5820)!1.1pt!(0.2193,0.5903)$) -- (0.2193,0.5903);
\draw[-{Stealth[length=2.6pt]}, line width=0.45pt, slicepurple!45] ($(0.1791,0.6742)!1.1pt!(0.1879,0.6856)$) -- (0.1879,0.6856);
\draw[-{Stealth[length=2.6pt]}, line width=0.45pt, slicepurple!45] ($(0.4326,0.2348)!1.1pt!(0.4500,0.2222)$) -- (0.4500,0.2222);
\draw[-{Stealth[length=2.6pt]}, line width=0.45pt, slicepurple!45] ($(0.4512,0.3224)!1.1pt!(0.4682,0.3005)$) -- (0.4682,0.3005);
\draw[-{Stealth[length=2.6pt]}, line width=0.45pt, slicepurple!45] ($(0.4416,0.3900)!1.1pt!(0.4580,0.3992)$) -- (0.4580,0.3992);
\draw[-{Stealth[length=2.6pt]}, line width=0.45pt, slicepurple!45] ($(0.3889,0.4479)!1.1pt!(0.3711,0.4481)$) -- (0.3711,0.4481);
\draw[-{Stealth[length=2.6pt]}, line width=0.45pt, slicepurple!45] ($(0.3443,0.5194)!1.1pt!(0.3255,0.5334)$) -- (0.3255,0.5334);
\draw[-{Stealth[length=2.6pt]}, line width=0.45pt, slicepurple!45] ($(0.2581,0.6037)!1.1pt!(0.2939,0.6032)$) -- (0.2939,0.6032);
\draw[-{Stealth[length=2.6pt]}, line width=0.45pt, slicepurple!45] ($(0.2400,0.6948)!1.1pt!(0.2783,0.6848)$) -- (0.2783,0.6848);
\draw[-{Stealth[length=2.6pt]}, line width=0.45pt, slicepurple!45] ($(0.4817,0.1804)!1.1pt!(0.5063,0.1400)$) -- (0.5063,0.1400);
\draw[-{Stealth[length=2.6pt]}, line width=0.45pt, slicepurple!45] ($(0.5006,0.2738)!1.1pt!(0.5149,0.2495)$) -- (0.5149,0.2495);
\draw[-{Stealth[length=2.6pt]}, line width=0.45pt, slicepurple!45] ($(0.5072,0.3496)!1.1pt!(0.5257,0.3329)$) -- (0.5257,0.3329);
\draw[-{Stealth[length=2.6pt]}, line width=0.45pt, slicepurple!45] ($(0.5021,0.4155)!1.1pt!(0.5184,0.4167)$) -- (0.5184,0.4167);
\draw[-{Stealth[length=2.6pt]}, line width=0.45pt, slicepurple!45] ($(0.4844,0.4775)!1.1pt!(0.4846,0.4921)$) -- (0.4846,0.4921);
\draw[-{Stealth[length=2.6pt]}, line width=0.45pt, slicepurple!45] ($(0.4574,0.5420)!1.1pt!(0.4710,0.5659)$) -- (0.4710,0.5659);
\draw[-{Stealth[length=2.6pt]}, line width=0.45pt, slicepurple!45] ($(0.3120,0.6253)!1.1pt!(0.3217,0.6832)$) -- (0.3217,0.6832);
\draw[-{Stealth[length=2.6pt]}, line width=0.45pt, slicepurple!45] ($(0.2844,0.7186)!1.1pt!(0.3008,0.6619)$) -- (0.3008,0.6619);
\draw[-{Stealth[length=2.6pt]}, line width=0.45pt, slicepurple!45] ($(0.5571,0.2337)!1.1pt!(0.5502,0.2074)$) -- (0.5502,0.2074);
\draw[-{Stealth[length=2.6pt]}, line width=0.45pt, slicepurple!45] ($(0.5462,0.2972)!1.1pt!(0.5821,0.2750)$) -- (0.5821,0.2750);
\draw[-{Stealth[length=2.6pt]}, line width=0.45pt, slicepurple!45] ($(0.5483,0.3689)!1.1pt!(0.5837,0.3556)$) -- (0.5837,0.3556);
\draw[-{Stealth[length=2.6pt]}, line width=0.45pt, slicepurple!45] ($(0.5424,0.4333)!1.1pt!(0.5750,0.4391)$) -- (0.5750,0.4391);
\draw[-{Stealth[length=2.6pt]}, line width=0.45pt, slicepurple!45] ($(0.5297,0.4954)!1.1pt!(0.5600,0.5142)$) -- (0.5600,0.5142);
\draw[-{Stealth[length=2.6pt]}, line width=0.45pt, slicepurple!45] ($(0.5040,0.5589)!1.1pt!(0.5323,0.5953)$) -- (0.5323,0.5953);
\draw[-{Stealth[length=2.6pt]}, line width=0.45pt, slicepurple!45] ($(0.3569,0.8160)!1.1pt!(0.3656,0.8295)$) -- (0.3656,0.8295);
\draw[-{Stealth[length=2.6pt]}, line width=0.45pt, slicepurple!45] ($(0.6081,0.2660)!1.1pt!(0.6373,0.2358)$) -- (0.6373,0.2358);
\draw[-{Stealth[length=2.6pt]}, line width=0.45pt, slicepurple!45] ($(0.5962,0.3249)!1.1pt!(0.5949,0.3714)$) -- (0.5949,0.3714);
\draw[-{Stealth[length=2.6pt]}, line width=0.45pt, slicepurple!45] ($(0.5766,0.3808)!1.1pt!(0.6214,0.3793)$) -- (0.6214,0.3793);
\draw[-{Stealth[length=2.6pt]}, line width=0.45pt, slicepurple!45] ($(0.5709,0.4467)!1.1pt!(0.6141,0.4560)$) -- (0.6141,0.4560);
\draw[-{Stealth[length=2.6pt]}, line width=0.45pt, slicepurple!45] ($(0.5558,0.5098)!1.1pt!(0.5933,0.5379)$) -- (0.5933,0.5379);
\draw[-{Stealth[length=2.6pt]}, line width=0.45pt, slicepurple!45] ($(0.4264,0.8243)!1.1pt!(0.4663,0.7878)$) -- (0.4663,0.7878);
\draw[-{Stealth[length=2.6pt]}, line width=0.45pt, slicepurple!45] ($(0.4127,0.9081)!1.1pt!(0.4449,0.9038)$) -- (0.4449,0.9038);
\draw[-{Stealth[length=2.6pt]}, line width=0.45pt, slicepurple!45] ($(0.6254,0.3361)!1.1pt!(0.6602,0.3800)$) -- (0.6602,0.3800);

%% file: figures/unproject/unproject-faint-pix.tex
\fill[slicepurple!45] (0.2380,0.6574) circle[radius=0.55pt];
\fill[slicepurple!45] (0.2380,0.7551) circle[radius=0.55pt];
\fill[slicepurple!45] (0.3358,0.1686) circle[radius=0.55pt];
\fill[slicepurple!45] (0.3358,0.2664) circle[radius=0.55pt];
\fill[slicepurple!45] (0.3358,0.3641) circle[radius=0.55pt];
\fill[slicepurple!45] (0.3358,0.4619) circle[radius=0.55pt];
\fill[slicepurple!45] (0.3358,0.5596) circle[radius=0.55pt];
\fill[slicepurple!45] (0.3358,0.6574) circle[radius=0.55pt];
\fill[slicepurple!45] (0.3358,0.7551) circle[radius=0.55pt];
\fill[slicepurple!45] (0.4335,0.0708) circle[radius=0.55pt];
\fill[slicepurple!45] (0.4335,0.1686) circle[radius=0.55pt];
\fill[slicepurple!45] (0.4335,0.2664) circle[radius=0.55pt];
\fill[slicepurple!45] (0.4335,0.3641) circle[radius=0.55pt];
\fill[slicepurple!45] (0.4335,0.4619) circle[radius=0.55pt];
\fill[slicepurple!45] (0.4335,0.5596) circle[radius=0.55pt];
\fill[slicepurple!45] (0.4335,0.6574) circle[radius=0.55pt];
\fill[slicepurple!45] (0.4335,0.7551) circle[radius=0.55pt];
\fill[slicepurple!45] (0.5313,0.0708) circle[radius=0.55pt];
\fill[slicepurple!45] (0.5313,0.1686) circle[radius=0.55pt];
\fill[slicepurple!45] (0.5313,0.2664) circle[radius=0.55pt];
\fill[slicepurple!45] (0.5313,0.3641) circle[radius=0.55pt];
\fill[slicepurple!45] (0.5313,0.4619) circle[radius=0.55pt];
\fill[slicepurple!45] (0.5313,0.5596) circle[radius=0.55pt];
\fill[slicepurple!45] (0.5313,0.8529) circle[radius=0.55pt];
\fill[slicepurple!45] (0.6290,0.0708) circle[radius=0.55pt];
\fill[slicepurple!45] (0.6290,0.1686) circle[radius=0.55pt];
\fill[slicepurple!45] (0.6290,0.2664) circle[radius=0.55pt];
\fill[slicepurple!45] (0.6290,0.3641) circle[radius=0.55pt];
\fill[slicepurple!45] (0.6290,0.4619) circle[radius=0.55pt];
\fill[slicepurple!45] (0.6290,0.8529) circle[radius=0.55pt];
\fill[slicepurple!45] (0.6290,0.9506) circle[radius=0.55pt];
\fill[slicepurple!45] (0.7268,0.1686) circle[radius=0.55pt];

%% file: figures/gate2/gate-snippet.tex
\begin{figure}
\centering
\definecolor{gatefront}{HTML}{9CCBE0}\definecolor{gateback}{HTML}{F4C6A0}
\begin{tikzpicture}[every node/.style={inner sep=0, outer sep=0}]
  \node[anchor=south west] (A) at (0,0) {\includegraphics[width=0.3\columnwidth]{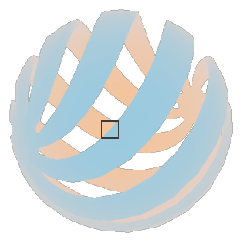}};
  \node[anchor=south west] (B) at ([xshift=0.05\columnwidth]A.south east) {\includegraphics[width=0.3\columnwidth]{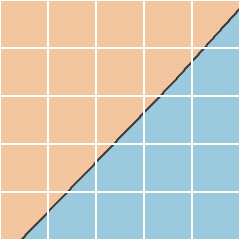}};
  \node[anchor=south west] (C) at ([xshift=0.05\columnwidth]B.south east) {\includegraphics[width=0.3\columnwidth]{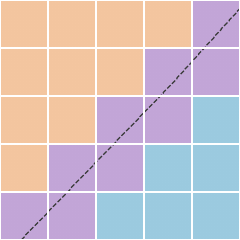}};
  \begin{scope}[shift={(A.south west)}, x={(A.south east)}, y={(A.north west)}]
    \coordinate (b1) at (0.4920,0.4984); \coordinate (b2) at (0.4920,0.4279);
  \end{scope}
  \draw[black!60, line width=0.3pt] (b1) -- (B.north west); \draw[black!60, line width=0.3pt] (b2) -- (B.south west);
  \begin{scope}[shift={(B.south west)}, x={($(B.south east)-(B.south west)$)}, y={($(B.north west)-(B.south west)$)}]
    \node[font=\footnotesize, fill=gateback, inner sep=0.8pt] at (0.35,0.80) {back surface};
    \node[font=\footnotesize, fill=gatefront, inner sep=0.8pt] at (0.65,0.15) {front surface};
  \end{scope}
  \begin{scope}[shift={(C.south west)}, x={($(C.south east)-(C.south west)$)}, y={($(C.north west)-(C.south west)$)}]
    \node[font=\footnotesize, fill=gateback, inner sep=0.8pt, align=center] (nb) at (0.25,0.80) {belongs to\\neither};
    \draw[-{Stealth[length=3pt, width=2.4pt]}, line width=0.5pt] ([xshift=-5pt]nb.south east) -- (0.47,0.53);
  \end{scope}
  \node[anchor=north west] (D) at ([yshift=-10pt]A.south west) {\includegraphics[width=0.3\columnwidth]{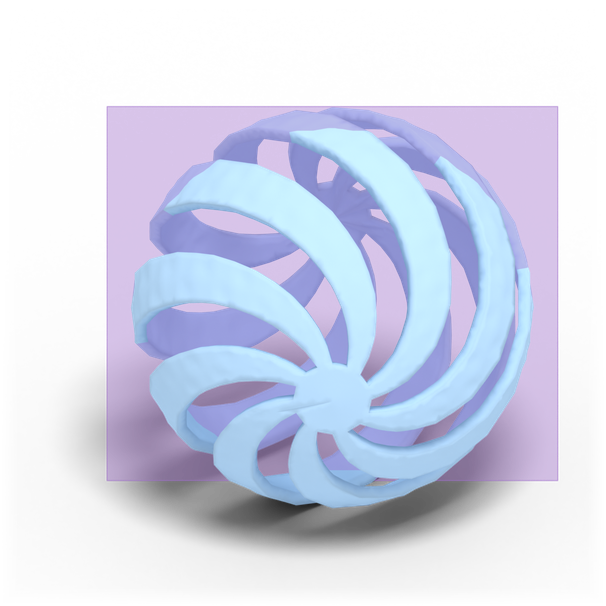}};
  \node[anchor=north west] (E) at ([yshift=-10pt]B.south west) {\includegraphics[width=0.3\columnwidth]{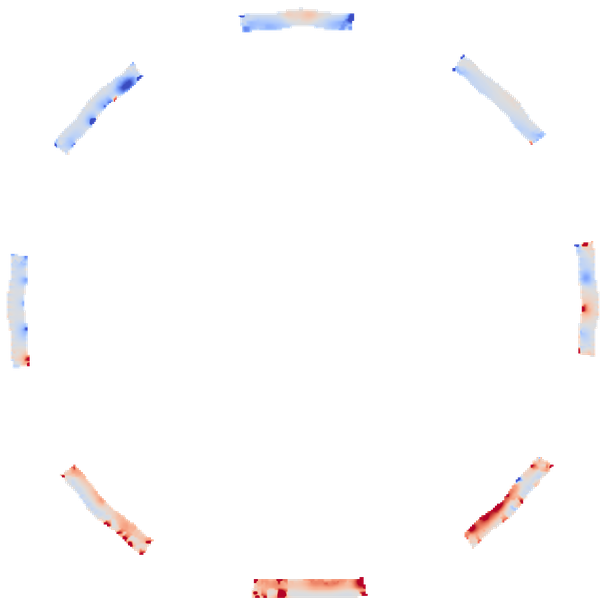}};
  \node[anchor=north west] (F) at ([yshift=-10pt]C.south west) {\includegraphics[width=0.3\columnwidth]{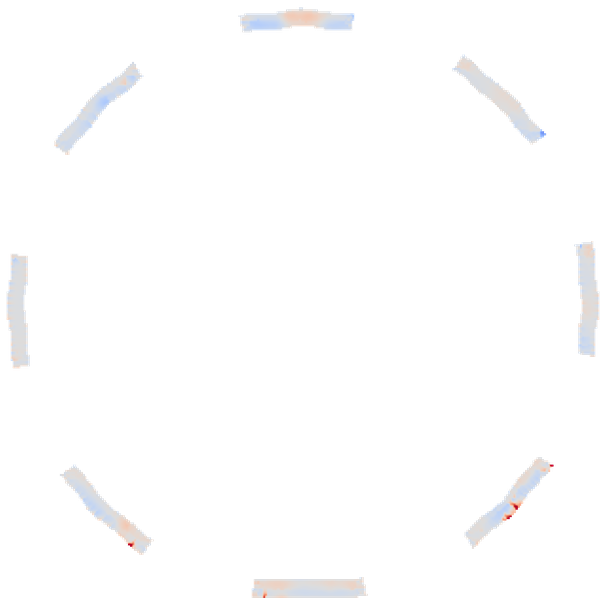}};
  \node[anchor=north, font=\footnotesize] (LD) at ([yshift=-3pt]D.south) {\textsc{Rind} and cut plane};
  \node[anchor=north, font=\footnotesize] at ([yshift=-3pt]E.south) {error without the filter};
  \node[anchor=north, font=\footnotesize] at ([yshift=-3pt]F.south) {error with the filter};
\end{tikzpicture}
\caption{Depth-slope filtering. Top: the rendered depth with a window of $5 \times 5$ pixels at the discontinuity between two surface patches, where blending makes the pixels' radiance and normal unreliable. Bottom: the interior error on the slice plane (left) without and with the filter.}
\label{fig:gate}
\end{figure}
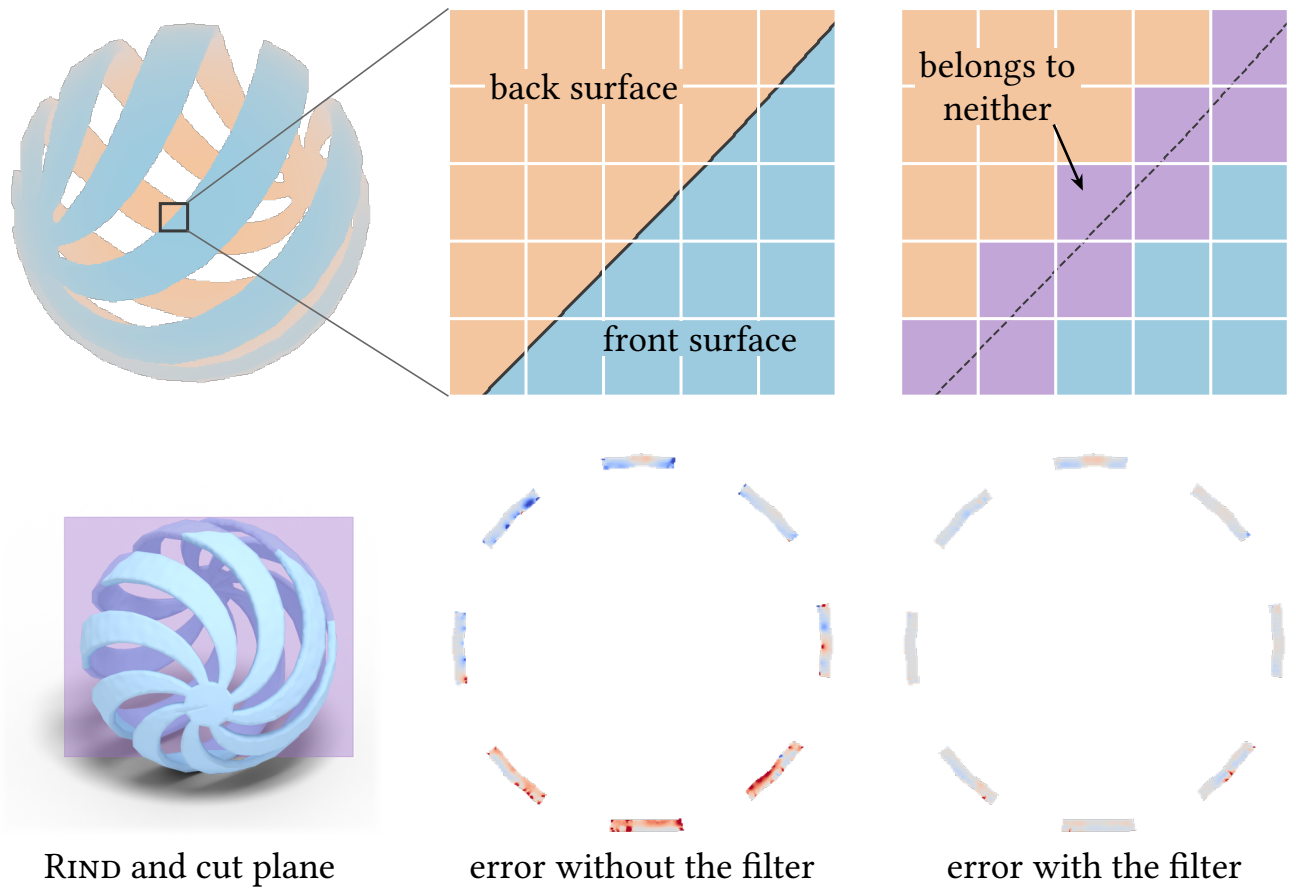

%% file: figures/subtraction/subtraction-snippet.tex
\begin{figure}
\centering
\begin{tikzpicture}[every node/.style={inner sep=0, outer sep=0}, pan/.style={draw=black!35, line width=0.3pt}, lab/.style={font=\footnotesize, anchor=north}]
  \def\wo{0.15\columnwidth}\def\wp{0.2\columnwidth}
  \node[anchor=south west] (O) at (0, 0) {\includegraphics[width=\wo]{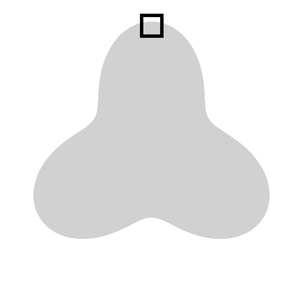}};
  \node[anchor=south west, pan] (A) at ([xshift=0.022\columnwidth]O.south east) {\includegraphics[width=\wp]{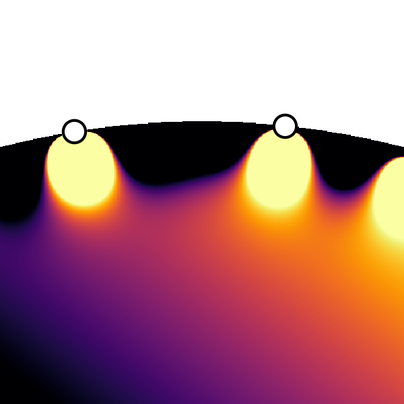}};
  \node[anchor=south west, pan] (B) at ([xshift=0.006\columnwidth]A.south east) {\includegraphics[width=\wp]{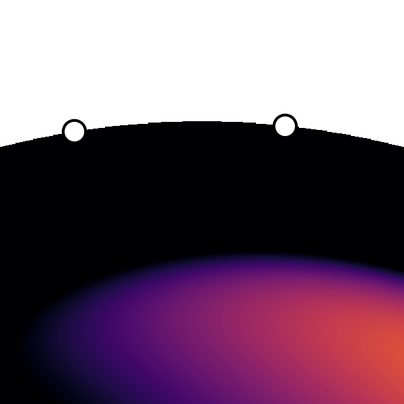}};
  \node[anchor=south west, pan] (C) at ([xshift=0.006\columnwidth]B.south east) {\includegraphics[width=\wp]{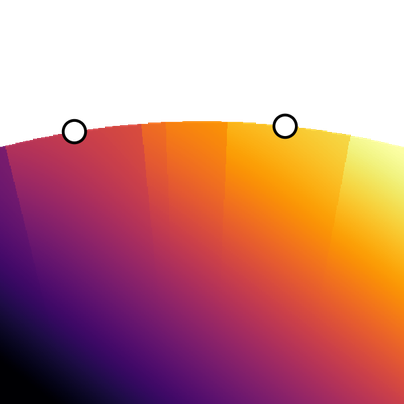}};
  \node[anchor=south west, pan] (D) at ([xshift=0.006\columnwidth]C.south east) {\includegraphics[width=\wp]{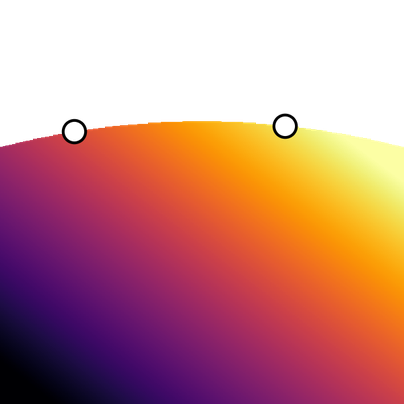}};
  \draw[black!55, line width=0.35pt] ($(O.south west) + (0.534*\wo, 0.95*\wo)$) -- (A.north west);
  \draw[black!55, line width=0.35pt] ($(O.south west) + (0.534*\wo, 0.882*\wo)$) -- (A.south west);
  \node[lab] at ([yshift=-2pt]O.south) {slice}; \node[lab] at ([yshift=-2pt]A.south) {plain sum}; \node[lab] at ([yshift=-2pt]B.south) {mollified}; \node[lab] at ([yshift=-2pt]C.south) {+ subtraction}; \node[lab] at ([yshift=-2pt]D.south) {reference};
\end{tikzpicture}
\caption{Interior temperature near the surface. The singularity in the plain point sum leaves a hot spot at every collocation point; mollifying the kernel smooths the singularity into a cold band; the density subtraction removes it up to faint seams where the nearest sample changes.}
\label{fig:subtraction}
\end{figure}
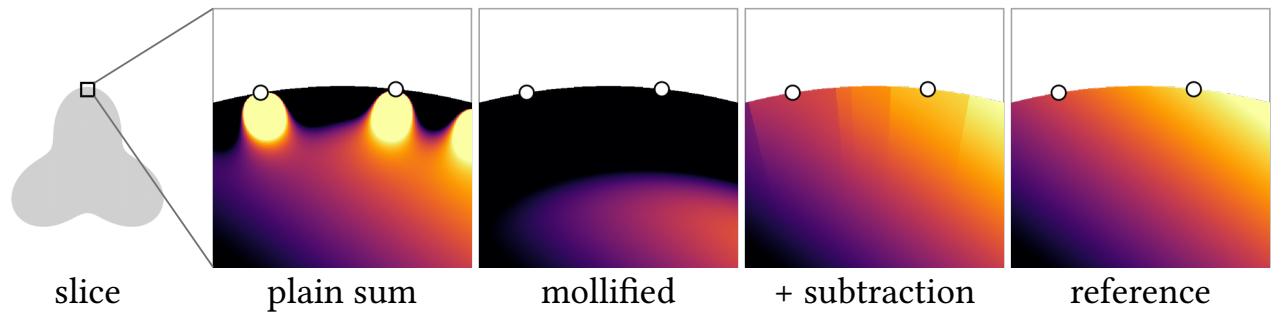

%% file: figures/rows/rows-inset.tex
\definecolor{rowsblueL}{RGB}{135,180,222}\definecolor{rowsblueD}{RGB}{30,90,150}\definecolor{rowspurpL}{RGB}{172,142,212}\definecolor{rowspurpD}{RGB}{96,56,150}%
\begin{tikzpicture}[x=1pt, y=1pt, font=\scriptsize, every node/.style={inner sep=0.6pt}]
  \fill[black!8] (38.00,38.00) rectangle (74.40,74.40);  \fill[black!8] (0,0) rectangle (36.40,36.40);
  \fill[rowsblueL!22] (0,38.00) rectangle (36.40,74.40);  \fill[rowspurpL!14] (38.00,0) rectangle (74.40,36.40);
  \fill[rowspurpL, fill opacity=0.850] (38.00,33.80) rectangle (40.60,36.40);
  \fill[rowspurpL, fill opacity=0.635] (40.60,33.80) rectangle (43.20,36.40);
  \fill[rowspurpL, fill opacity=0.375] (43.20,33.80) rectangle (45.80,36.40);
  \fill[rowspurpL, fill opacity=0.251] (45.80,33.80) rectangle (48.40,36.40);
  \fill[rowspurpL, fill opacity=0.188] (48.40,33.80) rectangle (51.00,36.40);
  \fill[rowspurpL, fill opacity=0.151] (51.00,33.80) rectangle (53.60,36.40);
  \fill[rowspurpL, fill opacity=0.126] (53.60,33.80) rectangle (56.20,36.40);
  \fill[rowspurpL, fill opacity=0.108] (56.20,33.80) rectangle (58.80,36.40);
  \fill[rowspurpL, fill opacity=0.094] (58.80,33.80) rectangle (61.40,36.40);
  \fill[rowspurpL, fill opacity=0.084] (61.40,33.80) rectangle (64.00,36.40);
  \fill[rowspurpL, fill opacity=0.075] (64.00,33.80) rectangle (66.60,36.40);
  \fill[rowspurpL, fill opacity=0.068] (66.60,33.80) rectangle (69.20,36.40);
  \fill[rowspurpL, fill opacity=0.063] (69.20,33.80) rectangle (71.80,36.40);
  \fill[rowspurpL, fill opacity=0.058] (71.80,33.80) rectangle (74.40,36.40);
  \fill[rowspurpL, fill opacity=0.635] (38.00,31.20) rectangle (40.60,33.80);
  \fill[rowspurpL, fill opacity=0.850] (40.60,31.20) rectangle (43.20,33.80);
  \fill[rowspurpL, fill opacity=0.635] (43.20,31.20) rectangle (45.80,33.80);
  \fill[rowspurpL, fill opacity=0.375] (45.80,31.20) rectangle (48.40,33.80);
  \fill[rowspurpL, fill opacity=0.251] (48.40,31.20) rectangle (51.00,33.80);
  \fill[rowspurpL, fill opacity=0.188] (51.00,31.20) rectangle (53.60,33.80);
  \fill[rowspurpL, fill opacity=0.151] (53.60,31.20) rectangle (56.20,33.80);
  \fill[rowspurpL, fill opacity=0.126] (56.20,31.20) rectangle (58.80,33.80);
  \fill[rowspurpL, fill opacity=0.108] (58.80,31.20) rectangle (61.40,33.80);
  \fill[rowspurpL, fill opacity=0.094] (61.40,31.20) rectangle (64.00,33.80);
  \fill[rowspurpL, fill opacity=0.084] (64.00,31.20) rectangle (66.60,33.80);
  \fill[rowspurpL, fill opacity=0.075] (66.60,31.20) rectangle (69.20,33.80);
  \fill[rowspurpL, fill opacity=0.068] (69.20,31.20) rectangle (71.80,33.80);
  \fill[rowspurpL, fill opacity=0.063] (71.80,31.20) rectangle (74.40,33.80);
  \fill[rowspurpL, fill opacity=0.375] (38.00,28.60) rectangle (40.60,31.20);
  \fill[rowspurpL, fill opacity=0.635] (40.60,28.60) rectangle (43.20,31.20);
  \fill[rowspurpL, fill opacity=0.850] (43.20,28.60) rectangle (45.80,31.20);
  \fill[rowspurpL, fill opacity=0.635] (45.80,28.60) rectangle (48.40,31.20);
  \fill[rowspurpL, fill opacity=0.375] (48.40,28.60) rectangle (51.00,31.20);
  \fill[rowspurpL, fill opacity=0.251] (51.00,28.60) rectangle (53.60,31.20);
  \fill[rowspurpL, fill opacity=0.188] (53.60,28.60) rectangle (56.20,31.20);
  \fill[rowspurpL, fill opacity=0.151] (56.20,28.60) rectangle (58.80,31.20);
  \fill[rowspurpL, fill opacity=0.126] (58.80,28.60) rectangle (61.40,31.20);
  \fill[rowspurpL, fill opacity=0.108] (61.40,28.60) rectangle (64.00,31.20);
  \fill[rowspurpL, fill opacity=0.094] (64.00,28.60) rectangle (66.60,31.20);
  \fill[rowspurpL, fill opacity=0.084] (66.60,28.60) rectangle (69.20,31.20);
  \fill[rowspurpL, fill opacity=0.075] (69.20,28.60) rectangle (71.80,31.20);
  \fill[rowspurpL, fill opacity=0.068] (71.80,28.60) rectangle (74.40,31.20);
  \fill[rowspurpL, fill opacity=0.251] (38.00,26.00) rectangle (40.60,28.60);
  \fill[rowspurpL, fill opacity=0.375] (40.60,26.00) rectangle (43.20,28.60);
  \fill[rowspurpL, fill opacity=0.635] (43.20,26.00) rectangle (45.80,28.60);
  \fill[rowspurpL, fill opacity=0.850] (45.80,26.00) rectangle (48.40,28.60);
  \fill[rowspurpL, fill opacity=0.635] (48.40,26.00) rectangle (51.00,28.60);
  \fill[rowspurpL, fill opacity=0.375] (51.00,26.00) rectangle (53.60,28.60);
  \fill[rowspurpL, fill opacity=0.251] (53.60,26.00) rectangle (56.20,28.60);
  \fill[rowspurpL, fill opacity=0.188] (56.20,26.00) rectangle (58.80,28.60);
  \fill[rowspurpL, fill opacity=0.151] (58.80,26.00) rectangle (61.40,28.60);
  \fill[rowspurpL, fill opacity=0.126] (61.40,26.00) rectangle (64.00,28.60);
  \fill[rowspurpL, fill opacity=0.108] (64.00,26.00) rectangle (66.60,28.60);
  \fill[rowspurpL, fill opacity=0.094] (66.60,26.00) rectangle (69.20,28.60);
  \fill[rowspurpL, fill opacity=0.084] (69.20,26.00) rectangle (71.80,28.60);
  \fill[rowspurpL, fill opacity=0.075] (71.80,26.00) rectangle (74.40,28.60);
  \fill[rowspurpL, fill opacity=0.188] (38.00,23.40) rectangle (40.60,26.00);
  \fill[rowspurpL, fill opacity=0.251] (40.60,23.40) rectangle (43.20,26.00);
  \fill[rowspurpL, fill opacity=0.375] (43.20,23.40) rectangle (45.80,26.00);
  \fill[rowspurpL, fill opacity=0.635] (45.80,23.40) rectangle (48.40,26.00);
  \fill[rowspurpL, fill opacity=0.850] (48.40,23.40) rectangle (51.00,26.00);
  \fill[rowspurpL, fill opacity=0.635] (51.00,23.40) rectangle (53.60,26.00);
  \fill[rowspurpL, fill opacity=0.375] (53.60,23.40) rectangle (56.20,26.00);
  \fill[rowspurpL, fill opacity=0.251] (56.20,23.40) rectangle (58.80,26.00);
  \fill[rowspurpL, fill opacity=0.188] (58.80,23.40) rectangle (61.40,26.00);
  \fill[rowspurpL, fill opacity=0.151] (61.40,23.40) rectangle (64.00,26.00);
  \fill[rowspurpL, fill opacity=0.126] (64.00,23.40) rectangle (66.60,26.00);
  \fill[rowspurpL, fill opacity=0.108] (66.60,23.40) rectangle (69.20,26.00);
  \fill[rowspurpL, fill opacity=0.094] (69.20,23.40) rectangle (71.80,26.00);
  \fill[rowspurpL, fill opacity=0.084] (71.80,23.40) rectangle (74.40,26.00);
  \fill[rowspurpL, fill opacity=0.151] (38.00,20.80) rectangle (40.60,23.40);
  \fill[rowspurpL, fill opacity=0.188] (40.60,20.80) rectangle (43.20,23.40);
  \fill[rowspurpL, fill opacity=0.251] (43.20,20.80) rectangle (45.80,23.40);
  \fill[rowspurpL, fill opacity=0.375] (45.80,20.80) rectangle (48.40,23.40);
  \fill[rowspurpL, fill opacity=0.635] (48.40,20.80) rectangle (51.00,23.40);
  \fill[rowspurpL, fill opacity=0.850] (51.00,20.80) rectangle (53.60,23.40);
  \fill[rowspurpL, fill opacity=0.635] (53.60,20.80) rectangle (56.20,23.40);
  \fill[rowspurpL, fill opacity=0.375] (56.20,20.80) rectangle (58.80,23.40);
  \fill[rowspurpL, fill opacity=0.251] (58.80,20.80) rectangle (61.40,23.40);
  \fill[rowspurpL, fill opacity=0.188] (61.40,20.80) rectangle (64.00,23.40);
  \fill[rowspurpL, fill opacity=0.151] (64.00,20.80) rectangle (66.60,23.40);
  \fill[rowspurpL, fill opacity=0.126] (66.60,20.80) rectangle (69.20,23.40);
  \fill[rowspurpL, fill opacity=0.108] (69.20,20.80) rectangle (71.80,23.40);
  \fill[rowspurpL, fill opacity=0.094] (71.80,20.80) rectangle (74.40,23.40);
  \fill[rowspurpL, fill opacity=0.126] (38.00,18.20) rectangle (40.60,20.80);
  \fill[rowspurpL, fill opacity=0.151] (40.60,18.20) rectangle (43.20,20.80);
  \fill[rowspurpL, fill opacity=0.188] (43.20,18.20) rectangle (45.80,20.80);
  \fill[rowspurpL, fill opacity=0.251] (45.80,18.20) rectangle (48.40,20.80);
  \fill[rowspurpL, fill opacity=0.375] (48.40,18.20) rectangle (51.00,20.80);
  \fill[rowspurpL, fill opacity=0.635] (51.00,18.20) rectangle (53.60,20.80);
  \fill[rowspurpL, fill opacity=0.850] (53.60,18.20) rectangle (56.20,20.80);
  \fill[rowspurpL, fill opacity=0.635] (56.20,18.20) rectangle (58.80,20.80);
  \fill[rowspurpL, fill opacity=0.375] (58.80,18.20) rectangle (61.40,20.80);
  \fill[rowspurpL, fill opacity=0.251] (61.40,18.20) rectangle (64.00,20.80);
  \fill[rowspurpL, fill opacity=0.188] (64.00,18.20) rectangle (66.60,20.80);
  \fill[rowspurpL, fill opacity=0.151] (66.60,18.20) rectangle (69.20,20.80);
  \fill[rowspurpL, fill opacity=0.126] (69.20,18.20) rectangle (71.80,20.80);
  \fill[rowspurpL, fill opacity=0.108] (71.80,18.20) rectangle (74.40,20.80);
  \fill[rowspurpL, fill opacity=0.108] (38.00,15.60) rectangle (40.60,18.20);
  \fill[rowspurpL, fill opacity=0.126] (40.60,15.60) rectangle (43.20,18.20);
  \fill[rowspurpL, fill opacity=0.151] (43.20,15.60) rectangle (45.80,18.20);
  \fill[rowspurpL, fill opacity=0.188] (45.80,15.60) rectangle (48.40,18.20);
  \fill[rowspurpL, fill opacity=0.251] (48.40,15.60) rectangle (51.00,18.20);
  \fill[rowspurpL, fill opacity=0.375] (51.00,15.60) rectangle (53.60,18.20);
  \fill[rowspurpL, fill opacity=0.635] (53.60,15.60) rectangle (56.20,18.20);
  \fill[rowspurpL, fill opacity=0.850] (56.20,15.60) rectangle (58.80,18.20);
  \fill[rowspurpL, fill opacity=0.635] (58.80,15.60) rectangle (61.40,18.20);
  \fill[rowspurpL, fill opacity=0.375] (61.40,15.60) rectangle (64.00,18.20);
  \fill[rowspurpL, fill opacity=0.251] (64.00,15.60) rectangle (66.60,18.20);
  \fill[rowspurpL, fill opacity=0.188] (66.60,15.60) rectangle (69.20,18.20);
  \fill[rowspurpL, fill opacity=0.151] (69.20,15.60) rectangle (71.80,18.20);
  \fill[rowspurpL, fill opacity=0.126] (71.80,15.60) rectangle (74.40,18.20);
  \fill[rowspurpL, fill opacity=0.094] (38.00,13.00) rectangle (40.60,15.60);
  \fill[rowspurpL, fill opacity=0.108] (40.60,13.00) rectangle (43.20,15.60);
  \fill[rowspurpL, fill opacity=0.126] (43.20,13.00) rectangle (45.80,15.60);
  \fill[rowspurpL, fill opacity=0.151] (45.80,13.00) rectangle (48.40,15.60);
  \fill[rowspurpL, fill opacity=0.188] (48.40,13.00) rectangle (51.00,15.60);
  \fill[rowspurpL, fill opacity=0.251] (51.00,13.00) rectangle (53.60,15.60);
  \fill[rowspurpL, fill opacity=0.375] (53.60,13.00) rectangle (56.20,15.60);
  \fill[rowspurpL, fill opacity=0.635] (56.20,13.00) rectangle (58.80,15.60);
  \fill[rowspurpL, fill opacity=0.850] (58.80,13.00) rectangle (61.40,15.60);
  \fill[rowspurpL, fill opacity=0.635] (61.40,13.00) rectangle (64.00,15.60);
  \fill[rowspurpL, fill opacity=0.375] (64.00,13.00) rectangle (66.60,15.60);
  \fill[rowspurpL, fill opacity=0.251] (66.60,13.00) rectangle (69.20,15.60);
  \fill[rowspurpL, fill opacity=0.188] (69.20,13.00) rectangle (71.80,15.60);
  \fill[rowspurpL, fill opacity=0.151] (71.80,13.00) rectangle (74.40,15.60);
  \fill[rowspurpL, fill opacity=0.084] (38.00,10.40) rectangle (40.60,13.00);
  \fill[rowspurpL, fill opacity=0.094] (40.60,10.40) rectangle (43.20,13.00);
  \fill[rowspurpL, fill opacity=0.108] (43.20,10.40) rectangle (45.80,13.00);
  \fill[rowspurpL, fill opacity=0.126] (45.80,10.40) rectangle (48.40,13.00);
  \fill[rowspurpL, fill opacity=0.151] (48.40,10.40) rectangle (51.00,13.00);
  \fill[rowspurpL, fill opacity=0.188] (51.00,10.40) rectangle (53.60,13.00);
  \fill[rowspurpL, fill opacity=0.251] (53.60,10.40) rectangle (56.20,13.00);
  \fill[rowspurpL, fill opacity=0.375] (56.20,10.40) rectangle (58.80,13.00);
  \fill[rowspurpL, fill opacity=0.635] (58.80,10.40) rectangle (61.40,13.00);
  \fill[rowspurpL, fill opacity=0.850] (61.40,10.40) rectangle (64.00,13.00);
  \fill[rowspurpL, fill opacity=0.635] (64.00,10.40) rectangle (66.60,13.00);
  \fill[rowspurpL, fill opacity=0.375] (66.60,10.40) rectangle (69.20,13.00);
  \fill[rowspurpL, fill opacity=0.251] (69.20,10.40) rectangle (71.80,13.00);
  \fill[rowspurpL, fill opacity=0.188] (71.80,10.40) rectangle (74.40,13.00);
  \fill[rowspurpL, fill opacity=0.075] (38.00,7.80) rectangle (40.60,10.40);
  \fill[rowspurpL, fill opacity=0.084] (40.60,7.80) rectangle (43.20,10.40);
  \fill[rowspurpL, fill opacity=0.094] (43.20,7.80) rectangle (45.80,10.40);
  \fill[rowspurpL, fill opacity=0.108] (45.80,7.80) rectangle (48.40,10.40);
  \fill[rowspurpL, fill opacity=0.126] (48.40,7.80) rectangle (51.00,10.40);
  \fill[rowspurpL, fill opacity=0.151] (51.00,7.80) rectangle (53.60,10.40);
  \fill[rowspurpL, fill opacity=0.188] (53.60,7.80) rectangle (56.20,10.40);
  \fill[rowspurpL, fill opacity=0.251] (56.20,7.80) rectangle (58.80,10.40);
  \fill[rowspurpL, fill opacity=0.375] (58.80,7.80) rectangle (61.40,10.40);
  \fill[rowspurpL, fill opacity=0.635] (61.40,7.80) rectangle (64.00,10.40);
  \fill[rowspurpL, fill opacity=0.850] (64.00,7.80) rectangle (66.60,10.40);
  \fill[rowspurpL, fill opacity=0.635] (66.60,7.80) rectangle (69.20,10.40);
  \fill[rowspurpL, fill opacity=0.375] (69.20,7.80) rectangle (71.80,10.40);
  \fill[rowspurpL, fill opacity=0.251] (71.80,7.80) rectangle (74.40,10.40);
  \fill[rowspurpL, fill opacity=0.068] (38.00,5.20) rectangle (40.60,7.80);
  \fill[rowspurpL, fill opacity=0.075] (40.60,5.20) rectangle (43.20,7.80);
  \fill[rowspurpL, fill opacity=0.084] (43.20,5.20) rectangle (45.80,7.80);
  \fill[rowspurpL, fill opacity=0.094] (45.80,5.20) rectangle (48.40,7.80);
  \fill[rowspurpL, fill opacity=0.108] (48.40,5.20) rectangle (51.00,7.80);
  \fill[rowspurpL, fill opacity=0.126] (51.00,5.20) rectangle (53.60,7.80);
  \fill[rowspurpL, fill opacity=0.151] (53.60,5.20) rectangle (56.20,7.80);
  \fill[rowspurpL, fill opacity=0.188] (56.20,5.20) rectangle (58.80,7.80);
  \fill[rowspurpL, fill opacity=0.251] (58.80,5.20) rectangle (61.40,7.80);
  \fill[rowspurpL, fill opacity=0.375] (61.40,5.20) rectangle (64.00,7.80);
  \fill[rowspurpL, fill opacity=0.635] (64.00,5.20) rectangle (66.60,7.80);
  \fill[rowspurpL, fill opacity=0.850] (66.60,5.20) rectangle (69.20,7.80);
  \fill[rowspurpL, fill opacity=0.635] (69.20,5.20) rectangle (71.80,7.80);
  \fill[rowspurpL, fill opacity=0.375] (71.80,5.20) rectangle (74.40,7.80);
  \fill[rowspurpL, fill opacity=0.063] (38.00,2.60) rectangle (40.60,5.20);
  \fill[rowspurpL, fill opacity=0.068] (40.60,2.60) rectangle (43.20,5.20);
  \fill[rowspurpL, fill opacity=0.075] (43.20,2.60) rectangle (45.80,5.20);
  \fill[rowspurpL, fill opacity=0.084] (45.80,2.60) rectangle (48.40,5.20);
  \fill[rowspurpL, fill opacity=0.094] (48.40,2.60) rectangle (51.00,5.20);
  \fill[rowspurpL, fill opacity=0.108] (51.00,2.60) rectangle (53.60,5.20);
  \fill[rowspurpL, fill opacity=0.126] (53.60,2.60) rectangle (56.20,5.20);
  \fill[rowspurpL, fill opacity=0.151] (56.20,2.60) rectangle (58.80,5.20);
  \fill[rowspurpL, fill opacity=0.188] (58.80,2.60) rectangle (61.40,5.20);
  \fill[rowspurpL, fill opacity=0.251] (61.40,2.60) rectangle (64.00,5.20);
  \fill[rowspurpL, fill opacity=0.375] (64.00,2.60) rectangle (66.60,5.20);
  \fill[rowspurpL, fill opacity=0.635] (66.60,2.60) rectangle (69.20,5.20);
  \fill[rowspurpL, fill opacity=0.850] (69.20,2.60) rectangle (71.80,5.20);
  \fill[rowspurpL, fill opacity=0.635] (71.80,2.60) rectangle (74.40,5.20);
  \fill[rowspurpL, fill opacity=0.058] (38.00,0.00) rectangle (40.60,2.60);
  \fill[rowspurpL, fill opacity=0.063] (40.60,0.00) rectangle (43.20,2.60);
  \fill[rowspurpL, fill opacity=0.068] (43.20,0.00) rectangle (45.80,2.60);
  \fill[rowspurpL, fill opacity=0.075] (45.80,0.00) rectangle (48.40,2.60);
  \fill[rowspurpL, fill opacity=0.084] (48.40,0.00) rectangle (51.00,2.60);
  \fill[rowspurpL, fill opacity=0.094] (51.00,0.00) rectangle (53.60,2.60);
  \fill[rowspurpL, fill opacity=0.108] (53.60,0.00) rectangle (56.20,2.60);
  \fill[rowspurpL, fill opacity=0.126] (56.20,0.00) rectangle (58.80,2.60);
  \fill[rowspurpL, fill opacity=0.151] (58.80,0.00) rectangle (61.40,2.60);
  \fill[rowspurpL, fill opacity=0.188] (61.40,0.00) rectangle (64.00,2.60);
  \fill[rowspurpL, fill opacity=0.251] (64.00,0.00) rectangle (66.60,2.60);
  \fill[rowspurpL, fill opacity=0.375] (66.60,0.00) rectangle (69.20,2.60);
  \fill[rowspurpL, fill opacity=0.635] (69.20,0.00) rectangle (71.80,2.60);
  \fill[rowspurpL, fill opacity=0.850] (71.80,0.00) rectangle (74.40,2.60);
  \fill[rowsblueD] (0.00,71.80) rectangle (2.60,74.40);
  \fill[rowsblueD] (2.60,69.20) rectangle (5.20,71.80);
  \fill[rowsblueD] (5.20,66.60) rectangle (7.80,69.20);
  \fill[rowsblueD] (7.80,64.00) rectangle (10.40,66.60);
  \fill[rowsblueD] (10.40,61.40) rectangle (13.00,64.00);
  \fill[rowsblueD] (13.00,58.80) rectangle (15.60,61.40);
  \fill[rowsblueD] (15.60,56.20) rectangle (18.20,58.80);
  \fill[rowsblueD] (18.20,53.60) rectangle (20.80,56.20);
  \fill[rowsblueD] (20.80,51.00) rectangle (23.40,53.60);
  \fill[rowsblueD] (23.40,48.40) rectangle (26.00,51.00);
  \fill[rowsblueD] (26.00,45.80) rectangle (28.60,48.40);
  \fill[rowsblueD] (28.60,43.20) rectangle (31.20,45.80);
  \fill[rowsblueD] (31.20,40.60) rectangle (33.80,43.20);
  \fill[rowsblueD] (33.80,38.00) rectangle (36.40,40.60);
  \draw[rowspurpD, line width=1.1pt, dash pattern=on 1.6pt off 1.0pt] (39.30,35.10) -- (73.10,1.30);
  \node[text=rowsblueD] at (24.75,64.94) {$\tfrac{1}{2}I + K$};
  \node[text=black!65] at (56.20,56.20) {$-V$};
  \node[text=black!65] at (18.20,18.20) {$K$};
  \node[text=rowspurpD, fill=white, fill opacity=0.75, text opacity=1, rounded corners=0.8pt] at (60.57,29.12) {$-(V + \gamma I)$};
  \fill[rowsblueL!55] (77.90,38.00) rectangle (84.40,74.40); \node[text=rowsblueD] at (81.15,56.20) {$u$};
  \fill[rowspurpL!55] (77.90,0) rectangle (84.40,36.40); \node[text=rowspurpD] at (81.15,18.20) {$q$};
  \node at (89.00,37.20) {$=$};
  \fill[black!8] (93.60,0) rectangle (100.10,74.40);
  \node[anchor=north, text=black!70] at (37.20,-1.2) {$M$}; \node[anchor=north, text=black!70] at (81.15,-1.2) {$\psi$}; \node[anchor=north, text=black!70] at (96.85,-1.2) {$g$};
\end{tikzpicture}%

%% file: figures/gradcheck/gradcheck-snippet.tex
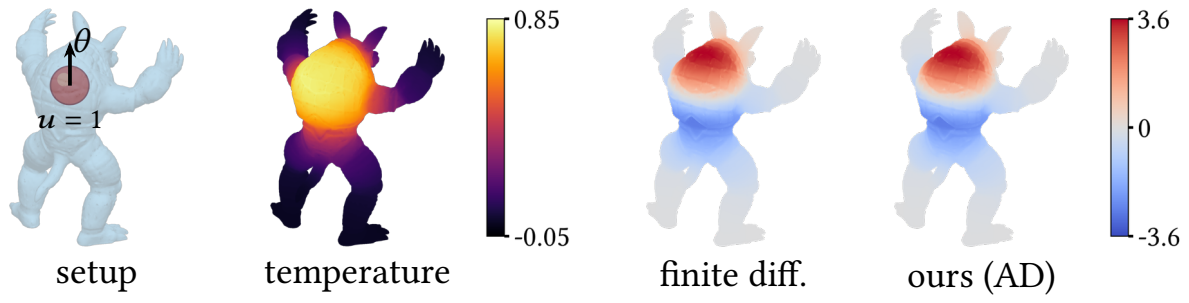
\begin{figure}
\centering
\begin{tikzpicture}[every node/.style={inner sep=0, outer sep=0}]
  \def\w{0.19\columnwidth}
  \node[anchor=south west] (A) at (0,0) {\includegraphics[width=\w]{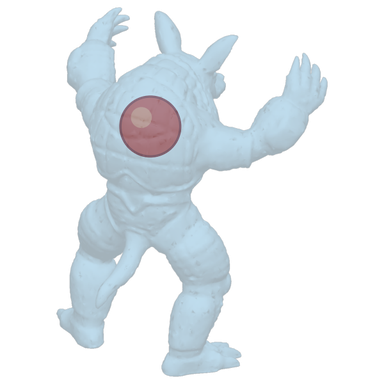}};
  \node[anchor=south west] (T) at ([xshift=0.012\columnwidth]A.south east) {\includegraphics[width=\w]{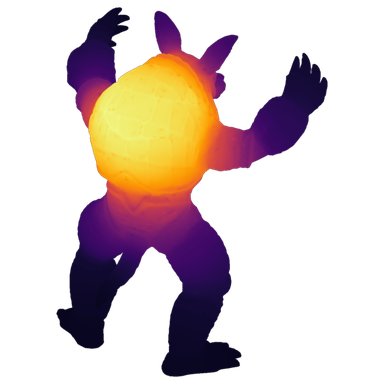}};
  \node[anchor=south west] (TC) at ([xshift=0.002\columnwidth]T.south east) {\includegraphics[height=\w]{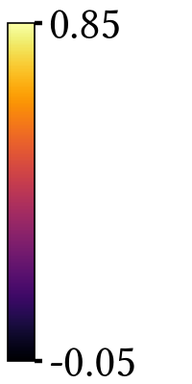}};
  \node[anchor=south west] (F) at ([xshift=0.004\columnwidth]TC.south east) {\includegraphics[width=\w]{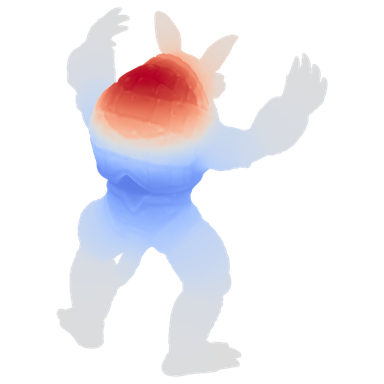}};
  \node[anchor=south west] (D) at ([xshift=0.002\columnwidth]F.south east) {\includegraphics[width=\w]{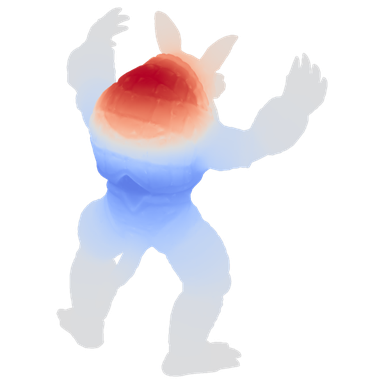}};
  \node[anchor=south west] (DC) at ([xshift=0.002\columnwidth]D.south east) {\includegraphics[height=\w]{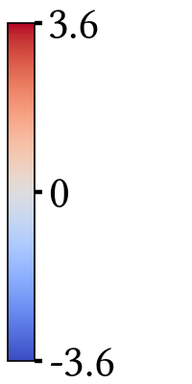}};
  \begin{scope}[shift={(A.south west)}, x={($(A.south east)-(A.south west)$)}, y={($(A.north west)-(A.south west)$)}]
    \draw[-{Stealth[length=3pt]}, line width=0.7pt] (0.3915,0.6705) -- (0.3915,0.8498) node[anchor=south west, font=\scriptsize, xshift=0.5pt, yshift=-2pt] {$\theta$};
    \node[font=\tiny, anchor=north] at (0.3915,0.5705) {$u = 1$};
  \end{scope}
  \foreach \n/\l in {A/setup, T/temperature, F/finite diff., D/ours (AD)} { \node[anchor=north, font=\footnotesize] at ([yshift=-2pt]\n.south) {\l}; }
\end{tikzpicture}
\caption{Verifying the gradient. A hidden sphere held at $u = 1$ heats the armadillo, which cools convectively over its surface ($u + 0.5\,\partial u / \partial \bn = 0$), and we differentiate the surface temperature with respect to an upward translation $\theta$ of the sphere.}
\label{fig:gradcheck}
\end{figure}

%% file: 4-results.tex
\section{Evaluation} \label{sec:results}

\paragraph{Implementation details}
Everything after the reconstruction is written in PyTorch \cite{paszkePyTorchImperativeStyle2019}.
The depth and normal maps are rendered by the RaDe-GS rasterizer \cite{zhangRaDeGSRasterizingDepth2024},
the extraction, the reads and the areas are batched tensor operations,
every step is differentiable by automatic differentiation, the solve through its adjoint.
The solve is matrix-free (\cref{par:solving}), so memory does not bound the collocation count; the tables extract $42$K to $61$K points with full coverage and $19$K to $28$K under one view.
All timings are recorded on one NVIDIA RTX 4090 (24\,GB) with $16$ cores of an AMD EPYC 9554; \cref{ssec:e2e-comparison} gives the run time of both pipelines.
To help reproduce our method, the datasets and code will be released upon acceptance.

\subsection{Datasets} \label{ssec:datasets}
\input{figures/slices/slices-combined-snippet}
Our synthetic dataset comprises eighteen meshes, in 3 groups:
\begin{itemize}\emergencystretch=1.5em
\item \emph{solids}: eight objects, \textsc{Armadillo}, \textsc{Bunny}, \textsc{Hand}, \textsc{Nefertiti}, \textsc{Spot} and \textsc{Springer} from Oded Stein's mesh repository \cite{steinOdedsteinMeshes2024} and \textsc{Botijo} and \textsc{Fertility} from the AIM@SHAPE repository;
\item \emph{thin structures}: six shapes from the mesh set of \citeauthor{mehtaLevelSetTheory2022}~\shortcite{mehtaLevelSetTheory2022}, chosen to stress surface topology: a genus-6 solid (\textsc{Genus-6}), a spiral-ribbon shell whose wall is $3\%$ of its extent thick (\textsc{Rind}), and four strut lattices, a Voronoi sphere (\textsc{Voronoi}) and three high-genus Voronoi-strut shells, a bunny (\textsc{V-bunny}), a kangaroo (\textsc{Kangaroo}) and a dinosaur (\textsc{Dino});
\item \emph{industrial parts}: four parts drawn at random from the industrial models of the \emph{Hex Me If You Can} benchmark \cite{beaufortHexMeIfYou2022}, each a single connected solid of genus one to seven, a bearing bracket (\textsc{Bracket}), a clamp arm (\textsc{Clamp}), a hinge bracket (\textsc{Hinge}) and a slotted housing (\textsc{Housing}), rendered as painted cast parts with non-linear harmonics as their surface temperatures.
\end{itemize}
Each mesh is prescribed a surface temperature by sampling a harmonic function, which we convert to radiance values and render from 300 viewpoints in Blender~\cite{blender45}, and in visible light from the same cameras; each experiment below states what it takes as its reference.
Radiance grows with the fourth power of the absolute temperature following the Stefan--Boltzmann law \cite[Ch. 1]{modestRadiativeHeatTransfer2021}, at emissivity of 1.0 (perfect black body emitter).
We assume perfect black body emitters, with zero thermal reflectance, to focus on conduction.
In addition to the synthetic scenes,
we present a real capture dataset, described in \cref{ssec:real-capture}.

\subsection{Temperature interpolation} \label{ssec:temperature-interpolation}
\input{tables/tab-overview}
\paragraph{Setup}
In this experiment, our goal is to recover the temperature inside the object given thermal images covering its entire surface.
We use the image dataset from the previous section.
Each scene is reconstructed from its $300$ visible-spectrum renders.
Here every thermal view is available, so we can prescribe the temperature (Dirichlet boundary value) over the whole surface.
The rendered surface temperature is prescribed by a harmonic function, normalized to $[0, 1]$ over the temperature range of the render.
Every temperature and error we state on the synthetic scenes is in this unit, a fraction of the scene's temperature range, so an error of $0.001$ is $0.1\%$ of the range; every length is in units of the object's largest extent, which we scale to $1$.
Being harmonic, the function is also the exact interior field, which serves as our reference: we report the mean absolute error on a uniform grid of points inside the true mesh.

\paragraph{Accuracy}
\Cref{tab:overview} gives the result on all eighteen scenes next to the mesh-based baseline of \cref{par:mesh-baseline}, and \cref{fig:slices} shows the cross sections of seven of them, the supplementary material those of all eighteen.
Our mean error is $0.17$\% of the temperature range, from $0.050$\% on \textsc{Springer} to $0.41$\% on \textsc{Kangaroo}.
On the slices of \cref{fig:slices} our error shows the cells of the nearest collocation point: the density subtraction of \mbox{\cref{eq:gs-eval-solutions}} takes the temperature of the collocation point nearest to the evaluation point, so there is a small error that changes from one point's cell to the next and leaves faint seams between them, inside as well as near the surface.
It grows only mildly toward the surface ($0.11$\% of the temperature range within half a spacing against $0.074$\% beyond two spacings on the solids), and it is largest in thin features, which make the strut shells our hardest scenes.
For results of all the scenes in our datasets, please see our supplementary material.

\subsection{Incomplete thermal capture} \label{ssec:incomplete-thermal-capture}
\input{figures/e2e/e2e-snippet}
\paragraph{Setup}
In this experiment, our goal is to recover both the temperature of the surface that the camera does not see and the temperature inside when the thermal observation is not complete.
We study this by using one thermal view per scene.
The Gaussians are still reconstructed from all $300$ visible-spectrum views; the boundary data come from the first thermal view of each scene alone.
On the seen points we prescribe the temperature as seen in the camera, and on the rest we impose an assumed boundary condition, either insulation or a convective exchange with the ambient, the Robin condition of \cref{eq:robin-problem} with $a = 1$, $b = 1$ and $g = 0.25$.
The harmonic function that sets the seen temperature does not satisfy this condition on the unseen part, so analytical solutions are not available in this setting.
Instead, we use fine finite-element solutions of the same problems on the true meshes (FEniCSx \cite{barattaDOLFINxNextGeneration2023} on an fTetWild tetrahedralization \cite{huFastTetrahedralMeshing2020}).

\paragraph{Accuracy}
We report the mean absolute error of the interior temperature and of the temperature recovered on the unobserved part of the surface; \cref{fig:slices} shows, for seven scenes (all eighteen in the supplementary material), which points the camera sees, the interior slice against the reference, and the recovered unseen surface, and \cref{tab:overview} gives the numbers for the convective exchange.
The unobserved surface carries the largest error, since its temperature is inferred from the assumed condition rather than measured: $0.82$\% of the temperature range in the mean, against $0.58$\% inside.
For results of all the scenes in our datasets, please see our supplementary material.

\subsection{Comparison with a mesh-based pipeline}
\label{ssec:e2e-comparison}
\label{par:mesh-baseline}
\paragraph{Setup}
The obvious alternative to our method is assembled from existing parts: reconstruct a mesh from the images, fit the boundary data to it, and hand it to a volumetric solver.
We build that pipeline from the strongest parts we know and give it the same images.
Its mesh comes from the same RaDe-GS model as our points: following \citeauthor{zhangRaDeGSRasterizingDepth2024}~\shortcite{zhangRaDeGSRasterizingDepth2024}, we fuse the median depth maps of the $300$ training views into a truncated signed distance field.
Of the image-based meshers of \cref{fig:meshers}, this one also serves the finite-element solve best: with full coverage, FEM on the meshes of MILo \cite{guedonMILoMeshIntheLoopGaussian2025} and TopoSurfel \cite{fan2026toposurfel} has $1.8$ and $1.9$ times its error in the mean over the eighteen scenes.
We use PyMeshLab \cite{cignoniMeshLab2025072025} to repair the surface mesh (duplicate and unreferenced elements, non-manifold edges and vertices, face orientation, holes of up to thirty edges), and fTetWild \cite{huFastTetrahedralMeshing2020} for the tetrahedralization (target edge length \mbox{$3\%$} of the bounding box diagonal).
We prescribe the boundary data in the same fashion as our pipeline.
The seen region is transferred from our cloud to the mesh faces, each seen vertex reads its temperature from the same thermal views, and the unseen faces receive the same boundary conditions.
Then, we use FEniCSx \cite{barattaDOLFINxNextGeneration2023} to solve the Laplace problem with linear elements; \cref{sec:mesh-pipeline} lists the other parameters.

\noindent\begin{minipage}[c]{\dimexpr\linewidth-118pt-6pt\relax}
\paragraph{Timing}
In time both are dominated by the reconstruction, the same $5$ minutes of RaDe-GS training; what follows takes $31$\,s for ours and $210$\,s for the mesh pipeline on \textsc{Armadillo} (inset).
\end{minipage}\hspace{6pt}\begin{minipage}[c]{118pt}\includegraphics{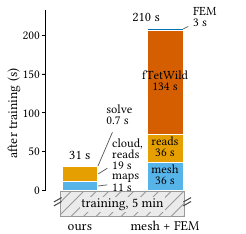}\end{minipage}\par
\noindent
Most of the mesh pipeline's time goes into the tetrahedralization, $134$\,s here and $78$ to $518$\,s across the scenes of \cref{tab:overview}, while our solver takes under a second; the run time of the tetrahedralization also grows quickly as its elements shrink (\cref{sec:mesh-pipeline}).
Since our solve needs no mesh, the inverse problems of \cref{ssec:inverse-problems} can reuse the primitives as they move, whereas a mesh pipeline would re-tetrahedralize at every step even if the change in geometry is small.

\paragraph{Accuracy}
The error is on par or lower (\cref{tab:overview}): with full coverage ours is lower on seventeen of the eighteen scenes, and under one view on fifteen inside and on all eighteen on the unobserved surface.
The errors are small fractions of the temperature range because the reference fields are smooth. On the solids, both pipelines stay below $0.2\%$ of the range with full coverage, so there the time matters more than the difference in error; on the strut lattices and \textsc{Rind}, the mesh pipeline's error is two to four times ours.
In the incomplete thermal capture setting, the errors grow, on the unobserved surface up to $1.9\%$ of the range for ours and $2.6\%$ for the mesh pipeline.
With full coverage the difference lies in the boundary data: given the exact field at its boundary nodes, FEM on the same meshes leaves at most $21\%$ of its error. A mesh needs a temperature at every vertex, also where the views see the surface only at a depth discontinuity, whose pixels mix a strut with what lies behind it.
A boundary element solve on these meshes skips the tetrahedralization but not the meshing, and inherits the same errors (\cref{sec:mesh-bem}).
\Cref{fig:e2e-comp} shows both pipelines on two scenes, with full coverage and under one view.
For comparison results of all the scenes in our datasets, please see our supplementary material.

\subsection{Real capture}
\label{ssec:real-capture}
\begin{figure}[t]
    \centering
    \input{figures/capture/capture-snippet}
    \caption{Real capture setup.
    (a) A rendering of the rig, the heated teapot on the hot plate of a turntable and the infrared camera on a rotating arm, with the $360$ camera positions over the hemisphere as COLMAP registers them (blue); (b) photo of the setup; (c) a registered pair, infrared at lower left and visible at upper right.
    }
    \label{fig:capture-setup}
\end{figure}

\input{figures/teapot/teapot-snippet}
\input{figures/pipe/pipe-snippet}
\paragraph{Capture setup}
We built a custom capture rig, a turntable and a rotating arm that carries a FLIR T560, which records a thermal and a visible-spectrum image from each of $360$ positions over the hemisphere (\cref{fig:capture-setup}).
Camera poses are recovered from the visible-spectrum images with COLMAP \cite{schonbergerStructurefromMotionRevisited2016}, the thermal images inherit the pose of the visible-spectrum image they were captured with, and object masks from Segment Anything V3 \cite{carionSAM3Segment2026} remove floaters during training.
We captured a solid cast plaster \textsc{Teapot} that carries three temperature probe inserts cast into its body, on the axis at $2.5$, $5$ and $7.5$\,cm above the base of the $10.5$\,cm object.
It was captured twice on the same rig, at $100$ and $150$\,W of hot-plate power, each capture $360$ image pairs at thermal steady state.
More details of the fabrication, capture and data processing of the teapot can be found in \cref{sec:fabrication}.

\paragraph{Boundary conditions}

Here, two parts of the teapot's boundary conditions are unknown: the temperature of the hot plate under it, which no camera sees, and the Robin coefficient with which its surface exchanges heat with the room.
We fit the first to the thermal images of each capture, calibrate the second once against the probes of one capture, and predict the interior of the other, whose probes check the prediction; the capture thus tests the whole chain, from the images to the interior.

First, we train RaDe-GS on the visible-spectrum images and extract the point cloud.
Each point reads its temperature from the thermal images, fused over the views that see it as on the synthetic scenes, after the camera's calibrated Planck-type model is inverted in each thermal image (\cref{sec:fabrication}).
Since there is no camera that can see the plate contact, we sample points on the cloud's convex hull close the bottom.
Then, the problem is a Robin exchange $u + b\,\partial u / \partial \bn = T_\mathrm{amb}$ with the room at $22$\,\textdegree C on the imaged surface, and a Dirichlet condition with one unknown constant temperature, the plate contact temperature, on the filled bottom.

\paragraph{Calibration on one heating power}
We calibrate the Robin coefficient $b$ on one capture against its three probes $\by_p$, with sensor readings $\hat{T}_p$, by minimizing the mean squared error at the probes,
\begin{align}
\label{eq:teapot-calibration}
b^\ast &= \operatorname*{arg\,min}_{b > 0}\; \frac{1}{3}\sum_{p=1}^{3} \big(u_b(\by_p) - \hat{T}_p\big)^2, \qquad u_b = u^0_b + c^\ast(b)\,u^1_b,\\
\label{eq:teapot-plate}
c^\ast(b) &= \operatorname*{arg\,min}_{c}\; \sum_{i} \big(u^0_b(\bx_i) + c\,u^1_b(\bx_i) - T_i\big)^2,
\end{align}
with $u^0_b$ and $u^1_b$ the solved temperatures for the exchange alone and for a unit plate temperature, and $T_i$ the read at the imaged point $\bx_i$: for each $b$ the plate temperature $c^\ast(b)$ is the least-squares fit of the solved surface temperature to the reads, in closed form, and we minimize over $b$ by gradient descent, L-BFGS on $\log b$ with the gradient from the adjoint of the solve.
The coefficients of the two captures differ slightly: the $150$\,W capture is fitted at $b^\ast = 7.4$\,cm with a mean absolute probe error of $0.38$\,\textdegree C on a $41$--$64$\,\textdegree C range, the $100$\,W capture at $6.6$\,cm with $0.23$\,\textdegree C on $28$--$37$\,\textdegree C (\cref{fig:teapot}, left).
We suspect the reason for the mismatch of the coefficient is that the hot plate also warms the air around the object, more so at $150$\,W, while the model assumes the ambient to be at the room's $22$\,\textdegree C: an ambient about $2$\,\textdegree C warmer at $150$\,W would explain the two coefficients.

\paragraph{Prediction on the other heating power}
The calibrated $b^\ast$ is then deployed unchanged on the other capture, whose images provide its own surface reads and its own plate constant.
Calibrated at $150$\,W and deployed at $100$\,W, the probes are predicted to $0.43$\,\textdegree C on average; in the other direction to $1.41$\,\textdegree C; see \cref{fig:teapot}.

\subsection{Inverse problems}
\label{ssec:inverse-problems}
\begin{figure*}[t]
\centering
\ifdefined\facadeSetupWidth\else\newlength{\facadeSetupWidth}\fi\setlength{\facadeSetupWidth}{3.15in}%
\setlength{\tabcolsep}{1pt}
\begin{tabular}{@{}c@{\hspace{5pt}}c@{}}
\raisebox{-\height}{\input{figures/facade3/setup-overlay}} &
\raisebox{-\height}{%
\begin{tabular}{@{}cccc@{}}
\raisebox{-0.5\height}{\includegraphics{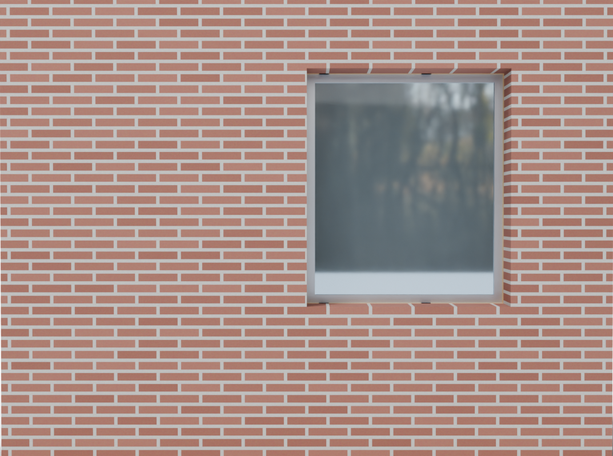}} & \raisebox{-0.5\height}{\includegraphics{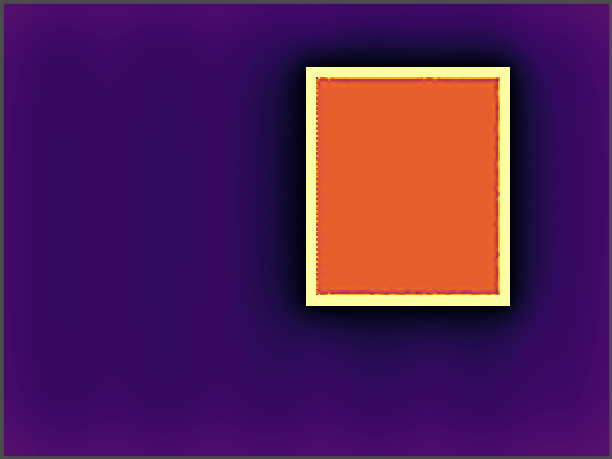}} & \raisebox{-0.5\height}{\includegraphics{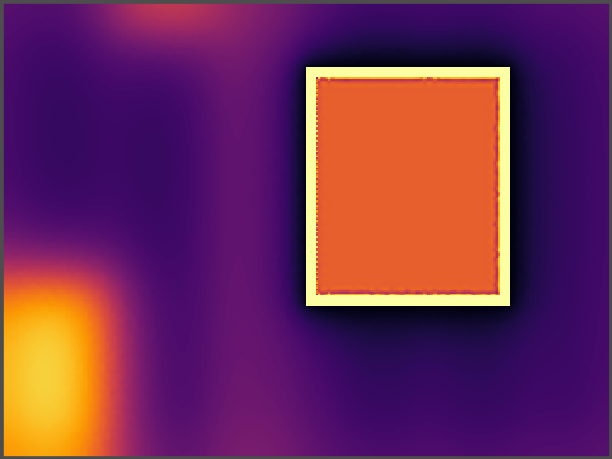}} & \raisebox{-0.5\height}{\includegraphics{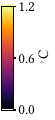}} \\
\footnotesize RGB photo & \footnotesize thermal image, intact & \footnotesize thermal image, defects & \\[3pt]
\raisebox{-0.5\height}{\includegraphics{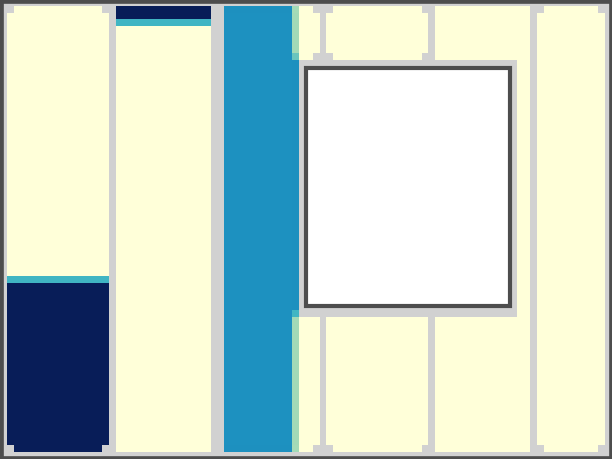}} & \raisebox{-0.5\height}{\includegraphics{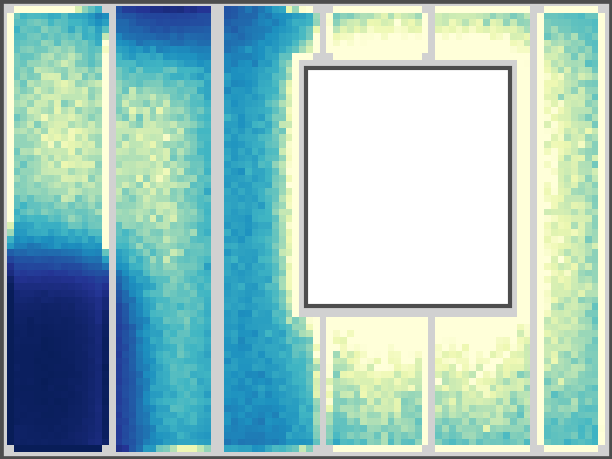}} & \raisebox{-0.5\height}{\includegraphics{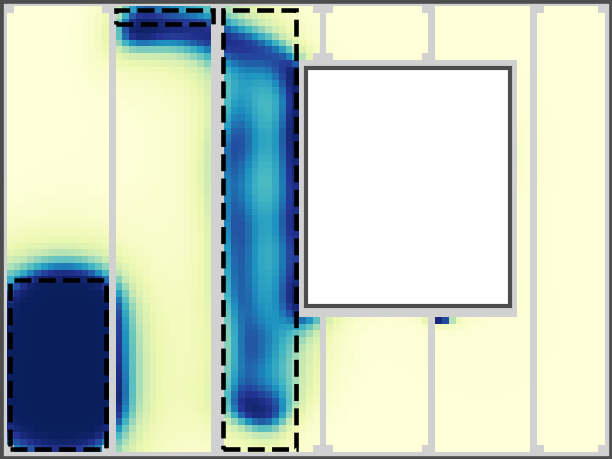}} & \raisebox{-0.5\height}{\includegraphics{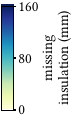}} \\
\parbox[t]{1.02in}{\centering\footnotesize true} & \parbox[t]{1.02in}{\centering\footnotesize per-pixel inversion\\\scriptsize without conduction} & \parbox[t]{1.02in}{\centering\footnotesize ours\\\scriptsize with conduction} & \\
\end{tabular}}
\end{tabular}
\caption{%
Insulation inspection.
We use thermal images taken from the street to find insulation defects hidden behind a brick wall, shown on the left with the wall cut away over them.
The RGB photo does not show them, and in the thermal images the defective wall differs from the intact one only faintly, because the brick blurs their signature.
We recover the insulation thickness behind the wall using our optimization algorithm, which models the conduction along the brick and recovers the panels close to the truth, while a per-pixel inversion without lateral conduction \cite{wangNewMethodPixellevel2023} interprets the blur as missing insulation.}
\label{fig:facade}
\end{figure*}
\paragraph{Heating pipe inspection}
\label{ssec:pipe}
The first application looks through a floor (\cref{fig:pipe}), after the radiant-heating mock-up of \citet{shinDesignRadiantFloor2015}: two counterflow spirals of PE-X pipe in a $60\,\mathrm{mm}$ screed, $30\,\mathrm{mm}$ below the floor, in a room of \mbox{$2.88 \times 5.52\,\mathrm{m}$} with their water at $32\,^\circ$C, room at $20\,^\circ$C.
We use 40 visible-spectrum and thermal images of the room.
The visible spectrum images are used to train Gaussian splats and to extract point cloud.
And the thermal views rendered from a finite-element solution, and are fused onto the extracted points.
We consider the floor and the pipe wall both Robin boundaries.
The floor exchanges heat with the room through the room's heat transfer coefficient.
The pipe's surface exchanges heat with the water through the $2\,\mathrm{mm}$ wall; the conductivity of PE-X divided by the wall's thickness takes the place of a heat transfer coefficient.
The unseen sides and the underside are assumed insulated.
With the shape of the loop known (from the construction drawings), we recover its pose $\theta$, including position, depth and rotation, from the floor reading alone: L-BFGS through the differentiable solve minimizes
\begin{gather}
\label{eq:pipe-loss}
\mathcal{L}(\theta) = \frac{1}{N}\sum_{j=1}^{N} \big(u_j(\theta) - T_j\big)^2,
\end{gather}
the mean squared difference between the solved floor temperatures $u_j$ and the reads $T_j$ at the $N$ floor points, every trial pose keeping the whole loop inside the screed.
The optimization converges in $24$ solves ($20$\,s) and the final configuration is within $3.7$\,mm of the true depth and $0.02^\circ$ of its rotation.

\input{figures/crack-free/crack-free-snippet}

\paragraph{Insulation inspection}
\label{ssec:facade}

This application looks for insulation defects behind a building facade (\cref{fig:facade}).
The facade consists of a $240$~mm brick wall with a window. Within the facade is a $160$~mm timber frame filled with wood-fiber panels that provide insulation.
There are three defects in the insulation panels:
a panel destroyed at its lower part,
a fill settled by \mbox{$80$~mm} under the top plate,
and a $60$~mm thick panel where it should be $160$~mm thick.
The wall is seen from the street only, the passive setting of \citeauthor{larbiyoucefInSituQuantitative2020}~\shortcite{larbiyoucefInSituQuantitative2020}, in $100$ photographs and $100$ thermal images rendered from a finite-element solution of the layered wall.
The wall is layered, and our Laplace model cannot directly model different materials in the same system.
So we consider the brick wall alone as the problem domain and model the frame behind the bricks as a boundary condition.
Following ISO 6946~\cite{iso6946}, a thin layer adds its thermal resistance to that of the surface, so each cell of the brick's internal face exchanges heat with the inside of the building through a Robin condition whose coefficient is the inverse of the resistance in front of it.
This thermal resistance combines the known timber fraction and a panel of unknown thickness $t$ as parallel paths, each in series with the plaster and the internal surface resistance.
The external face exchanges heat with the outdoor air through the external surface resistance, which combines convection and radiation together.
The brick's external face is extracted from RaDe-GS on the photographs and projected onto locally fitted planes because the reconstruction scatters a flat face by $13$~mm.
The camera sees nothing else of the brick, so we assume its other faces flat and place them from the wall's dimensions, which an inspector could take from the building's plans.
Like the external face, these faces are represented by sampling collocation points, without splats or a mesh.
The points of the internal side are the cells whose panel thickness $t$ we fit.
The conduction in the brick diffuses the thermal defect in the thermal image, so the destroyed panel warms the external face by only $0.75\,^\circ$C and the thin panel by $0.15$ (\cref{fig:facade}, top).
We fit $t$ per cell by minimizing, with Adam through the solve,
\begin{align}
\label{eq:facade-loss}
\mathcal{L}(t) = \frac{1}{N}\sum_{i=1}^{N} \big(u_i(t) - T_i\big)^2,
\end{align}
the mean squared difference between the solved temperatures $u_i$ at the $N$ points of the street face and the reads $T_i$.
Where the frame's timber posts fill a cell, its panel thickness barely affects the solve, so we leave such cells out of the map and of the error.
For comparison, we use the per-pixel model from quantitative thermography.
\citet{wangNewMethodPixellevel2023} state it for thermography from outdoors: a one-dimensional steady model of the heat crossing the wall maps its thermal transmittance pixel by pixel from the outdoor surface temperature.
Our reference solves the same model at each pixel, but for the panel thickness, since the other layers are known;
this per-pixel inversion method lacks the lateral conduction in the wall and reads the blur as insulation, so it underestimates the intact wall and overestimates the thin panel (\cref{fig:facade}, bottom).
Our solve through the brick undoes the blur and recovers the footprints and thicknesses of the panels, with a mean thickness error of $14$~mm against the per-pixel inversion's $43$~mm, without a thermal image of an intact wall.

\paragraph{Crack detection}
\label{ssec:crack}
Active thermography is a common non-de\-struc\-tive defect detection method~\cite{maldagueTheoryPracticeInfrared2001}. In this application, we locate a sub-surface defect that is not visible in RGB images (\cref{fig:crack-free}), using steady-state thermal images.
A propeller from the HexMe collection~\cite{beaufortHexMeIfYou2022}, scaled to unit length, carries a thin crescent-shaped crack under the surface of its hub wall; the shaft heats the bore to $207$~K above the ambient, the outer surface cools into the ambient, and the crack conducts no heat.
In units of that temperature difference, with the ambient at $0$ and lengths in units of the propeller, the bore is a Dirichlet boundary at $u = 1$, the outer surface a Robin boundary with $\partial u / \partial \bn = -3u$, and the crack an internal Neumann boundary.
In thermal views rendered from a finite-element solution of the cut solid, the hub surface above the crack runs colder, by up to $31$~K in the view of \cref{fig:crack-free}, top.
The crack changes nothing outside, so one RGB reconstruction of the intact part serves both parts; it comes from a textured capture, since the reconstruction of an untextured capture sinks the hub deeper than the crack lies.
The outer surface is extracted as in \cref{par:extraction}, and the bore comes from the drawing.

We fit what the crack changes, as an inspector compares a part with an intact one.
We image an intact part the same way, and the difference between our solves with and without the crack should match the difference between the thermal images of the cracked and the intact part.
The model's error then cancels to first order, and the true crack is a clear minimum, $0.24$~K against $1.58$~K with no crack.
There is still an error of scale, though: on the reconstructed part, the solved difference for the true crack is slightly weaker than the measured one, mostly because about $1$\% of the reconstructed points lie about a spacing inside the wall, and the solve treats them as surface.
We therefore compare the two differences up to a scale, so that the fit matches the pattern of the difference, which locates the crack, and not its strength.

The crack's shape is unknown, and we recover it with a \emph{Gaussian splat flow}.
We represent the crack in the same way as we represent the outer surface, with Gaussian splats, but with a set of its own inside the part, which the flow moves while the reconstruction of the outer surface stays fixed.
Each splat has its own position, orientation and size, and the crack's Neumann points come from the extraction of \cref{par:extraction} applied to renders of these splats from virtual cameras around the crack; since the extraction makes every point a weighted average of the surface points of its pixels, recomputing that average with the selection of pixels held fixed carries the gradient of the solve back to the splats.
The flow minimizes the root mean square (RMS) of the differential mismatch above over the observed points $i$ of the outer surface, plus a barrier,
\begin{align}
\label{eq:crack-loss}
\mathcal{L} = \Big(\frac{1}{N}\sum_{i=1}^{N} \big(\alpha\,(u_i - u^0_i) - (T_i - T^0_i)\big)^2\Big)^{1/2} + B,
\end{align}
with $u_i$ and $u^0_i$ the solved temperatures at point $i$ with and without the crack, $T_i$ and $T^0_i$ the reads fused at that point from the thermal images of the cracked and the intact part,
$\alpha$ the amplitude, the least-squares scale that best maps the solved difference onto the measured one, recomputed in closed form at every step, since it changes as the crack takes shape, and kept between \mbox{$0.7$} and \mbox{$2$}, so that a crack far too small or too large cannot be scaled to the measured strength,
and $B$ is a log barrier that keeps the splats inside the hub wall, \mbox{$B = \eta \sum_{w} \big(s_w - 1 - \log s_w\big)$}, with $\eta = 10^{-4}$: for each wall \mbox{$w$} of the hub, its outer surface, its ends and its bore, \mbox{$s_w$} is the smallest clearance of a splat center from \mbox{$w$} beyond a margin of \mbox{$\varepsilon$}, in units of \mbox{$\varepsilon/4$} and capped at \mbox{$1$}, so that the term vanishes away from the walls.
Both terms are needed.
With $\alpha$ fixed at $1$ the flow stalls at a footprint overlap of $0.39$.
If left free, $\alpha$ stays at its bound of $2$ until the crack has taken shape and then settles at $1.5$, the factor of the true crack.
Without the wall barrier the flow stalls early:
once a splat center reaches the margin of the wall, every further step is rejected, while the barrier turns the descent away from the wall before that.
Another strategy we use to make the optimization stable is smoothing the gradient, as we notice independent gradient steps tear the thin cavity apart.
Following \citeauthor{ecksteinGeneralizedSurfaceFlows2007}~\shortcite{ecksteinGeneralizedSurfaceFlows2007}, we smooth the gradient by changing the inner product of the flow: each step moves all splats by one displacement field, the gradient convolved with a Gaussian whose width shrinks from $0.03$ to $0.012$ as the flow stalls, and carries every splat's frame and size along; following the splitting strategy in the original 3D Gaussian splatting algorithm by \citeauthor{kerbl3DGaussianSplatting2023a}~\shortcite{kerbl3DGaussianSplatting2023a}, splats stretched too far are split.
From a disk of diameter $0.03$ beside the crack, sharing no material with it, the RMS mismatch falls from $1.52$ to $0.21$~K in $178$ steps (\cref{fig:splat-flow}); the centroid lands within $0.0023$ of the truth with a footprint overlap of $0.77$.
The shrinking width matters: with the width fixed at $0.012$, the same disk stalls with an overlap of $0.20$.
The mock-up is steady and its coefficients are known; it shows that a defect invisible from outside is recovered, shape included, from thermal images once a reference part removes the reconstruction's own bias.

\subsection{Convergence and ablations}
\input{figures/convergence-seq/convseq-snippet}
\label{ssec:density}
\label{ssec:scalability}
\label{ssec:eps-study}

\paragraph{Convergence}
Under refinement the full-coverage error converges at about first order in the spacing on clean points sampled from the true mesh, on every scene (orders $0.9$ to $1.7$).
The extracted cloud follows the clean points until the images run out of information and then levels off on a scene-specific shelf (\cref{fig:capture-convergence}, \textsc{Hand}), lowest on the solids ($0.071$\% of the temperature range against $0.045$\% on clean points on \textsc{Armadillo}) and highest on the thin structures, up to $0.44$\% on \textsc{Kangaroo}.

\paragraph{The regularization radius}
The radius $\varepsilon$ of the mollified kernels is the one solver parameter. In \cref{fig:eps-sweep} (left), we sweep it from $0.05h$ to $6h$ on the eighteen scenes.
The direct formulation is flat around $h$ (the spacing in the point cloud): with full coverage the mean of the scenes' relative errors from $0.3h$ to $1.5h$ is within $1\%$ of that at $h$, and under one view $h$ is the best radius or within $2\%$ of it on sixteen of the eighteen scenes, so we use that in our method.

\input{figures/eps-sweep-snippet}

\paragraph{The frontality threshold}
The frontality test and the depth-slope filter share one threshold, $70^\circ$ from the normal; \cref{fig:front-sweep} (right) sweeps it from $37^\circ$ to $87^\circ$.
With full coverage, where every point is also seen frontally, stricter thresholds lower the mean error by $9\%$; under one view they leave the obliquely seen surface to the neighbor fill and double the error at $49^\circ$.
One threshold serves both settings within $10\%$ of either optimum.

\begin{figure}[h]
\centering
\setlength{\tabcolsep}{1pt}%
\newcommand{\bclab}[2]{\footnotesize\begin{tabular}[t]{@{}c@{}}#1\\$#2$\end{tabular}}%
$\vcenter{\hbox{\begin{tabular}{@{}c@{}}
\begin{tabular}{@{}cc@{}}
\includegraphics[width=0.27\linewidth]{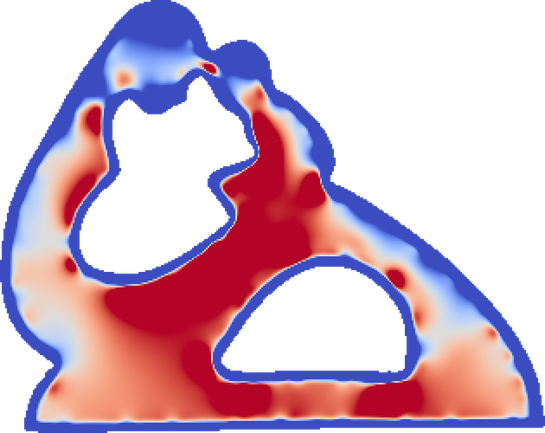} & \includegraphics[width=0.27\linewidth]{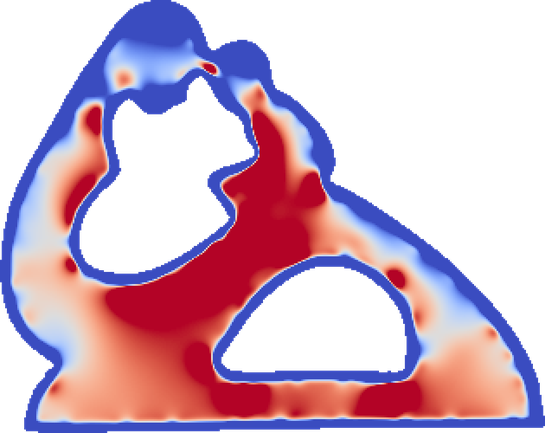} \\
\bclab{double layer}{0.056} & \bclab{direct}{0.056} \\
\end{tabular}\\[2pt]
\begin{tabular}{@{}ccc@{}}
\includegraphics[width=0.27\linewidth]{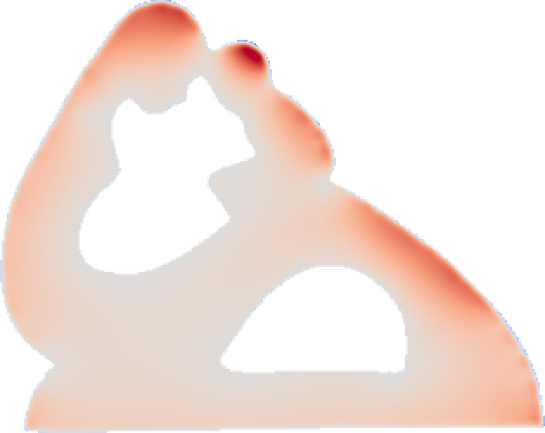} & \includegraphics[width=0.27\linewidth]{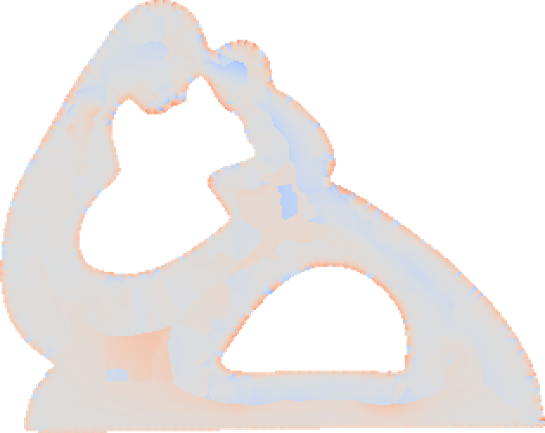} & \includegraphics[width=0.27\linewidth]{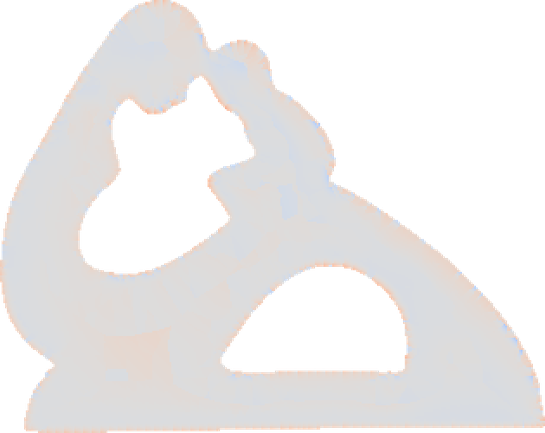} \\
\bclab{single layer}{0.0017} & \bclab{double layer + subtr.}{0.00049} & \bclab{direct + subtr. (ours)}{0.00028} \\
\end{tabular}
\end{tabular}}}$\hspace{3pt}%
$\vcenter{\hbox{\includegraphics{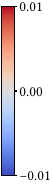}}}$
\caption{
BIE formulations on the extracted cloud of \textsc{Fertility}, full coverage. We show the error maps of 5 different BIE formulations in the same setup.}
\label{fig:bie-comp}
\end{figure}

\paragraph{Leave-one-out}
\label{ssec:abl-leave-one-out}
\input{tables/tab-ablation}
\Cref{tab:ablation} removes one component at a time, on all eighteen scenes and in both settings; removing any of the components raises the mean error in both.
Two components are not optional.
The first is the density subtraction as in \cref{eq:gs-eval-solutions}. Without it, the interior error is 21 times larger with full coverage and 8 times under one view; this can also be seen in \cref{fig:bie-comp}.
The second is the damping of the full-coverage system. Without it, the noise on the surface increases and the error nearly doubles.
The depth-slope filter is the largest of the rest with full coverage, where its absence nearly doubles the error.
Outside the table, letting a single sample make a point in the merge of \cref{par:extraction} raises the mean error by a third with full coverage ($0.17$\% to $0.22$\% of the temperature range) and by a quarter under one thermal view setting.

\paragraph{Different BIE formulations}
Our discretization applies equally to the single layer potential (SLP), the double layer potential (DLP) and the direct formulation (\cref{sec:bie-formulations}); \cref{fig:bie-comp} compares them on \textsc{Fertility} with full coverage and an analytical harmonic solution.
As \cref{sec:bie-formulations} discusses, in the smooth setting, all three represent the same temperature inside $\Omega$ (\cref{fig:bie-fields}), so the error comes from the discretization of different kernels, which here span more than two orders of magnitude.
Evaluated as plain point sums, with the mollified kernels but without the density subtraction, the DLP and the direct formulation are more than thirty times worse than the SLP ($0.056$ against $0.0017$): both evaluate the interior through the double-layer kernel $\partial G / \partial \bn \sim 1/r^2$, which the point quadrature resolves only down to the spacing, so within half a spacing of the surface their error is of the order of the field.
However, with density subtraction,
the direct formulation becomes more accurate than the SLP and the DLP, all the way to the surface.
Therefore, we use it for both the solve and the evaluation.

\paragraph{Gaussian representation}
\label{ssec:abl-representation}
The maps we extract from can come from any Gaussian reconstruction that renders depth and normals; with those of 2DGS \cite{huang2DGaussianSplatting2024} and the rest of the pipeline unchanged the mean error rises from $0.0017$ to $0.0029$ with full coverage and from $0.0058$ to $0.0078$ under one view (\cref{tab:ablation}), most on the solids and the industrial parts, while on the thin structures neither reconstruction is consistently better.
TopoSurfel \cite{fan2026toposurfel} co-evolves the surfels with a differentiable proxy mesh and fits tighter surfaces, which lowers the full-coverage error on the solids, the industrial parts and the two thin structures without struts ($0.0011$ against $0.0014$ over those fourteen scenes), but its proxy mesh closes the openings between thin struts with membranes, which leave the inner faces behind them largely without points (\cref{fig:topo-clouds}); on the four strut lattices the error grows more than fivefold, and we keep RaDe-GS and note TopoSurfel as the better front end for solids.
\input{figures/topo/topo-snippet}

%% file: figures/slices/slices-combined-snippet.tex
\begin{figure*}[t]
\centering
\setlength{\tabcolsep}{1pt}
\ifdefined\panelw\else\newlength{\panelw}\fi\setlength{\panelw}{0.074\textwidth}
\ifdefined\slicelabw\else\newlength{\slicelabw}\fi\providecommand{\slicelab}[1]{\settowidth{\slicelabw}{\scriptsize\textsc{#1}}\raisebox{\dimexpr(\panelw-\slicelabw)/2\relax}{\rotatebox{90}{\scriptsize\textsc{#1}}}}
\begin{tabular}{@{}c@{}cccc@{\hspace{4pt}}ccccccc@{}}
 & & \multicolumn{3}{c}{\makebox[\dimexpr 3\panelw+6pt\relax][c]{\hrulefill\hspace{4pt}\footnotesize full thermal coverage\hspace{4pt}\hrulefill}} & \multicolumn{7}{c}{\makebox[\dimexpr 7\panelw+14pt\relax][c]{\hrulefill\hspace{4pt}\footnotesize one thermal view, convection on the unseen surface\hspace{4pt}\hrulefill}} \\[1pt]
 & \scriptsize mesh, cut & \scriptsize reference & \scriptsize ours & \scriptsize ours $-$ ref & \scriptsize\makebox[0pt]{\textcolor[HTML]{B89CDA}{\rule{4pt}{4pt}}\,seen\enspace \textcolor[HTML]{9CC3E6}{\rule{4pt}{4pt}}\,unseen} & \scriptsize reference & \scriptsize ours & \scriptsize ours $-$ ref & \scriptsize ref.\ surface & \scriptsize ours, surface & \scriptsize ours $-$ ref \\
\slicelab{Bunny} & \includegraphics[width=0.074\textwidth]{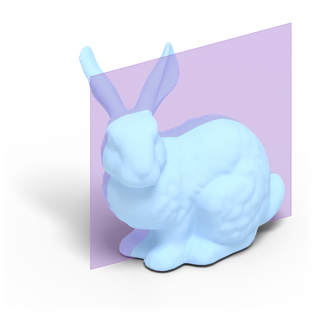} & \includegraphics[width=0.074\textwidth]{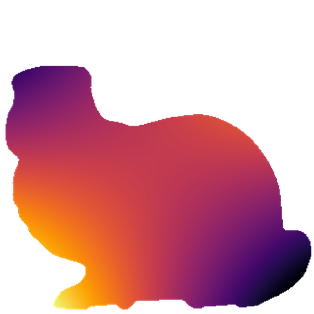} & \includegraphics[width=0.074\textwidth]{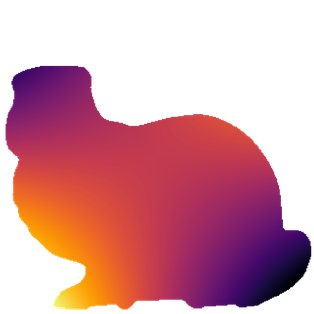} & \includegraphics[width=0.074\textwidth]{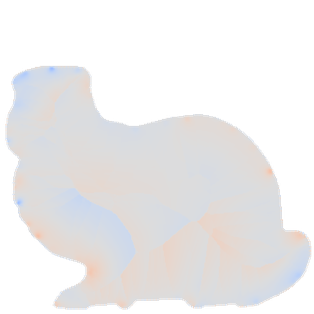} & \includegraphics[width=0.074\textwidth]{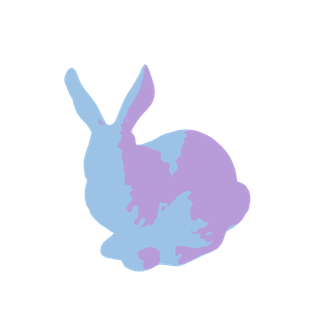} & \includegraphics[width=0.074\textwidth]{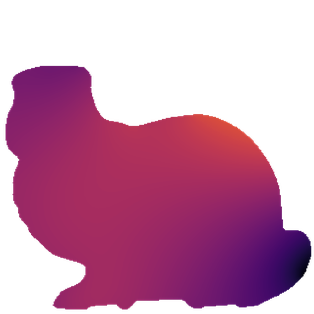} & \includegraphics[width=0.074\textwidth]{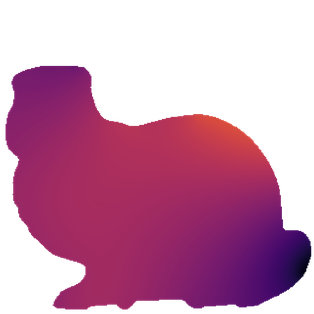} & \includegraphics[width=0.074\textwidth]{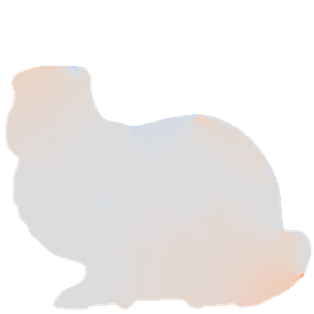} & \includegraphics[width=0.074\textwidth]{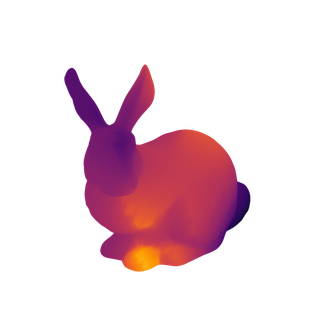} & \includegraphics[width=0.074\textwidth]{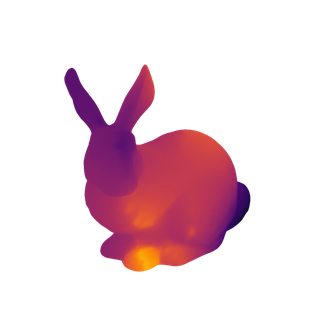} & \includegraphics[width=0.074\textwidth]{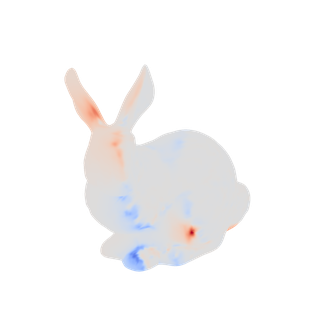} \\
\slicelab{Fertility} & \includegraphics[width=0.074\textwidth]{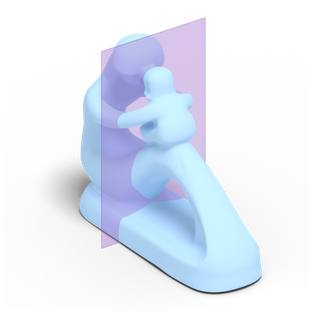} & \includegraphics[width=0.074\textwidth]{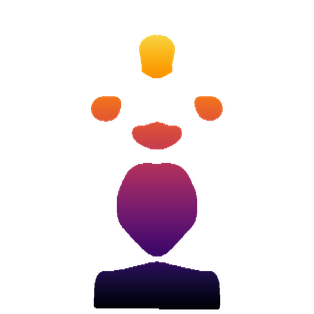} & \includegraphics[width=0.074\textwidth]{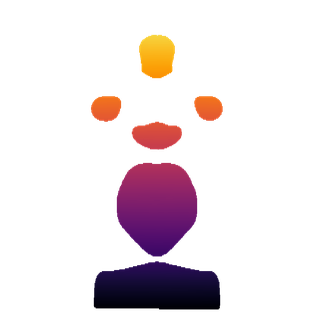} & \includegraphics[width=0.074\textwidth]{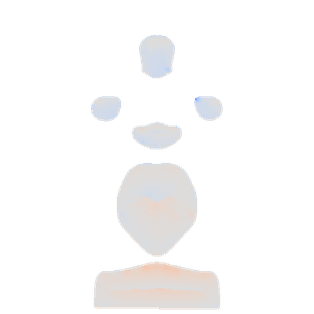} & \includegraphics[width=0.074\textwidth]{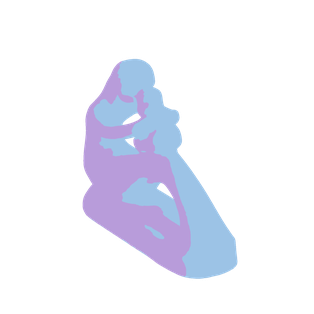} & \includegraphics[width=0.074\textwidth]{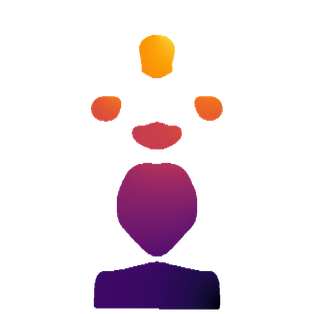} & \includegraphics[width=0.074\textwidth]{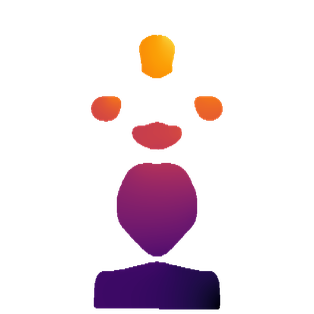} & \includegraphics[width=0.074\textwidth]{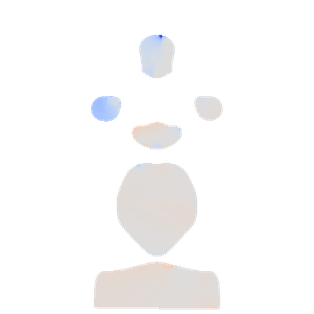} & \includegraphics[width=0.074\textwidth]{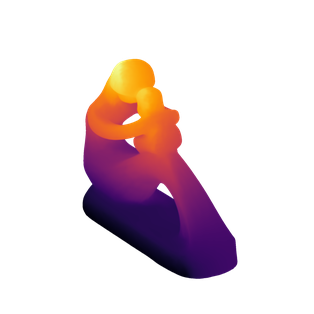} & \includegraphics[width=0.074\textwidth]{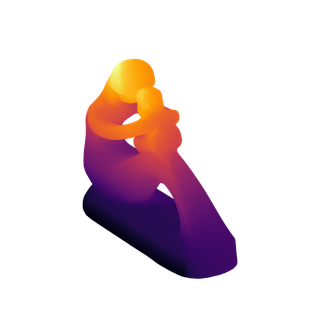} & \includegraphics[width=0.074\textwidth]{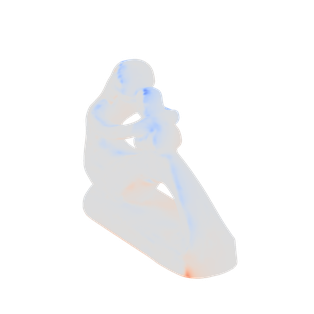} \\
\slicelab{Nefertiti} & \includegraphics[width=0.074\textwidth]{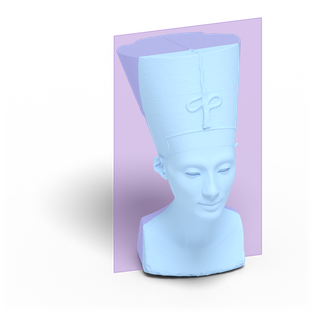} & \includegraphics[width=0.074\textwidth]{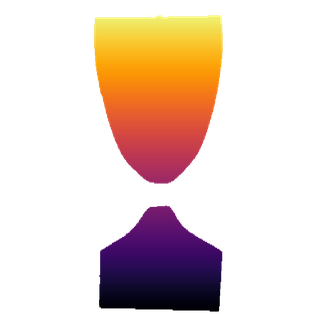} & \includegraphics[width=0.074\textwidth]{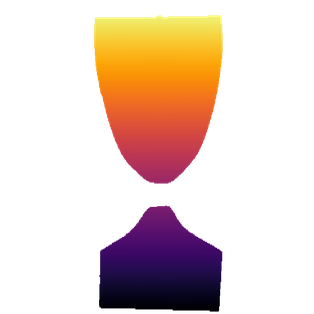} & \includegraphics[width=0.074\textwidth]{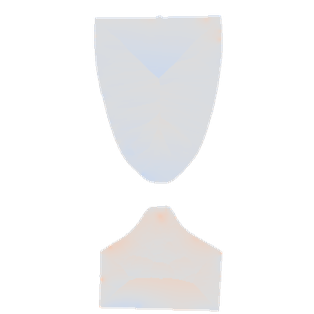} & \includegraphics[width=0.074\textwidth]{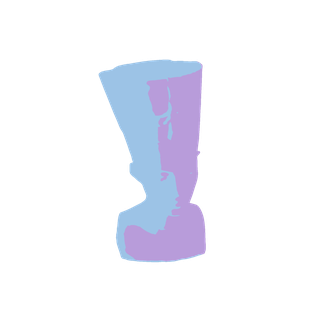} & \includegraphics[width=0.074\textwidth]{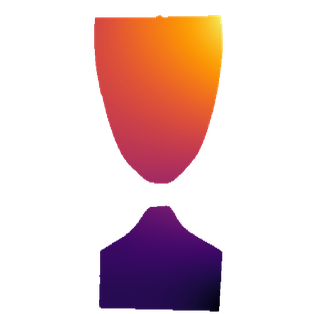} & \includegraphics[width=0.074\textwidth]{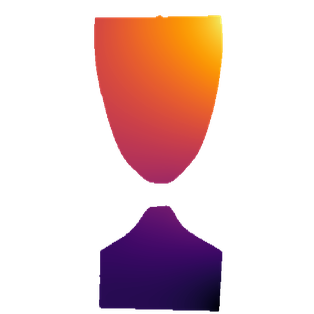} & \includegraphics[width=0.074\textwidth]{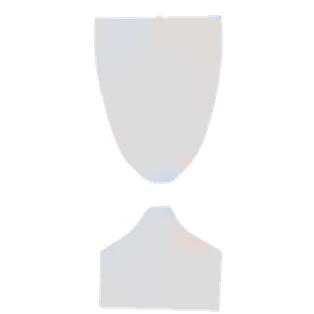} & \includegraphics[width=0.074\textwidth]{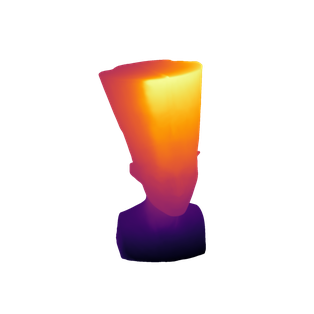} & \includegraphics[width=0.074\textwidth]{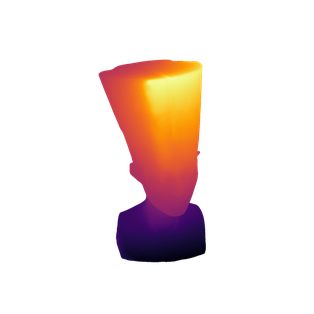} & \includegraphics[width=0.074\textwidth]{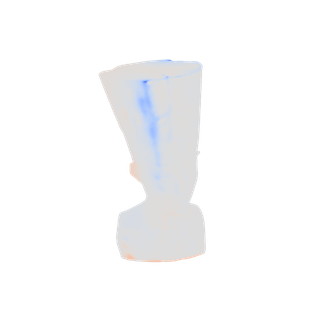} \\[3pt]
\slicelab{Rind} & \includegraphics[width=0.074\textwidth]{figures/slices/rind-mesh-sv} & \includegraphics[width=0.074\textwidth]{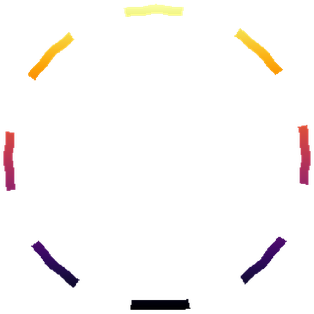} & \includegraphics[width=0.074\textwidth]{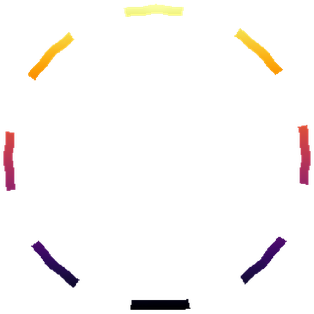} & \includegraphics[width=0.074\textwidth]{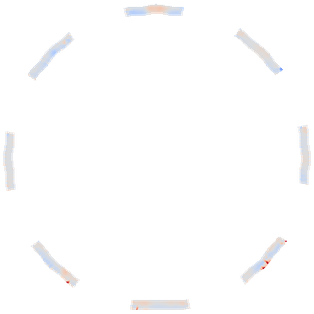} & \includegraphics[width=0.074\textwidth]{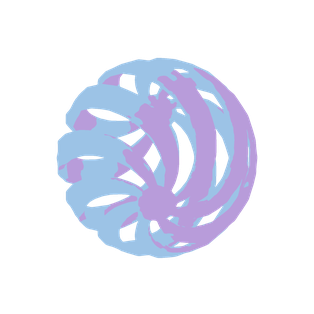} & \includegraphics[width=0.074\textwidth]{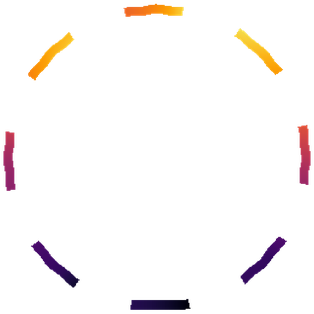} & \includegraphics[width=0.074\textwidth]{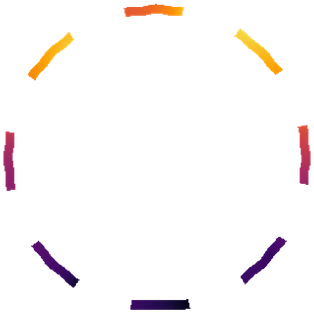} & \includegraphics[width=0.074\textwidth]{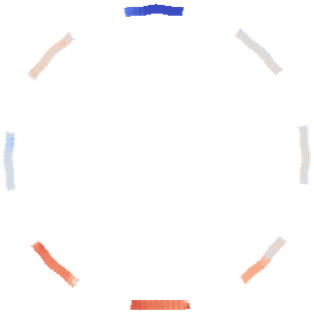} & \includegraphics[width=0.074\textwidth]{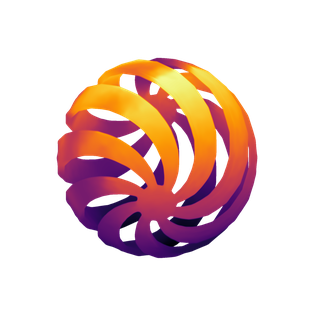} & \includegraphics[width=0.074\textwidth]{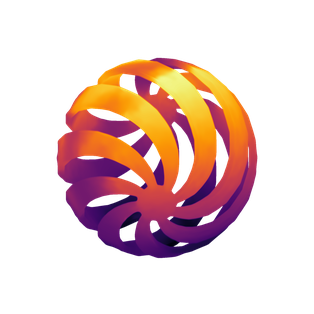} & \includegraphics[width=0.074\textwidth]{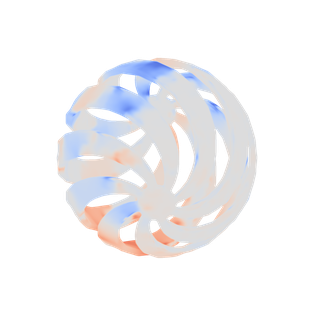} \\
\slicelab{Dino} & \includegraphics[width=0.074\textwidth]{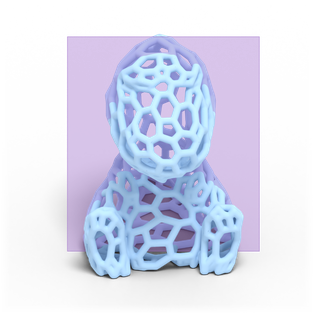} & \includegraphics[width=0.074\textwidth]{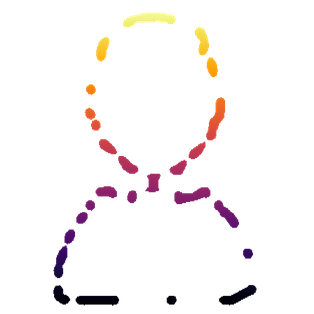} & \includegraphics[width=0.074\textwidth]{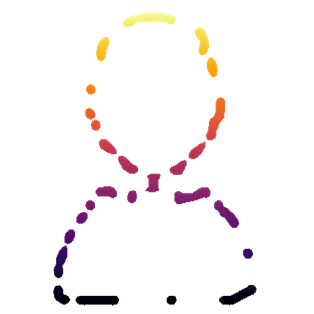} & \includegraphics[width=0.074\textwidth]{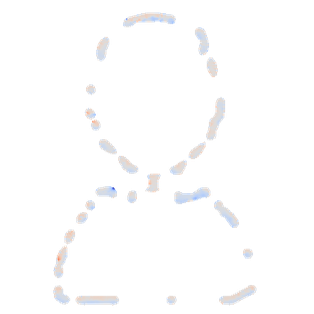} & \includegraphics[width=0.074\textwidth]{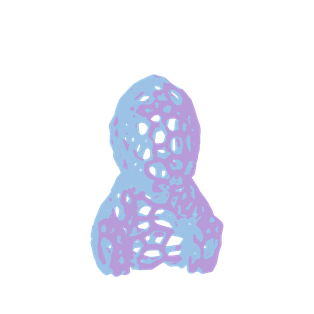} & \includegraphics[width=0.074\textwidth]{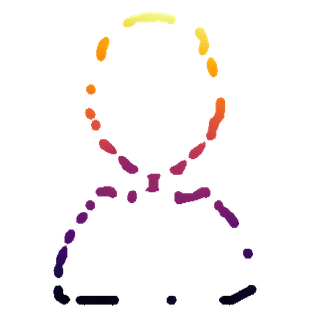} & \includegraphics[width=0.074\textwidth]{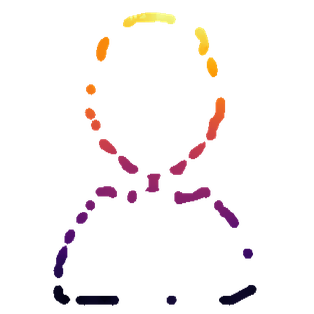} & \includegraphics[width=0.074\textwidth]{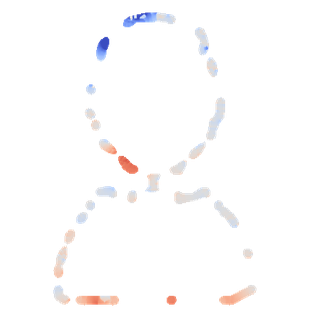} & \includegraphics[width=0.074\textwidth]{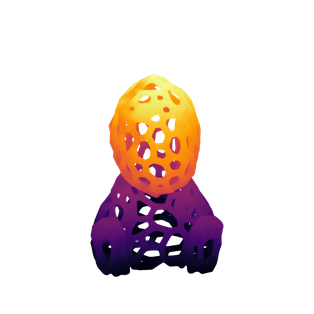} & \includegraphics[width=0.074\textwidth]{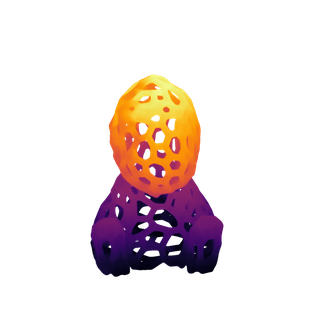} & \includegraphics[width=0.074\textwidth]{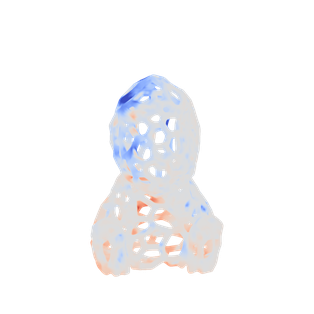} \\[3pt]
\slicelab{Hinge} & \includegraphics[width=0.074\textwidth]{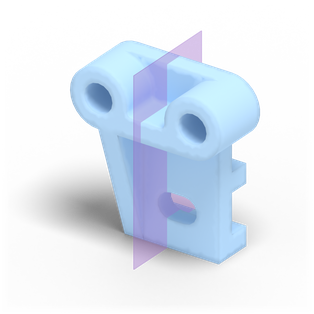} & \includegraphics[width=0.074\textwidth]{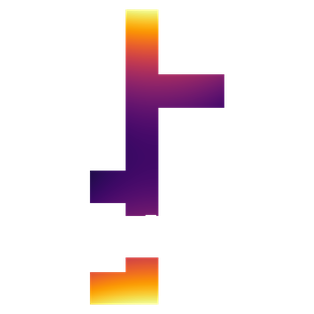} & \includegraphics[width=0.074\textwidth]{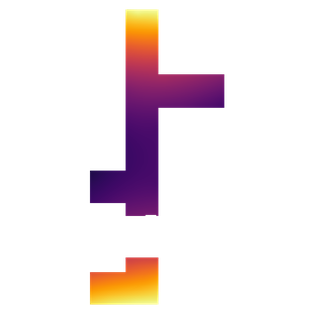} & \includegraphics[width=0.074\textwidth]{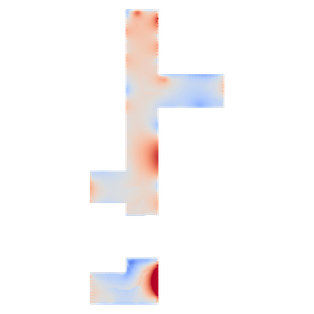} & \includegraphics[width=0.074\textwidth]{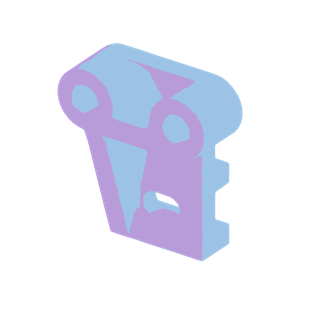} & \includegraphics[width=0.074\textwidth]{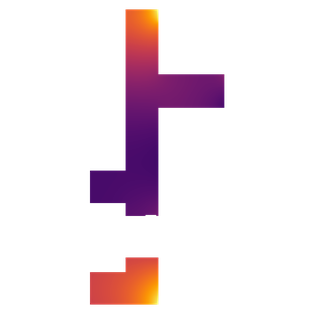} & \includegraphics[width=0.074\textwidth]{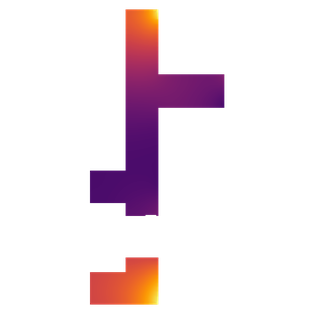} & \includegraphics[width=0.074\textwidth]{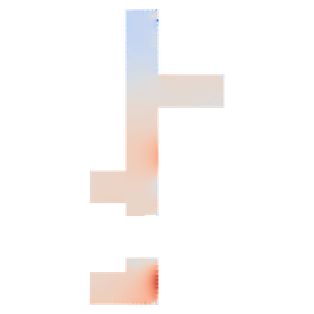} & \includegraphics[width=0.074\textwidth]{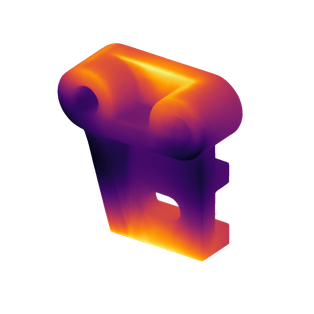} & \includegraphics[width=0.074\textwidth]{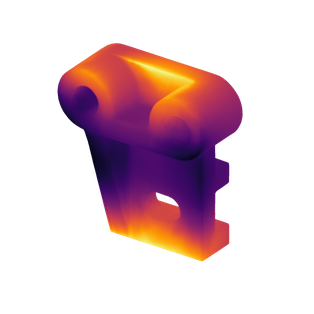} & \includegraphics[width=0.074\textwidth]{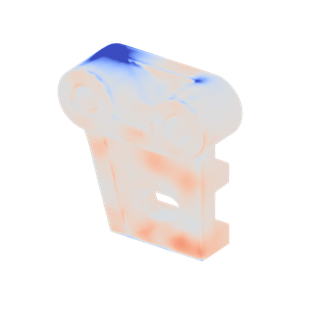} \\
\slicelab{Housing} & \includegraphics[width=0.074\textwidth]{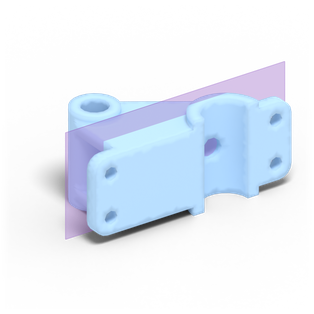} & \includegraphics[width=0.074\textwidth]{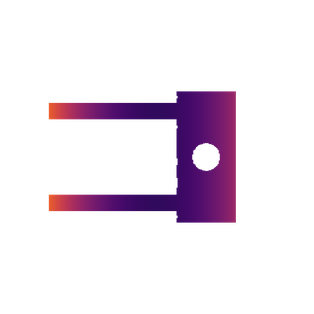} & \includegraphics[width=0.074\textwidth]{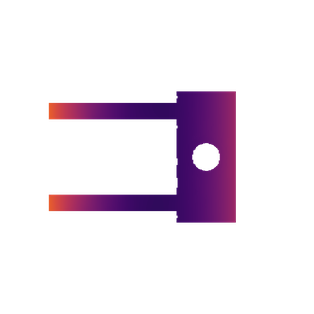} & \includegraphics[width=0.074\textwidth]{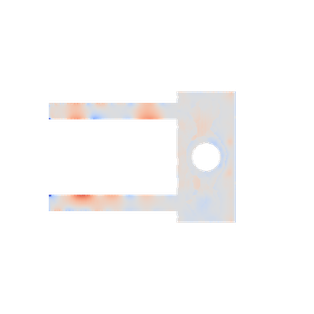} & \includegraphics[width=0.074\textwidth]{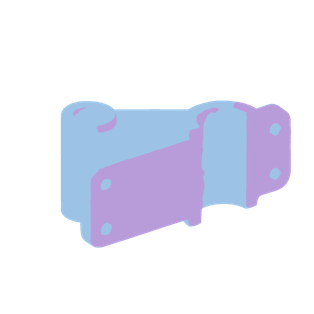} & \includegraphics[width=0.074\textwidth]{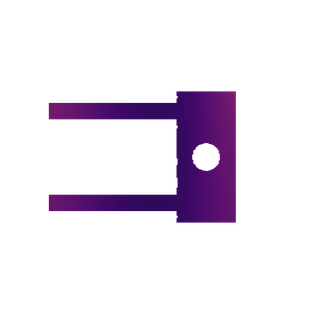} & \includegraphics[width=0.074\textwidth]{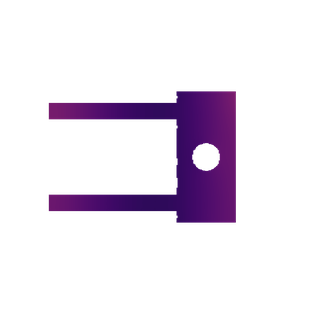} & \includegraphics[width=0.074\textwidth]{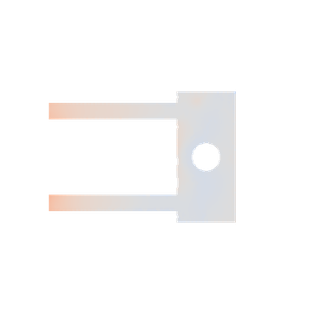} & \includegraphics[width=0.074\textwidth]{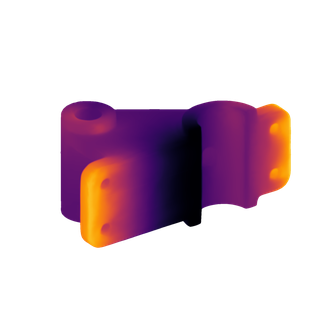} & \includegraphics[width=0.074\textwidth]{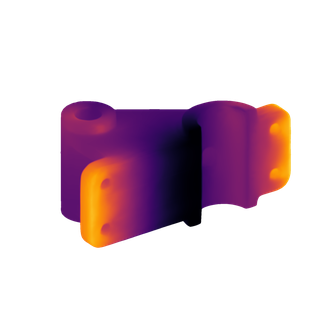} & \includegraphics[width=0.074\textwidth]{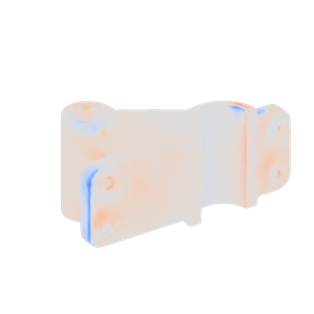} \\
 &  & \multicolumn{2}{c}{\includegraphics{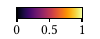}} & \includegraphics{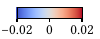} &  & \multicolumn{2}{c}{\includegraphics{figures/slices/shared-temperature}} & \includegraphics{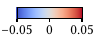} & \multicolumn{2}{c}{\includegraphics{figures/slices/shared-temperature}} & \includegraphics{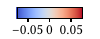} \\
\end{tabular}
\caption{%
Seven of the eighteen scenes under both protocols: three solids, two thin structures and two industrial parts. Leftmost: the true mesh and the slice plane (purple). Full thermal coverage: reference, ours and signed error on the slice. One thermal view: the seen (purple) and unseen (blue) points, then reference, ours and error inside and on the surface, where the blue region is the one that counts. \Cref{tab:overview} gives the numbers. For the rest of the scenes, please see our supplementary material.
}
\label{fig:slices}
\end{figure*}

%% file: tables/tab-overview.tex
\begin{table}
\centering
\footnotesize
\setlength{\tabcolsep}{3pt}
\caption{Mean absolute error per scene, ours against the mesh-based pipeline (\emph{mesh}, \cref{par:mesh-baseline}), the better of each pair in bold. Full coverage: interior error against the analytic field. Single view: one thermal view with convective exchange elsewhere, error on the unobserved surface and in the interior against a FEM solve on the true mesh.}
\label{tab:overview}
\begin{tabular}{@{}lrrrrrr@{}}
\toprule
 & \multicolumn{2}{c}{full coverage} & \multicolumn{4}{c}{single view} \\
\cmidrule(lr){2-3} \cmidrule(lr){4-7}
 & \multicolumn{2}{c}{interior} & \multicolumn{2}{c}{\begin{tabular}[b]{@{}c@{}}surface\end{tabular}} & \multicolumn{2}{c}{interior} \\
\cmidrule(lr){2-3} \cmidrule(lr){4-5} \cmidrule(lr){6-7}
scene & ours & mesh & ours & mesh & ours & mesh \\
\midrule
\textsc{Armadillo} & \textbf{0.00073} & 0.00083 & \textbf{0.00647} & 0.01057 & \textbf{0.00450} & 0.00622 \\
\textsc{Botijo} & \textbf{0.00105} & 0.00172 & \textbf{0.00997} & 0.01493 & \textbf{0.00453} & 0.00923 \\
\textsc{Bunny} & \textbf{0.00062} & 0.00082 & \textbf{0.00516} & 0.00758 & \textbf{0.00208} & 0.00377 \\
\textsc{Hand} & \textbf{0.00083} & 0.00095 & \textbf{0.00465} & 0.00971 & \textbf{0.00173} & 0.00471 \\
\textsc{Fertility} & \textbf{0.00078} & 0.00131 & \textbf{0.00484} & 0.00672 & \textbf{0.00167} & 0.00449 \\
\textsc{Nefertiti} & \textbf{0.00058} & 0.00128 & \textbf{0.00354} & 0.00533 & \textbf{0.00123} & 0.00575 \\
\textsc{Spot} & 0.00151 & \textbf{0.00141} & \textbf{0.00620} & 0.00731 & 0.00352 & \textbf{0.00300} \\
\textsc{Springer} & \textbf{0.00050} & 0.00107 & \textbf{0.00281} & 0.00630 & \textbf{0.00115} & 0.00349 \\
\textsc{Genus-6} & \textbf{0.00259} & 0.00309 & \textbf{0.00892} & 0.01062 & \textbf{0.00677} & 0.00758 \\
\textsc{Voronoi} & \textbf{0.00225} & 0.00479 & \textbf{0.01921} & 0.02591 & \textbf{0.01581} & 0.02136 \\
\textsc{Rind} & \textbf{0.00126} & 0.00308 & \textbf{0.01141} & 0.02127 & \textbf{0.00905} & 0.01750 \\
\textsc{V-bunny} & \textbf{0.00186} & 0.00763 & \textbf{0.01770} & 0.02387 & \textbf{0.01443} & 0.01999 \\
\textsc{Kangaroo} & \textbf{0.00410} & 0.00826 & \textbf{0.01426} & 0.01557 & \textbf{0.01243} & 0.01476 \\
\textsc{Dino} & \textbf{0.00191} & 0.00709 & \textbf{0.01069} & 0.01863 & \textbf{0.00891} & 0.01580 \\
\textsc{Bracket} & \textbf{0.00152} & 0.00157 & \textbf{0.00501} & 0.00510 & 0.00378 & \textbf{0.00308} \\
\textsc{Clamp} & \textbf{0.00120} & 0.00128 & \textbf{0.00311} & 0.00572 & \textbf{0.00183} & 0.00253 \\
\textsc{Hinge} & \textbf{0.00379} & 0.00414 & \textbf{0.00963} & 0.01018 & 0.00764 & \textbf{0.00529} \\
\textsc{Housing} & \textbf{0.00301} & 0.00319 & \textbf{0.00400} & 0.00736 & \textbf{0.00355} & 0.00483 \\
\midrule
mean (18 scenes) & \textbf{0.00167} & 0.00297 & \textbf{0.00820} & 0.01181 & \textbf{0.00581} & 0.00852 \\
\bottomrule
\end{tabular}
\end{table}

%% file: figures/e2e/e2e-snippet.tex
\begin{figure}
\centering
\setlength{\tabcolsep}{0.5pt}%
\ifdefined\eepw\else\newlength{\eepw}\fi\setlength{\eepw}{0.147\columnwidth}%
\ifdefined\eerlw\else\newlength{\eerlw}\fi\providecommand{\eerowlab}[2]{\settowidth{\eerlw}{\scriptsize\begin{tabular}{@{}c@{}}#1\\#2\end{tabular}}\raisebox{\dimexpr(\eepw-\eerlw)/2\relax}{\rotatebox{90}{\scriptsize\begin{tabular}{@{}c@{}}#1\\#2\end{tabular}}}}%
\begin{tabular}{@{}c@{\hspace{2pt}}ccccccc@{}}
 & \scriptsize reference & \scriptsize\makebox[\eepw][c]{mesh\,+\,FEM} & \scriptsize ours &  & \scriptsize\makebox[\eepw][c]{FEM $-$ ref} & \scriptsize\makebox[\eepw][c]{ours $-$ ref} &  \\
\multicolumn{8}{@{}l@{}}{\footnotesize\textsc{Fertility}} \\[0.5pt]
\eerowlab{full}{interior} & \includegraphics[width=\eepw]{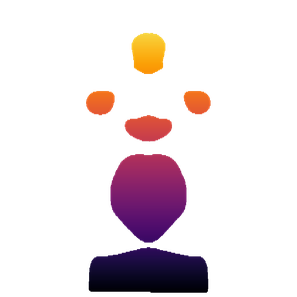} & \includegraphics[width=\eepw]{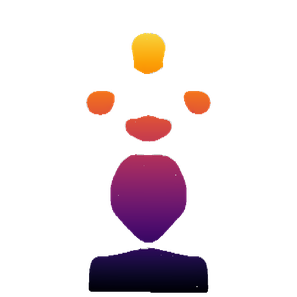} & \includegraphics[width=\eepw]{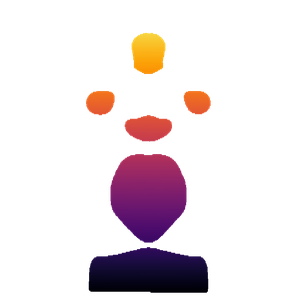} & \includegraphics{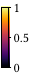} & \includegraphics[width=\eepw]{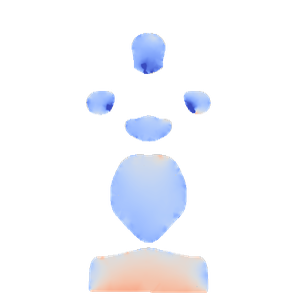} & \includegraphics[width=\eepw]{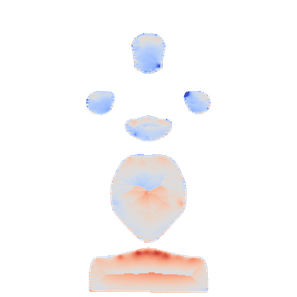} & \includegraphics{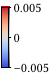} \\[1pt]
\eerowlab{one view}{surface} & \includegraphics[width=\eepw]{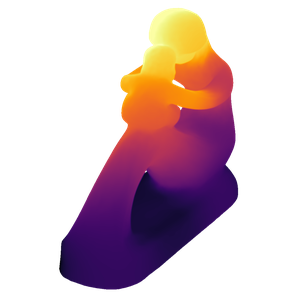} & \includegraphics[width=\eepw]{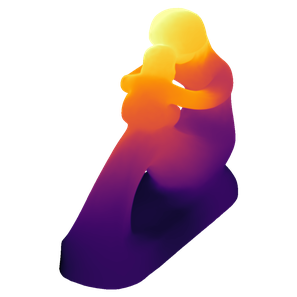} & \includegraphics[width=\eepw]{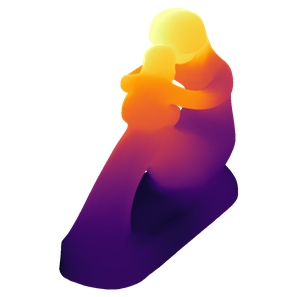} & \includegraphics{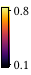} & \includegraphics[width=\eepw]{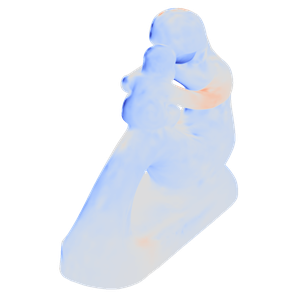} & \includegraphics[width=\eepw]{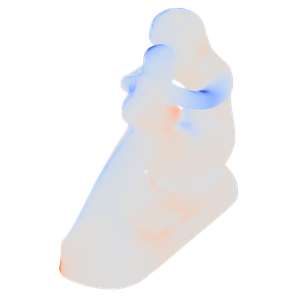} & \includegraphics{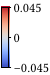} \\[1pt]
\eerowlab{one view}{interior} & \includegraphics[width=\eepw]{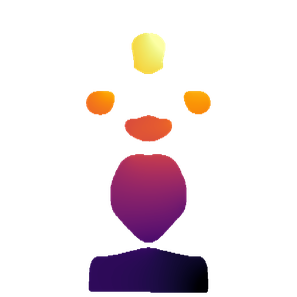} & \includegraphics[width=\eepw]{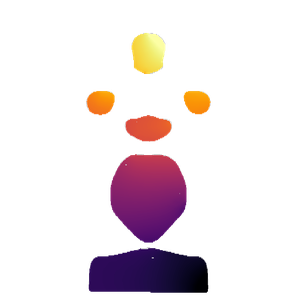} & \includegraphics[width=\eepw]{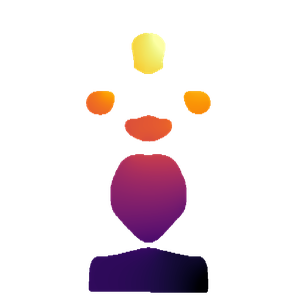} & \includegraphics{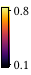} & \includegraphics[width=\eepw]{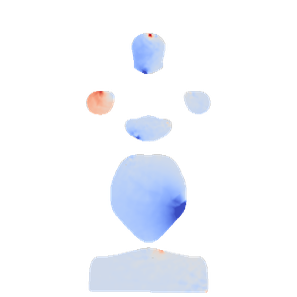} & \includegraphics[width=\eepw]{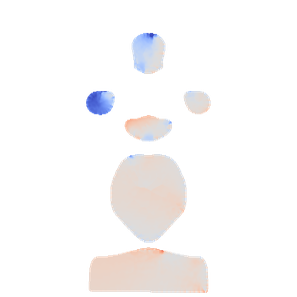} & \includegraphics{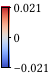} \\[1pt]
\multicolumn{8}{@{}l@{}}{\footnotesize\textsc{Rind}} \\[0.5pt]
\eerowlab{full}{interior} & \includegraphics[width=\eepw]{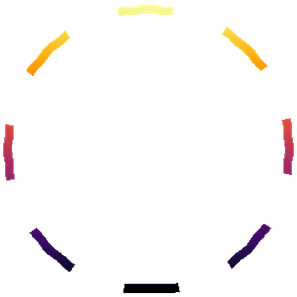} & \includegraphics[width=\eepw]{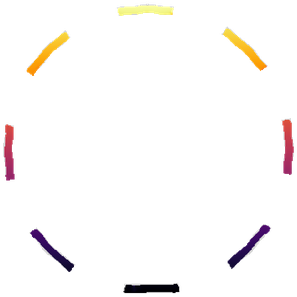} & \includegraphics[width=\eepw]{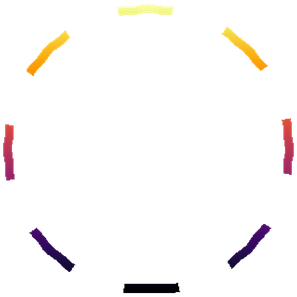} & \includegraphics{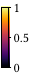} & \includegraphics[width=\eepw]{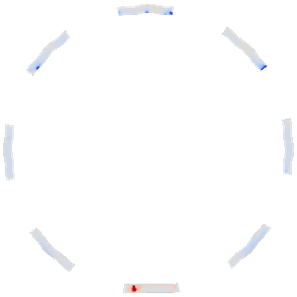} & \includegraphics[width=\eepw]{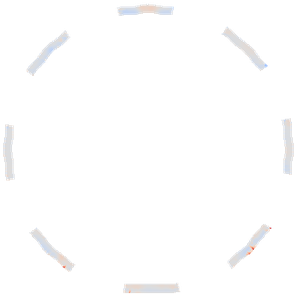} & \includegraphics{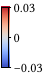} \\[1pt]
\eerowlab{one view}{surface} & \includegraphics[width=\eepw]{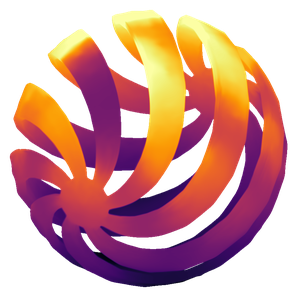} & \includegraphics[width=\eepw]{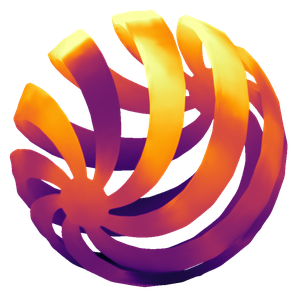} & \includegraphics[width=\eepw]{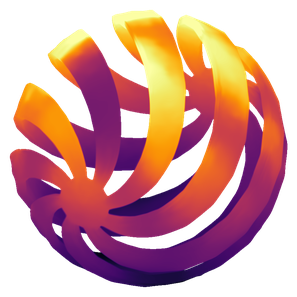} & \includegraphics{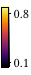} & \includegraphics[width=\eepw]{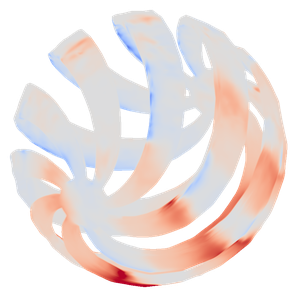} & \includegraphics[width=\eepw]{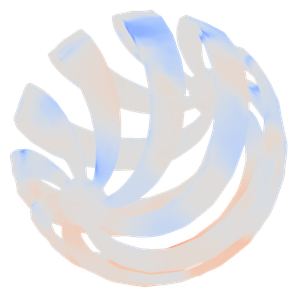} & \includegraphics{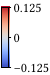} \\[1pt]
\eerowlab{one view}{interior} & \includegraphics[width=\eepw]{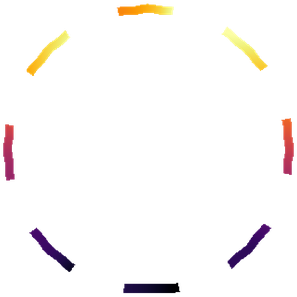} & \includegraphics[width=\eepw]{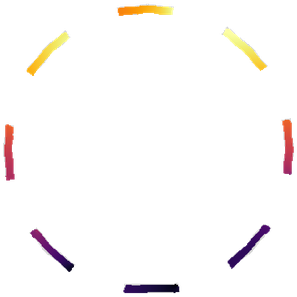} & \includegraphics[width=\eepw]{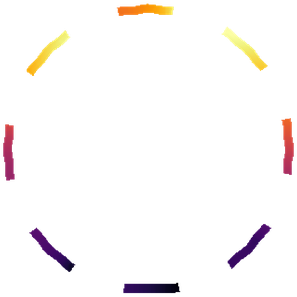} & \includegraphics{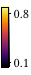} & \includegraphics[width=\eepw]{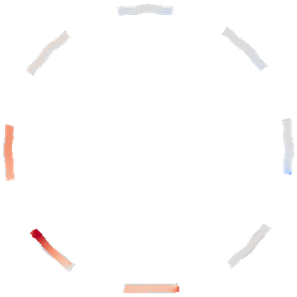} & \includegraphics[width=\eepw]{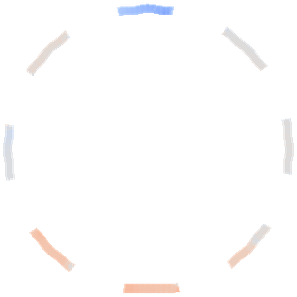} & \includegraphics{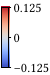} \\[1pt]
\end{tabular}
\caption{%
End-to-end comparison with the mesh-based pipeline on \textsc{Fertility} and \textsc{Rind}, with full thermal coverage and under one thermal view: the reference, the mesh pipeline and ours, and the errors of both.}
\label{fig:e2e-comp}
\end{figure}

%% file: figures/capture/capture-snippet.tex
\begin{tikzpicture}[every node/.style={inner sep=0, outer sep=0}, lab/.style={font=\footnotesize, inner sep=1pt, fill=white, fill opacity=0.8, text opacity=1},
  let/.style={font=\footnotesize, inner sep=2pt}, lead/.style={-{Stealth[length=3.2pt, width=2.6pt]}, black, line width=0.55pt}, arr/.style={-{Stealth[length=3.2pt, width=2.6pt]}, black, line width=0.55pt}]
  \node[anchor=north west] (A) at (0, 0) {\includegraphics[width=168.97pt]{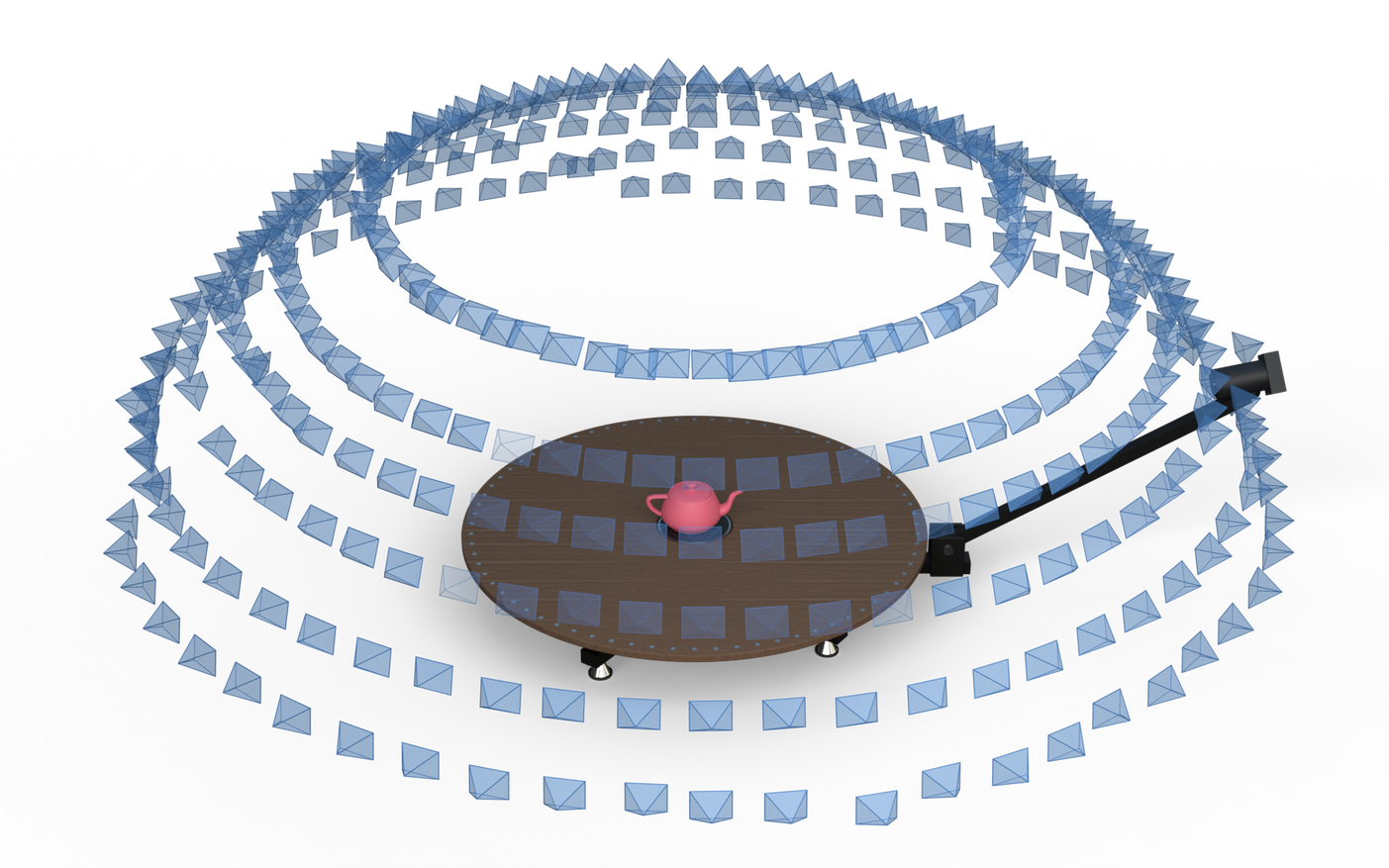}};
  \node[anchor=north west] (B) at (170.97pt, 0) {\includegraphics[width=72.17pt]{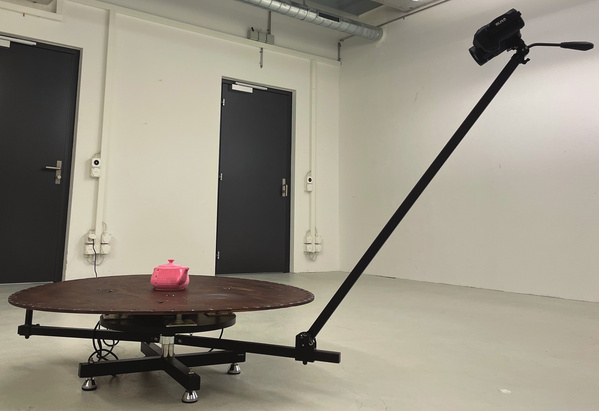}};
  \node[anchor=north west] (C) at ([yshift=-2pt]B.south west) {\includegraphics[width=72.17pt]{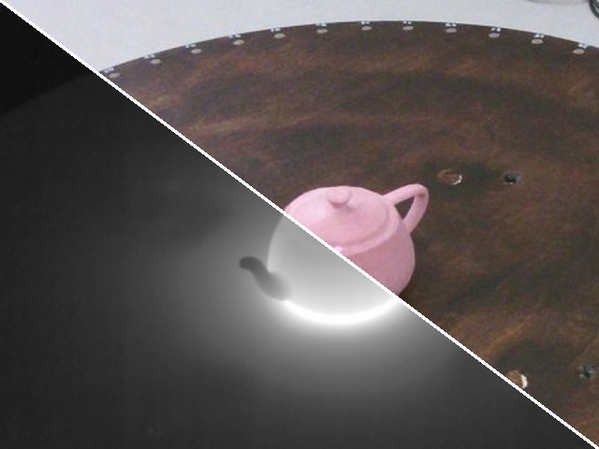}};
  \begin{scope}[shift={(A.north west)}]
    \node[lab, anchor=south east] (lcam) at (164.75pt,-39.38pt) {Infrared camera};
    \node[lab, anchor=south] (lpot) at (62.52pt,-45.05pt) {Heated object};
    \draw[lead] (lpot.south) -- (84.49pt,-58.36pt);
    \node[lab, anchor=west] (lhp) at (110.02pt,-71.33pt) {Hot plate};
    \draw[lead] ([xshift=-6pt]lcam.south) -- (151.65pt,-44.97pt);
    \draw[arr] (lhp.west) to[bend left=20] (90.08pt,-64.99pt);
    \node[lab, anchor=north west, align=left] (lcp) at (2pt,-2pt) {COLMAP\\camera poses};
    \draw[lead] (lcp.east) -- (53.38pt,-13.20pt);
    \draw[lead] (lcp.south) -- (44.46pt,-27.88pt);
    \draw[lead] (lcp.south) -- (33.34pt,-35.26pt);
  \end{scope}
  \node[let, anchor=south west] at (A.south west) {(a)};
  \node[let, anchor=south west] at (B.south west) {(b)};
  \node[let, anchor=south west, text=white] at (C.south west) {(c)};
\end{tikzpicture}%

%% file: figures/teapot/teapot-snippet.tex
\begin{figure}
\centering
\input{figures/teapot/teapot-proto-numbers}%
\definecolor{tpA}{HTML}{603896}\definecolor{tpB}{HTML}{1E5A96}%
\begin{tikzpicture}[every node/.style={inner sep=0, outer sep=0}, lab/.style={font=\footnotesize, align=center}, pw/.style={font=\scriptsize, anchor=north west},
  bst/.style={font=\scriptsize, anchor=north}, arr/.style={-{Stealth[length=3.5pt]}, line width=0.55pt}]
  \node[anchor=west] (TA1) at (0, 0) {\includegraphics[width=37pt]{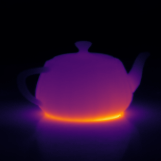}};
  \node[anchor=west] (CU1) at ([xshift=2pt]TA1.east) {\includegraphics{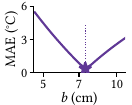}};
  \node[anchor=west] (SL1) at ([xshift=9pt]CU1.east) {\includegraphics{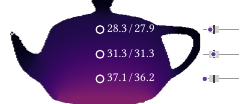}};
  \node[anchor=west] (TA2) at (0, -55pt) {\includegraphics[width=37pt]{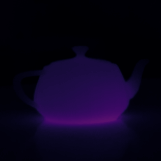}};
  \node[anchor=west] (CU2) at ([xshift=2pt]TA2.east) {\includegraphics{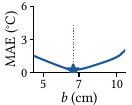}};
  \node[anchor=west] (SL2) at ([xshift=9pt]CU2.east) {\includegraphics{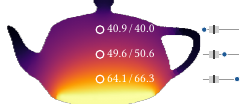}};
  \foreach \n/\w in {TA1/150, TA2/100} {\node[pw, text=white] at ([xshift=1.5pt, yshift=-1.5pt]\n.north west) {\w\,W};}
  \foreach \n/\w in {SL1/100, SL2/150} {\node[pw] at (\n.north west) {\w\,W};}
  \node[bst, text=tpA] at ([xshift=\tpLabXA, yshift=\tpLabYA]CU1.south west) {$b^\ast = \tpBstarA$\,cm};
  \node[bst, text=tpB] at ([xshift=\tpLabXB, yshift=\tpLabYB]CU2.south west) {$b^\ast = \tpBstarB$\,cm};
  \draw[arr] ([xshift=1.5pt]CU1.east) -- ([xshift=-1.5pt]SL1.west);
  \draw[arr] ([xshift=1.5pt]CU2.east) -- ([xshift=-1.5pt]SL2.west);
  \node[anchor=west] at ([xshift=3pt]$(SL1.east)!0.5!(SL2.east)$) {\includegraphics{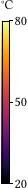}};
  \node[anchor=north east] at ([yshift=-1pt]SL2.south east) {\includegraphics{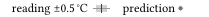}};
  \coordinate (HY) at ([yshift=3pt]CU1.north);
  \coordinate (H1) at ($(TA1.west)!0.5!(CU1.east)$); \coordinate (H2) at ($(SL1.west)!0.5!(SL1.east)$);
  \node[lab, anchor=south] at (H1 |- HY) {calibrate $b$ on one heating power};
  \node[lab, anchor=south] at (H2 |- HY) {predict the other with $b^\ast$};
\end{tikzpicture}
\caption{The real \textsc{Teapot}, calibrated on one heating power and predicted on the other, in both directions. In each row, left: a thermal image of the calibration capture and the error at its three cast-in probes against the Robin coefficient, starred at the optimum $b^\ast$. Right: the interior of the other capture on the axial plane, predicted from its own thermal images with the transferred $b^\ast$, each probe labeled measured / predicted in \textdegree C.
The probes' rated accuracy against the reading is annotated on the right.
}
\label{fig:teapot}
\end{figure}

%% file: figures/teapot/teapot-proto-numbers.tex
\def\tpBstarA{7.4}\def\tpCalA{0.38}\def\tpDepA{0.43}\def\tpOwnA{0.23}\def\tpLabXA{40.94pt}\def\tpLabYA{46.50pt}%
\def\tpBstarB{6.6}\def\tpCalB{0.23}\def\tpDepB{1.41}\def\tpOwnB{0.38}\def\tpLabXB{35.15pt}\def\tpLabYB{46.50pt}%

%% file: figures/pipe/pipe-snippet.tex
\begin{figure*}
\centering
\ifdefined\pipeSetupWidth\else\newlength{\pipeSetupWidth}\fi\setlength{\pipeSetupWidth}{0.36\textwidth}%
\newcommand{\piperow}[1]{\rotatebox{90}{\makebox[0.068\textwidth][c]{\scriptsize #1}}}%
\setlength{\tabcolsep}{3pt}
\begin{tabular}{@{}c@{\hspace{6pt}}c@{\hspace{12pt}}c@{}}
\raisebox{-0.5\height}{\input{figures/pipe/setup-overlay}} &
\raisebox{-0.5\height}{\begin{tikzpicture}[remember picture]
  \node[anchor=south west, inner sep=0] (pipefloor) at (0,0) {\includegraphics[width=0.116\textwidth]{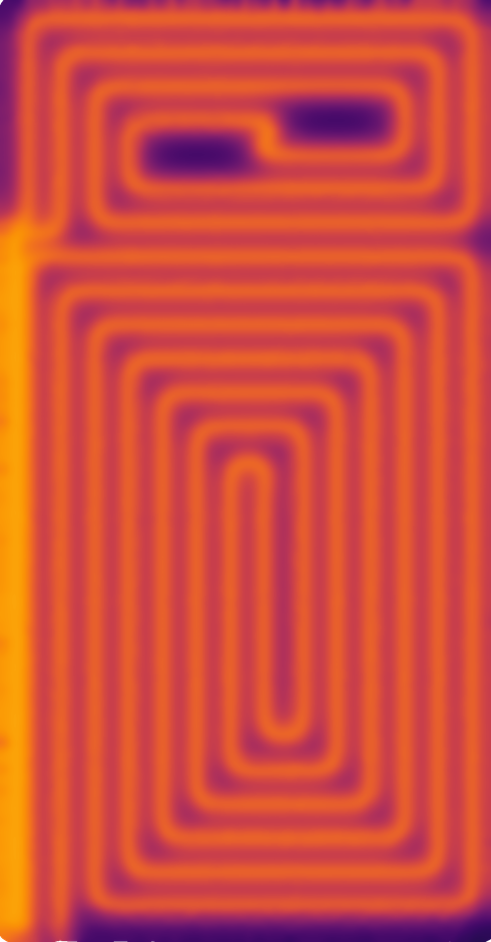}};
  \begin{scope}[shift={(pipefloor.south west)}, x={($(pipefloor.south east)-(pipefloor.south west)$)}, y={($(pipefloor.north west)-(pipefloor.south west)$)}]
    \draw[white, line width=0.6pt] (0.75,0.0711) rectangle (0.9063,0.1553);
    \coordinate (pipezoomtop) at (0.9063,0.1553); \coordinate (pipezoombot) at (0.9063,0.0711);
  \end{scope}
\end{tikzpicture}} &
\raisebox{-0.5\height}{\setlength{\tabcolsep}{0.5pt}\begin{tabular}[b]{@{}c@{\hspace{2pt}}cccccc@{\hspace{2pt}}c@{}}
 & \footnotesize start & \multicolumn{3}{c}{\footnotesize iterates $\longrightarrow$} & \footnotesize recovered & \footnotesize truth & \\
\tikz[remember picture, overlay]\coordinate (pipegridleft);\piperow{plan} & \tikz[remember picture, baseline=(pipegridtop.base)]\node[inner sep=0, anchor=base] (pipegridtop) {\includegraphics[width=0.068\textwidth]{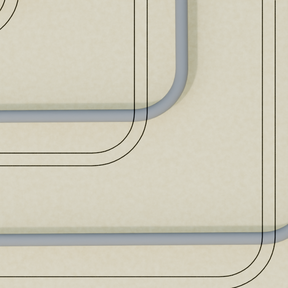}}; & \includegraphics[width=0.068\textwidth]{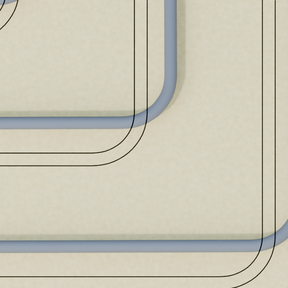} & \includegraphics[width=0.068\textwidth]{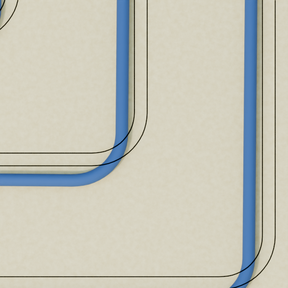} & \includegraphics[width=0.068\textwidth]{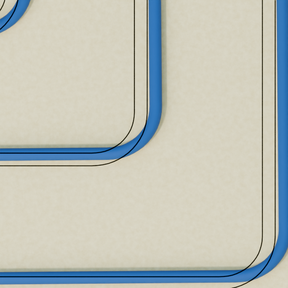} & \includegraphics[width=0.068\textwidth]{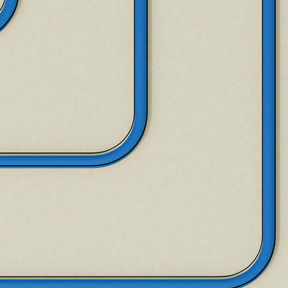} & \includegraphics[width=0.068\textwidth]{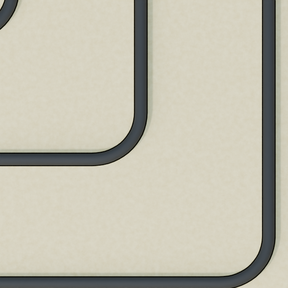} & \\
\piperow{section} & \includegraphics[width=0.068\textwidth]{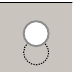} & \includegraphics[width=0.068\textwidth]{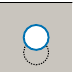} & \includegraphics[width=0.068\textwidth]{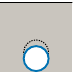} & \includegraphics[width=0.068\textwidth]{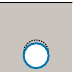} & \includegraphics[width=0.068\textwidth]{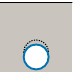} & \includegraphics[width=0.068\textwidth]{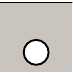} & \\
\piperow{floor} & \tikz[remember picture, baseline=(pipegridbot.base)]\node[inner sep=0, anchor=base] (pipegridbot) {\includegraphics[width=0.068\textwidth]{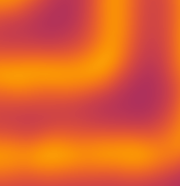}}; & \includegraphics[width=0.068\textwidth]{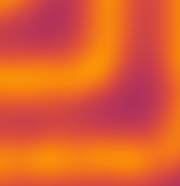} & \includegraphics[width=0.068\textwidth]{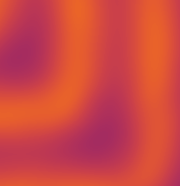} & \includegraphics[width=0.068\textwidth]{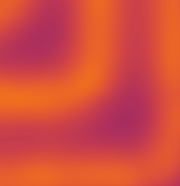} & \includegraphics[width=0.068\textwidth]{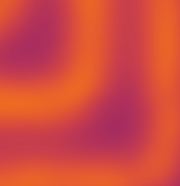} & \includegraphics[width=0.068\textwidth]{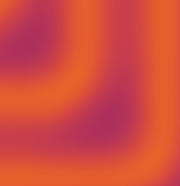} & \raisebox{1.5pt}{\includegraphics{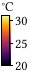}} \\
 & \scriptsize $1.67$~K & \scriptsize $1.42$~K & \scriptsize $0.58$~K & \scriptsize $0.43$~K & \scriptsize $0.34$~K & \scriptsize measured & \\
\end{tabular}} \\
\footnotesize the floor and the thermal camera & \footnotesize measured floor & \footnotesize optimization trajectory (rotation, translation, and depth) \\
\end{tabular}%
\begin{tikzpicture}[remember picture, overlay]
  \draw[black!60, line width=0.3pt] (pipezoomtop) -- (pipegridleft |- pipegridtop.north west);
  \draw[black!60, line width=0.3pt] (pipezoombot) -- (pipegridleft |- pipegridbot.south west);
\end{tikzpicture}
\caption{%
Heating pipe inspection.
We use thermal images to help inspect whether the heating pipe under the floor is installed at the correct orientation and depth.
We recover the pipe's pose and depth from the floor temperature, reconstructed using our optimization algorithm, from the start through the iterates to the truth.}
\label{fig:pipe}
\end{figure*}

%% file: figures/pipe/setup-overlay.tex
\begin{tikzpicture}
\node[anchor=south west, inner sep=0] (img) at (0, 0) {\includegraphics[width=\pipeSetupWidth]{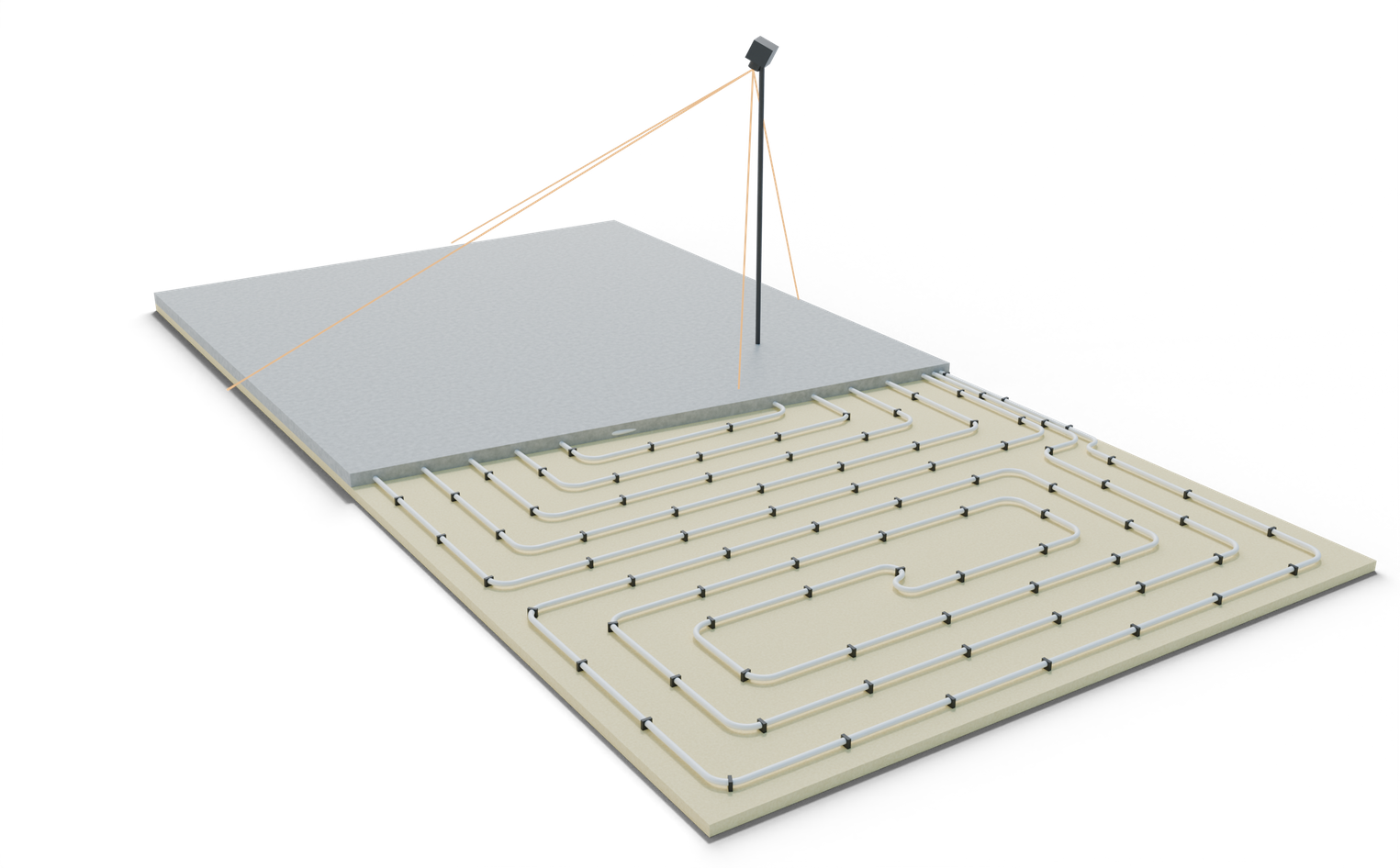}};
\begin{scope}[x={(img.south east)}, y={(img.north west)}, font=\scriptsize]
\fill[black!70] (0.6488, 0.3232) circle (0.9pt); \draw[black!60, line width=0.35pt] (0.6488, 0.3232) -- (0.735, 0.100) node[anchor=west, inner sep=1pt, fill=white, fill opacity=0.75, text opacity=1, text=black] {PE-X pipe, $\varnothing 20$ mm};
\fill[black!70] (0.3109, 0.4646) circle (0.9pt); \draw[black!60, line width=0.35pt] (0.3109, 0.4646) -- (0.010, 0.640) node[anchor=west, inner sep=1pt, fill=white, fill opacity=0.75, text opacity=1, text=black] {mortar screed, 60 mm};
\fill[black!70] (0.9187, 0.3068) circle (0.9pt); \draw[black!60, line width=0.35pt] (0.9187, 0.3068) -- (0.800, 0.550) node[anchor=west, inner sep=1pt, fill=white, fill opacity=0.75, text opacity=1, text=black] {insulation};
\fill[black!70] (0.5444, 0.9401) circle (0.9pt); \draw[black!60, line width=0.35pt] (0.5444, 0.9401) -- (0.600, 0.955) node[anchor=west, inner sep=1pt, fill=white, fill opacity=0.75, text opacity=1, text=black] {thermal camera};
\draw[black!80, line width=0.7pt, dash pattern=on 0.9pt off 1.8pt] (0.6784, 0.5819) -- (0.9853, 0.3713) -- (0.5068, 0.0709) -- (0.2499, 0.4541);
\end{scope}
\end{tikzpicture}

%% file: figures/facade3/setup-overlay.tex
\begin{tikzpicture}
\node[anchor=south west, inner sep=0] (img) at (0, 0) {\includegraphics[width=\facadeSetupWidth]{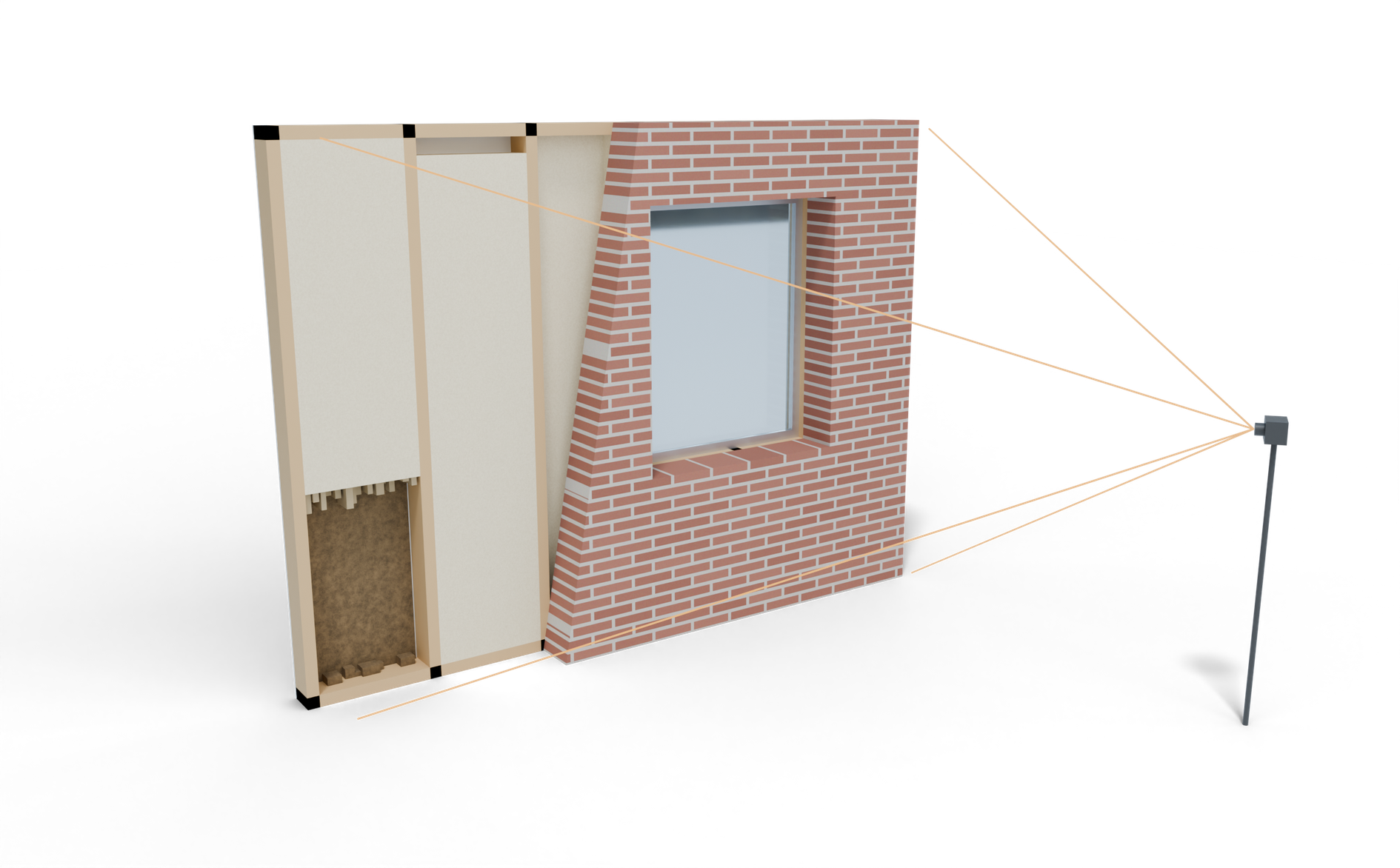}};
\begin{scope}[x={(img.south east)}, y={(img.north west)}, font=\footnotesize]
\fill[black!70] (0.3345, 0.8325) circle (0.9pt); \draw[black!60, line width=0.35pt] (0.3345, 0.8325) -- (0.020, 0.970) node[anchor=west, inner sep=1pt, fill=white, fill opacity=0.75, text opacity=1, text=black] {fill settled, 80 mm gap};
\fill[black!70] (0.3980, 0.5490) circle (0.9pt); \draw[black!60, line width=0.35pt] (0.3980, 0.5490) -- (0.360, 0.970) node[anchor=west, inner sep=1pt, fill=white, fill opacity=0.75, text opacity=1, text=black] {60 mm panel};
\fill[black!70] (0.2552, 0.3354) circle (0.9pt); \draw[black!60, line width=0.35pt] (0.2552, 0.3354) -- (0.020, 0.140) node[anchor=west, inner sep=1pt, fill=white, fill opacity=0.75, text opacity=1, text=black] {panel destroyed by moisture};
\draw[black!80, line width=0.7pt, dash pattern=on 0.9pt off 1.8pt] (0.4565, 0.8583) -- (0.2025, 0.8541) -- (0.2329, 0.1605) -- (0.4044, 0.2337);
\end{scope}
\end{tikzpicture}

%% file: figures/crack-free/crack-free-snippet.tex
\begin{figure}
\centering
\setlength{\tabcolsep}{2pt}
\begin{tabular}{@{}ccccc@{}}
\raisebox{-0.5\height}{\includegraphics[width=0.125\textwidth]{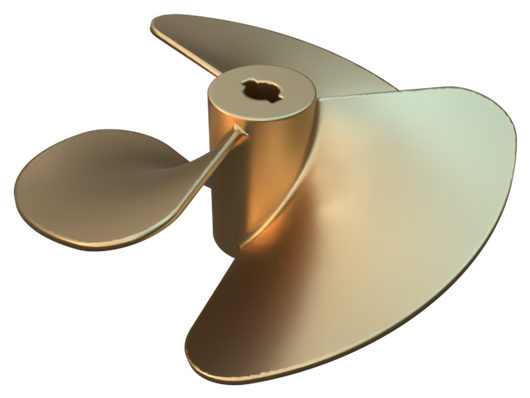}} & \raisebox{-0.5\height}{\includegraphics[width=0.125\textwidth]{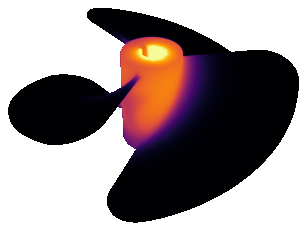}} & \raisebox{-0.5\height}{\includegraphics{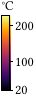}} & \raisebox{-0.5\height}{\includegraphics[width=0.125\textwidth]{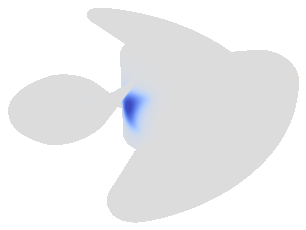}} & \raisebox{-0.5\height}{\includegraphics{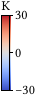}} \\
\footnotesize RGB & \makebox[0pt]{\footnotesize thermal, cracked part} & & \makebox[0pt]{\footnotesize cracked minus intact} & \\[3pt]
\end{tabular}
\par\smallskip
\begin{tikzpicture}[every node/.style={inner sep=0, outer sep=0}, lab/.style={font=\footnotesize, anchor=base}, val/.style={font=\scriptsize, anchor=base}]
  \pgfmathsetlengthmacro{\pw}{(\columnwidth-6pt-8pt)/5}\def\gap{2pt}
  \node[anchor=south west] (A) at (0,0) {\includegraphics[width=\pw]{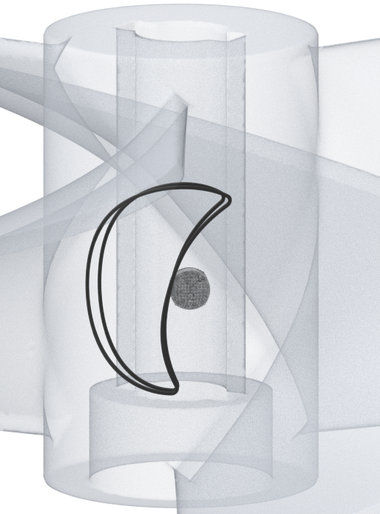}};
  \node[anchor=south west] (B) at ([xshift=\gap]A.south east) {\includegraphics[width=\pw]{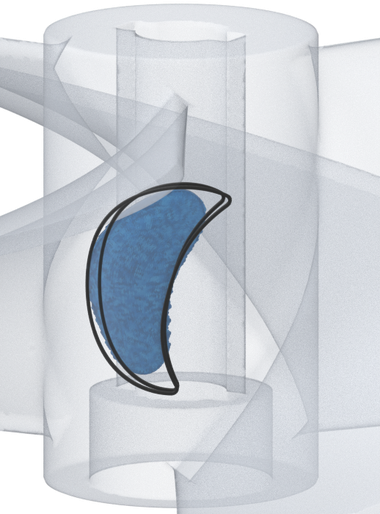}};
  \node[anchor=south west] (C) at ([xshift=8pt]B.south east) {\includegraphics[width=\pw]{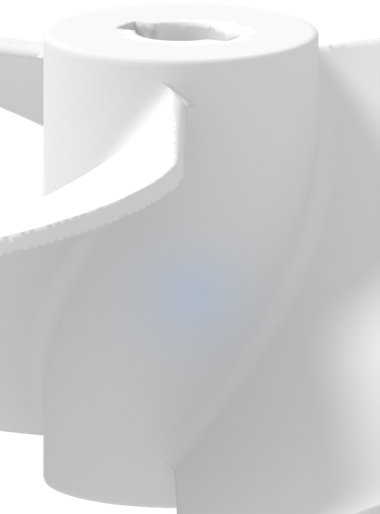}};
  \node[anchor=south west] (D) at ([xshift=\gap]C.south east) {\includegraphics[width=\pw]{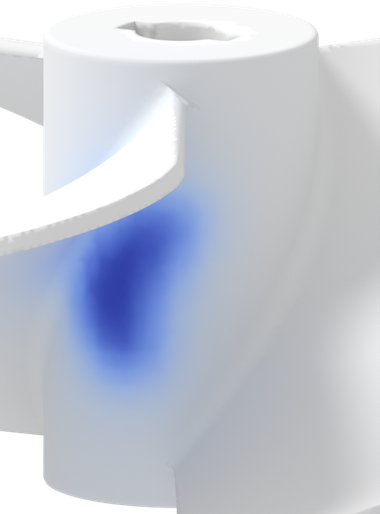}};
  \node[anchor=south west] (E) at ([xshift=\gap]D.south east) {\includegraphics[width=\pw]{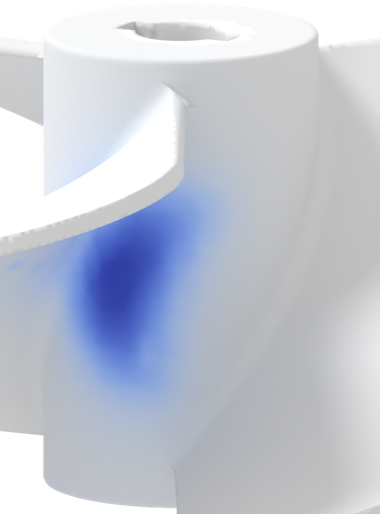}};
  \foreach \n/\t in {A/initial, B/recovered, C/initial, D/recovered, E/measured} {\node[lab] at ([yshift=3pt]\n.north) {\t};}
  \node[lab] at ([yshift=13pt]$(A.north)!0.5!(B.north)$) {splats in the hub};
  \node[lab] at ([yshift=13pt]D.north) {thermal difference on the surface};
  \draw[black!40, line width=0.4pt] ([yshift=11pt]A.north west) -- ([yshift=11pt]B.north east);
  \draw[black!40, line width=0.4pt] ([yshift=11pt]C.north west) -- ([yshift=11pt]E.north east);
  \node[val, text=black!60] at ([yshift=-7.5pt]A.south) {$528$ splats}; \node[val, text=black!60] at ([yshift=-7.5pt]B.south) {$5{,}237$ splats};
  \node[val] at ([yshift=-7.5pt]C.south) {$1.52$\,K}; \node[val] at ([yshift=-7.5pt]D.south) {$0.21$\,K};
\end{tikzpicture}
\caption{%
Identifying crack by Gaussian splat flow.
We use thermal images to help find a crack hidden inside a propeller hub, which looks intact in RGB photo.
We recover the crack's shape (\cref{fig:splat-flow})
from the difference between the thermal images of the cracked and the intact part.
The crack is represented using Gaussian splats throughout the optimization, without any need for meshes.
Bottom: the splats inside the hub (black outline: the true crack) and the crack's signature on the hub's surface in our solve, at the start and as recovered, against the measured difference.
}
\label{fig:crack-free}
\end{figure}

%% file: figures/convergence-seq/convseq-snippet.tex
\begin{figure}
\centering
\input{figures/convergence-seq/row-anchors}%
\begin{tikzpicture}[every node/.style={inner sep=0, outer sep=0}]
  \node[anchor=north west] (R) at (0,0) {\includegraphics[width=\columnwidth]{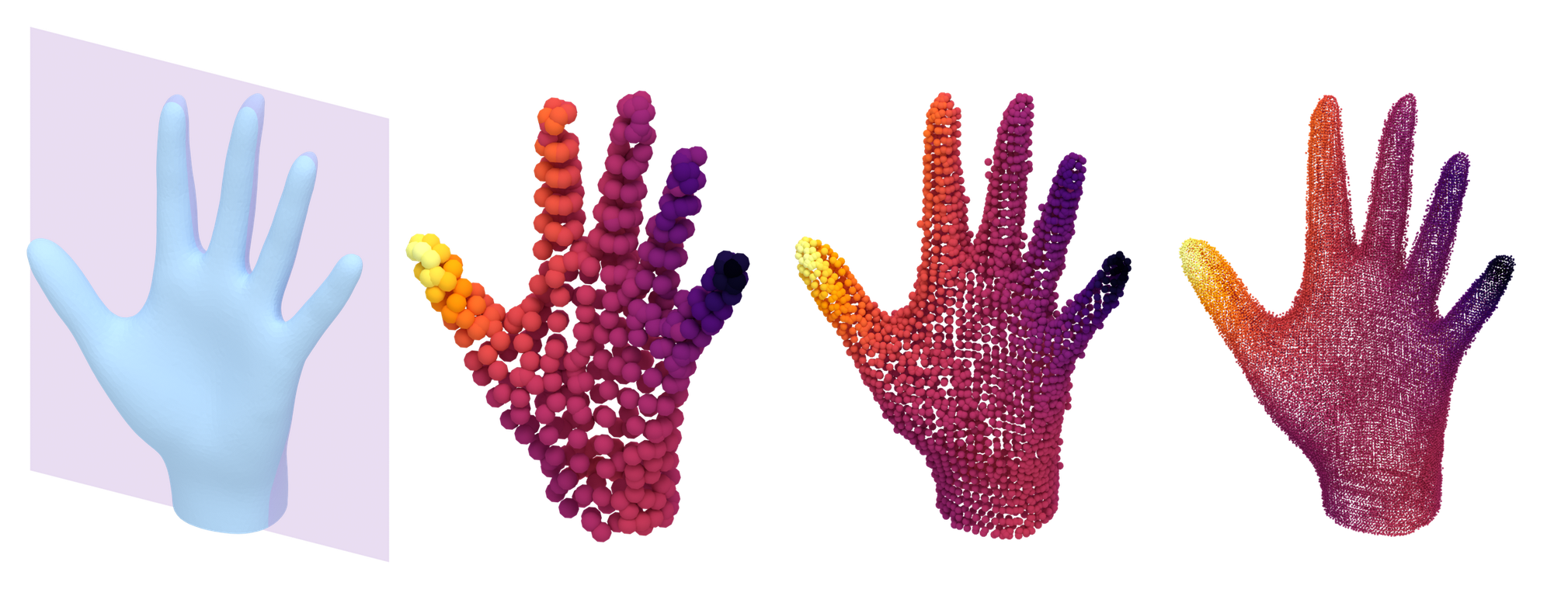}};
  \begin{scope}[shift={(R.south west)}, x={($(R.south east)-(R.south west)$)}, y={($(R.north west)-(R.south west)$)}]
    \node[anchor=south west] at (\csLB - 0.008, 0.02) {\includegraphics[width=0.066\columnwidth]{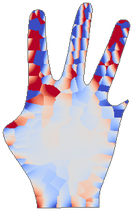}};
    \node[anchor=south west] at (\csLC - 0.008, 0.02) {\includegraphics[width=0.066\columnwidth]{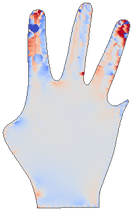}};
    \node[anchor=south west] at (\csLD - 0.008, 0.02) {\includegraphics[width=0.066\columnwidth]{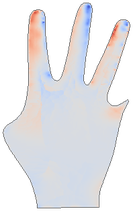}};
    \foreach \c/\t in {\csCA/\textsc{Hand}, \csCB/{$N = 496$}, \csCC/{$N = 3.5$K}, \csCD/{$N = 45.6$K}} {\node[font=\footnotesize, anchor=base] at ([yshift=3pt]\c, 1) {\t};}
    \foreach \c/\t in {\csCB/{MAE $4.2 \times 10^{-3}$}, \csCC/{MAE $1.3 \times 10^{-3}$}, \csCD/{MAE $8.3 \times 10^{-4}$}} {\node[font=\scriptsize, anchor=base] at ([yshift=-7pt]\c, 0) {\t};}
    \node[font=\scriptsize, anchor=base] (CB) at ([yshift=-7pt]\csCA, 0) {$-0.01$\,\raisebox{0.15ex}{\includegraphics[width=22pt, height=4pt]{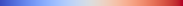}}\,$0.01$};
  \end{scope}
  \node[anchor=north west] at ([yshift=-11pt]R.south west) {\includegraphics{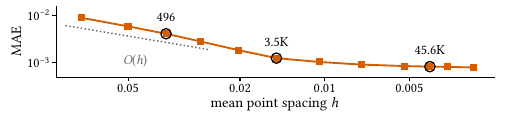}};
\end{tikzpicture}
\caption{%
Convergence under collocation refinement.
We extract more and more collocation points from the same reconstruction of \textsc{Hand} and solve with full thermal coverage.
Each cloud shows its solved surface temperature and, in its corner, the interior error on the slice.
As the cloud grows denser, the error decreases.
The interior MAE falls at about first order against the mean spacing $h$.}
\label{fig:capture-convergence}
\end{figure}

%% file: figures/convergence-seq/row-anchors.tex
\def\csCA{0.1242}\def\csLA{0.0171}%
\def\csCB{0.3691}\def\csLB{0.2620}%
\def\csCC{0.6141}\def\csLC{0.5069}%
\def\csCD{0.8590}\def\csLD{0.7519}%
\def\csAspect{0.3758}%

%% file: figures/eps-sweep-snippet.tex
\begin{figure}
\centering
\includegraphics{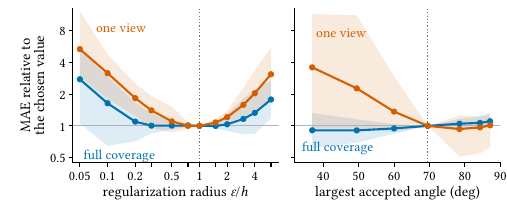}
\caption{%
The regularization radius $\varepsilon$ (left) and the frontality threshold (right) on the eighteen scenes: the error inside with full coverage and on the unobserved surface under one view, relative to that at the chosen value; the mean of the scenes (line) and their range (shaded).
The mean is flat around $\varepsilon = h$; full coverage favors a stricter frontality threshold, one view a more permissive one.}
\label{fig:eps-sweep}\label{fig:front-sweep}
\end{figure}

%% file: tables/tab-ablation.tex
\begin{table}
\centering
\footnotesize
\setlength{\tabcolsep}{2pt}
\caption{
Leave-one-out ablation: each row removes one component of the full pipeline. Means over the eighteen scenes of the interior MAE with full coverage, and of the interior and unobserved-surface MAE under one thermal view with convection on the unseen surface. Best of each column in bold.}
\label{tab:ablation}
\begin{tabular}{@{}lccc@{}}
\toprule
 & full coverage & \multicolumn{2}{c}{single view} \\
\cmidrule(lr){2-2}\cmidrule(l){3-4}
variant & interior & interior & surface \\
\midrule
full pipeline (ours) & \textbf{0.00167} & \textbf{0.00581} & \textbf{0.00820} \\
\midrule
without the depth-slope filter & 0.00309 & 0.00709 & 0.00988 \\
maps from 2DGS instead of RaDe-GS & 0.00290 & 0.00783 & 0.01267 \\
without the membership filter & 0.00199 & 0.00610 & 0.00850 \\
without density subtraction & 0.03473 & 0.04597 & 0.00820 \\
without damping ($\gamma = 0$) & 0.00303 & 0.00715 & 0.00878 \\
\bottomrule
\end{tabular}
\end{table}

%% file: figures/topo/topo-snippet.tex
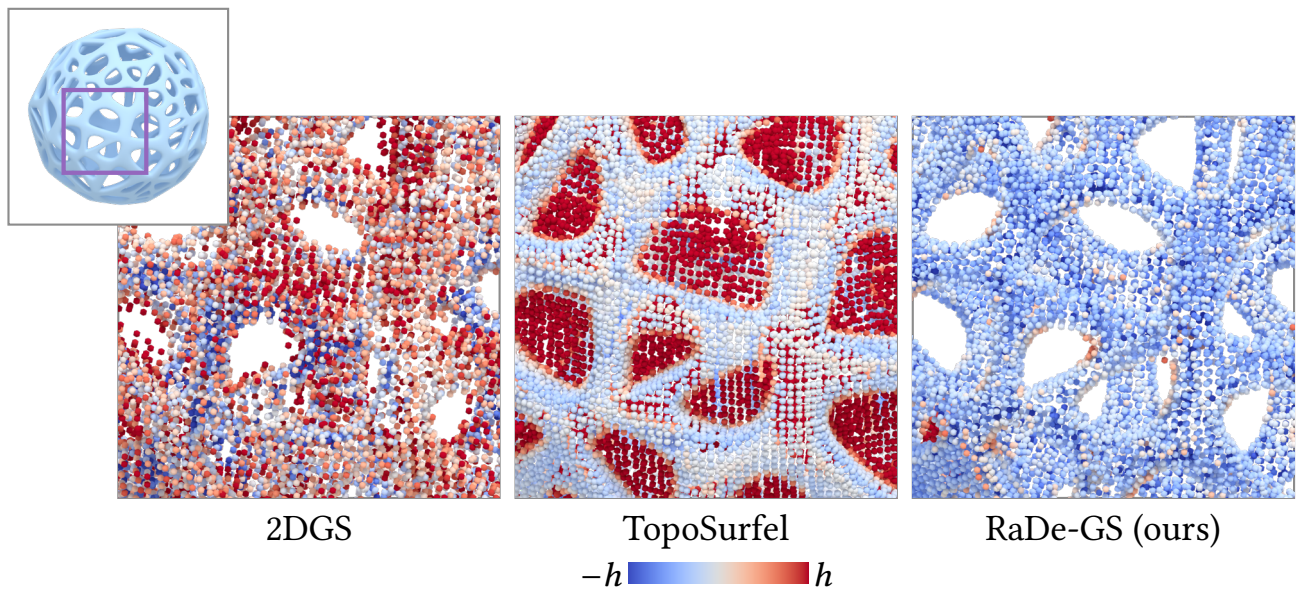
\begin{figure}
\centering
\definecolor{tpurple}{HTML}{9467BD}%
\begin{tikzpicture}[every node/.style={inner sep=0, outer sep=0}]
  \pgfmathsetlengthmacro{\insetS}{0.17\columnwidth}\pgfmathsetlengthmacro{\pw}{(\columnwidth - 0.5*\insetS - 6pt - 1.2pt)/3}
  \node[anchor=north west, draw=black!45, line width=0.4pt] (A) at (0.5*\insetS,0) {\includegraphics[width=\pw]{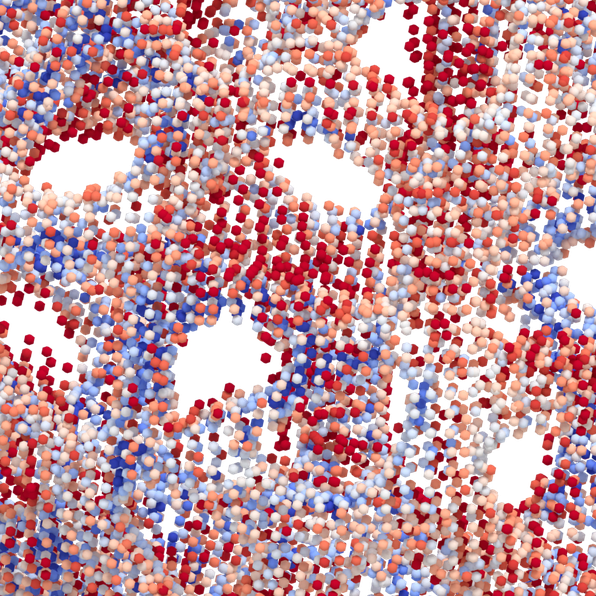}};
  \node[anchor=north west, draw=black!45, line width=0.4pt] (B) at ([xshift=3pt]A.north east) {\includegraphics[width=\pw]{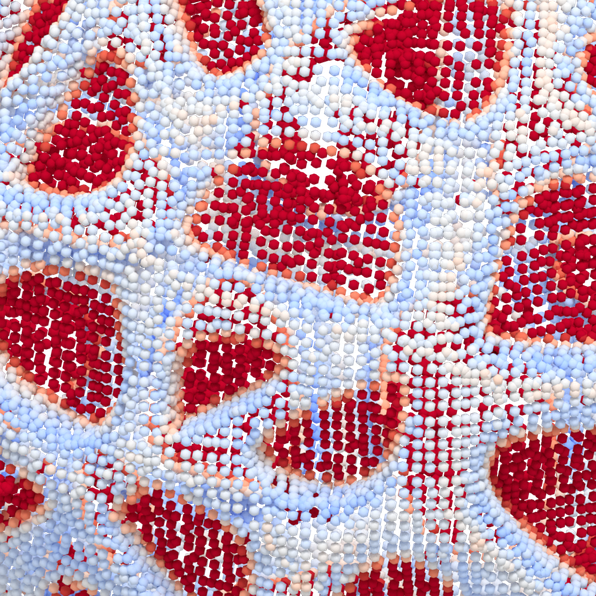}};
  \node[anchor=north west, draw=black!45, line width=0.4pt] (C) at ([xshift=3pt]B.north east) {\includegraphics[width=\pw]{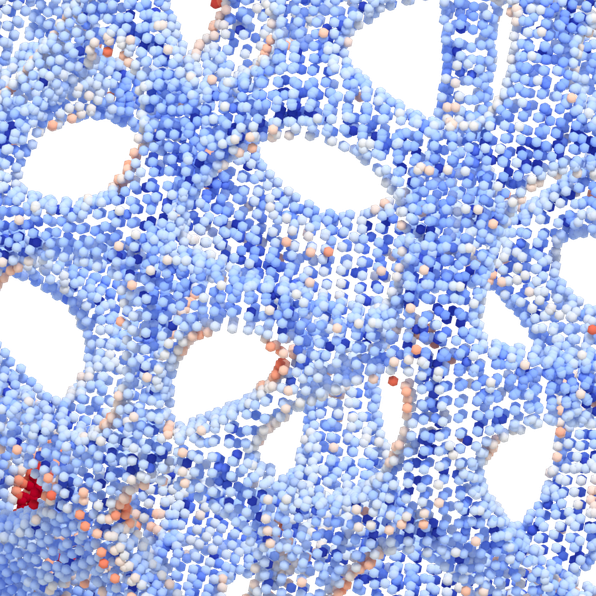}};
  \node[anchor=center, fill=white, draw=black!45, line width=0.4pt] (V) at (A.north west) {\includegraphics[width=\insetS]{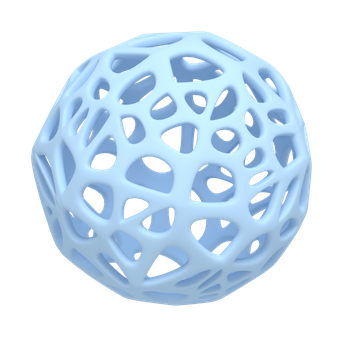}};
  \begin{scope}[shift={(V.south west)}, x={($(V.south east)-(V.south west)$)}, y={($(V.north west)-(V.south west)$)}]
    \draw[tpurple, line width=0.8pt] (0.2510,0.2402) rectangle ++(0.3765,0.3835);
  \end{scope}
  \node[font=\footnotesize, anchor=base] (LA) at ([yshift=-8pt]A.south) {2DGS};
  \node[font=\footnotesize, anchor=base] at ([yshift=-8pt]B.south) {TopoSurfel};
  \node[font=\footnotesize, anchor=base] at ([yshift=-8pt]C.south) {RaDe-GS (ours)};
  \node[font=\scriptsize, anchor=base] at ([yshift=-8.5pt]B.south |- LA.base) {$-h$\,\raisebox{0.15ex}{\includegraphics[width=34pt, height=4pt]{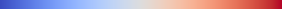}}\,$h$};
\end{tikzpicture}
\caption{%
Collocation points of our method using three Gaussian reconstructions, colored by their signed distance to the true surface.
2DGS scatters points to both sides of the struts, TopoSurfel closes the openings between them with membranes, and RaDe-GS follows the struts the best among the three.}
\label{fig:topo-clouds}
\end{figure}

%% file: 5-conclusion.tex
\section{Conclusions}
\label{sec:conclusion}

\paragraph{Summary}
We present CiGS, a method that infers the interior temperature of a solid from registered visible-spectrum and thermal images by solving a direct boundary integral equation on a point cloud extracted from a Gaussian reconstruction, without meshing the surface or the volume.
We have two steps that make our method accurate, a depth-slope filter that keeps the pixels at depth discontinuities out of the cloud, and an interior evaluation that integrates the constant part of the density exactly.
CiGS needs less time after the reconstruction than tetrahedralization alone in a mesh pipeline, and its error is similar or lower.
Our pipeline is differentiable and that allows us to solve three inverse problems, from a buried pipe to a crack in a hub, with no remeshing as the geometry moves.

\paragraph{Limitations and future work}
The underlying model is the Laplace equation,
and therefore, the object must be homogeneous, isotropic, source-free and at steady state.
Real objects are assembled from several materials, yet as long as each material is homogeneous the problem stays on boundaries, with integral equations on the boundary of each material coupled where they meet.
Our heating pipe example has different materials, but needs no such coupling by assuming the heating pipe at a known temperature.
Coupling several materials of unknown temperature and identifying these interfaces inside an object with multiple materials could be a direction going forward for more realistic scenarios.
Objects in real life also rarely stay in a steady state.
We could use a thermal video to help analyze the time-dependent effect and material properties in thermal conduction.
Most of the error on our hardest scenes, the strut lattices, comes from the reconstruction, so every advance in reconstructing thin structures carries over to our solve.
The surface is also assumed to be purely emissive, so the solver could be coupled with HiGS \cite{zhenyuanInverseRadiative2025}, to solve conduction and radiation jointly.
Looking ahead, we anticipate that resolving these outstanding issues will significantly broaden the use of thermal inspections in many real-world applications.

%% file: 7-appendix.tex
\section{Alternative BIE formulations}
\label{sec:bie-formulations}
In the main text, we use the direct BIE formulation that carries physical quantities, namely the temperature $u$ and the flux $q$ on $\Gamma$.
By Green's representation theorem, we can write the temperature inside from $u$ and $q$; outside $\Omega$ the same formula gives zero (\cref{fig:bie-fields}, right).
Here we give the two indirect BIE formulations that \cref{fig:bie-comp} compares.
Both formulations posit a fictitious source layer on $\Gamma$, whose density is not a temperature: its potential reproduces $u$ inside $\Omega$ and continues outside, smoothly for the single layer and with a jump across $\Gamma$ for the double layer (\cref{fig:bie-fields}); we present them for the pure Dirichlet problem, which is what \cref{fig:bie-comp} uses, and refer the reader to \citeauthor{sauterBoundaryElementMethods2011}~\shortcite[Sec.~3.4]{sauterBoundaryElementMethods2011} for the general case.
\input{figures/bie-fields/bie-fields-snippet}

\paragraph{Single layer potential formulation}
A source density $\sigma: \Gamma \to \mathbb{R}$, which is not a temperature, represents the solution as
\begin{align}
\label{eq:slp-evaluate-solutions}
u(\bx) = \int_{\Gamma} G(\bx, \by)\, \sigma(\by) \;\dd A_\by,
\end{align}
and since the single layer potential is continuous across $\Gamma$, the Dirichlet data give the equation, $\int_\Gamma G(\bx, \by)\,\sigma(\by)\,\dd A_\by = g(\bx)$ for $\bx \in \Gamma$, whose discrete form is $\sum_j V_{ij}\,\sigma_j = g_i$ with the self-term of \cref{eq:regularized-green}.
On a Robin or Neumann boundary its rows carry the normal derivative of the single layer, the adjoint of the double-layer kernel, $\adjpoisson{\bx}{\by}$, with a jump term $\tfrac{1}{2}\sigma$, and it is on those rows that the solver fails. In a system with only Dirichlet boundaries, the single-layer formulation is about as accurate as the direct formulation (ours).

\paragraph{Double layer potential formulation}
We consider an unknown potential moment function $\mu: \Gamma \to \mathbb{R}$,
the solution $u$ can be written as
\begin{align}
\label{eq:dlp-evaluate-solutions}
u(\bx) = -\int_{\Gamma}
\poisson{\bx}{\by} \mu(\by) \;\dd A_\by.
\end{align}
To solve for the unknown $\mu$,
we solve the following system:
for all $\bx \in \Gamma$,
\begin{align}
\label{eq:dlp-system}
g(\bx) - \frac{1}{2}\mu(\bx) + \int_{\Gamma} \poisson{\bx}{\by} \mu(\by) \;\dd A_\by = 0,
\end{align}
where the $-\frac{1}{2}\mu(\bx)$ term arises from the jump relation of the double layer potential across $\Gamma$.
Its discrete form uses the same operator $K$ as \cref{eq:gs-single-layer-potenial}, rigid-body diagonal included, $\tfrac{1}{2}\mu_i - \sum_j K_{ij}\,\mu_j = g_i$, which needs no damping, since the unknown also stands outside the sum; its interior evaluation is $-\sum_j K(\bx, \bx_j)\,\mu_j A_j$ and suffers the same close-evaluation failure as \cref{eq:evaluate-solution}, which the same density subtraction repairs, $u(\bx) = -\sum_j \poisson{\bx}{\bx_j}(\mu_j - \mu_{j^\ast}) A_j + \mu_{j^\ast}$.
On a Robin or Neumann boundary its rows would carry the normal derivative of the double layer, the hypersingular operator, which we do not discretize; in \cref{fig:bie-fields} it therefore takes the boundary temperature of the direct solve as its Dirichlet data.
On manufactured data it is then second only to the direct formulation (\cref{fig:bie-comp}); on image-read data with full coverage, in a separate sweep of the formulations at $24$K points, its interior error is $1.3$ to $2.3$ times the direct formulation's on seventeen of the eighteen scenes and eleven times on the perforated \textsc{Kangaroo} ($4.8$\% of the temperature range against $0.45$\%), which is why we do not use it.

\section{Fabrication, capture, and processing of the \textsc{Teapot}}
\label{sec:fabrication}
We fabricated the Utah \textsc{Teapot} from plaster (gypsum). The mould was modeled in CAD (Rhino) and 3D printed (BambuLab), then used to cast the object as a single solid pour with no internal cavities.
A high-temperature dye was mixed directly into the plaster prior to casting to achieve a uniform color throughout the material, avoiding the emissivity inconsistencies associated with surface paint coatings. The resulting object has a high-emissivity matte finish, suitable for quantitative thermal imaging.

The object was placed in the center of the capture table, with a power-controlled hot plate positioned beneath the model as the heat source.
We performed two hemispherical captures at heating power levels of 100\,W and 150\,W.
All captures were performed at the thermal steady state, defined as a temperature change rate less than $0.1$\,\textdegree C/min, ensuring that the thermal field was stable and representative of the steady-state heat conduction behavior of the object.
\paragraph{Interior temperature sensors}
Casting the object in-house makes an interior measurement possible, which is what lets us check the recovered field rather than only inspect it.
In the \textsc{Teapot} model we cast in three temperature sensors (Shelly DS18B20, range $-55$ to $125$\,\textdegree C, accuracy $0.5$\,\textdegree C, resolution $0.0625$\,\textdegree C) at known interior locations, reading them out once the steady-state criterion above was met.
The sensors are connected to an Arduino UNO over the 1-Wire bus, with firmware built in the Arduino IDE using the OneWire and DallasTemperature libraries.

\paragraph{Boundary data}
After the capture at $100$ and $150$\,W, we co-register both captures in one COLMAP frame.
Then, to obtain the temperature of the object's surface, we follow \citeauthor{minkinaInfraredThermography2009}~\shortcite{minkinaInfraredThermography2009} and invert the camera's calibrated Planck-type model in each thermal image, with that image's emissivity, object distance and reflected temperature. No camera sees the plate contact, so we close the hole at the bottom with samples of the cloud's convex hull that no depth map sees and that lie inside the visual hull of the object masks, oriented outward, about a thousand points per capture.

\section{Details of the FEM mesh-based pipeline}
\label{sec:mesh-pipeline}

Besides the parameters in \cref{ssec:e2e-comparison}, the mesh-based pipeline uses:
\begin{itemize}
\item \emph{fusion}: voxels of $0.2\%$ of the object's extent, truncated at $8$ voxels, keeping the largest connected component of the zero level set;
\item \emph{boundary data}: each mesh face takes the seen label of the nearest cloud point, and a seen vertex the median temperature of the four most frontal views in which a depth test on the mesh finds it visible;
\item \emph{linear solver}: a direct LU factorization.
\end{itemize}
For the propeller of \cref{ssec:crack}, fTetWild takes $3$ minutes for $86$K tetrahedra at an edge length of $1\%$ of the bounding-box diagonal, $13$ minutes for $0.51$M at $0.5\%$, the size of the crack's reference solution, and $20$ minutes for $1.4$M at $0.35\%$ ($16$ cores of an AMD EPYC 7763, one run each).

\section{Boundary element solver on the reconstructed meshes}
\label{sec:mesh-bem}
A boundary element solve on the meshes of the image-based meshers of \cref{fig:meshers} avoids the tetrahedralization of the mesh pipeline (\cref{ssec:e2e-comparison}), but not the meshing.
We compare against such a solve on \textsc{Voronoi} with full coverage, accelerated as Chen et al.~\shortcite{chenLightningfastBoundaryElement2025} propose and collocated at the mesh vertices with their normals and areas and the boundary data of the finite element pipeline.
We run it with the kernels, preconditioner and evaluation of our method (\cref{par:solving}), so that the two differ only in the geometry: the mesh vertices against our points.
As the meshers produce them, the meshes are poor boundary element meshes: with $0.5$ to $1.6$ million vertices, $10$ to $30$ times our point count, $7$ to $9\%$ of their triangles are slivers with an angle below $5^\circ$, and the mesh of TopoSurfel has about $2{,}000$ holes and a hundred floating pieces.
We therefore repair them, as we do for the finite element pipeline.
Then, we remesh them isotropically to about our point count, $56$K to $61$K vertices.
On the remeshed meshes of RaDe-GS, MILo and TopoSurfel, the boundary element solve has an error of $0.48$\%, $0.84$\% and $4.3$\% of the temperature range, $2$ to $19$ times our  $0.23$\% (\cref{fig:mesh-bem}).
A boundary element solve thus inherits the errors of its mesh: the material the mesh loses where it thins the struts or adds where it closes the openings between them, and the boundary data read at depth discontinuities (\cref{ssec:e2e-comparison}).

\input{figures/mesh-bem/mesh-bem-snippet}

\paragraph{Timing}
With the setup of \cref{ssec:e2e-comparison}, on \textsc{Armadillo}, the BEM pipeline shares the first stages of the finite element pipeline.
RaDe-GS outputs splats, not a mesh: the mesh is fused from the depth maps that the model renders at the training views and then repaired ($36$\,s), and the boundary data are transferred to it ($36$\,s).
We then remesh it to our point count ($33$\,s) and solve and evaluate the interior ($0.7$\,s), $105$\,s in all after the reconstruction, against $31$\,s for ours and $210$\,s for the finite element pipeline.

\clearpage

%% file: figures/bie-fields/bie-fields-snippet.tex
\ifdefined\bfKeySolid\else\newsavebox{\bfKeySolid}\newsavebox{\bfKeyDashed}\fi
\sbox{\bfKeySolid}{\tikz[baseline=-0.55ex]\draw[line width=0.42pt] (0,0) -- (9pt,0);}
\sbox{\bfKeyDashed}{\tikz[baseline=-0.55ex]\draw[line width=0.42pt, dash pattern=on 1.45pt off 0.95pt] (0,0) -- (9pt,0);}
\begin{figure}[hb]
\centering
\begin{tikzpicture}[every node/.style={inner sep=0, outer sep=0}, lab/.style={font=\footnotesize, anchor=south}, prop/.style={font=\footnotesize, anchor=north}]
  \node[anchor=north west] (S) at (0, 0) {\includegraphics[width=69pt]{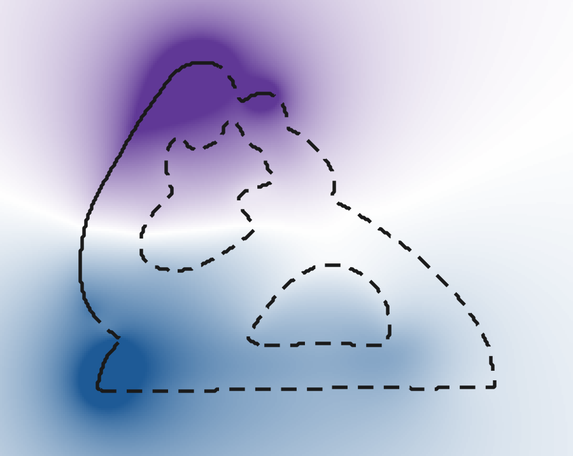}};
  \node[anchor=north west] (D) at ([xshift=4pt]S.north east) {\includegraphics[width=69pt]{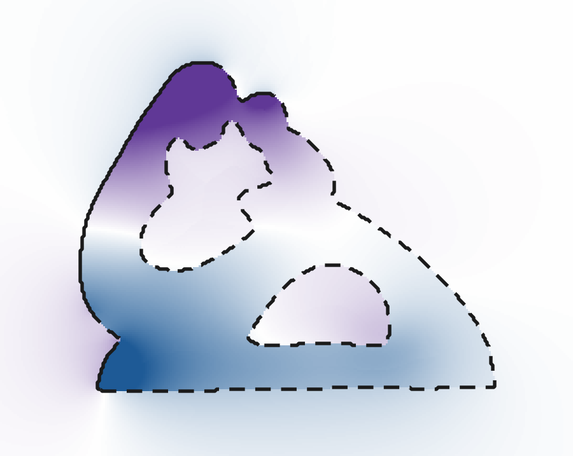}};
  \node[anchor=north west] (R) at ([xshift=4pt]D.north east) {\includegraphics[width=69pt]{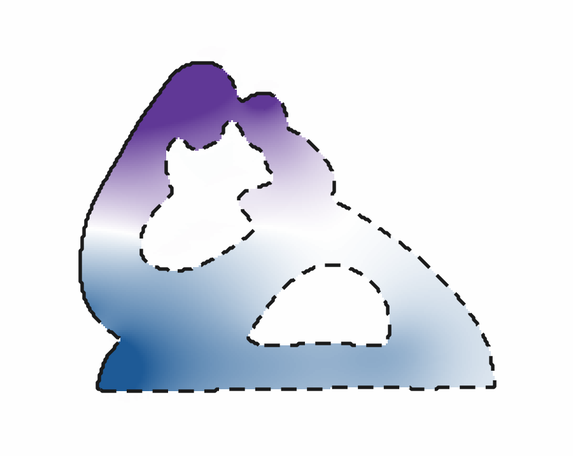}};
  \node[anchor=west] at ([xshift=3pt]R.east) {\includegraphics{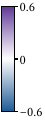}};
  \node[lab] at ([yshift=2.5pt]S.north) {single layer, $\sigma$};
  \node[lab] at ([yshift=2.5pt]D.north) {double layer, $\mu$};
  \node[lab] at ([yshift=2.5pt]R.north) {direct, $u$ and $q$};
  \node[prop] at ([yshift=-2.5pt]S.south) {continuous across $\Gamma$};
  \node[prop] at ([yshift=-2.5pt]D.south) {jumps by $\mu$ across $\Gamma$};
  \node[prop] at ([yshift=-2.5pt]R.south) {zero outside $\Omega$};
  \node[prop] at ([yshift=-13pt]D.south) {\usebox{\bfKeySolid}~observed (Dirichlet)\hspace{1.2em}\usebox{\bfKeyDashed}~unobserved (Robin)};
\end{tikzpicture}
\caption{The three BIE formulations on a mixed problem with Dirichlet and Robin boundary conditions. Inside, all three formulations give the same solution; outside, the single-layer potential continues across the surface, the double-layer potential jumps by its density, and the direct formulation vanishes.}
\label{fig:bie-fields}
\end{figure}
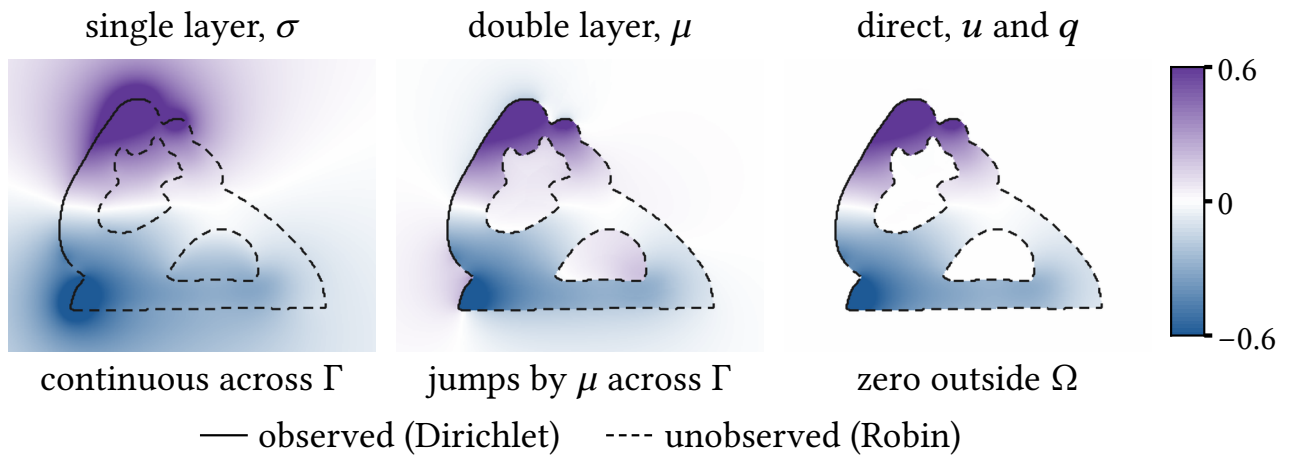

%% file: figures/mesh-bem/mesh-bem-snippet.tex
\begin{figure}[t]
\centering
\definecolor{mwpurple}{HTML}{9467BD}%
\begin{tikzpicture}[every node/.style={inner sep=0, outer sep=0}, pn/.style={anchor=north west, draw=black!30, line width=0.4pt},
  rl/.style={font=\footnotesize, rotate=90, anchor=base}, mae/.style={font=\scriptsize, anchor=base}]
  \pgfmathsetlengthmacro{\pw}{(\columnwidth-21pt)/5}\def\gap{2pt}
  \node[anchor=north west] (T0) at (0,0) {\includegraphics[width=\pw]{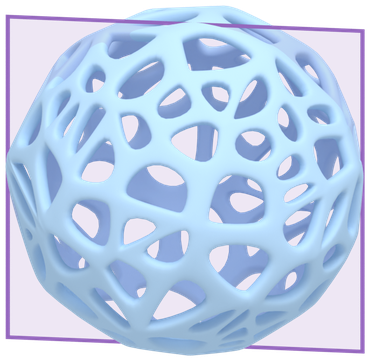}};
  \node[pn] (T1) at ([yshift=-\gap]T0.south west) {\includegraphics[width=\pw]{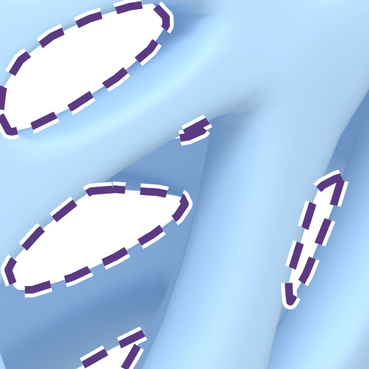}};
  \node[anchor=north west] (T2) at ([yshift=-\gap]T1.south west) {\includegraphics[width=\pw, height=\pw]{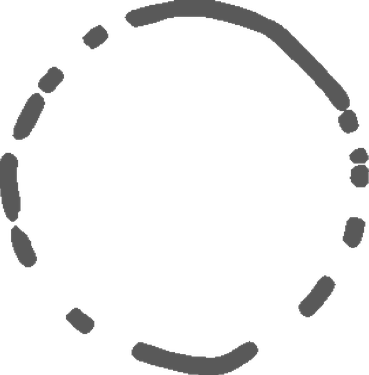}};
    \node[anchor=north west] (R0) at ([xshift=\gap]T0.north east) {\includegraphics[width=\pw]{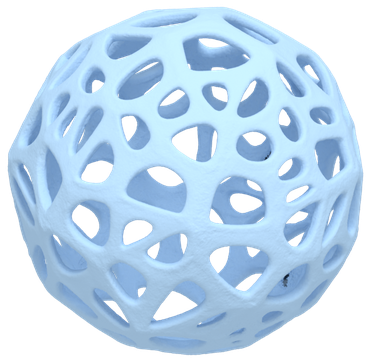}};
    \node[pn] (R1) at ([yshift=-\gap]R0.south west) {\includegraphics[width=\pw]{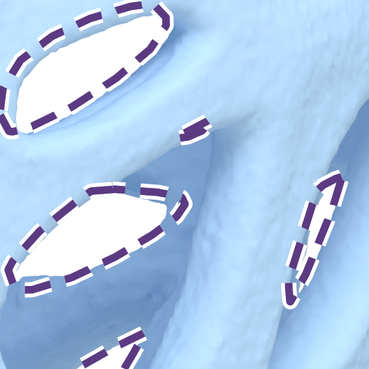}};
    \node[anchor=north west] (M0) at ([xshift=\gap]R0.north east) {\includegraphics[width=\pw]{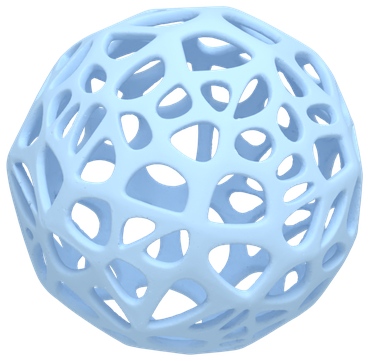}};
    \node[pn] (M1) at ([yshift=-\gap]M0.south west) {\includegraphics[width=\pw]{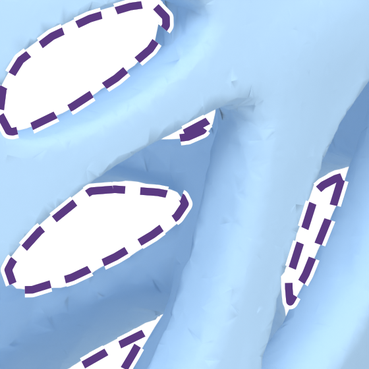}};
    \node[anchor=north west] (P0) at ([xshift=\gap]M0.north east) {\includegraphics[width=\pw]{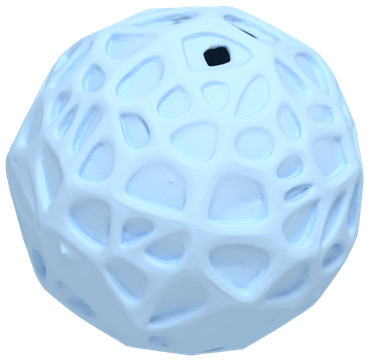}};
    \node[pn] (P1) at ([yshift=-\gap]P0.south west) {\includegraphics[width=\pw]{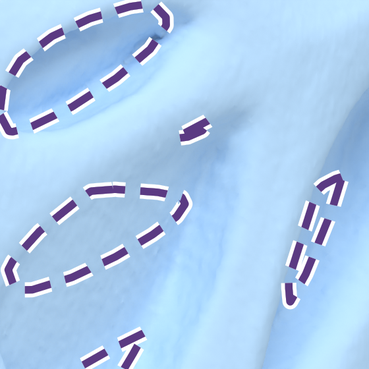}};
    \node[anchor=north west] (O0) at ([xshift=\gap]P0.north east) {\includegraphics[width=\pw]{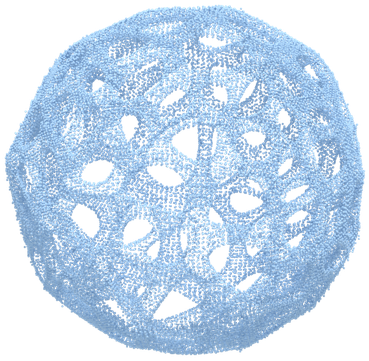}};
    \node[pn] (O1) at ([yshift=-\gap]O0.south west) {\includegraphics[width=\pw]{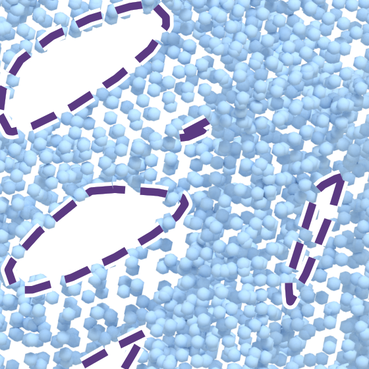}};
  \node[anchor=north west] (R2) at ([yshift=-\gap]R1.south west) {\includegraphics[width=\pw]{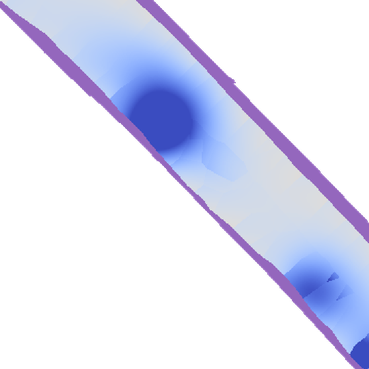}};
  \node[anchor=north west] (M2) at ([yshift=-\gap]M1.south west) {\includegraphics[width=\pw]{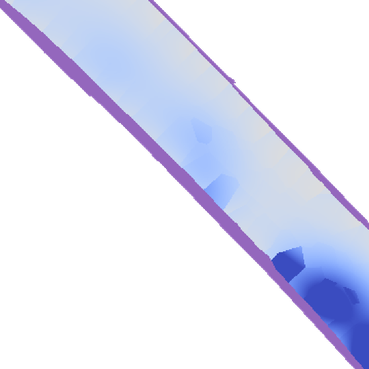}};
  \node[anchor=north west] (P2) at ([yshift=-\gap]P1.south west) {\includegraphics[width=\pw]{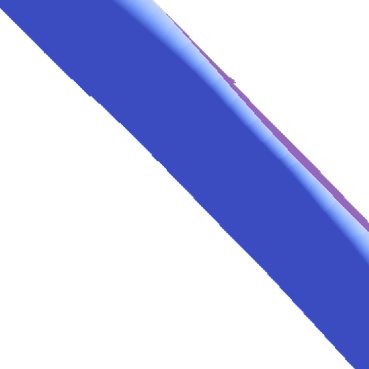}};
  \node[anchor=north west] (O2) at ([yshift=-\gap]O1.south west) {\includegraphics[width=\pw]{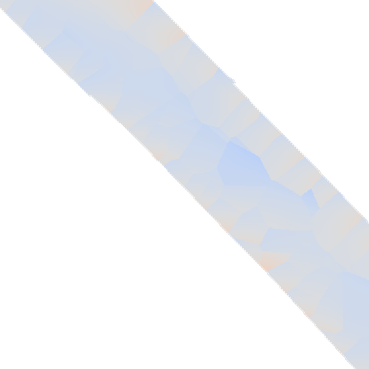}};
  \foreach \n in {T2, R2, M2, P2, O2} {\draw[mwpurple, line width=0.5pt] (\n.south west) rectangle (\n.north east);}
  \foreach \c in {T, R, M, P, O} {
    \begin{scope}[shift={(\c0.south west)}, x={($(\c0.south east)-(\c0.south west)$)}, y={($(\c0.north west)-(\c0.south west)$)}]
      \draw[black!60, line width=0.5pt] (0.7693,0.3433) rectangle ++(0.1460,0.1493);
      \coordinate (wl-\c) at (0.7693,0.3433); \coordinate (wr-\c) at (0.9153,0.3433);
    \end{scope}
  }
  \draw[black!60, line width=0.5pt, densely dotted] (wl-T) -- (T1.north west) (wr-T) -- (T1.north east);
  \begin{scope}[shift={(T2.south west)}, x={($(T2.south east)-(T2.south west)$)}, y={($(T2.north west)-(T2.south west)$)}]
    \draw[mwpurple, line width=0.6pt] (0.7627,0.7592) rectangle ++(0.1220,0.1200);
    \coordinate (st) at (0.8847,0.8792); \coordinate (sb) at (0.8847,0.7592);
  \end{scope}
  \draw[mwpurple, line width=0.5pt, densely dotted] (st) -- (R2.north west) (sb) -- (R2.south west);
  \foreach \p in {R2, M2} {
    \begin{scope}[shift={(\p.south west)}, x={($(\p.south east)-(\p.south west)$)}, y={($(\p.north west)-(\p.south west)$)}]
      \node[font=\scriptsize, anchor=south west, align=left, inner sep=0] at (0.03,0.03) {geometry\\[-1.5pt]error};
      \draw[-{Stealth[length=2.2pt, width=2pt]}, line width=0.45pt] (0.48,0.285) -- (0.48,0.508);
    \end{scope}
  }
  \foreach \n/\t in {T0/{ground truth}, R0/{RaDe-GS}, M0/{MILo}, P0/{TopoSurfel}, O0/{ours}} {\node[font=\footnotesize, anchor=base] at ([yshift=3pt]\n.north) {\t};}
  \foreach \n/\t in {T0/{full view}, T1/{close-up}, T2/{BEM}} {\node[rl] at ([xshift=-3pt]\n.west) {\t};}
  \foreach \n/\t in {R2/{$0.0048$}, M2/{$0.0084$}, P2/{$0.043$}, O2/{$0.0023$}} {\node[mae] at ([yshift=-7.5pt]\n.south) {\t};}
  \node[mae] at ([yshift=-7.5pt]T2.south) {MAE};
  \node[anchor=north] (CB) at ([yshift=-11pt]$(M2.south)!0.5!(P2.south)$) {\includegraphics{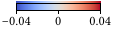}};
  \node[font=\scriptsize] at ($(T2.south west)!0.5!(T2.north east)$) {slice};
  \node[font=\scriptsize, anchor=east] at ([xshift=-4pt, yshift=-3pt]CB.north west) {temperature error};
\end{tikzpicture}
\caption{A boundary element solve on the meshes of the three image-based meshers, against ours on \textsc{Voronoi}. It inherits the meshing errors.}
\label{fig:mesh-bem}
\end{figure}
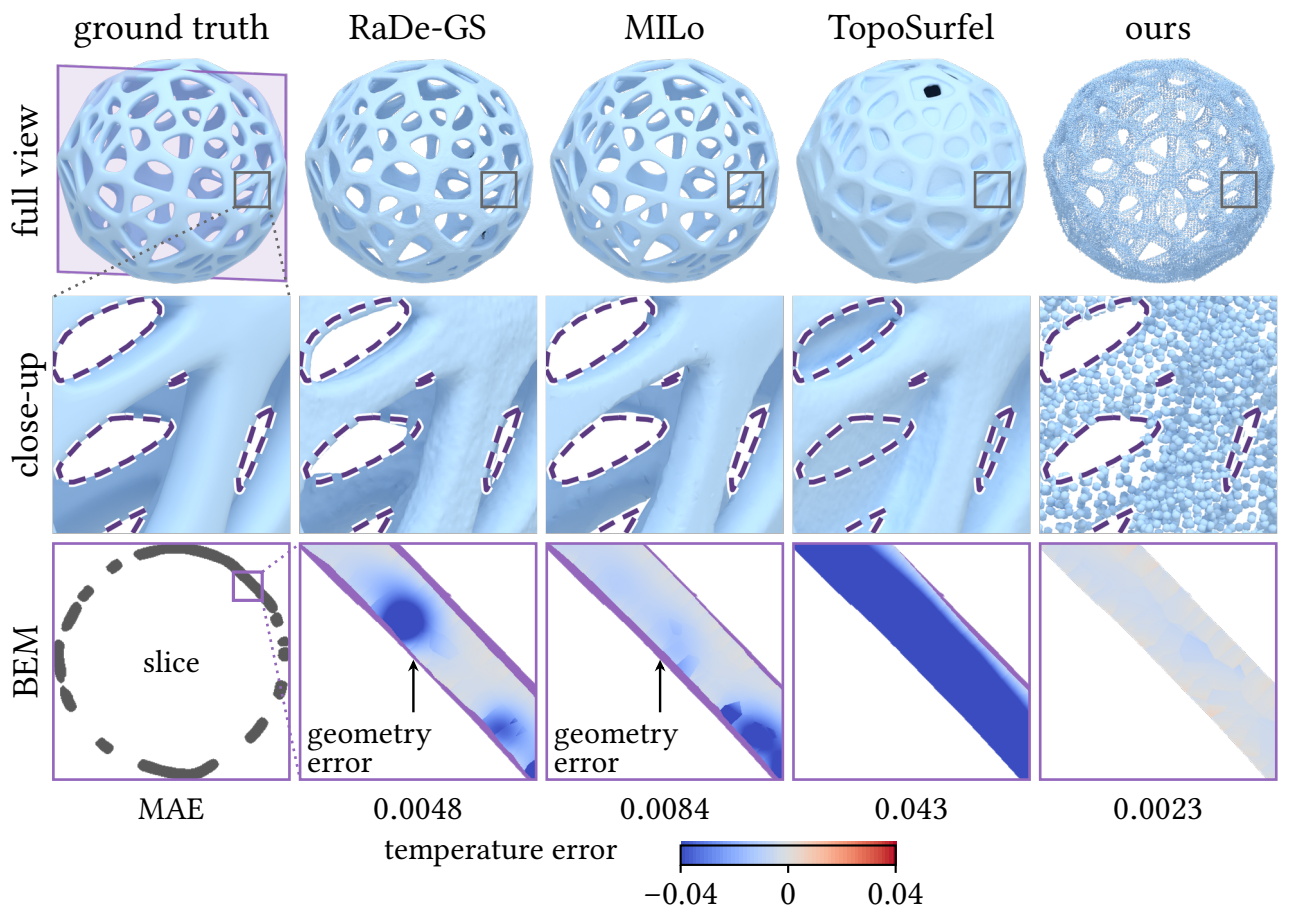

%% file: main.bbl

\begin{thebibliography}{82}


\ifx \showCODEN    \undefined \def \showCODEN     #1{\unskip}     \fi
\ifx \showISBNx    \undefined \def \showISBNx     #1{\unskip}     \fi
\ifx \showISBNxiii \undefined \def \showISBNxiii  #1{\unskip}     \fi
\ifx \showISSN     \undefined \def \showISSN      #1{\unskip}     \fi
\ifx \showLCCN     \undefined \def \showLCCN      #1{\unskip}     \fi
\ifx \shownote     \undefined \def \shownote      #1{#1}          \fi
\ifx \showarticletitle \undefined \def \showarticletitle #1{#1}   \fi
\ifx \showURL      \undefined \def \showURL       {\relax}        \fi
\providecommand\bibfield[2]{#2}
\providecommand\bibinfo[2]{#2}
\providecommand\natexlab[1]{#1}
\providecommand\showeprint[2][]{arXiv:#2}

\bibitem[Aguerre and Fern{\'a}ndez(2021)]%
        {pedroaguerreThRendRayTracing2021}
\bibfield{author}{\bibinfo{person}{Jos{\'e}~Pedro Aguerre} {and}
  \bibinfo{person}{Eduardo Fern{\'a}ndez}.} \bibinfo{year}{2021}\natexlab{}.
\newblock \showarticletitle{{ThRend}: a ray tracing module for infrared
  rendering of urban scenes} \emph{(\bibinfo{series}{Building {Simulation}},
  Vol.~\bibinfo{volume}{17})}. \bibinfo{publisher}{IBPSA},
  \bibinfo{pages}{1588--1595}.
\newblock
\showISBNx{978-1-7750520-2-9}
\href{https://doi.org/10.26868/25222708.2021.30435}{doi:\nolinkurl{10.26868/25222708.2021.30435}}


\bibitem[ANSYS(2024)]%
        {ansysAnsys2024}
\bibfield{author}{\bibinfo{person}{Inc. ANSYS}.}
  \bibinfo{year}{2024}\natexlab{}.
\newblock \bibinfo{title}{Ansys}.
\newblock
\urldef\tempurl%
\url{https://www.ansys.com}
\showURL{%
\tempurl}


\bibitem[Baratta et~al\mbox{.}(2023)]%
        {barattaDOLFINxNextGeneration2023}
\bibfield{author}{\bibinfo{person}{Igor~A. Baratta}, \bibinfo{person}{Joseph~P.
  Dean}, \bibinfo{person}{Jørgen~S. Dokken}, \bibinfo{person}{Michal Habera},
  \bibinfo{person}{Jack~S. Hale}, \bibinfo{person}{Chris~N. Richardson},
  \bibinfo{person}{Marie~E. Rognes}, \bibinfo{person}{Matthew~W. Scroggs},
  \bibinfo{person}{Nathan Sime}, {and} \bibinfo{person}{Garth~N. Wells}.}
  \bibinfo{year}{2023}\natexlab{}.
\newblock \bibinfo{title}{{DOLFINx}: {The} next generation {FEniCS} problem
  solving environment}.
\newblock
\href{https://doi.org/10.5281/zenodo.10447666}{doi:\nolinkurl{10.5281/zenodo.10447666}}


\bibitem[Barill et~al\mbox{.}(2018)]%
        {barillFastWindingNumbers2018}
\bibfield{author}{\bibinfo{person}{Gavin Barill}, \bibinfo{person}{Neil~G.
  Dickson}, \bibinfo{person}{Ryan Schmidt}, \bibinfo{person}{David I.~W.
  Levin}, {and} \bibinfo{person}{Alec Jacobson}.}
  \bibinfo{year}{2018}\natexlab{}.
\newblock \showarticletitle{Fast winding numbers for soups and clouds}.
\newblock \bibinfo{journal}{\emph{ACM Transactions on Graphics}}
  \bibinfo{volume}{37}, \bibinfo{number}{4} (\bibinfo{year}{2018}),
  \bibinfo{pages}{43:1--43:12}.
\newblock
\href{https://doi.org/10.1145/3197517.3201337}{doi:\nolinkurl{10.1145/3197517.3201337}}


\bibitem[Bati et~al\mbox{.}(2023)]%
        {batiCouplingConductionConvection2023}
\bibfield{author}{\bibinfo{person}{Mégane Bati}, \bibinfo{person}{Stéphane
  Blanco}, \bibinfo{person}{Christophe Coustet}, \bibinfo{person}{Vincent
  Eymet}, \bibinfo{person}{Vincent Forest}, \bibinfo{person}{Richard Fournier},
  \bibinfo{person}{Jacques Gautrais}, \bibinfo{person}{Nicolas Mellado},
  \bibinfo{person}{Mathias Paulin}, {and} \bibinfo{person}{Benjamin Piaud}.}
  \bibinfo{year}{2023}\natexlab{}.
\newblock \showarticletitle{Coupling {Conduction}, {Convection} and {Radiative}
  {Transfer} in a {Single} {Path}-{Space}: {Application} to {Infrared}
  {Rendering}}.
\newblock \bibinfo{journal}{\emph{ACM Transactions on Graphics}}
  \bibinfo{volume}{42}, \bibinfo{number}{4} (\bibinfo{date}{Aug.}
  \bibinfo{year}{2023}), \bibinfo{pages}{1--20}.
\newblock
\showISSN{0730-0301, 1557-7368}
\href{https://doi.org/10.1145/3592121}{doi:\nolinkurl{10.1145/3592121}}


\bibitem[Beale and Tlupova(2024)]%
        {bealeExtrapolatedRegularization2024}
\bibfield{author}{\bibinfo{person}{J.~Thomas Beale} {and}
  \bibinfo{person}{Svetlana Tlupova}.} \bibinfo{year}{2024}\natexlab{}.
\newblock \showarticletitle{Extrapolated regularization of nearly singular
  integrals on surfaces}.
\newblock \bibinfo{journal}{\emph{Advances in Computational Mathematics}}
  \bibinfo{volume}{50} (\bibinfo{year}{2024}), \bibinfo{pages}{61}.
\newblock
\href{https://doi.org/10.1007/s10444-024-10161-4}{doi:\nolinkurl{10.1007/s10444-024-10161-4}}


\bibitem[Beale et~al\mbox{.}(2016)]%
        {bealeSimpleMethodComputing2016}
\bibfield{author}{\bibinfo{person}{J.~Thomas Beale}, \bibinfo{person}{Wenjun
  Ying}, {and} \bibinfo{person}{Jason~R. Wilson}.}
  \bibinfo{year}{2016}\natexlab{}.
\newblock \showarticletitle{A Simple Method for Computing Singular or Nearly
  Singular Integrals on Closed Surfaces}.
\newblock \bibinfo{journal}{\emph{Communications in Computational Physics}}
  \bibinfo{volume}{20}, \bibinfo{number}{3} (\bibinfo{year}{2016}),
  \bibinfo{pages}{733--753}.
\newblock
\href{https://doi.org/10.4208/cicp.030815.240216a}{doi:\nolinkurl{10.4208/cicp.030815.240216a}}


\bibitem[Beaufort et~al\mbox{.}(2022)]%
        {beaufortHexMeIfYou2022}
\bibfield{author}{\bibinfo{person}{Pierre-Alexandre Beaufort},
  \bibinfo{person}{Maxence Reberol}, \bibinfo{person}{Denis Kalmykov},
  \bibinfo{person}{Heng Liu}, \bibinfo{person}{Franck Ledoux}, {and}
  \bibinfo{person}{David Bommes}.} \bibinfo{year}{2022}\natexlab{}.
\newblock \showarticletitle{Hex {{Me If You Can}}}.
\newblock \bibinfo{journal}{\emph{Computer Graphics Forum}}
  \bibinfo{volume}{41}, \bibinfo{number}{5} (\bibinfo{year}{2022}),
  \bibinfo{pages}{125--134}.
\newblock
\href{https://doi.org/10.1111/cgf.14608}{doi:\nolinkurl{10.1111/cgf.14608}}


\bibitem[Beck et~al\mbox{.}(1985)]%
        {beckInverseHeatConduction1985}
\bibfield{author}{\bibinfo{person}{James~V. Beck}, \bibinfo{person}{Ben
  Blackwell}, {and} \bibinfo{person}{Charles~R. St.~Clair, Jr.}}
  \bibinfo{year}{1985}\natexlab{}.
\newblock \bibinfo{booktitle}{\emph{Inverse Heat Conduction: Ill-Posed
  Problems}}.
\newblock \bibinfo{publisher}{Wiley-Interscience}, \bibinfo{address}{New York}.
\newblock


\bibitem[Bian et~al\mbox{.}(2026)]%
        {bianModelingDepthAmbiguity2026}
\bibfield{author}{\bibinfo{person}{Siyuan Bian}, \bibinfo{person}{Congrong Xu},
  {and} \bibinfo{person}{Jun Gao}.} \bibinfo{year}{2026}\natexlab{}.
\newblock \bibinfo{title}{Modeling Depth Ambiguity: A Mixture-Density
  Representation for Flying-Point-Free Depth Estimation}.
\newblock
\showeprint[arxiv]{2606.02552}~[cs.CV]
\urldef\tempurl%
\url{https://arxiv.org/abs/2606.02552}
\showURL{%
\tempurl}


\bibitem[{Blender Foundation}(2025)]%
        {blender45}
\bibfield{author}{\bibinfo{person}{{Blender Foundation}}.}
  \bibinfo{year}{2025}\natexlab{}.
\newblock \bibinfo{title}{Blender 4.5}.
\newblock \bibinfo{howpublished}{\url{https://www.blender.org}}.
\newblock


\bibitem[Brebbia et~al\mbox{.}(1984)]%
        {brebbiaBoundaryElementTechniques1984}
\bibfield{author}{\bibinfo{person}{Carlos~A. Brebbia},
  \bibinfo{person}{Jos{\'e} C.~F. Telles}, {and} \bibinfo{person}{Luiz~C.
  Wrobel}.} \bibinfo{year}{1984}\natexlab{}.
\newblock \bibinfo{booktitle}{\emph{Boundary Element Techniques: Theory and
  Applications in Engineering}}.
\newblock \bibinfo{publisher}{Springer}, \bibinfo{address}{Berlin, Heidelberg}.
\newblock
\href{https://doi.org/10.1007/978-3-642-48860-3}{doi:\nolinkurl{10.1007/978-3-642-48860-3}}


\bibitem[Carion et~al\mbox{.}(2026)]%
        {carionSAM3Segment2026}
\bibfield{author}{\bibinfo{person}{Nicolas Carion}, \bibinfo{person}{Laura
  Gustafson}, \bibinfo{person}{Yuan-Ting Hu}, \bibinfo{person}{Shoubhik
  Debnath}, \bibinfo{person}{Ronghang Hu}, \bibinfo{person}{Didac Suris},
  \bibinfo{person}{Chaitanya Ryali}, \bibinfo{person}{Kalyan~Vasudev Alwala},
  \bibinfo{person}{Haitham Khedr}, \bibinfo{person}{Andrew Huang},
  \bibinfo{person}{Jie Lei}, \bibinfo{person}{Tengyu Ma},
  \bibinfo{person}{Baishan Guo}, \bibinfo{person}{Arpit Kalla},
  \bibinfo{person}{Markus Marks}, \bibinfo{person}{Joseph Greer},
  \bibinfo{person}{Meng Wang}, \bibinfo{person}{Peize Sun},
  \bibinfo{person}{Roman Rädle}, \bibinfo{person}{Triantafyllos Afouras},
  \bibinfo{person}{Effrosyni Mavroudi}, \bibinfo{person}{Katherine Xu},
  \bibinfo{person}{Tsung-Han Wu}, \bibinfo{person}{Yu Zhou},
  \bibinfo{person}{Liliane Momeni}, \bibinfo{person}{Rishi Hazra},
  \bibinfo{person}{Shuangrui Ding}, \bibinfo{person}{Sagar Vaze},
  \bibinfo{person}{Francois Porcher}, \bibinfo{person}{Feng Li},
  \bibinfo{person}{Siyuan Li}, \bibinfo{person}{Aishwarya Kamath},
  \bibinfo{person}{Ho~Kei Cheng}, \bibinfo{person}{Piotr Dollár},
  \bibinfo{person}{Nikhila Ravi}, \bibinfo{person}{Kate Saenko},
  \bibinfo{person}{Pengchuan Zhang}, {and} \bibinfo{person}{Christoph
  Feichtenhofer}.} \bibinfo{year}{2026}\natexlab{}.
\newblock \bibinfo{title}{{SAM} 3: {Segment} {Anything} with {Concepts}}.
\newblock
\shownote{arXiv:2511.16719 [cs]}.
\newblock
\href{https://doi.org/10.48550/arXiv.2511.16719}{doi:\nolinkurl{10.48550/arXiv.2511.16719}}


\bibitem[Carslaw and Jaeger(1959)]%
        {carslawConductionHeatSolids1959}
\bibfield{author}{\bibinfo{person}{H.~S. Carslaw} {and} \bibinfo{person}{J.~C
  Jaeger}.} \bibinfo{year}{1959}\natexlab{}.
\newblock \bibinfo{booktitle}{\emph{Conduction {Of} {Heat} {In} {Solids}}
  (\bibinfo{edition}{2nd} ed.)}.
\newblock \bibinfo{publisher}{Oxford University Press, U.S.A.},
  \bibinfo{address}{Oxford}.
\newblock


\bibitem[Carvalho et~al\mbox{.}(2020)]%
        {carvalhoAsymptoticApproximationsClose2020}
\bibfield{author}{\bibinfo{person}{Camille Carvalho}, \bibinfo{person}{Shilpa
  Khatri}, {and} \bibinfo{person}{Arnold~D. Kim}.}
  \bibinfo{year}{2020}\natexlab{}.
\newblock \showarticletitle{Asymptotic Approximations for the Close Evaluation
  of Double-Layer Potentials}.
\newblock \bibinfo{journal}{\emph{SIAM Journal on Scientific Computing}}
  \bibinfo{volume}{42}, \bibinfo{number}{1} (\bibinfo{year}{2020}),
  \bibinfo{pages}{A504--A533}.
\newblock
\href{https://doi.org/10.1137/18M1218698}{doi:\nolinkurl{10.1137/18M1218698}}


\bibitem[Chati et~al\mbox{.}(1999)]%
        {chatiBoundaryNodeMethod1999}
\bibfield{author}{\bibinfo{person}{Mandar~K. Chati}, \bibinfo{person}{Subrata
  Mukherjee}, {and} \bibinfo{person}{Yu~Xie Mukherjee}.}
  \bibinfo{year}{1999}\natexlab{}.
\newblock \showarticletitle{The boundary node method for three-dimensional
  linear elasticity}.
\newblock \bibinfo{journal}{\emph{Internat. J. Numer. Methods Engrg.}}
  \bibinfo{volume}{46}, \bibinfo{number}{8} (\bibinfo{year}{1999}),
  \bibinfo{pages}{1163--1184}.
\newblock


\bibitem[Chen et~al\mbox{.}(2024a)]%
        {chenPGSRPlanarBased2024}
\bibfield{author}{\bibinfo{person}{Danpeng Chen}, \bibinfo{person}{Hai Li},
  \bibinfo{person}{Weicai Ye}, \bibinfo{person}{Yifan Wang},
  \bibinfo{person}{Weijian Xie}, \bibinfo{person}{Shangjin Zhai},
  \bibinfo{person}{Nan Wang}, \bibinfo{person}{Haomin Liu},
  \bibinfo{person}{Hujun Bao}, {and} \bibinfo{person}{Guofeng Zhang}.}
  \bibinfo{year}{2024}\natexlab{a}.
\newblock \showarticletitle{{PGSR}: Planar-based {Gaussian} splatting for
  efficient and high-fidelity surface reconstruction}.
\newblock \bibinfo{journal}{\emph{IEEE Transactions on Visualization and
  Computer Graphics}} (\bibinfo{year}{2024}).
\newblock
\href{https://doi.org/10.1109/TVCG.2024.3494046}{doi:\nolinkurl{10.1109/TVCG.2024.3494046}}


\bibitem[Chen et~al\mbox{.}(2024b)]%
        {chen3DReconstructionFast2024}
\bibfield{author}{\bibinfo{person}{Hanyu Chen}, \bibinfo{person}{Bailey
  Miller}, {and} \bibinfo{person}{Ioannis Gkioulekas}.}
  \bibinfo{year}{2024}\natexlab{b}.
\newblock \showarticletitle{{3D} {Reconstruction} with {Fast} {Dipole} {Sums}}.
\newblock \bibinfo{journal}{\emph{ACM Trans. Graph.}} \bibinfo{volume}{43},
  \bibinfo{number}{6} (\bibinfo{date}{Nov.} \bibinfo{year}{2024}),
  \bibinfo{pages}{192:1--192:19}.
\newblock
\showISSN{0730-0301}
\href{https://doi.org/10.1145/3687914}{doi:\nolinkurl{10.1145/3687914}}


\bibitem[Chen et~al\mbox{.}(2024c)]%
        {chenLightningfastMethodFundamental2024}
\bibfield{author}{\bibinfo{person}{Jiong Chen}, \bibinfo{person}{Florian
  Schaefer}, {and} \bibinfo{person}{Mathieu Desbrun}.}
  \bibinfo{year}{2024}\natexlab{c}.
\newblock \showarticletitle{Lightning-fast {Method} of {Fundamental}
  {Solutions}}.
\newblock \bibinfo{journal}{\emph{ACM Trans. Graph.}} \bibinfo{volume}{43},
  \bibinfo{number}{4} (\bibinfo{date}{July} \bibinfo{year}{2024}),
  \bibinfo{pages}{77:1--77:16}.
\newblock
\showISSN{0730-0301}
\href{https://doi.org/10.1145/3658199}{doi:\nolinkurl{10.1145/3658199}}


\bibitem[Chen et~al\mbox{.}(2025)]%
        {chenLightningfastBoundaryElement2025}
\bibfield{author}{\bibinfo{person}{Jiong Chen}, \bibinfo{person}{Florian
  Schäfer}, {and} \bibinfo{person}{Mathieu Desbrun}.}
  \bibinfo{year}{2025}\natexlab{}.
\newblock \showarticletitle{Lightning-fast {Boundary} {Element} {Method}}.
\newblock \bibinfo{journal}{\emph{ACM Transactions on Graphics}}
  \bibinfo{volume}{44}, \bibinfo{number}{4} (\bibinfo{date}{Aug.}
  \bibinfo{year}{2025}), \bibinfo{pages}{1--14}.
\newblock
\href{https://doi.org/10.1145/3731196}{doi:\nolinkurl{10.1145/3731196}}


\bibitem[Chen and Wang(2010)]%
        {chenSingularBoundaryMethod2010}
\bibfield{author}{\bibinfo{person}{Wen Chen} {and} \bibinfo{person}{Fuzhang
  Wang}.} \bibinfo{year}{2010}\natexlab{}.
\newblock \showarticletitle{A method of fundamental solutions without
  fictitious boundary}.
\newblock \bibinfo{journal}{\emph{Engineering Analysis with Boundary Elements}}
  \bibinfo{volume}{34}, \bibinfo{number}{5} (\bibinfo{year}{2010}),
  \bibinfo{pages}{530--532}.
\newblock
\href{https://doi.org/10.1016/j.enganabound.2009.12.002}{doi:\nolinkurl{10.1016/j.enganabound.2009.12.002}}


\bibitem[Chew et~al\mbox{.}(2025)]%
        {chewAcousticReliefs2025}
\bibfield{author}{\bibinfo{person}{Jeremy Chew}, \bibinfo{person}{Michal
  Piovarci}, \bibinfo{person}{Kangrui Xue}, \bibinfo{person}{Doug James}, {and}
  \bibinfo{person}{Bernd Bickel}.} \bibinfo{year}{2025}\natexlab{}.
\newblock \showarticletitle{Acoustic Reliefs}.
\newblock \bibinfo{journal}{\emph{ACM Transactions on Graphics}}
  \bibinfo{volume}{44}, \bibinfo{number}{6} (\bibinfo{year}{2025}),
  \bibinfo{pages}{1--23}.
\newblock
\href{https://doi.org/10.1145/3763287}{doi:\nolinkurl{10.1145/3763287}}


\bibitem[Cignoni et~al\mbox{.}(2025)]%
        {cignoniMeshLab2025072025}
\bibfield{author}{\bibinfo{person}{Paolo Cignoni}, \bibinfo{person}{Alessandro
  Muntoni}, \bibinfo{person}{granzuglia}, \bibinfo{person}{Marco Callieri},
  \bibinfo{person}{Massimiliano Corsini}, \bibinfo{person}{Matteo Dellepiane},
  \bibinfo{person}{ganovelli}, \bibinfo{person}{Rylie Pavlik},
  \bibinfo{person}{Nico Pietroni}, \bibinfo{person}{Gianpaolo Palma},
  \bibinfo{person}{Gabriele Pappalardo}, \bibinfo{person}{Luigi Malomo},
  \bibinfo{person}{Federico Ponchio}, \bibinfo{person}{jmespadero},
  \bibinfo{person}{lvk88}, \bibinfo{person}{Oleg Alexandrov},
  \bibinfo{person}{Andrea Maggiordomo}, \bibinfo{person}{Paolo Fasano},
  \bibinfo{person}{PeC-KAYO}, \bibinfo{person}{Daniele Panozzo},
  \bibinfo{person}{Michele Morisco}, \bibinfo{person}{davidboening},
  \bibinfo{person}{Marco~Di Benedetto}, \bibinfo{person}{Lex van~der Sluijs},
  \bibinfo{person}{Thomas Alderighi}, \bibinfo{person}{Alberto Mardegan},
  \bibinfo{person}{luzpaz}, \bibinfo{person}{StefanBruens}, {and}
  \bibinfo{person}{Chelsea Jaggi}.} \bibinfo{year}{2025}\natexlab{}.
\newblock \bibinfo{title}{{MeshLab}-2025.07}.
\newblock
\href{https://doi.org/10.5281/zenodo.16307665}{doi:\nolinkurl{10.5281/zenodo.16307665}}


\bibitem[{COMSOL}(2024)]%
        {comsol}
\bibfield{author}{\bibinfo{person}{{COMSOL}}.} \bibinfo{year}{2024}\natexlab{}.
\newblock \bibinfo{booktitle}{\emph{COMSOL Multiphysics}}.
\newblock
\urldef\tempurl%
\url{https://www.comsol.com}
\showURL{%
\tempurl}


\bibitem[Dai et~al\mbox{.}(2024)]%
        {daiHighqualitySurfaceReconstruction2024}
\bibfield{author}{\bibinfo{person}{Pinxuan Dai}, \bibinfo{person}{Jiamin Xu},
  \bibinfo{person}{Wenxiang Xie}, \bibinfo{person}{Xinguo Liu},
  \bibinfo{person}{Huamin Wang}, {and} \bibinfo{person}{Weiwei Xu}.}
  \bibinfo{year}{2024}\natexlab{}.
\newblock \showarticletitle{High-Quality Surface Reconstruction Using
  {Gaussian} Surfels}. In \bibinfo{booktitle}{\emph{ACM SIGGRAPH 2024
  Conference Papers}}.
\newblock
\href{https://doi.org/10.1145/3641519.3657441}{doi:\nolinkurl{10.1145/3641519.3657441}}


\bibitem[Eckstein et~al\mbox{.}(2007)]%
        {ecksteinGeneralizedSurfaceFlows2007}
\bibfield{author}{\bibinfo{person}{Ilya Eckstein},
  \bibinfo{person}{Jean-Philippe Pons}, \bibinfo{person}{Yiying Tong},
  \bibinfo{person}{C.-C.~Jay Kuo}, {and} \bibinfo{person}{Mathieu Desbrun}.}
  \bibinfo{year}{2007}\natexlab{}.
\newblock \showarticletitle{Generalized Surface Flows for Mesh Processing}. In
  \bibinfo{booktitle}{\emph{Proceedings of the Fifth {Eurographics} Symposium
  on Geometry Processing}}. \bibinfo{pages}{183--192}.
\newblock


\bibitem[Fan et~al\mbox{.}(2026)]%
        {fan2026toposurfel}
\bibfield{author}{\bibinfo{person}{Chuanjin Fan}, \bibinfo{person}{Wenjie
  Chang}, \bibinfo{person}{Bohao Liao}, \bibinfo{person}{Yujia Chen},
  \bibinfo{person}{Wenfei Yang}, {and} \bibinfo{person}{Tianzhu Zhang}.}
  \bibinfo{year}{2026}\natexlab{}.
\newblock \showarticletitle{{TopoSurfel}: Closing the Loop between {Gaussian}
  Surfels and Meshes for Surface Reconstruction}.
\newblock \bibinfo{journal}{\emph{arXiv preprint arXiv:2608.20687}}
  (\bibinfo{year}{2026}).
\newblock
\urldef\tempurl%
\url{https://fan-treasure.github.io/TopoSurfel_page/}
\showURL{%
\tempurl}


\bibitem[Freude et~al\mbox{.}(2023)]%
        {freudePrecomputedRadiativeHeat2023}
\bibfield{author}{\bibinfo{person}{C. Freude}, \bibinfo{person}{D. Hahn},
  \bibinfo{person}{F. Rist}, \bibinfo{person}{L. Lipp}, {and}
  \bibinfo{person}{M. Wimmer}.} \bibinfo{year}{2023}\natexlab{}.
\newblock \showarticletitle{Precomputed {Radiative} {Heat} {Transport} for
  {Efficient} {Thermal} {Simulation}}.
\newblock \bibinfo{journal}{\emph{Computer Graphics Forum}}
  \bibinfo{volume}{42}, \bibinfo{number}{7} (\bibinfo{year}{2023}),
  \bibinfo{pages}{e14957}.
\newblock
\showISSN{1467-8659}
\href{https://doi.org/10.1111/cgf.14957}{doi:\nolinkurl{10.1111/cgf.14957}}


\bibitem[Freude et~al\mbox{.}(2025)]%
        {freudeInverseSimulationRadiative2025}
\bibfield{author}{\bibinfo{person}{C. Freude}, \bibinfo{person}{L. Lipp},
  \bibinfo{person}{M. Zezulka}, \bibinfo{person}{F. Rist}, \bibinfo{person}{M.
  Wimmer}, {and} \bibinfo{person}{D. Hahn}.} \bibinfo{year}{2025}\natexlab{}.
\newblock \showarticletitle{Inverse {Simulation} of {Radiative} {Thermal}
  {Transport}}.
\newblock \bibinfo{journal}{\emph{Computer Graphics Forum}}
  \bibinfo{volume}{44}, \bibinfo{number}{2} (\bibinfo{year}{2025}),
  \bibinfo{pages}{e70048}.
\newblock
\showISSN{1467-8659}
\href{https://doi.org/10.1111/cgf.70048}{doi:\nolinkurl{10.1111/cgf.70048}}


\bibitem[Gu{\'e}don and Lepetit(2024)]%
        {guedonSuGaRSurfaceAligned2024}
\bibfield{author}{\bibinfo{person}{Antoine Gu{\'e}don} {and}
  \bibinfo{person}{Vincent Lepetit}.} \bibinfo{year}{2024}\natexlab{}.
\newblock \showarticletitle{{SuGaR}: Surface-aligned {Gaussian} splatting for
  efficient {3D} mesh reconstruction and high-quality mesh rendering}. In
  \bibinfo{booktitle}{\emph{Proceedings of the IEEE/CVF Conference on Computer
  Vision and Pattern Recognition (CVPR)}}.
\newblock


\bibitem[Guédon et~al\mbox{.}(2025)]%
        {guedonMILoMeshIntheLoopGaussian2025}
\bibfield{author}{\bibinfo{person}{Antoine Guédon}, \bibinfo{person}{Diego
  Gomez}, \bibinfo{person}{Nissim Maruani}, \bibinfo{person}{Bingchen Gong},
  \bibinfo{person}{George Drettakis}, {and} \bibinfo{person}{Maks Ovsjanikov}.}
  \bibinfo{year}{2025}\natexlab{}.
\newblock \showarticletitle{{MILo}: {Mesh}-{In}-the-{Loop} {Gaussian}
  {Splatting} for {Detailed} and {Efficient} {Surface} {Reconstruction}}.
\newblock \bibinfo{journal}{\emph{ACM Trans. Graph.}} \bibinfo{volume}{44},
  \bibinfo{number}{6} (\bibinfo{date}{Dec.} \bibinfo{year}{2025}),
  \bibinfo{pages}{223:1--223:15}.
\newblock
\href{https://doi.org/10.1145/3763339}{doi:\nolinkurl{10.1145/3763339}}


\bibitem[Hassan et~al\mbox{.}(2025)]%
        {hassanThermoNeRFMultimodalNeural2025}
\bibfield{author}{\bibinfo{person}{Mariam Hassan}, \bibinfo{person}{Florent
  Forest}, \bibinfo{person}{Olga Fink}, {and} \bibinfo{person}{Malcolm
  Mielle}.} \bibinfo{year}{2025}\natexlab{}.
\newblock \showarticletitle{{ThermoNeRF}: {A} multimodal {Neural} {Radiance}
  {Field} for joint {RGB}-thermal novel view synthesis of building facades}.
\newblock \bibinfo{journal}{\emph{Adv. Eng. Inform.}}  \bibinfo{volume}{65}
  (\bibinfo{date}{May} \bibinfo{year}{2025}), \bibinfo{pages}{103345}.
\newblock
\showISSN{1474-0346}
\href{https://doi.org/10.1016/j.aei.2025.103345}{doi:\nolinkurl{10.1016/j.aei.2025.103345}}


\bibitem[Hestenes and Stiefel(1952)]%
        {hestenesMethodsConjugateGradients1952}
\bibfield{author}{\bibinfo{person}{Magnus~R. Hestenes} {and}
  \bibinfo{person}{Eduard Stiefel}.} \bibinfo{year}{1952}\natexlab{}.
\newblock \showarticletitle{Methods of Conjugate Gradients for Solving Linear
  Systems}.
\newblock \bibinfo{journal}{\emph{J. Res. Nat. Bur. Standards}}
  \bibinfo{volume}{49}, \bibinfo{number}{6} (\bibinfo{year}{1952}),
  \bibinfo{pages}{409--436}.
\newblock
\href{https://doi.org/10.6028/jres.049.044}{doi:\nolinkurl{10.6028/jres.049.044}}


\bibitem[Hu et~al\mbox{.}(2020)]%
        {huFastTetrahedralMeshing2020}
\bibfield{author}{\bibinfo{person}{Yixin Hu}, \bibinfo{person}{Teseo
  Schneider}, \bibinfo{person}{Bolun Wang}, \bibinfo{person}{Denis Zorin},
  {and} \bibinfo{person}{Daniele Panozzo}.} \bibinfo{year}{2020}\natexlab{}.
\newblock \showarticletitle{Fast tetrahedral meshing in the wild}.
\newblock \bibinfo{journal}{\emph{ACM Trans. Graph.}} \bibinfo{volume}{39},
  \bibinfo{number}{4} (\bibinfo{date}{Aug.} \bibinfo{year}{2020}),
  \bibinfo{pages}{117:1--117:18}.
\newblock
\showISSN{0730-0301}
\href{https://doi.org/10.1145/3386569.3392385}{doi:\nolinkurl{10.1145/3386569.3392385}}


\bibitem[Huang et~al\mbox{.}(2024)]%
        {huang2DGaussianSplatting2024}
\bibfield{author}{\bibinfo{person}{Binbin Huang}, \bibinfo{person}{Zehao Yu},
  \bibinfo{person}{Anpei Chen}, \bibinfo{person}{Andreas Geiger}, {and}
  \bibinfo{person}{Shenghua Gao}.} \bibinfo{year}{2024}\natexlab{}.
\newblock \showarticletitle{{2D} {Gaussian} {Splatting} for {Geometrically}
  {Accurate} {Radiance} {Fields}}. In \bibinfo{booktitle}{\emph{ACM SIGGRAPH
  2024 Conference Papers}}. \bibinfo{publisher}{Association for Computing
  Machinery}.
\newblock
\href{https://doi.org/10.1145/3641519.3657428}{doi:\nolinkurl{10.1145/3641519.3657428}}


\bibitem[Incropera et~al\mbox{.}(2007)]%
        {incroperaFundamentalsHeatMass2007}
\bibfield{author}{\bibinfo{person}{Frank~P. Incropera},
  \bibinfo{person}{David~P. DeWitt}, \bibinfo{person}{Theodore~L. Bergman},
  {and} \bibinfo{person}{Adrienne~S. Lavine}.} \bibinfo{year}{2007}\natexlab{}.
\newblock \bibinfo{booktitle}{\emph{Fundamentals of Heat and Mass Transfer}
  (\bibinfo{edition}{6th} ed.)}.
\newblock \bibinfo{publisher}{John Wiley \& Sons}.
\newblock


\bibitem[{International Organization for Standardization}(2017)]%
        {iso6946}
\bibfield{author}{\bibinfo{person}{{International Organization for
  Standardization}}.} \bibinfo{year}{2017}\natexlab{}.
\newblock \bibinfo{booktitle}{\emph{{ISO} 6946:2017 {Building} components and
  building elements -- {Thermal} resistance and thermal transmittance --
  {Calculation} methods}}.
\newblock \bibinfo{type}{{T}echnical {R}eport}. \bibinfo{institution}{ISO},
  \bibinfo{address}{Geneva}.
\newblock


\bibitem[James et~al\mbox{.}(2006)]%
        {jamesPrecomputedAcousticTransfer2006}
\bibfield{author}{\bibinfo{person}{Doug~L. James}, \bibinfo{person}{Jernej
  Barbi\v{c}}, {and} \bibinfo{person}{Dinesh~K. Pai}.}
  \bibinfo{year}{2006}\natexlab{}.
\newblock \showarticletitle{Precomputed Acoustic Transfer: Output-Sensitive,
  Accurate Sound Generation for Geometrically Complex Vibration Sources}.
\newblock \bibinfo{journal}{\emph{ACM Transactions on Graphics}}
  \bibinfo{volume}{25}, \bibinfo{number}{3} (\bibinfo{year}{2006}),
  \bibinfo{pages}{987--995}.
\newblock
\href{https://doi.org/10.1145/1141911.1141983}{doi:\nolinkurl{10.1145/1141911.1141983}}


\bibitem[James and Pai(1999)]%
        {jamesArtDefoAccurateReal1999}
\bibfield{author}{\bibinfo{person}{Doug~L. James} {and}
  \bibinfo{person}{Dinesh~K. Pai}.} \bibinfo{year}{1999}\natexlab{}.
\newblock \showarticletitle{{ArtDefo}: {Accurate} {Real} {Time} {Deformable}
  {Objects}}. In \bibinfo{booktitle}{\emph{Proceedings of the 26th Annual
  Conference on Computer Graphics and Interactive Techniques (SIGGRAPH '99)}}.
  \bibinfo{pages}{65--72}.
\newblock
\href{https://doi.org/10.1145/311535.311542}{doi:\nolinkurl{10.1145/311535.311542}}


\bibitem[Kerbl et~al\mbox{.}(2023)]%
        {kerbl3DGaussianSplatting2023a}
\bibfield{author}{\bibinfo{person}{Bernhard Kerbl}, \bibinfo{person}{Georgios
  Kopanas}, \bibinfo{person}{Thomas Leimkuehler}, {and} \bibinfo{person}{George
  Drettakis}.} \bibinfo{year}{2023}\natexlab{}.
\newblock \showarticletitle{{3D} {Gaussian} {Splatting} for {Real}-{Time}
  {Radiance} {Field} {Rendering}}.
\newblock \bibinfo{journal}{\emph{ACM Transactions on Graphics}}
  \bibinfo{volume}{42}, \bibinfo{number}{4} (\bibinfo{date}{July}
  \bibinfo{year}{2023}), \bibinfo{pages}{139:1--139:14}.
\newblock
\showISSN{0730-0301}
\href{https://doi.org/10.1145/3592433}{doi:\nolinkurl{10.1145/3592433}}


\bibitem[Kress(2014)]%
        {kressLinearIntegralEquations2014}
\bibfield{author}{\bibinfo{person}{Rainer Kress}.}
  \bibinfo{year}{2014}\natexlab{}.
\newblock \bibinfo{booktitle}{\emph{Linear Integral Equations}
  (\bibinfo{edition}{3rd} ed.)}. \bibinfo{series}{Applied Mathematical
  Sciences}, Vol.~\bibinfo{volume}{82}.
\newblock \bibinfo{publisher}{Springer}.
\newblock
\href{https://doi.org/10.1007/978-1-4614-9593-2}{doi:\nolinkurl{10.1007/978-1-4614-9593-2}}


\bibitem[Larbi~Youcef et~al\mbox{.}(2022)]%
        {larbiyoucefInSituQuantitative2020}
\bibfield{author}{\bibinfo{person}{Mohamed Hafid~Abdelhak Larbi~Youcef},
  \bibinfo{person}{Vincent Feuillet}, \bibinfo{person}{Laurent Ibos}, {and}
  \bibinfo{person}{Yves Candau}.} \bibinfo{year}{2022}\natexlab{}.
\newblock \showarticletitle{In situ quantitative diagnosis of insulated
  building walls using passive infrared thermography}.
\newblock \bibinfo{journal}{\emph{Quantitative InfraRed Thermography Journal}}
  \bibinfo{volume}{19}, \bibinfo{number}{1} (\bibinfo{year}{2022}),
  \bibinfo{pages}{41--69}.
\newblock
\href{https://doi.org/10.1080/17686733.2020.1805939}{doi:\nolinkurl{10.1080/17686733.2020.1805939}}


\bibitem[Lesnic et~al\mbox{.}(1997)]%
        {lesnicIterativeBEMCauchy1997}
\bibfield{author}{\bibinfo{person}{Daniel Lesnic}, \bibinfo{person}{Lionel
  Elliott}, {and} \bibinfo{person}{Derek~B. Ingham}.}
  \bibinfo{year}{1997}\natexlab{}.
\newblock \showarticletitle{An iterative boundary element method for solving
  numerically the {Cauchy} problem for the {Laplace} equation}.
\newblock \bibinfo{journal}{\emph{Engineering Analysis with Boundary Elements}}
  \bibinfo{volume}{20}, \bibinfo{number}{2} (\bibinfo{year}{1997}),
  \bibinfo{pages}{123--133}.
\newblock


\bibitem[Li and Aluru(2002)]%
        {liBoundaryCloudMethod2002}
\bibfield{author}{\bibinfo{person}{Gang Li} {and} \bibinfo{person}{N.~R.
  Aluru}.} \bibinfo{year}{2002}\natexlab{}.
\newblock \showarticletitle{Boundary cloud method: a combined scattered
  point/boundary integral approach for boundary-only analysis}.
\newblock \bibinfo{journal}{\emph{Computer Methods in Applied Mechanics and
  Engineering}} \bibinfo{volume}{191}, \bibinfo{number}{21--22}
  (\bibinfo{year}{2002}), \bibinfo{pages}{2337--2370}.
\newblock
\href{https://doi.org/10.1016/S0045-7825(01)00415-7}{doi:\nolinkurl{10.1016/S0045-7825(01)00415-7}}


\bibitem[Lin et~al\mbox{.}(2023)]%
        {linSurfaceReconstructionGauss2023}
\bibfield{author}{\bibinfo{person}{Siyou Lin}, \bibinfo{person}{Dong Xiao},
  \bibinfo{person}{Zuoqiang Shi}, {and} \bibinfo{person}{Bin Wang}.}
  \bibinfo{year}{2023}\natexlab{}.
\newblock \showarticletitle{Surface reconstruction from point clouds without
  normals by parametrizing the {Gauss} formula}.
\newblock \bibinfo{journal}{\emph{ACM Transactions on Graphics}}
  \bibinfo{volume}{42}, \bibinfo{number}{2} (\bibinfo{year}{2023}),
  \bibinfo{pages}{14:1--14:19}.
\newblock
\href{https://doi.org/10.1145/3554730}{doi:\nolinkurl{10.1145/3554730}}


\bibitem[Lu et~al\mbox{.}(2025)]%
        {luThermalGaussianThermal3D2024}
\bibfield{author}{\bibinfo{person}{Rongfeng Lu}, \bibinfo{person}{Hangyu Chen},
  \bibinfo{person}{Zunjie Zhu}, \bibinfo{person}{Yuhang Qin},
  \bibinfo{person}{Ming Lu}, \bibinfo{person}{Le Zhang},
  \bibinfo{person}{Chenggang Yan}, {and} \bibinfo{person}{Anke Xue}.}
  \bibinfo{year}{2025}\natexlab{}.
\newblock \showarticletitle{{ThermalGaussian}: {Thermal} {3D} {Gaussian}
  {Splatting}}. In \bibinfo{booktitle}{\emph{International {Conference} on
  {Learning} {Representations} ({ICLR})}}.
\newblock
\urldef\tempurl%
\url{https://proceedings.iclr.cc/paper_files/paper/2025/hash/03bdba50e3741ac5e3eaa0e55423587e-Abstract-Conference.html}
\showURL{%
\tempurl}


\bibitem[Lu et~al\mbox{.}(2019)]%
        {luSurfaceReconstructionGauss2019}
\bibfield{author}{\bibinfo{person}{Wenjia Lu}, \bibinfo{person}{Zuoqiang Shi},
  \bibinfo{person}{Jian Sun}, {and} \bibinfo{person}{Bin Wang}.}
  \bibinfo{year}{2019}\natexlab{}.
\newblock \showarticletitle{Surface reconstruction based on the modified
  {Gauss} formula}.
\newblock \bibinfo{journal}{\emph{ACM Transactions on Graphics}}
  \bibinfo{volume}{38}, \bibinfo{number}{1} (\bibinfo{year}{2019}),
  \bibinfo{pages}{2:1--2:18}.
\newblock
\href{https://doi.org/10.1145/3233984}{doi:\nolinkurl{10.1145/3233984}}


\bibitem[Maldague(2001)]%
        {maldagueTheoryPracticeInfrared2001}
\bibfield{author}{\bibinfo{person}{Xavier P.~V. Maldague}.}
  \bibinfo{year}{2001}\natexlab{}.
\newblock \bibinfo{booktitle}{\emph{Theory and Practice of Infrared Technology
  for Nondestructive Testing}}.
\newblock \bibinfo{publisher}{Wiley}, \bibinfo{address}{New York}.
\newblock


\bibitem[Mehta et~al\mbox{.}(2022)]%
        {mehtaLevelSetTheory2022}
\bibfield{author}{\bibinfo{person}{Ishit Mehta}, \bibinfo{person}{Manmohan
  Chandraker}, {and} \bibinfo{person}{Ravi Ramamoorthi}.}
  \bibinfo{year}{2022}\natexlab{}.
\newblock \showarticletitle{A Level Set Theory for Neural Implicit Evolution
  under Explicit Flows}. In \bibinfo{booktitle}{\emph{European Conference on
  Computer Vision (ECCV)}}. \bibinfo{pages}{711--729}.
\newblock
\href{https://doi.org/10.1007/978-3-031-20086-1_41}{doi:\nolinkurl{10.1007/978-3-031-20086-1_41}}


\bibitem[Mildenhall et~al\mbox{.}(2020)]%
        {mildenhallNeRFRepresentingScenes2020}
\bibfield{author}{\bibinfo{person}{Ben Mildenhall}, \bibinfo{person}{Pratul~P.
  Srinivasan}, \bibinfo{person}{Matthew Tancik}, \bibinfo{person}{Jonathan~T.
  Barron}, \bibinfo{person}{Ravi Ramamoorthi}, {and} \bibinfo{person}{Ren Ng}.}
  \bibinfo{year}{2020}\natexlab{}.
\newblock \showarticletitle{{NeRF}: {Representing} {Scenes} as {Neural}
  {Radiance} {Fields} for {View} {Synthesis}}. In
  \bibinfo{booktitle}{\emph{Computer {Vision} – {ECCV} 2020}}.
\newblock
\href{https://doi.org/10.1007/978-3-030-58452-8_24}{doi:\nolinkurl{10.1007/978-3-030-58452-8_24}}


\bibitem[Miller et~al\mbox{.}(2023)]%
        {millerBoundaryValueCaching2023}
\bibfield{author}{\bibinfo{person}{Bailey Miller}, \bibinfo{person}{Rohan
  Sawhney}, \bibinfo{person}{Keenan Crane}, {and} \bibinfo{person}{Ioannis
  Gkioulekas}.} \bibinfo{year}{2023}\natexlab{}.
\newblock \showarticletitle{Boundary {Value} {Caching} for {Walk} on
  {Spheres}}.
\newblock \bibinfo{journal}{\emph{ACM Trans. Graph.}} \bibinfo{volume}{42},
  \bibinfo{number}{4} (\bibinfo{date}{July} \bibinfo{year}{2023}),
  \bibinfo{pages}{82:1--82:11}.
\newblock
\showISSN{0730-0301}
\href{https://doi.org/10.1145/3592400}{doi:\nolinkurl{10.1145/3592400}}


\bibitem[Miller et~al\mbox{.}(2024a)]%
        {millerDifferentialWalkSpheres2024}
\bibfield{author}{\bibinfo{person}{Bailey Miller}, \bibinfo{person}{Rohan
  Sawhney}, \bibinfo{person}{Keenan Crane}, {and} \bibinfo{person}{Ioannis
  Gkioulekas}.} \bibinfo{year}{2024}\natexlab{a}.
\newblock \bibinfo{title}{Differential {Walk} on {Spheres}}.
\newblock
\shownote{arXiv:2405.12964}.
\newblock
\urldef\tempurl%
\url{http://arxiv.org/abs/2405.12964}
\showURL{%
\tempurl}


\bibitem[Miller et~al\mbox{.}(2024b)]%
        {millerWalkinRobinWalk2024}
\bibfield{author}{\bibinfo{person}{Bailey Miller}, \bibinfo{person}{Rohan
  Sawhney}, \bibinfo{person}{Keenan Crane}, {and} \bibinfo{person}{Ioannis
  Gkioulekas}.} \bibinfo{year}{2024}\natexlab{b}.
\newblock \showarticletitle{Walkin’ {Robin}: {Walk} on {Stars} with {Robin}
  {Boundary} {Conditions}}.
\newblock \bibinfo{journal}{\emph{ACM Trans. Graph.}} \bibinfo{volume}{43},
  \bibinfo{number}{4} (\bibinfo{date}{July} \bibinfo{year}{2024}),
  \bibinfo{pages}{41:1--41:18}.
\newblock
\showISSN{0730-0301}
\href{https://doi.org/10.1145/3658153}{doi:\nolinkurl{10.1145/3658153}}


\bibitem[Minkina and Dudzik(2009)]%
        {minkinaInfraredThermography2009}
\bibfield{author}{\bibinfo{person}{Waldemar Minkina} {and}
  \bibinfo{person}{Sebastian Dudzik}.} \bibinfo{year}{2009}\natexlab{}.
\newblock \bibinfo{booktitle}{\emph{Infrared Thermography: Errors and
  Uncertainties}}.
\newblock \bibinfo{publisher}{John Wiley \& Sons}.
\newblock


\bibitem[Modest and Mazumder(2021)]%
        {modestRadiativeHeatTransfer2021}
\bibfield{author}{\bibinfo{person}{Michael~F. Modest} {and}
  \bibinfo{person}{Sandip Mazumder}.} \bibinfo{year}{2021}\natexlab{}.
\newblock \bibinfo{booktitle}{\emph{Radiative Heat Transfer}
  (\bibinfo{edition}{4th} ed.)}.
\newblock \bibinfo{publisher}{Academic Press}.
\newblock
\href{https://doi.org/10.1016/C2018-0-03206-5}{doi:\nolinkurl{10.1016/C2018-0-03206-5}}


\bibitem[Muller(1956)]%
        {mullerContinuousMonteCarlo1956}
\bibfield{author}{\bibinfo{person}{Mervin~E. Muller}.}
  \bibinfo{year}{1956}\natexlab{}.
\newblock \showarticletitle{Some {Continuous} {Monte} {Carlo} {Methods} for the
  {Dirichlet} {Problem}}.
\newblock \bibinfo{journal}{\emph{The Annals of Mathematical Statistics}}
  \bibinfo{volume}{27}, \bibinfo{number}{3} (\bibinfo{year}{1956}),
  \bibinfo{pages}{569--589}.
\newblock
\showISSN{0003-4851}
\urldef\tempurl%
\url{https://www.jstor.org/stable/2237369}
\showURL{%
\tempurl}


\bibitem[Nabizadeh et~al\mbox{.}(2021)]%
        {nabizadehKelvinTransformationsSimulations2021}
\bibfield{author}{\bibinfo{person}{Mohammad~Sina Nabizadeh},
  \bibinfo{person}{Ravi Ramamoorthi}, {and} \bibinfo{person}{Albert Chern}.}
  \bibinfo{year}{2021}\natexlab{}.
\newblock \showarticletitle{Kelvin transformations for simulations on infinite
  domains}.
\newblock \bibinfo{journal}{\emph{ACM Trans. Graph.}} \bibinfo{volume}{40},
  \bibinfo{number}{4} (\bibinfo{date}{July} \bibinfo{year}{2021}),
  \bibinfo{pages}{97:1--97:15}.
\newblock
\showISSN{0730-0301}
\href{https://doi.org/10.1145/3450626.3459809}{doi:\nolinkurl{10.1145/3450626.3459809}}


\bibitem[Narayanan et~al\mbox{.}(2024)]%
        {narayananShapeHeatConduction2024}
\bibfield{author}{\bibinfo{person}{Sriram Narayanan}, \bibinfo{person}{Mani
  Ramanagopal}, \bibinfo{person}{Mark Sheinin}, \bibinfo{person}{Aswin~C.
  Sankaranarayanan}, {and} \bibinfo{person}{Srinivasa~G. Narasimhan}.}
  \bibinfo{year}{2024}\natexlab{}.
\newblock \showarticletitle{Shape from {Heat} {Conduction}}. In
  \bibinfo{booktitle}{\emph{Computer {Vision} – {ECCV} 2024: 18th {European}
  {Conference}, {Milan}, {Italy}, {September} 29–{October} 4, 2024,
  {Proceedings}, {Part} {XXXVIII}}}. \bibinfo{publisher}{Springer-Verlag},
  \bibinfo{address}{Berlin, Heidelberg}, \bibinfo{pages}{426--444}.
\newblock
\showISBNx{978-3-031-72919-5}
\href{https://doi.org/10.1007/978-3-031-72920-1_24}{doi:\nolinkurl{10.1007/978-3-031-72920-1_24}}


\bibitem[Paszke et~al\mbox{.}(2019)]%
        {paszkePyTorchImperativeStyle2019}
\bibfield{author}{\bibinfo{person}{Adam Paszke}, \bibinfo{person}{Sam Gross},
  \bibinfo{person}{Francisco Massa}, \bibinfo{person}{Adam Lerer},
  \bibinfo{person}{James Bradbury}, \bibinfo{person}{Gregory Chanan},
  \bibinfo{person}{Trevor Killeen}, \bibinfo{person}{Zeming Lin},
  \bibinfo{person}{Natalia Gimelshein}, \bibinfo{person}{Luca Antiga},
  \bibinfo{person}{Alban Desmaison}, \bibinfo{person}{Andreas K{\"o}pf},
  \bibinfo{person}{Edward Yang}, \bibinfo{person}{Zachary DeVito},
  \bibinfo{person}{Martin Raison}, \bibinfo{person}{Alykhan Tejani},
  \bibinfo{person}{Sasank Chilamkurthy}, \bibinfo{person}{Benoit Steiner},
  \bibinfo{person}{Lu Fang}, \bibinfo{person}{Junjie Bai}, {and}
  \bibinfo{person}{Soumith Chintala}.} \bibinfo{year}{2019}\natexlab{}.
\newblock \showarticletitle{{PyTorch}: An imperative style, high-performance
  deep learning library}. In \bibinfo{booktitle}{\emph{Advances in Neural
  Information Processing Systems}}, Vol.~\bibinfo{volume}{32}.
\newblock


\bibitem[P{\'e}rez-Arancibia et~al\mbox{.}(2019)]%
        {perezarancibiaHarmonicDensityInterpolation2019}
\bibfield{author}{\bibinfo{person}{Carlos P{\'e}rez-Arancibia},
  \bibinfo{person}{Luiz~M. Faria}, {and} \bibinfo{person}{Catalin Turc}.}
  \bibinfo{year}{2019}\natexlab{}.
\newblock \showarticletitle{Harmonic Density Interpolation Methods for
  High-Order Evaluation of {{Laplace}} Layer Potentials in {{2D}} and {{3D}}}.
\newblock \bibinfo{journal}{\emph{J. Comput. Phys.}}  \bibinfo{volume}{376}
  (\bibinfo{year}{2019}), \bibinfo{pages}{411--434}.
\newblock
\href{https://doi.org/10.1016/j.jcp.2018.10.002}{doi:\nolinkurl{10.1016/j.jcp.2018.10.002}}


\bibitem[Pfister et~al\mbox{.}(2000)]%
        {pfisterSurfelsSurfaceElements2000}
\bibfield{author}{\bibinfo{person}{Hanspeter Pfister},
  \bibinfo{person}{Matthias Zwicker}, \bibinfo{person}{Jeroen van Baar}, {and}
  \bibinfo{person}{Markus Gross}.} \bibinfo{year}{2000}\natexlab{}.
\newblock \showarticletitle{Surfels: Surface Elements as Rendering Primitives}.
  In \bibinfo{booktitle}{\emph{Proceedings of the 27th Annual Conference on
  Computer Graphics and Interactive Techniques (SIGGRAPH '00)}}.
  \bibinfo{pages}{335--342}.
\newblock
\href{https://doi.org/10.1145/344779.344940}{doi:\nolinkurl{10.1145/344779.344940}}


\bibitem[Qi et~al\mbox{.}(2022)]%
        {qiBidirectionalFormulationWalk2022}
\bibfield{author}{\bibinfo{person}{Yang Qi}, \bibinfo{person}{Dario Seyb},
  \bibinfo{person}{Benedikt Bitterli}, {and} \bibinfo{person}{Wojciech
  Jarosz}.} \bibinfo{year}{2022}\natexlab{}.
\newblock \showarticletitle{A bidirectional formulation for {Walk} on
  {Spheres}}.
\newblock \bibinfo{journal}{\emph{Computer Graphics Forum}}
  \bibinfo{volume}{41}, \bibinfo{number}{4} (\bibinfo{year}{2022}),
  \bibinfo{pages}{51--62}.
\newblock
\showISSN{1467-8659}
\href{https://doi.org/10.1111/cgf.14586}{doi:\nolinkurl{10.1111/cgf.14586}}


\bibitem[Råback et~al\mbox{.}(2024)]%
        {elmer}
\bibfield{author}{\bibinfo{person}{Peter Råback}, \bibinfo{person}{Mika
  Malinen}, \bibinfo{person}{Juha Ruokolainen}, \bibinfo{person}{Antti
  Pursula}, {and} \bibinfo{person}{Thomas Zwinger}.}
  \bibinfo{year}{2024}\natexlab{}.
\newblock \bibinfo{booktitle}{\emph{Elmer FEM}}.
\newblock
\urldef\tempurl%
\url{https://www.elmerfem.org}
\showURL{%
\tempurl}


\bibitem[Saad and Schultz(1986)]%
        {saadGMRESGeneralizedMinimal1986}
\bibfield{author}{\bibinfo{person}{Youcef Saad} {and}
  \bibinfo{person}{Martin~H. Schultz}.} \bibinfo{year}{1986}\natexlab{}.
\newblock \showarticletitle{{GMRES}: A Generalized Minimal Residual Algorithm
  for Solving Nonsymmetric Linear Systems}.
\newblock \bibinfo{journal}{\emph{SIAM J. Sci. Statist. Comput.}}
  \bibinfo{volume}{7}, \bibinfo{number}{3} (\bibinfo{year}{1986}),
  \bibinfo{pages}{856--869}.
\newblock
\href{https://doi.org/10.1137/0907058}{doi:\nolinkurl{10.1137/0907058}}


\bibitem[Sauter and Schwab(2011)]%
        {sauterBoundaryElementMethods2011}
\bibfield{author}{\bibinfo{person}{Stefan~A. Sauter} {and}
  \bibinfo{person}{Christoph Schwab}.} \bibinfo{year}{2011}\natexlab{}.
\newblock \bibinfo{booktitle}{\emph{Boundary Element Methods}}.
  \bibinfo{series}{Springer Series in Computational Mathematics},
  Vol.~\bibinfo{volume}{39}.
\newblock \bibinfo{publisher}{Springer}.
\newblock
\href{https://doi.org/10.1007/978-3-540-68093-2}{doi:\nolinkurl{10.1007/978-3-540-68093-2}}


\bibitem[Sawhney and Crane(2020)]%
        {sawhneyMonteCarloGeometry2020}
\bibfield{author}{\bibinfo{person}{Rohan Sawhney} {and} \bibinfo{person}{Keenan
  Crane}.} \bibinfo{year}{2020}\natexlab{}.
\newblock \showarticletitle{Monte {Carlo} Geometry Processing: A Grid-Free
  Approach to {PDE}-Based Methods on Volumetric Domains}.
\newblock \bibinfo{journal}{\emph{ACM Trans. Graph.}} \bibinfo{volume}{39},
  \bibinfo{number}{4} (\bibinfo{year}{2020}).
\newblock
\href{https://doi.org/10.1145/3386569.3392374}{doi:\nolinkurl{10.1145/3386569.3392374}}


\bibitem[Sawhney et~al\mbox{.}(2023)]%
        {sawhneyWalkStarsGridFree2023}
\bibfield{author}{\bibinfo{person}{Rohan Sawhney}, \bibinfo{person}{Bailey
  Miller}, \bibinfo{person}{Ioannis Gkioulekas}, {and} \bibinfo{person}{Keenan
  Crane}.} \bibinfo{year}{2023}\natexlab{}.
\newblock \showarticletitle{Walk on {Stars}: {A} {Grid}-{Free} {Monte} {Carlo}
  {Method} for {PDEs} with {Neumann} {Boundary} {Conditions}}.
\newblock \bibinfo{journal}{\emph{ACM Transactions on Graphics}}
  \bibinfo{volume}{42}, \bibinfo{number}{4} (\bibinfo{date}{July}
  \bibinfo{year}{2023}), \bibinfo{pages}{80:1--80:20}.
\newblock
\showISSN{0730-0301}
\href{https://doi.org/10.1145/3592398}{doi:\nolinkurl{10.1145/3592398}}


\bibitem[Sawhney et~al\mbox{.}(2022)]%
        {sawhneyGridfreeMonteCarlo2022}
\bibfield{author}{\bibinfo{person}{Rohan Sawhney}, \bibinfo{person}{Dario
  Seyb}, \bibinfo{person}{Wojciech Jarosz}, {and} \bibinfo{person}{Keenan
  Crane}.} \bibinfo{year}{2022}\natexlab{}.
\newblock \showarticletitle{Grid-free {Monte} {Carlo} for {PDEs} with spatially
  varying coefficients}.
\newblock \bibinfo{journal}{\emph{ACM Transactions on Graphics}}
  \bibinfo{volume}{41}, \bibinfo{number}{4} (\bibinfo{date}{July}
  \bibinfo{year}{2022}), \bibinfo{pages}{53:1--53:17}.
\newblock
\showISSN{0730-0301}
\href{https://doi.org/10.1145/3528223.3530134}{doi:\nolinkurl{10.1145/3528223.3530134}}


\bibitem[Sch{\"a}fer et~al\mbox{.}(2021)]%
        {schaferSparseCholeskyFactorization2021}
\bibfield{author}{\bibinfo{person}{Florian Sch{\"a}fer},
  \bibinfo{person}{Matthias Katzfuss}, {and} \bibinfo{person}{Houman Owhadi}.}
  \bibinfo{year}{2021}\natexlab{}.
\newblock \showarticletitle{Sparse {Cholesky} Factorization by
  {Kullback--Leibler} Minimization}.
\newblock \bibinfo{journal}{\emph{SIAM Journal on Scientific Computing}}
  \bibinfo{volume}{43}, \bibinfo{number}{3} (\bibinfo{year}{2021}),
  \bibinfo{pages}{A2019--A2046}.
\newblock
\href{https://doi.org/10.1137/20M1336254}{doi:\nolinkurl{10.1137/20M1336254}}


\bibitem[Sch\"{o}nberger and Frahm(2016)]%
        {schonbergerStructurefromMotionRevisited2016}
\bibfield{author}{\bibinfo{person}{Johannes~L. Sch\"{o}nberger} {and}
  \bibinfo{person}{Jan-Michael Frahm}.} \bibinfo{year}{2016}\natexlab{}.
\newblock \showarticletitle{Structure-from-{Motion} {Revisited}}. In
  \bibinfo{booktitle}{\emph{Proceedings of the {IEEE} Conference on Computer
  Vision and Pattern Recognition ({CVPR})}}. \bibinfo{pages}{4104--4113}.
\newblock
\href{https://doi.org/10.1109/CVPR.2016.445}{doi:\nolinkurl{10.1109/CVPR.2016.445}}


\bibitem[Shin et~al\mbox{.}(2015)]%
        {shinDesignRadiantFloor2015}
\bibfield{author}{\bibinfo{person}{Mi~Su Shin}, \bibinfo{person}{Kyu~Nam Rhee},
  \bibinfo{person}{Seong~Ryong Ryu}, \bibinfo{person}{Myoung~Souk Yeo}, {and}
  \bibinfo{person}{Kwang~Woo Kim}.} \bibinfo{year}{2015}\natexlab{}.
\newblock \showarticletitle{Design of radiant floor heating panel in view of
  floor surface temperatures}.
\newblock \bibinfo{journal}{\emph{Building and Environment}}
  \bibinfo{volume}{92} (\bibinfo{year}{2015}), \bibinfo{pages}{559--577}.
\newblock
\href{https://doi.org/10.1016/j.buildenv.2015.05.006}{doi:\nolinkurl{10.1016/j.buildenv.2015.05.006}}


\bibitem[Stein(2024)]%
        {steinOdedsteinMeshes2024}
\bibfield{author}{\bibinfo{person}{Oded Stein}.}
  \bibinfo{year}{2024}\natexlab{}.
\newblock \bibinfo{title}{odedstein-meshes: A Computer Graphics Example Mesh
  Repository}.
\newblock \bibinfo{howpublished}{\url{https://github.com/odedstein/meshes}}.
\newblock


\bibitem[Stuart and Pound(2025)]%
        {stuart3DGSToPC2025}
\bibfield{author}{\bibinfo{person}{Lewis A.~G. Stuart} {and}
  \bibinfo{person}{Michael~P. Pound}.} \bibinfo{year}{2025}\natexlab{}.
\newblock \bibinfo{title}{{3DGS-to-PC}: Convert a {3D} {Gaussian} splatting
  scene into a dense point cloud}.
\newblock
\showeprint[arxiv]{2501.07478}


\bibitem[Sugimoto et~al\mbox{.}(2023)]%
        {sugimotoPracticalWalkonBoundaryMethod2023}
\bibfield{author}{\bibinfo{person}{Ryusuke Sugimoto}, \bibinfo{person}{Terry
  Chen}, \bibinfo{person}{Yiti Jiang}, \bibinfo{person}{Christopher Batty},
  {and} \bibinfo{person}{Toshiya Hachisuka}.} \bibinfo{year}{2023}\natexlab{}.
\newblock \showarticletitle{A {Practical} {Walk}-on-{Boundary} {Method} for
  {Boundary} {Value} {Problems}}.
\newblock \bibinfo{journal}{\emph{ACM Trans. Graph.}} \bibinfo{volume}{42},
  \bibinfo{number}{4} (\bibinfo{date}{July} \bibinfo{year}{2023}),
  \bibinfo{pages}{81:1--81:16}.
\newblock
\showISSN{0730-0301}
\href{https://doi.org/10.1145/3592109}{doi:\nolinkurl{10.1145/3592109}}


\bibitem[Wang et~al\mbox{.}(2024)]%
        {wangNewMethodPixellevel2023}
\bibfield{author}{\bibinfo{person}{Zihao Wang}, \bibinfo{person}{Yu Hou}, {and}
  \bibinfo{person}{Lucio Soibelman}.} \bibinfo{year}{2024}\natexlab{}.
\newblock \showarticletitle{A new method of pixel-level in situ {U}-value
  measurement for building envelopes based on infrared thermography}. In
  \bibinfo{booktitle}{\emph{Computing in Civil Engineering 2023}}.
  \bibinfo{publisher}{ASCE}, \bibinfo{pages}{697--704}.
\newblock
\href{https://doi.org/10.1061/9780784485248.084}{doi:\nolinkurl{10.1061/9780784485248.084}}


\bibitem[Xie et~al\mbox{.}(2024)]%
        {xiePhysGaussianPhysicsIntegrated2024}
\bibfield{author}{\bibinfo{person}{Tianyi Xie}, \bibinfo{person}{Zeshun Zong},
  \bibinfo{person}{Yuxing Qiu}, \bibinfo{person}{Xuan Li},
  \bibinfo{person}{Yutao Feng}, \bibinfo{person}{Yin Yang}, {and}
  \bibinfo{person}{Chenfanfu Jiang}.} \bibinfo{year}{2024}\natexlab{}.
\newblock \showarticletitle{{PhysGaussian}: Physics-integrated {3D} {Gaussians}
  for generative dynamics}. In \bibinfo{booktitle}{\emph{Proceedings of the
  IEEE/CVF Conference on Computer Vision and Pattern Recognition (CVPR)}}.
\newblock


\bibitem[Yang et~al\mbox{.}(2025)]%
        {yangNTRGaussianNighttime2025}
\bibfield{author}{\bibinfo{person}{Kun Yang}, \bibinfo{person}{Yuxiang Liu},
  \bibinfo{person}{Zeyu Cui}, \bibinfo{person}{Yu Liu}, \bibinfo{person}{Maojun
  Zhang}, \bibinfo{person}{Shen Yan}, {and} \bibinfo{person}{Qing Wang}.}
  \bibinfo{year}{2025}\natexlab{}.
\newblock \showarticletitle{{NTR-Gaussian}: Nighttime dynamic thermal
  reconstruction with {4D} {Gaussian} splatting based on thermodynamics}. In
  \bibinfo{booktitle}{\emph{Proceedings of the IEEE/CVF Conference on Computer
  Vision and Pattern Recognition (CVPR)}}. \bibinfo{pages}{691--700}.
\newblock


\bibitem[Ye et~al\mbox{.}(2024)]%
        {yeThermalNeRFNeuralRadiance2024}
\bibfield{author}{\bibinfo{person}{Tianxiang Ye}, \bibinfo{person}{Qi Wu},
  \bibinfo{person}{Junyuan Deng}, \bibinfo{person}{Guoqing Liu},
  \bibinfo{person}{Liu Liu}, \bibinfo{person}{Songpengcheng Xia},
  \bibinfo{person}{Liang Pang}, \bibinfo{person}{Wenxian Yu}, {and}
  \bibinfo{person}{Ling Pei}.} \bibinfo{year}{2024}\natexlab{}.
\newblock \bibinfo{title}{Thermal-{NeRF}: {Neural} {Radiance} {Fields} from an
  {Infrared} {Camera}}.
\newblock
\shownote{arXiv:2403.10340 [cs]}.
\newblock
\href{https://doi.org/10.48550/arXiv.2403.10340}{doi:\nolinkurl{10.48550/arXiv.2403.10340}}


\bibitem[Yilmazer et~al\mbox{.}(2024)]%
        {yilmazerSolvingInversePDE}
\bibfield{author}{\bibinfo{person}{Ekrem~Fatih Yilmazer},
  \bibinfo{person}{Delio Vicini}, {and} \bibinfo{person}{Wenzel Jakob}.}
  \bibinfo{year}{2024}\natexlab{}.
\newblock \showarticletitle{Solving {Inverse} {PDE} {Problems} using
  {Grid}-{Free} {Monte} {Carlo} {Estimators}}.
\newblock \bibinfo{journal}{\emph{Transactions on Graphics (Proceedings of
  SIGGRAPH Asia 2024)}} \bibinfo{volume}{43}, \bibinfo{number}{6}
  (\bibinfo{year}{2024}).
\newblock
\href{https://doi.org/10.1145/3687990}{doi:\nolinkurl{10.1145/3687990}}


\bibitem[Zhang et~al\mbox{.}(2026)]%
        {zhangRaDeGSRasterizingDepth2024}
\bibfield{author}{\bibinfo{person}{Baowen Zhang}, \bibinfo{person}{Chuan Fang},
  \bibinfo{person}{Rakesh Shrestha}, \bibinfo{person}{Yixun Liang},
  \bibinfo{person}{Xiao-Xiao Long}, {and} \bibinfo{person}{Ping Tan}.}
  \bibinfo{year}{2026}\natexlab{}.
\newblock \showarticletitle{{RaDe-GS}: Rasterizing Depth in {Gaussian}
  Splatting}.
\newblock \bibinfo{journal}{\emph{ACM Trans. Graph.}} \bibinfo{volume}{45},
  \bibinfo{number}{2} (\bibinfo{date}{March} \bibinfo{year}{2026}),
  \bibinfo{pages}{1--14}.
\newblock
\showISSN{0730-0301}
\href{https://doi.org/10.1145/3789201}{doi:\nolinkurl{10.1145/3789201}}


\bibitem[Zhenyuan et~al\mbox{.}(2025)]%
        {zhenyuanInverseRadiative2025}
\bibfield{author}{\bibinfo{person}{Liu Zhenyuan}, \bibinfo{person}{Bharath
  Seshadri}, \bibinfo{person}{George Kopanas}, {and} \bibinfo{person}{Bernd
  Bickel}.} \bibinfo{year}{2025}\natexlab{}.
\newblock \showarticletitle{Inverse {Radiative} {Transport} for {Infrared}
  {Scenes} with {Gaussian} {Primitives}}. In
  \bibinfo{booktitle}{\emph{Proceedings of the {SIGGRAPH} {Asia} 2025
  {Conference} {Papers}}} \emph{(\bibinfo{series}{{SA} {Conference} {Papers}
  '25})}. \bibinfo{publisher}{Association for Computing Machinery},
  \bibinfo{address}{New York, NY, USA}, \bibinfo{pages}{1--12}.
\newblock
\showISBNx{979-8-4007-2137-3}
\href{https://doi.org/10.1145/3757377.3763938}{doi:\nolinkurl{10.1145/3757377.3763938}}


\bibitem[Zwicker et~al\mbox{.}(2001)]%
        {zwickerSurfaceSplatting2001}
\bibfield{author}{\bibinfo{person}{Matthias Zwicker},
  \bibinfo{person}{Hanspeter Pfister}, \bibinfo{person}{Jeroen van Baar}, {and}
  \bibinfo{person}{Markus Gross}.} \bibinfo{year}{2001}\natexlab{}.
\newblock \showarticletitle{Surface Splatting}. In
  \bibinfo{booktitle}{\emph{Proceedings of the 28th Annual Conference on
  Computer Graphics and Interactive Techniques (SIGGRAPH '01)}}.
  \bibinfo{pages}{371--378}.
\newblock
\href{https://doi.org/10.1145/383259.383300}{doi:\nolinkurl{10.1145/383259.383300}}


\end{thebibliography}
